\documentclass{article}

\newif\ifarxiv
\arxivtrue

\newcommand{\confnotice}{}

\ifarxiv
    \usepackage[final,nonatbib]{neurips_2020} %
\else
    \usepackage[final,nonatbib]{neurips_2020} %
\fi

\ifarxiv
    \newcommand{\ext}{jpg} %
\else
    \newcommand{\ext}{png}
\fi

\usepackage[utf8]{inputenc} %
\usepackage[T1]{fontenc}    %
\usepackage{hyperref}       %
\usepackage{url}            %
\usepackage{booktabs}       %
\usepackage{amsfonts}       %
\usepackage{nicefrac}       %
\usepackage{microtype}      %
\usepackage{graphicx}
\usepackage{amsmath}
\usepackage[noend]{algpseudocode}
\usepackage{algorithmicx}
\usepackage{algorithm}
\usepackage{subcaption}
\usepackage{calc}
\usepackage{tikz}
\usepackage{array}
\usepackage{multirow}
\usepackage{tabu}
\usepackage{bm}

\graphicspath{{./figures/}}

\usepackage{color}
\usepackage[normalem]{ulem}  %
\usepackage[outline]{contour}
\usepackage{afterpage}
\usepackage{pifont}
\definecolor{olive}{rgb}{0.5, 0.5, 0.0}
\definecolor{maroon}{rgb}{0.69, 0.19, 0.38}
\definecolor{celestialblue}{rgb}{0.29, 0.59, 0.82}
\definecolor{darkgreen}{rgb}{0.0, 0.5, 0.0}
\definecolor{grey}{rgb}{0.5,0.5,0.5}
\definecolor{darkblue}{rgb}{0.19, 0.19, 0.62}

\newcommand{\FINAL}[1]{#1}

\def\clap#1{\hbox to 0pt{\hss #1\hss}}%

\newcommand{\real}{\mathbb{R}}
\newcommand{\integer}{\mathbb{Z}}
\newcommand{\proby}{\mathbf{y}}
\newcommand{\probx}{\mathbf{x}}
\newcommand{\probz}{\mathbf{z}}
\newcommand{\probn}{\mathbf{n}}
\newcommand{\aug}{\mathcal{T}}
\newcommand{\augu}{\mathcal{U}}
\newcommand{\augr}{\mathcal{R}}
\newcommand{\augg}{\mathcal{G}}
\newcommand{\augp}{\mathcal{P}}

\newcommand{\augeye}{\mathcal{I}}

\newcommand{\h}{0mm}
\newcommand{\hh}{0mm}
\newcommand{\hhh}{0mm}
\newcommand{\hhhh}{0mm}
\newcommand{\vv}{0mm}
\newcommand{\vvv}{0mm}
\newcommand{\vvvv}{0mm}

\newcommand{\s}{\hphantom{0}}
\newcommand{\cs}{\hspace{0}}

\newcommand{\subtabletop}[4]{\parbox[b][#2]{#1}{\footnotesize\centering\scalebox{#3}{#4}\vfill}} %

\title{Training Generative Adversarial Networks with Limited Data}

\author{%
	Tero Karras\\NVIDIA%
	\\\ifarxiv\else\texttt{tkarras@nvidia.com}\fi
	\And
	Miika Aittala\\NVIDIA%
	\\\ifarxiv\else\texttt{maittala@nvidia.com}\fi
	\And
	Janne Hellsten\\NVIDIA%
	\\\ifarxiv\else\texttt{jhellsten@nvidia.com}\fi
	\And
	Samuli Laine\\NVIDIA%
	\\\ifarxiv\else\texttt{slaine@nvidia.com}\fi
	\And
	Jaakko Lehtinen\\NVIDIA and Aalto University%
	\\\ifarxiv\else\texttt{jlehtinen@nvidia.com}\fi
	\And
	Timo Aila\\NVIDIA%
	\\\ifarxiv\else\texttt{taila@nvidia.com}\fi
}

\newcommand{\figOverfit}[1]{
\begin{figure}[tb]
\footnotesize%
\renewcommand{\h}{0.4\linewidth}%
\renewcommand{\hh}{0.3\linewidth}%
\includegraphics[width=\h]{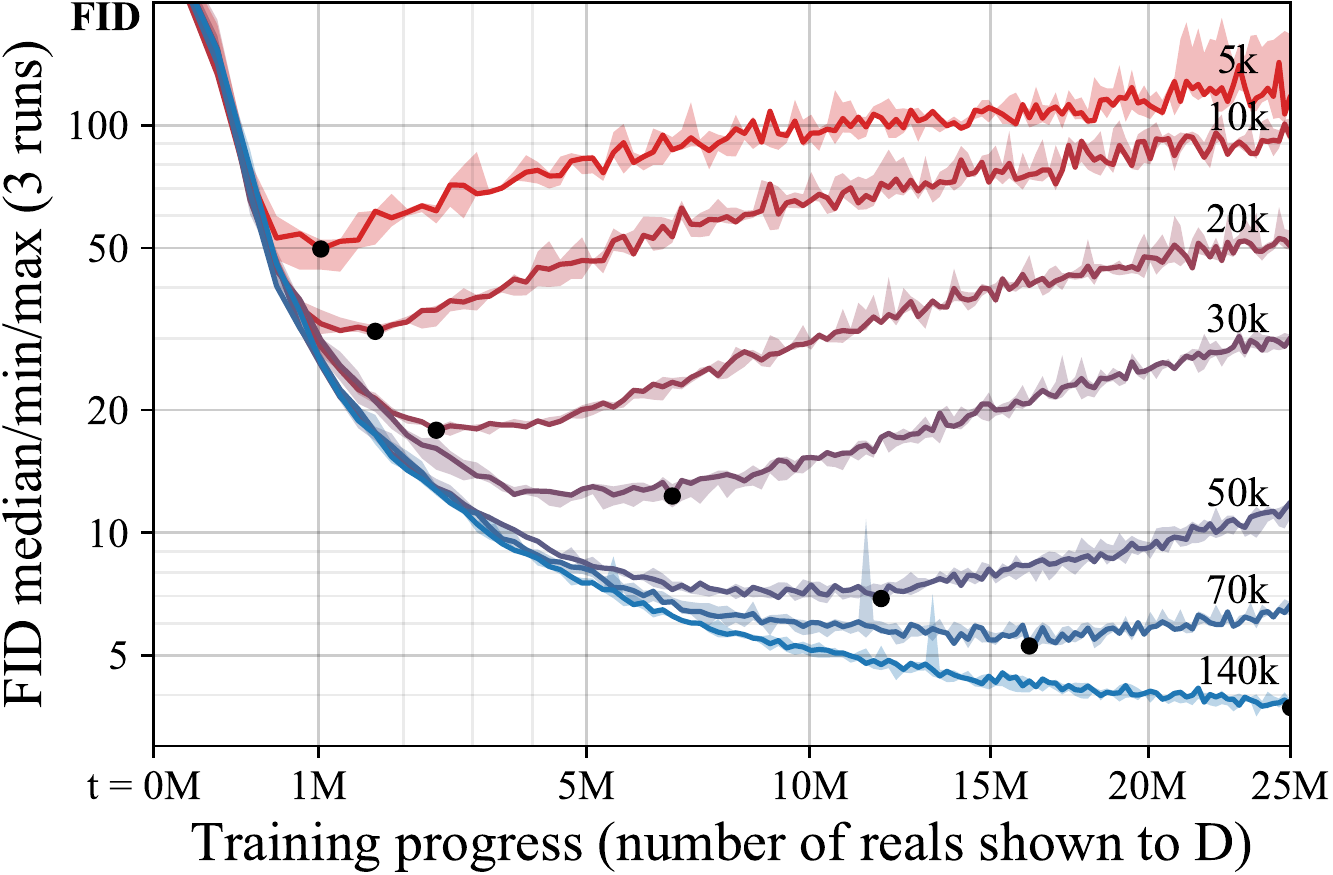}\hfill%
\includegraphics[width=\hh]{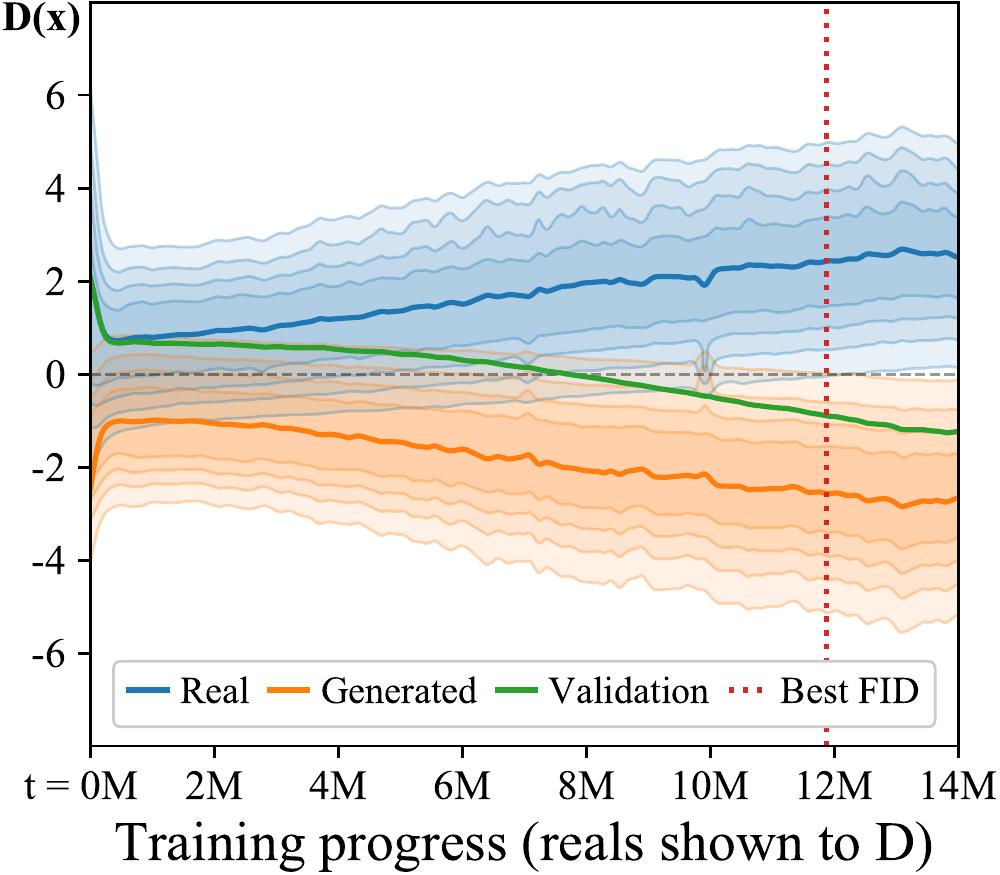}\hfill%
\includegraphics[width=\hh]{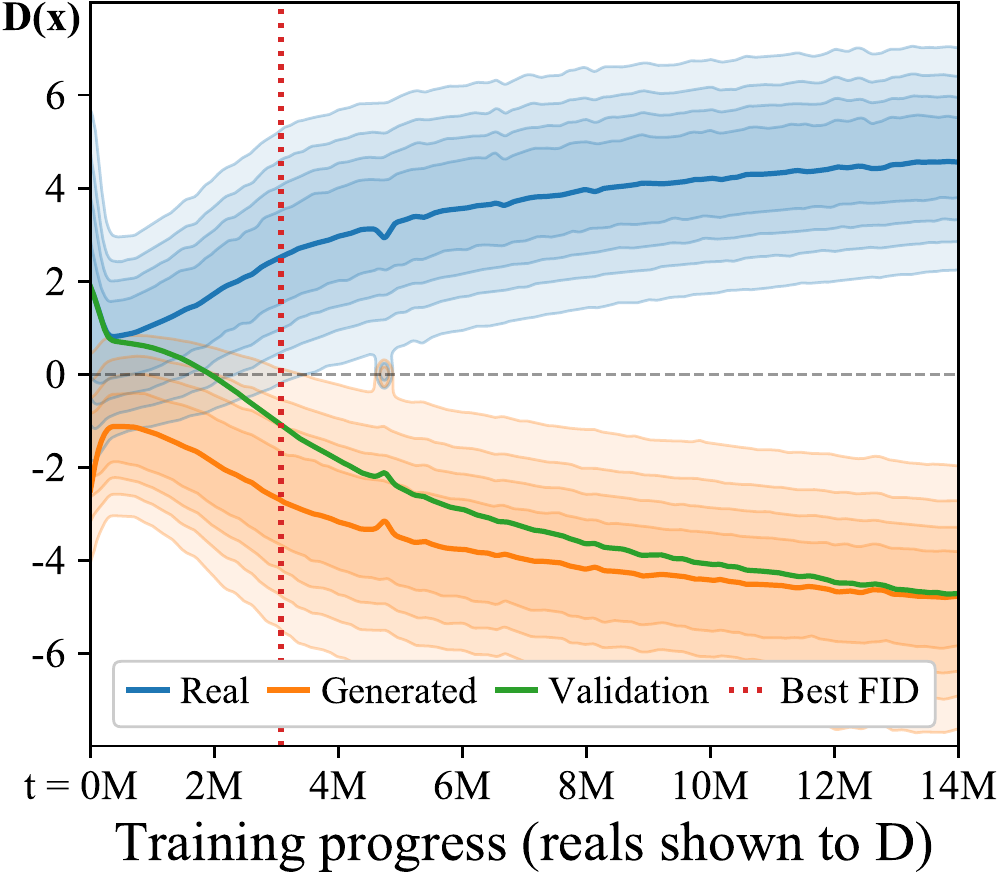}\vspace{0.8mm}\\
\makebox[\h][c]{(a) Convergence of FFHQ ($256\times256$)}\hfill%
\makebox[\hh][c]{(b) Discriminator outputs, 50k}\hfill%
\makebox[\hh][c]{(c) Discriminator outputs, 20k}%
\caption{
(a) Convergence with different training set sizes.
``140k'' means that we amplified the 70k dataset by $2\times$ through $x$-flips; we do not use data amplification in any other case.
(b,c) Evolution of discriminator outputs during training.
Each vertical slice shows a histogram of $D(x)$, i.e., raw logits.
}
\label{#1}
\end{figure}
}

\newcommand{\figDiagramsAndAugmentExamples}[1]{
\begin{figure}[t]
\footnotesize%
\centering%
\renewcommand{\h}{38mm}%
\renewcommand{\hh}{\h/\real{14}*\real{9.05}}%
\renewcommand{\hhh}{10.7mm}%
\renewcommand{\hhhh}{4mm}%
\renewcommand{\vv}{27.8mm}%
\parbox[b][\vv]{\h}{%
\includegraphics[width=\h,trim={0 0 0 4}]{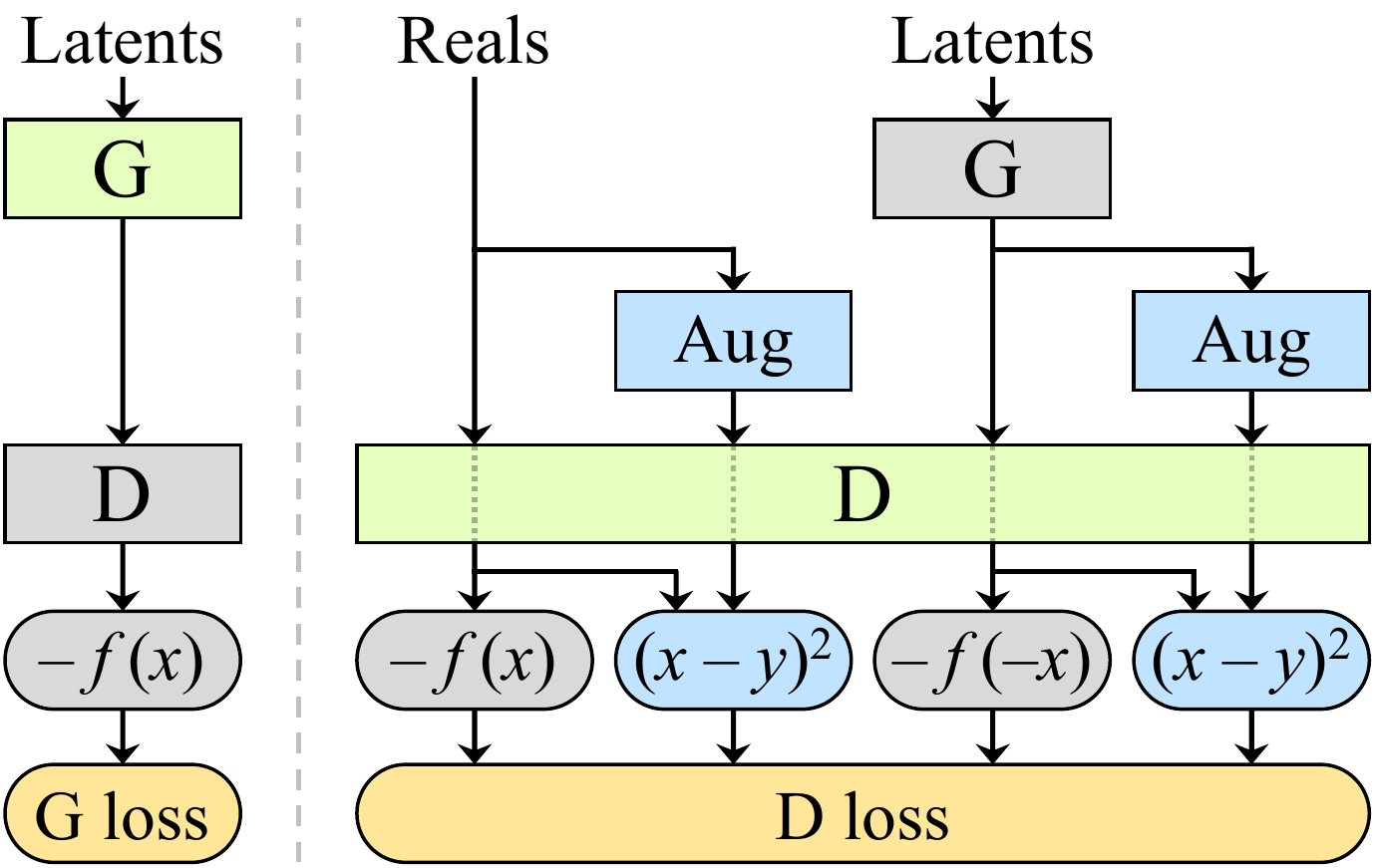}%
\vfill\makebox[\linewidth][c]{(a) bCR (previous work)}}%
\hfill\begin{tikzpicture}\draw (0,0) -- (0,\vv);\end{tikzpicture}\hfill%
\parbox[b][\vv]{\hh}{%
\includegraphics[width=\hh,trim={0 0 0 4}]{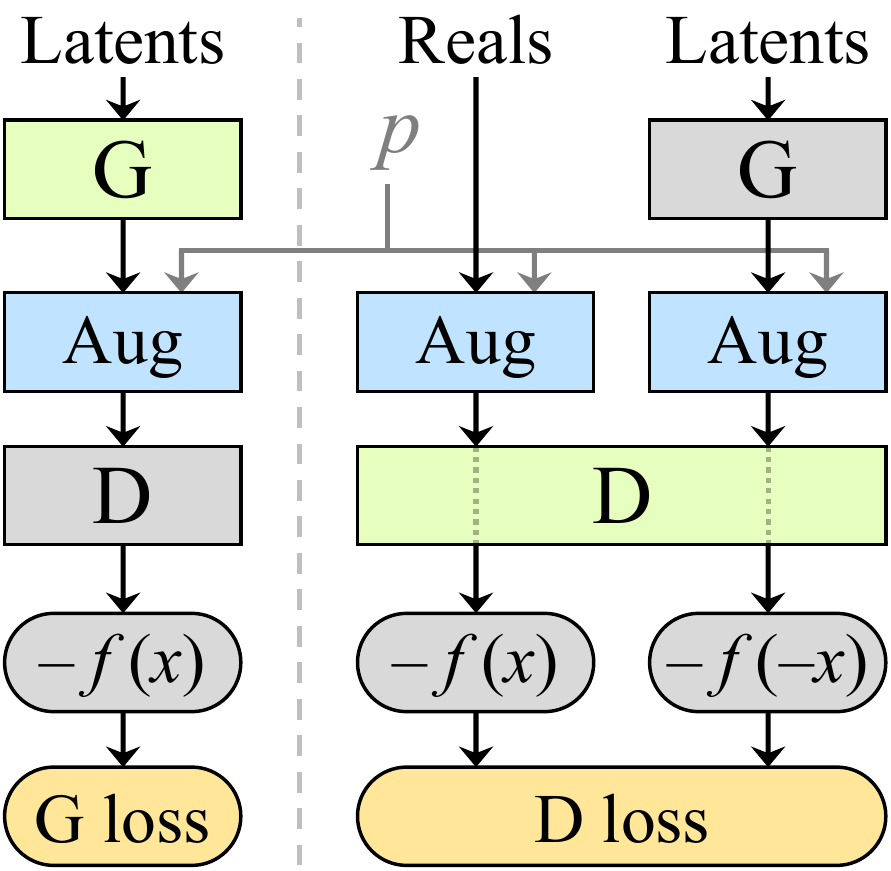}%
\vfill\makebox[\linewidth][c]{(b) \FINAL{Ours}}}%
\hfill\begin{tikzpicture}\draw (0,0) -- (0,\vv);\end{tikzpicture}\hfill%
\parbox[b][\vv]{\hhh*\real{6}+\hhhh}{%
\makebox[\hhh][c]{\tiny$p=0$}%
\makebox[\hhh][c]{\tiny$p=0.1$}\hfill%
\makebox[\hhh][c]{\tiny$p=0.2$}\hfill%
\makebox[\hhh][c]{\tiny$p=0.3$}\hfill%
\makebox[\hhh][c]{\tiny$p=0.5$}\hfill%
\makebox[\hhh][c]{\tiny$p=0.8$}%
\\
\includegraphics[width=\hhh]{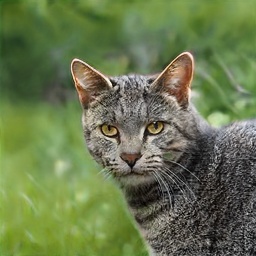}\hfill%
\includegraphics[width=\hhh]{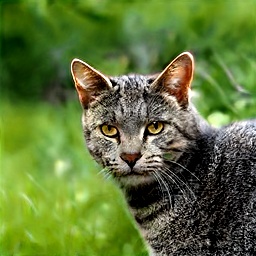}\hfill%
\includegraphics[width=\hhh]{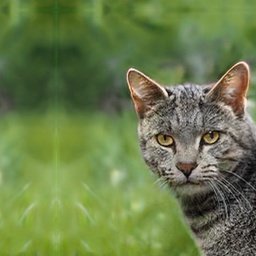}\hfill%
\includegraphics[width=\hhh]{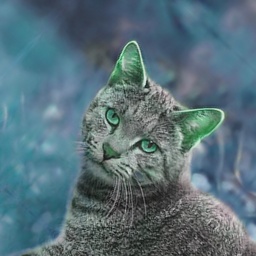}\hfill%
\includegraphics[width=\hhh]{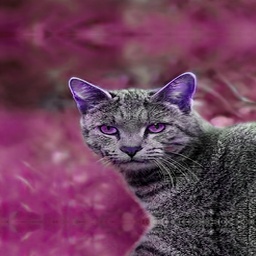}\hfill%
\includegraphics[width=\hhh]{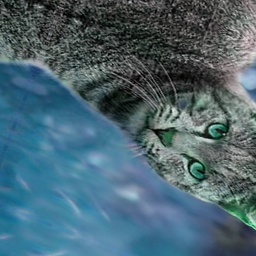}%
\vspace{-0.35mm}\\
\includegraphics[width=\hhh]{generated_images/augment_examples_simple/strength0_1x1_rep0.\ext}\hfill%
\includegraphics[width=\hhh]{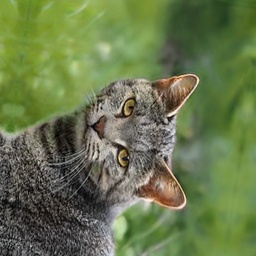}\hfill%
\includegraphics[width=\hhh]{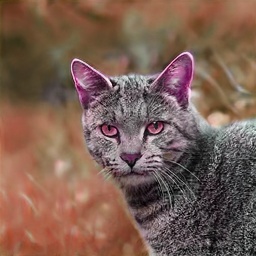}\hfill%
\includegraphics[width=\hhh]{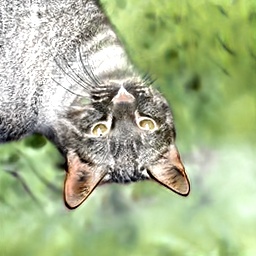}\hfill%
\includegraphics[width=\hhh]{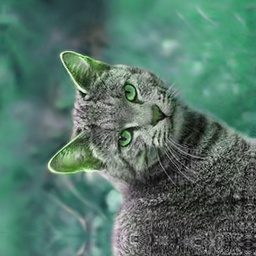}\hfill%
\includegraphics[width=\hhh]{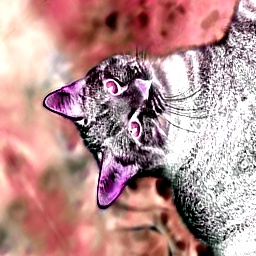}%
\vfill\makebox[\linewidth][c]{(c) Effect of augmentation probability $p$}}%
\caption{
(a,b) Flowcharts for balanced consistency regularization (bCR)~\cite{zhao2020improved} and our stochastic discriminator augmentations.
The blue elements highlight operations related to augmentations, while the rest implement standard GAN training with generator $G$ and discriminator $D$~\cite{Goodfellow2014}.
\FINAL{The orange elements indicate the loss function and the green boxes mark the network being trained.}  
We use the non-saturating logistic loss~\cite{Goodfellow2014} $f(x)=\log\left(\mathrm{sigmoid}(x)\right)$.
(c)
We apply a diverse set of augmentations to every image that the discriminator sees, controlled by an augmentation probability~$p$.
}
\label{#1}
\end{figure}
}

\newcommand{\leaklabel}[1]{\makebox(0,0)[r]{\raisebox{2.5mm}{\makebox[\h]{\scriptsize\color{white}\contourlength{0.15mm}\contour{black}{\bf\hfill#1\hfill}}}}}

\newcommand{\figLeakGroups}[1]{
\begin{figure}[t]
\footnotesize%
\centering%
\renewcommand{\h}{18mm}%
\renewcommand{\vv}{45.6mm}%
\renewcommand{\vvv}{22mm}%
\parbox[b][\vv]{0.265\linewidth}{%
\includegraphics[width=\h]{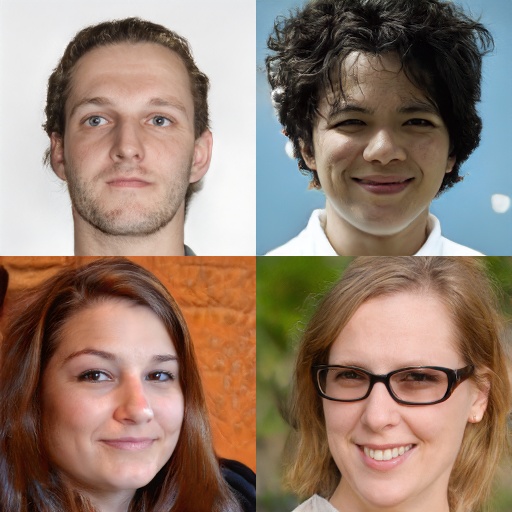}\leaklabel{A}\hfill%
\includegraphics[width=\h]{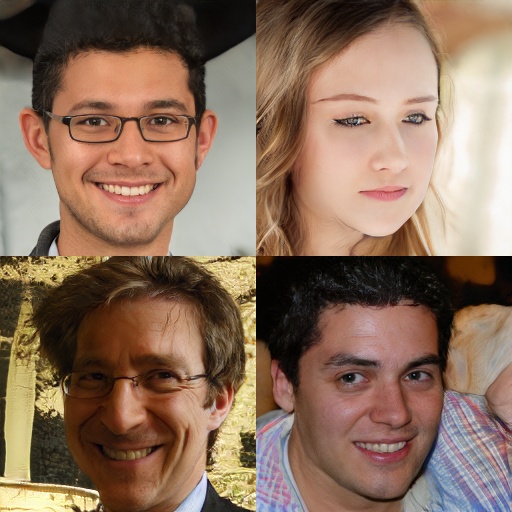}\leaklabel{B}\vfill%
\includegraphics[width=\linewidth,height=\vvv]{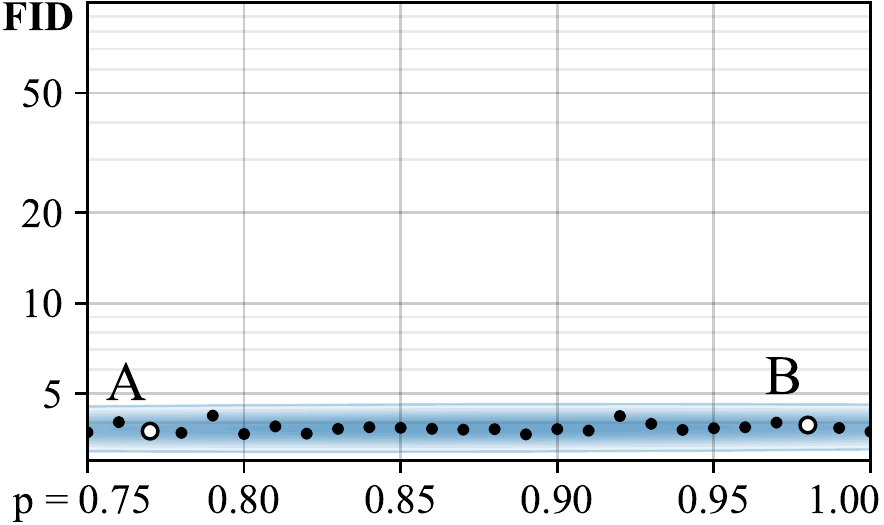}\\
\makebox[\linewidth][c]{(a) Isotropic image scaling}}%
\hfill\begin{tikzpicture}\draw (0,0) -- (0,\vv);\end{tikzpicture}\hfill%
\parbox[b][\vv]{0.400\linewidth}{%
\includegraphics[width=\h]{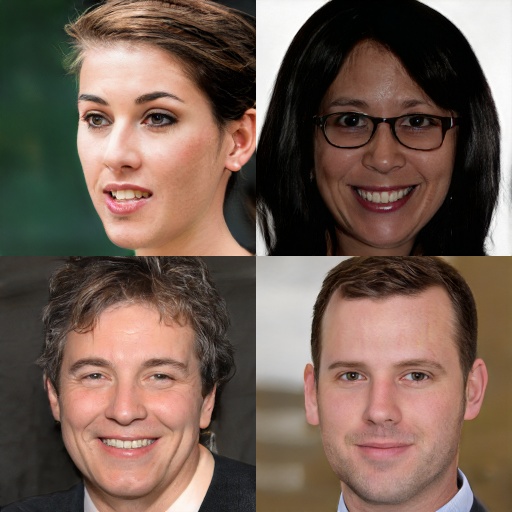}\leaklabel{C}\hfill%
\includegraphics[width=\h]{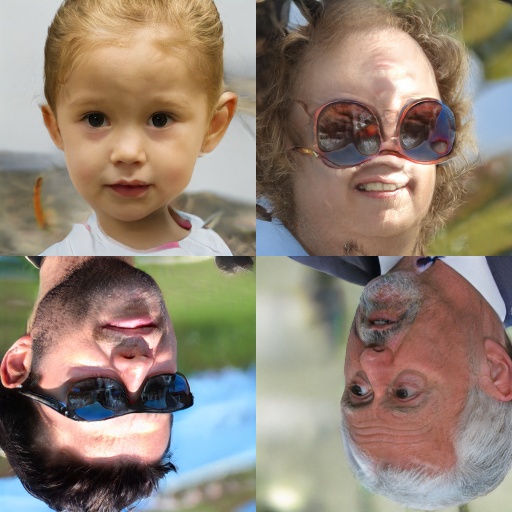}\leaklabel{D}\hfill%
\includegraphics[width=\h]{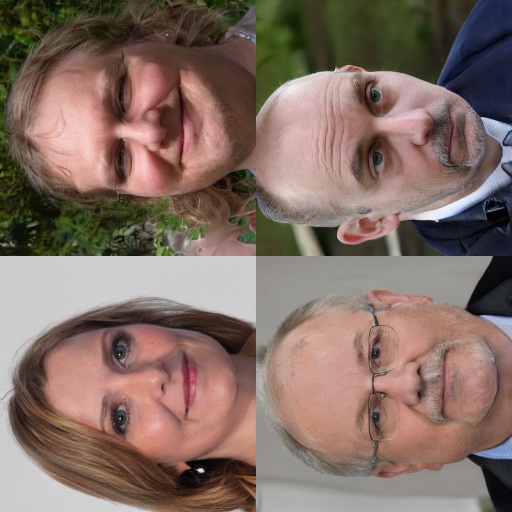}\leaklabel{E}\vfill%
\includegraphics[width=\linewidth,height=\vvv]{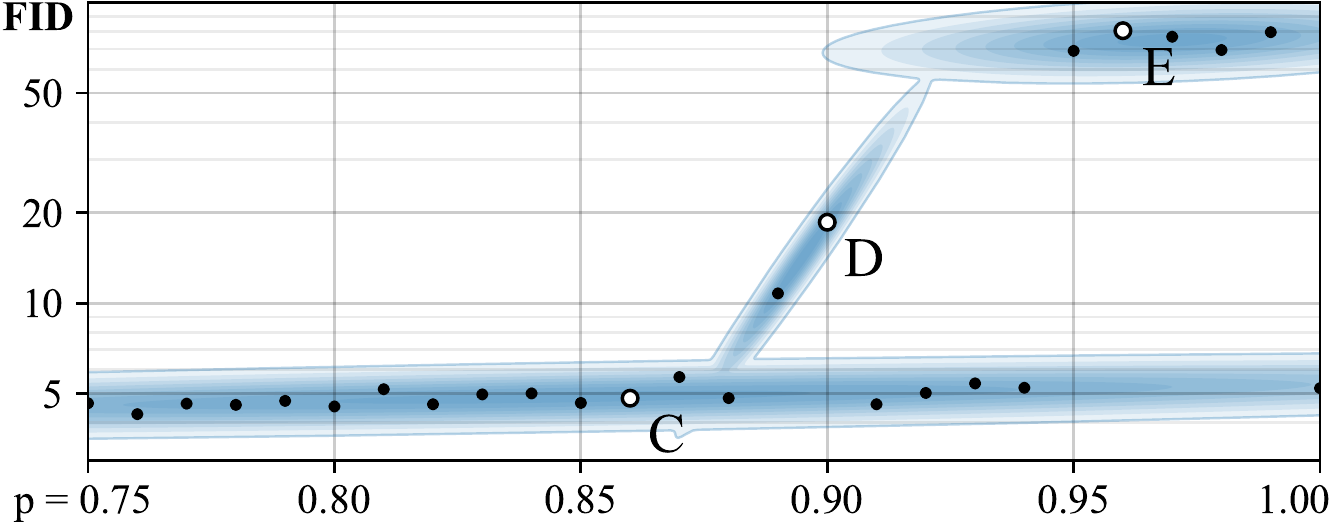}\\
\makebox[\linewidth][c]{(b) Random 90$^{\circ}$ rotations}}%
\hfill\begin{tikzpicture}\draw (0,0) -- (0,\vv);\end{tikzpicture}\hfill%
\parbox[b][\vv]{0.265\linewidth}{%
\includegraphics[width=\h]{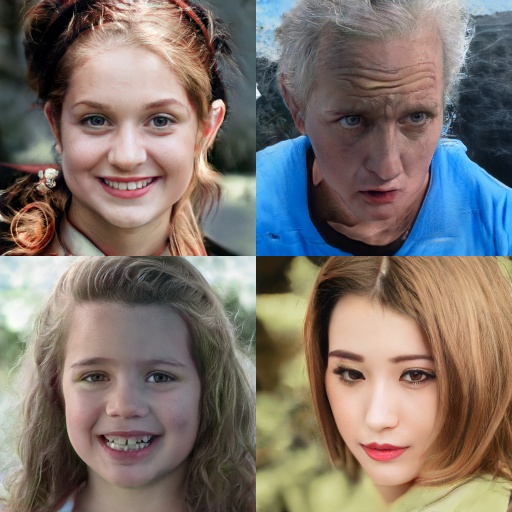}\leaklabel{F}\hfill%
\includegraphics[width=\h]{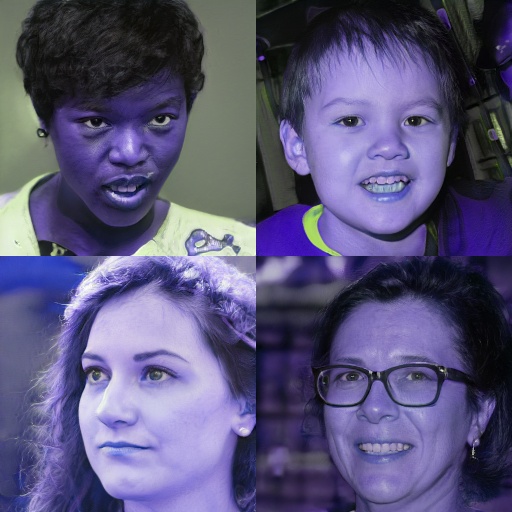}\leaklabel{G}\vfill%
\includegraphics[width=\linewidth,height=\vvv]{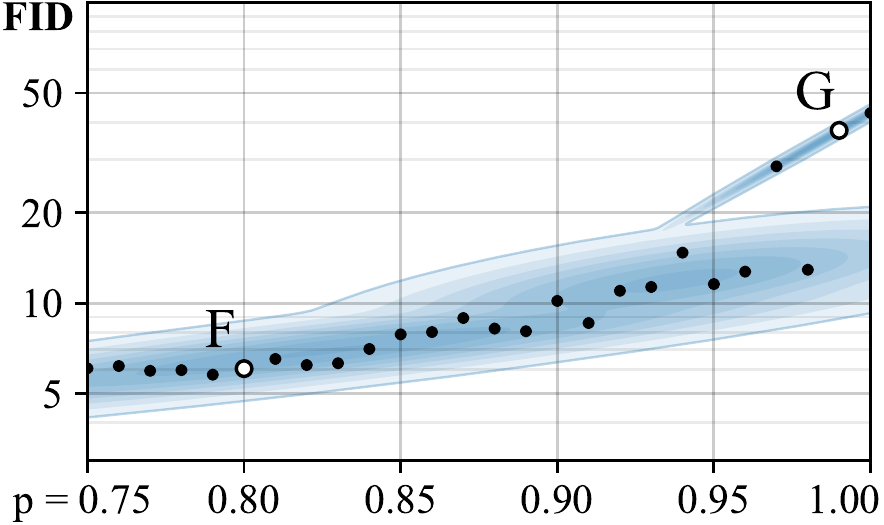}\\
\makebox[\linewidth][c]{(c) Color transformations}}%
\caption{
Leaking behavior of three example augmentations, shown as FID w.r.t.~the probability of executing the augmentation.
Each dot represents a complete training run, and the blue Gaussian mixture is a visualization aid.
The top row shows generated example images from selected training runs, indicated by uppercase letters in the plots.
}
\label{#1}
\end{figure}
}

\newcommand{\figFixedSweepsSimple}[1]{
\begin{figure}[t]
\footnotesize%
\renewcommand{\h}{0.228\linewidth}%
\renewcommand{\hh}{0.30\linewidth}%
\includegraphics[width=\h]{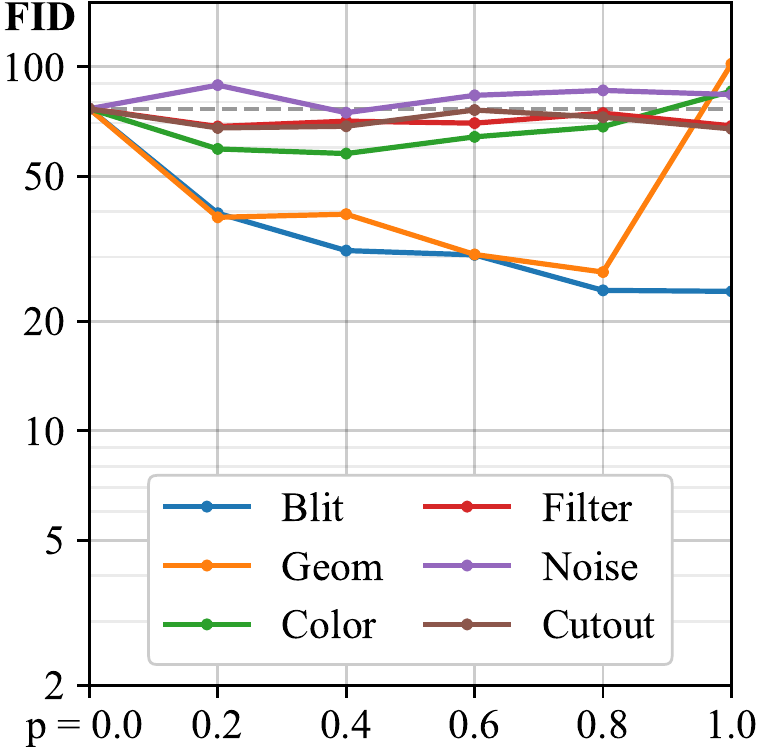}\hfill%
\includegraphics[width=\h]{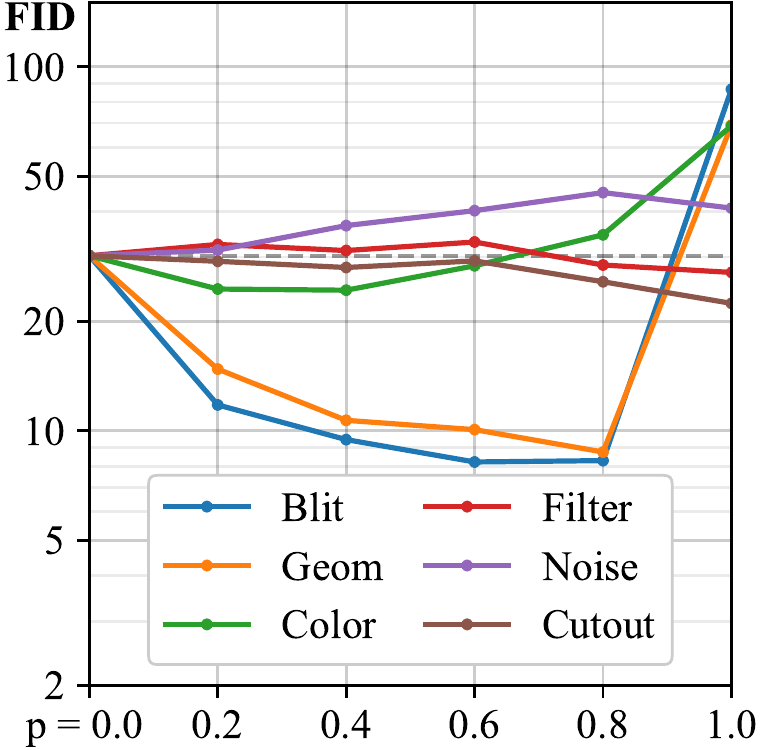}\hfill%
\includegraphics[width=\h]{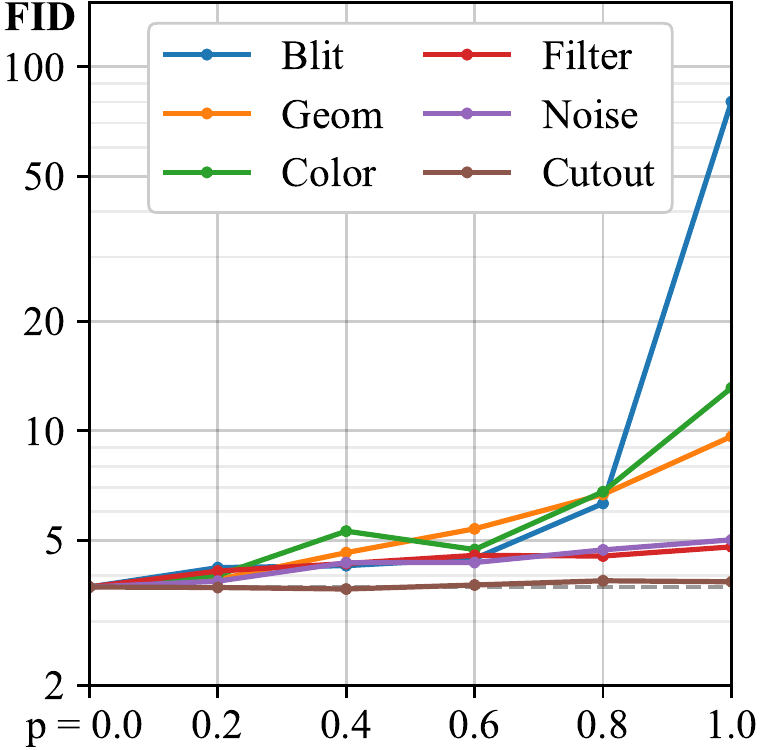}\hfill%
\includegraphics[width=\hh]{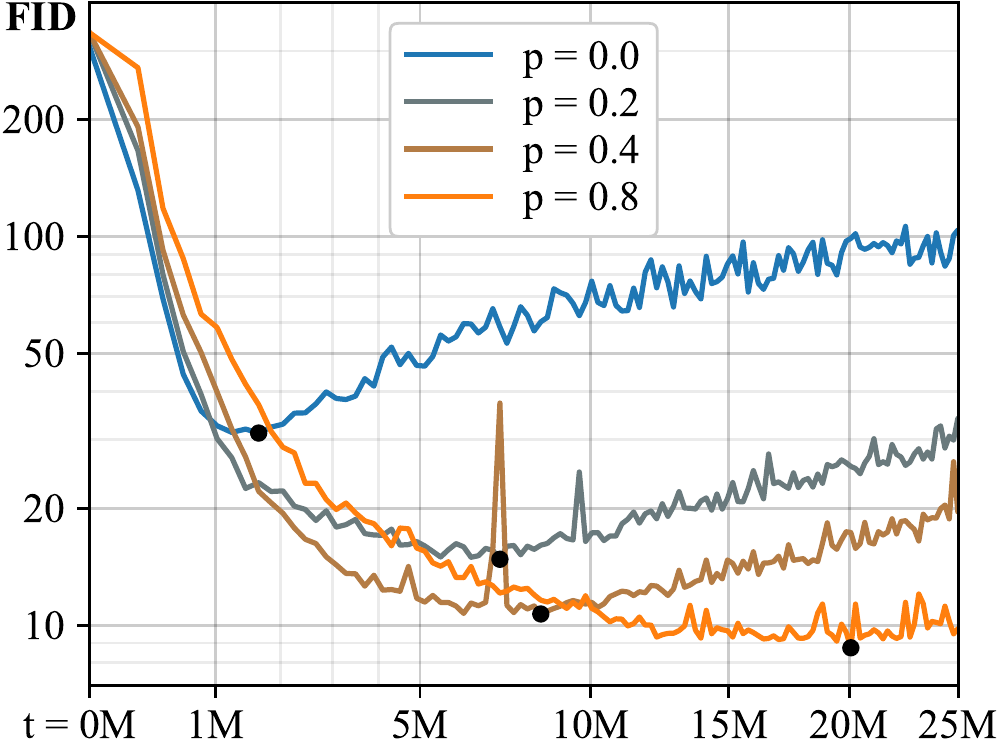}\\
\makebox[\h][c]{(a) FFHQ-2k}\hfill%
\makebox[\h][c]{(b) FFHQ-10k}\hfill%
\makebox[\h][c]{(c) FFHQ-140k}\hfill%
\makebox[\hh][c]{(d) Convergence, 10k, Geom}%
\caption{
(a-c)
Impact of $p$ for different augmentation categories and dataset sizes.
The dashed gray line indicates baseline FID without augmentations.
(d)
Convergence curves for selected values of $p$ using geometric augmentations with 10k training images.
}
\label{#1}
\end{figure}
}

\newcommand{\figAdaptiveSweeps}[1]{
\begin{figure}[t]
\footnotesize%
\renewcommand{\h}{0.21\linewidth}%
\renewcommand{\hh}{0.28\linewidth}%
\includegraphics[width=\h]{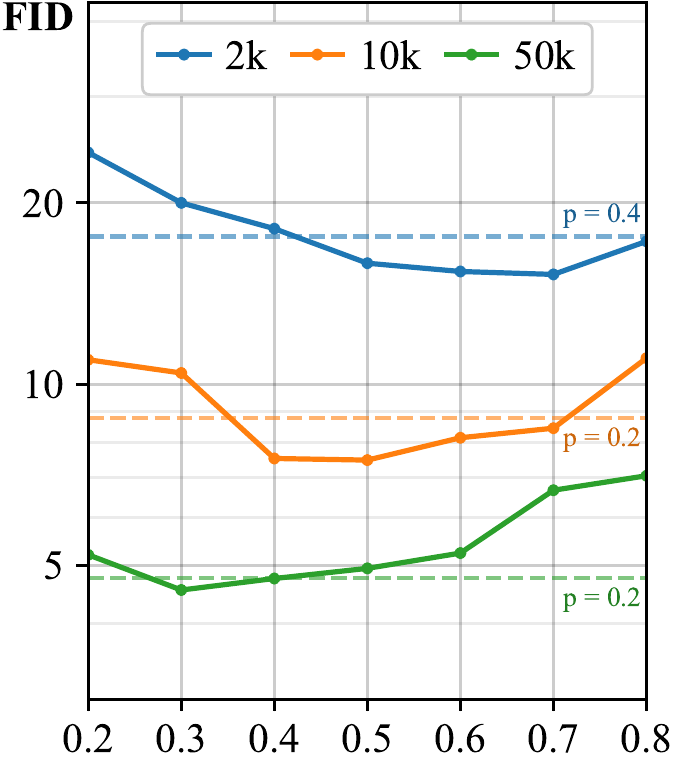}\hfill%
\includegraphics[width=\h]{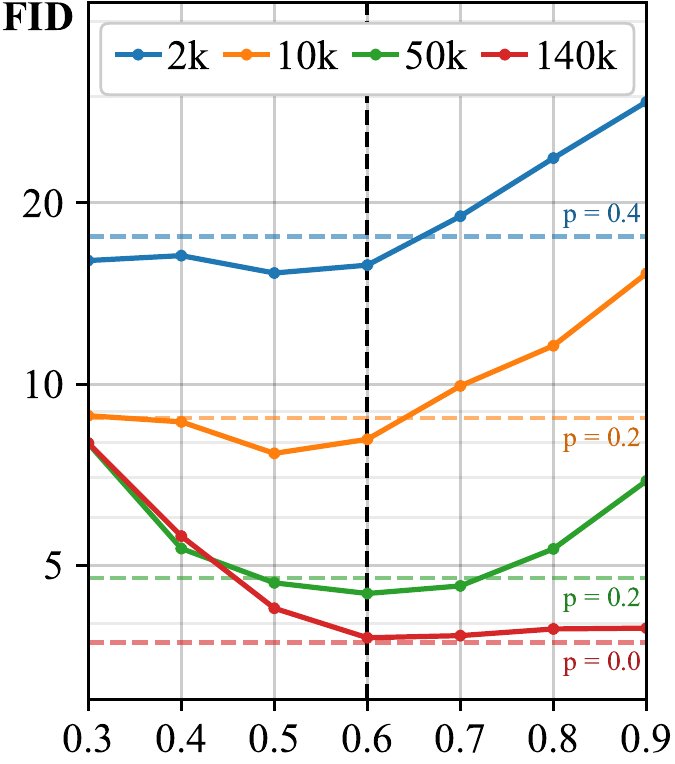}\hfill%
\includegraphics[width=\hh]{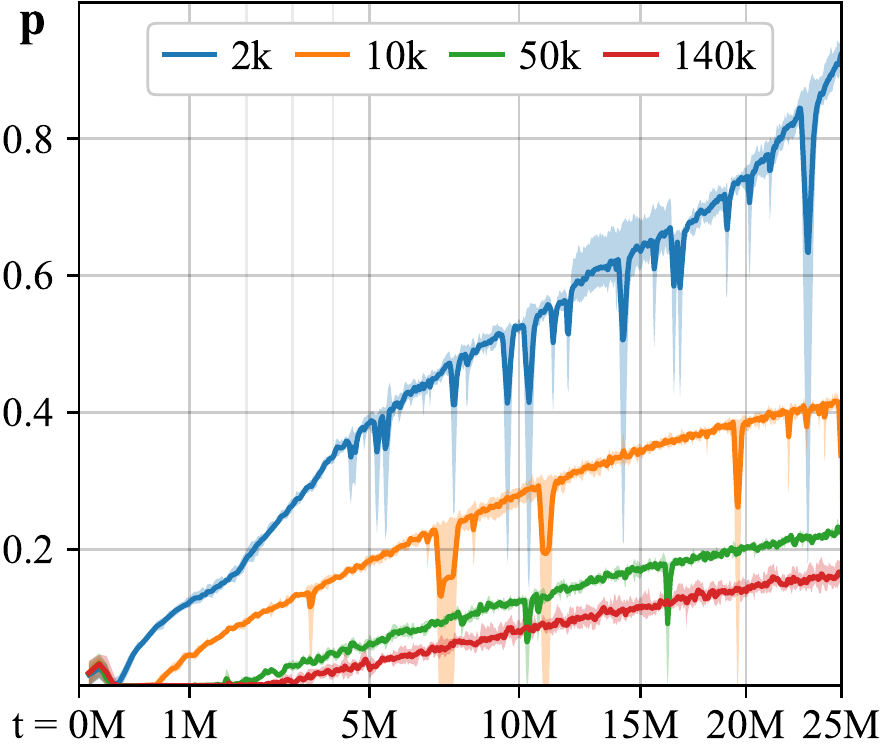}\hfill%
\includegraphics[width=\hh]{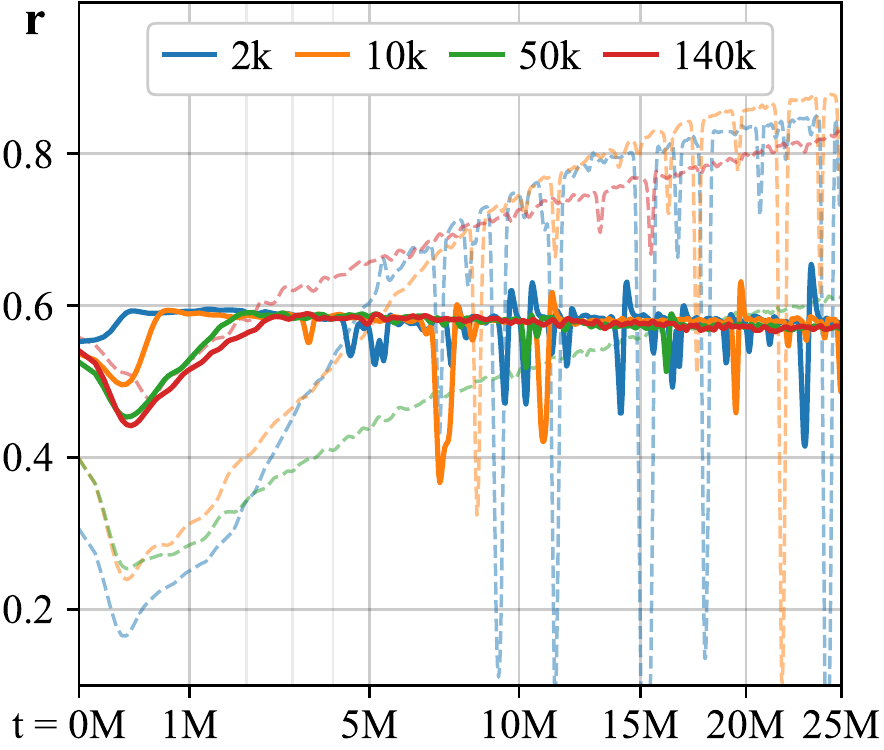}\\
\makebox[\h][c]{(a) $r_v$ target sweep}\hfill%
\makebox[\h][c]{(b) $r_t$ target sweep}\hfill%
\makebox[\hh][c]{(c) Evolution of $p$ over training}\hfill%
\makebox[\hh][c]{(d) Evolution of $r_t$}%
\caption{
Behavior of our adaptive augmentation strength heuristics in FFHQ. %
(a,b) FID for different training set sizes as a function of the target value for $r_v$ and $r_t$.
The dashed horizontal lines indicate the best fixed augmentation probability $p$ found using grid search, and the dashed vertical line marks the target value we will use in subsequent tests.
(c) Evolution of $p$ over the course of training using heuristic $r_t$. %
(d) Evolution of $r_t$ values over training. Dashes correspond to the fixed $p$ values in (b).
}
\vspace*{-1mm}	%
\label{#1}
\end{figure}
}

\newcommand{\figNoOverfit}[1]{
\begin{figure}[tb]
\footnotesize%
\renewcommand{\h}{0.35\linewidth}%
\renewcommand{\hh}{0.3\linewidth}%
\renewcommand{\hhh}{0.34\linewidth}%
\renewcommand{\vvv}{\hh*\real{3.13}/\real{4}}%
\includegraphics[width=\h]{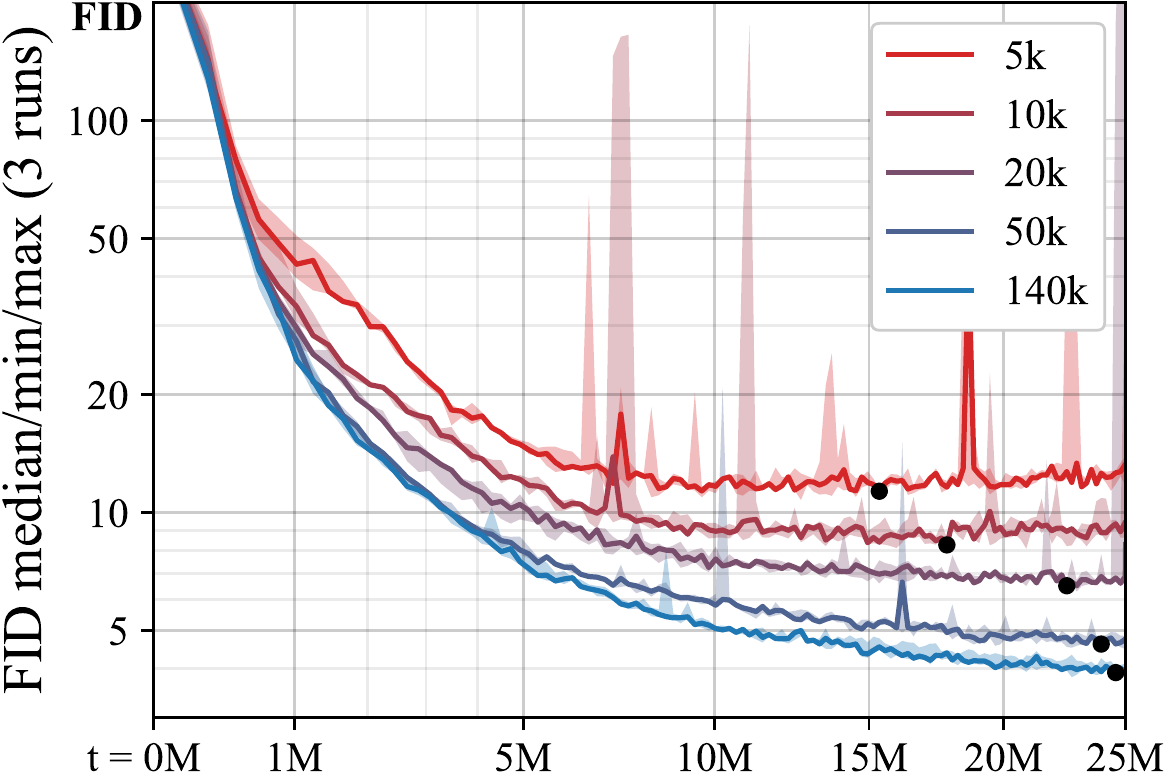}\hfill%
\includegraphics[width=\hh]{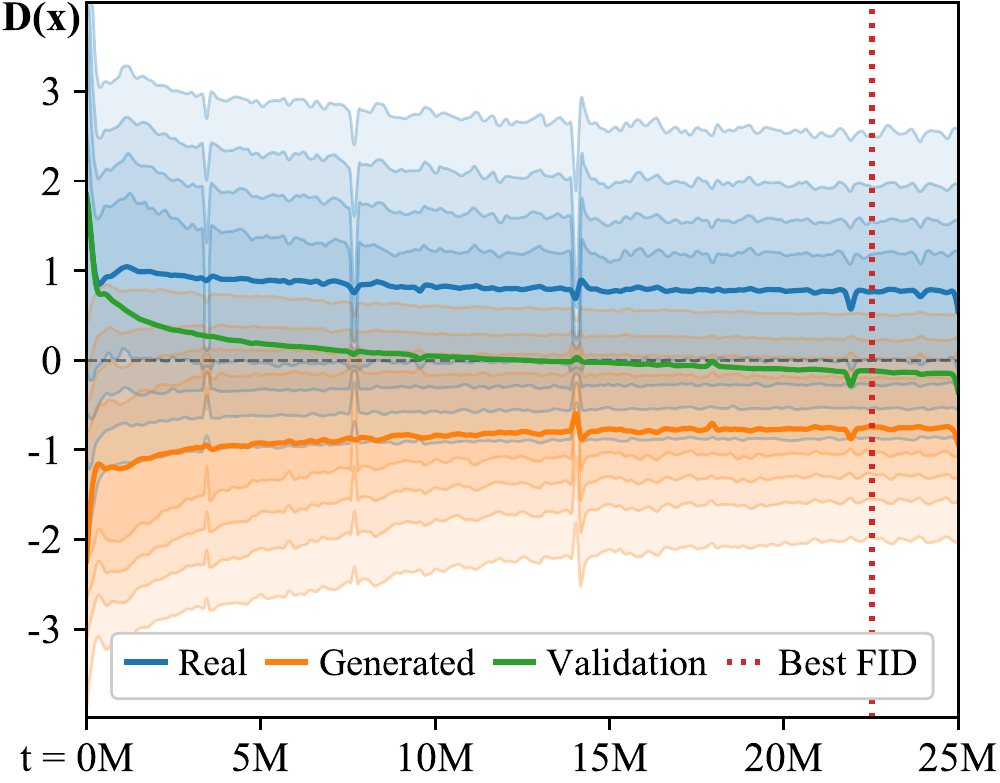}\hfill%
\parbox[b][\vvv]{\hhh}{%
\renewcommand{\hhhh}{0.94\linewidth}%
\renewcommand{\vvvv}{\hhhh/\real{3}}%
\rotatebox{90}{\makebox[\vvvv][c]{\scriptsize No augment}}\hfill%
\includegraphics[width=\hhhh]{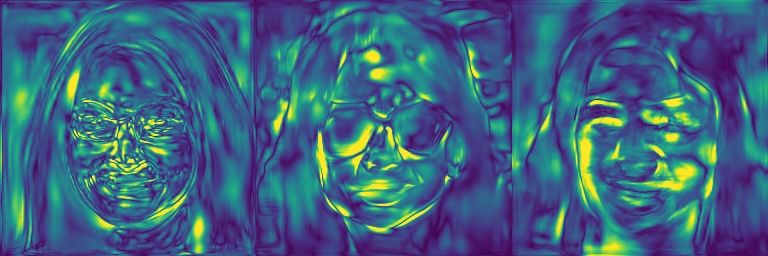}\vfill%
\rotatebox{90}{\makebox[\vvvv][c]{\scriptsize With ADA}}\hfill%
\includegraphics[width=\hhhh]{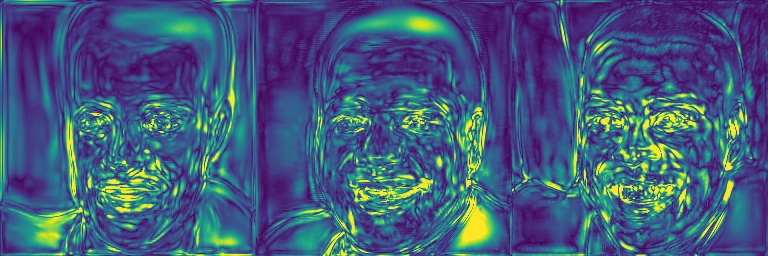}\vspace{-1.3mm}\\
\rotatebox{90}{\makebox[0mm][c]{\phantom{\scriptsize w/}}}\hfill%
\makebox[\vvvv][c]{\tiny 1M}%
\makebox[\vvvv][c]{\tiny 5M}%
\makebox[\vvvv][c]{\tiny 25M}\vspace{0.4mm}%
}
\makebox[\h][c]{(a) With adaptive augmentation}\hfill%
\makebox[\hh][c]{(b) Discriminator outputs, 20k}\hfill%
\makebox[\hhh][c]{(c) Discriminator gradients, 10k}\hfill%
\caption{
(a) Training curves for FFHQ with different training set sizes using adaptive augmentation.
(b) %
The supports of real and generated images continue to overlap. 
(c) Example magnitudes of the gradients the generator receives from the discriminator as the training progresses.
}
\label{#1}
\end{figure}
}

\newcommand{\tabMainSweep}{%
\newcolumntype{x}{>{\centering\arraybackslash\hspace{0pt}}p{11.5mm}}%
\renewcommand{\cs}{\hspace{0.8mm}}%
\tabulinesep=0.494mm%
\tabulinestyle{0.17mm}%
\begin{tabu}{|@{\cs}c@{\cs}|@{\hspace{2mm}}r@{\hspace{1.5mm}}|@{\cs}x@{}x@{}x@{\cs}|}
\tabucline{-}
\multicolumn{2}{|l|@{\cs}}{} & & & \\[-3.0mm]
\multicolumn{2}{|l|@{\cs}}{\bf Dataset} & {\bf Baseline} & {\bf ADA} & {\bf + bCR} \\[-3.4mm]
\multicolumn{2}{|l|@{\cs}}{} & & & \\
\tabucline{-}
\multirow{6}{*}{\rotatebox{90}{\bf\textsc{FFHQ}}}
& 1k    & 100.16         & {\bf21.29}  & 22.61        \\
& 5k    & \s49.68        & 10.96       & {\bf10.58}   \\
& 10k   & \s30.74        & \s8.13      & {\bf\s7.53}  \\
& 30k   & \s12.31        & \s5.46      & {\bf\s4.57}  \\
& 70k   & \s\s5.28       & \s4.30      & {\bf\s3.91}  \\
& 140k  & \s\s3.71       & \s3.81      & {\bf\s3.62}  \\
\tabucline{-}
\multirow{6}{*}{\rotatebox{90}{\bf\textsc{LSUN Cat}}}
& 1k    & 186.91         & 43.25       & {\bf38.82}   \\
& 5k    & \s96.44        & 16.95       & {\bf16.80}   \\
& 10k   & \s50.66        & 13.13       & {\bf12.90}   \\
& 30k   & \s15.90        & 10.50       & {\bf\s9.68}  \\
& 100k  & {\bf\s\s8.56}  & \s9.26      & \s8.73       \\
& 200k  & {\bf\s\s7.98}  & \s9.22      & \s9.03       \\
\tabucline{-}
\end{tabu}%
}

\newcommand{\meanlabel}[2]{\makebox(0,0)[r]{\raisebox{33.5mm}{\makebox[\linewidth]{\tiny\color{white}\hspace{1mm}\contourlength{0.15mm}\contour{black}{\bf #1}\hfill\contour{black}{\bf #2}\hspace{1mm}}}}}

\newcommand{\figMainSweep}[1]{
\begin{figure}[t]
\footnotesize%
\renewcommand{\h}{0.280\linewidth}%
\renewcommand{\hh}{0.295\linewidth}%
\renewcommand{\hhh}{0.1321\linewidth}%
\renewcommand{\hhhh}{1.8mm}%
\renewcommand{\vv}{0.288\linewidth}%
\includegraphics[width=\h]{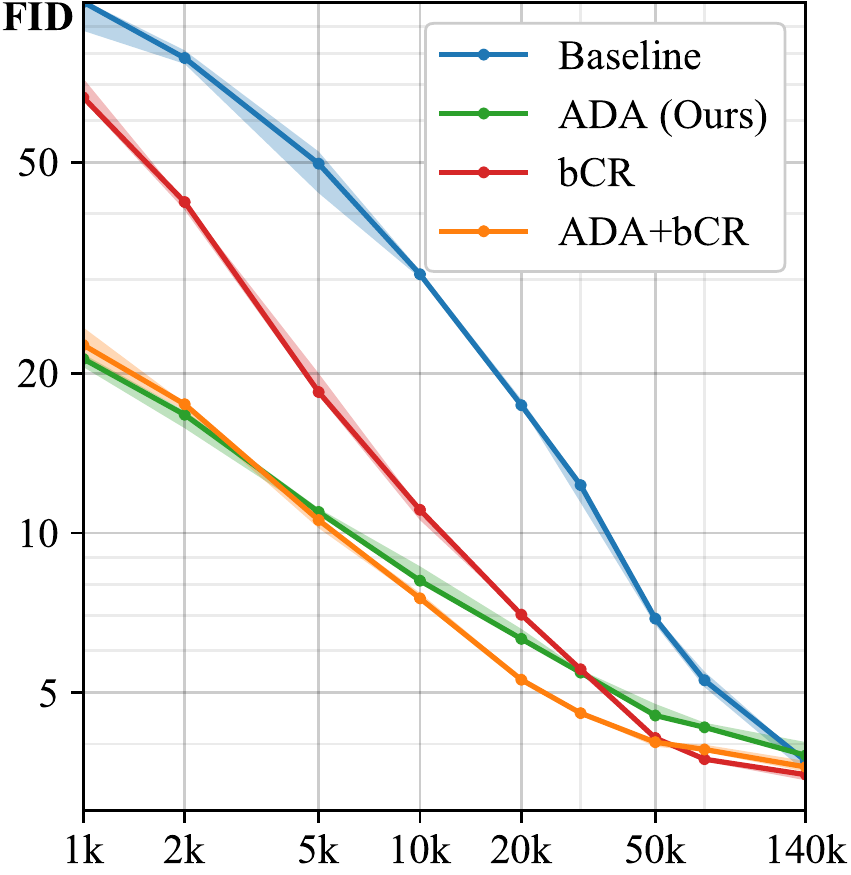}\hfill%
\includegraphics[width=\h]{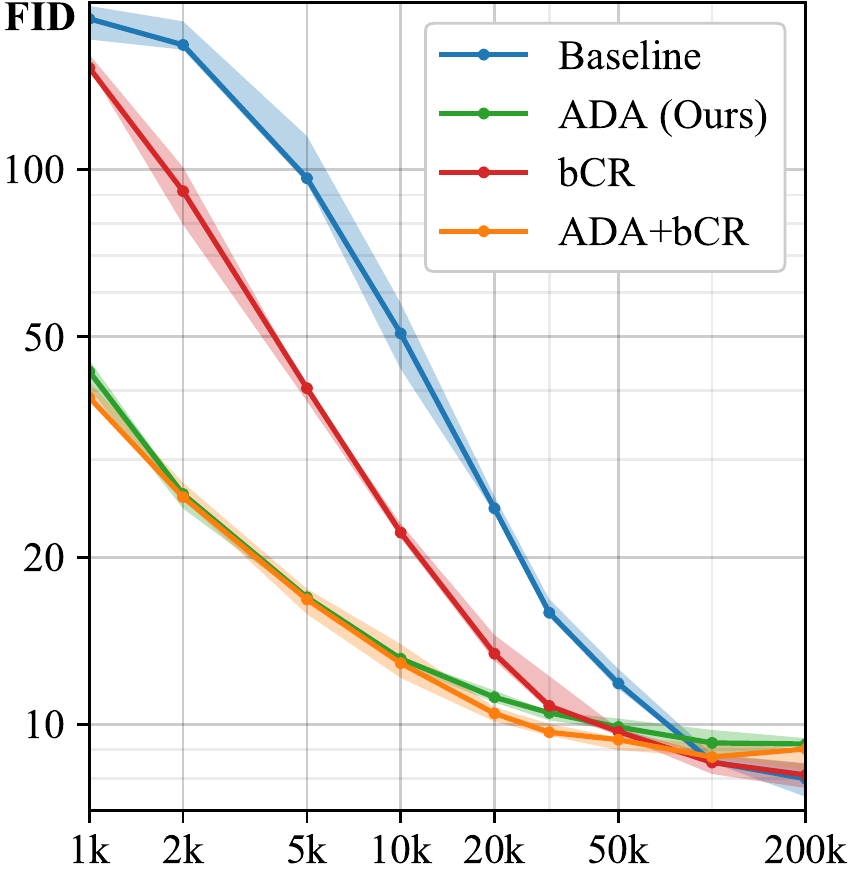}\hfill%
\subtabletop{\hh}{\vv}{0.75}{\tabMainSweep}\hfill\hspace{\hhhh}%
\parbox[b][\vv]{\hhh}{\vfill%
\renewcommand{\hhhh}{0.492\linewidth}%
\includegraphics[width=\hhhh,trim={0 0 130.5 0},clip]{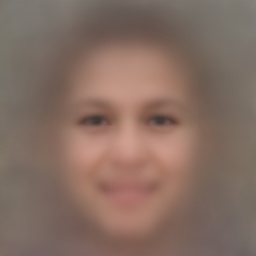}\hfill%
\includegraphics[width=\hhhh,trim={130.5 0 0 0},clip]{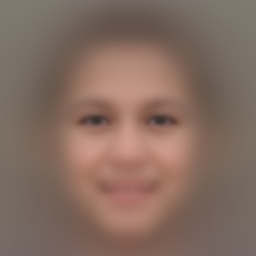}%
\meanlabel{ADA}{Real}\vspace{-0.05mm}\\
\includegraphics[width=\hhhh,trim={0 0 130.5 0},clip]{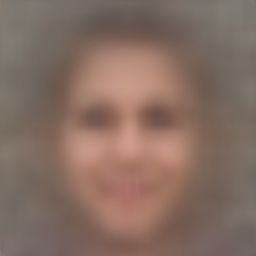}\hfill%
\includegraphics[width=\hhhh,trim={130.5 0 0 0},clip]{generated_images/average_images_ffhq/real.\ext}%
\meanlabel{bCR}{Real}\vspace{2.9mm}%
}\\
\makebox[\h][c]{(a) \textsc{FFHQ ($256\times256$)}}\hfill%
\makebox[\h][c]{(b) \textsc{LSUN Cat ($256\times256$)}}\hfill%
\makebox[\hh][c]{(c) Median FID}\hfill\hspace{\hhhh}%
\makebox[\hhh][r]{(d) Mean image}%
\caption{
(a-c) FID as a function of training set size, reported as median/min/max over 3 training runs.
(d) Average of 10k random images generated using the networks trained with 5k subset of FFHQ. ADA matches the average of real data, whereas the $xy$-translation augmentation in bCR~\cite{zhao2020improved} has leaked to the generated images, significantly blurring the average image. %
}
\vspace*{0mm}	%
\label{#1}
\end{figure}
}

\newcommand{\std}[1]{\makebox[6.3mm][r]{\scriptsize\raisebox{0.25mm}{\scalebox{0.7}{$\pm$}}\hspace{0.2mm}#1}}
\newcommand{\nostd}[1]{\makebox[6.3mm][r]{}}

\newcommand{\tabComparisonMethodsStd}{%
\newcolumntype{x}{>{\centering\arraybackslash\hspace{0pt}}p{13.5mm}}%
\newcolumntype{y}{>{\centering\arraybackslash\hspace{0pt}}p{12.0mm}}%
\renewcommand{\cs}{\hspace{4mm}}%
\tabulinesep=0.705mm%
\tabulinestyle{0.17mm}%
\begin{tabu}{|l|@{\cs}x@{\cs}x@{\cs}y@{\cs}|}
\tabucline{-}
& & & \\[-2.4mm]
{\bf FFHQ \raisebox{0.15mm}{\scalebox{0.9}{($256\times256$)}}} & {\bf 2k} & {\bf 10k} & {\bf 140k} \\[-2.8mm]
& & & \\
\tabucline{-}
Baseline                                      & 78.80\std{2.31}       & 30.73\std{0.48}        & 3.66\std{0.10}       \\
\tabucline{-}
PA-GAN         \hfill\cite{Khoreva2019}       & 56.49\std{7.28}       & 27.71\std{2.77}        & 3.78\std{0.06}       \\
WGAN-GP        \hfill\cite{Gulrajani2017}     & 79.19\std{6.30}       & 35.68\std{1.27}        & 6.54\std{0.37}       \\
zCR            \hfill\cite{zhao2020improved}  & 71.61\std{9.64}       & 23.02\std{2.09}        & {\bf3.45}\std{0.19}  \\
Auxiliary rotation  \hfill\cite{Chen2018aux}       & 66.64\std{3.64}       & 25.37\std{1.45}        & 4.16\std{0.05}       \\
Spectral norm  \hfill\cite{Miyato2018B}       & 88.71\std{3.18}       & 38.58\std{3.14}        & 4.60\std{0.19}       \\
\tabucline{-}
Shallow mapping                               & 71.35\std{7.20}       & 27.71\std{1.96}        & 3.59\std{0.22}       \\
Adaptive dropout                              & 67.23\std{4.76}       & 23.33\std{0.98}        & 4.16\std{0.05}       \\
ADA (Ours)                                    & {\bf16.49}\std{0.65}  & {\bf\s8.29}\std{0.31}  & 3.88\std{0.13}       \\
\tabucline{-}
\end{tabu}%
}

\newcommand{\figComparisonMethods}[1]{
\begin{figure}[t]
\footnotesize%
\renewcommand{\h}{0.45\linewidth}%
\renewcommand{\hh}{0.247\linewidth}%
\renewcommand{\hhh}{0.017\linewidth}%
\renewcommand{\vv}{32.9mm}%
\renewcommand{\vvv}{0.75}%
\subtabletop{\h}{\vv}{\vvv}{\tabComparisonMethodsStd}\hfill%
\parbox[b]{\hh}{\hspace*{4mm}\makebox[30mm][c]{\FINAL{Baseline}}\vspace{0.7mm}\\%
\includegraphics[width=\linewidth]{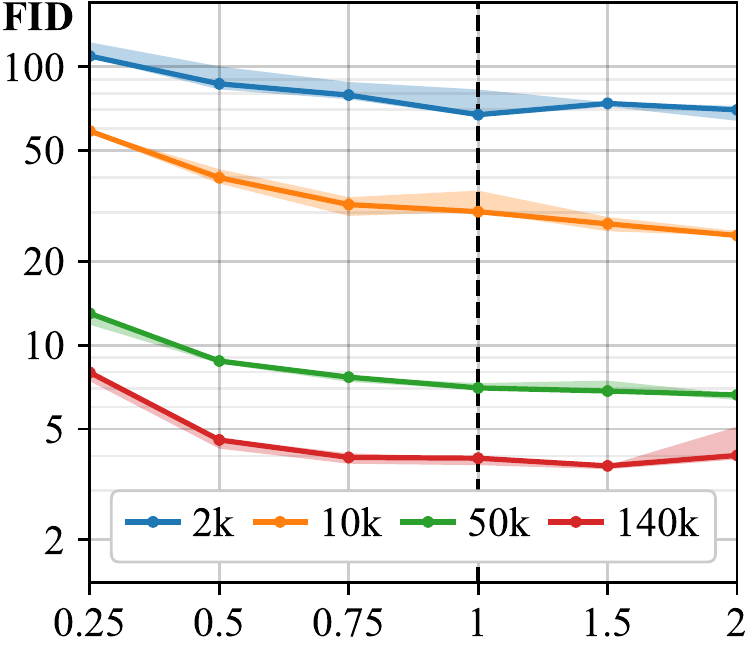}}%
\hspace{\hhh}%
\parbox[b]{\hh}{\hspace*{4mm}\makebox[30mm][c]{\FINAL{ADA}}\vspace{0.7mm}\\%
\includegraphics[width=\linewidth]{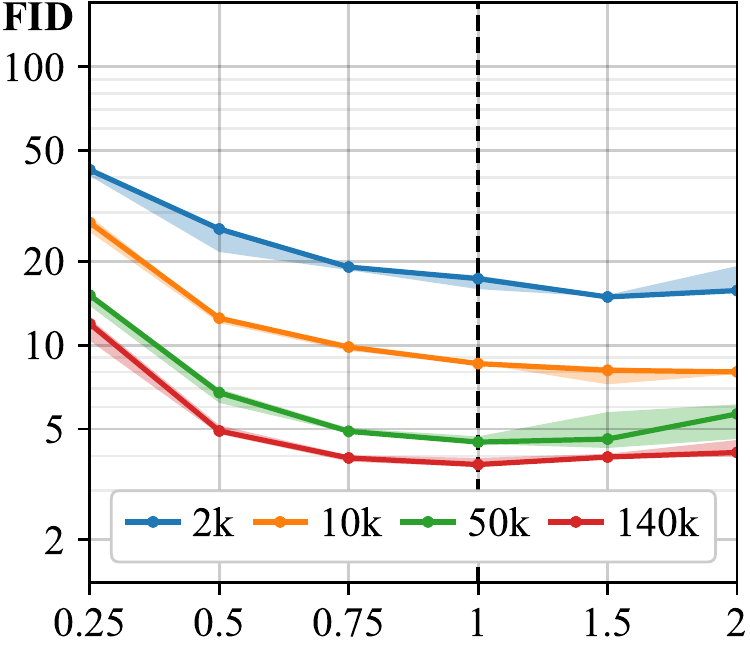}}\\
\makebox[\h][c]{(a) Comparison methods}\hfill%
\makebox[0.50\linewidth][c]{\FINAL{(b) Discriminator capacity sweeps}}%
\caption{
(a) We report the mean and standard deviation for each comparison method, calculated over \FINAL{3} training runs.
\FINAL{(b) FID as a function of discriminator capacity, reported as median/min/max over 3 training runs.
We scale the number of feature maps uniformly across all layers by a given factor ($x$-axis).
The baseline configuration (no scaling) is indicated by the dashed vertical line.}
}
\label{#1}
\end{figure}
}

\newcommand{\figTransferLearning}[1]{
\begin{figure}[t]
\footnotesize%
\renewcommand{\h}{0.3\linewidth}%
\renewcommand{\hh}{0.2\linewidth}%
\includegraphics[width=\h]{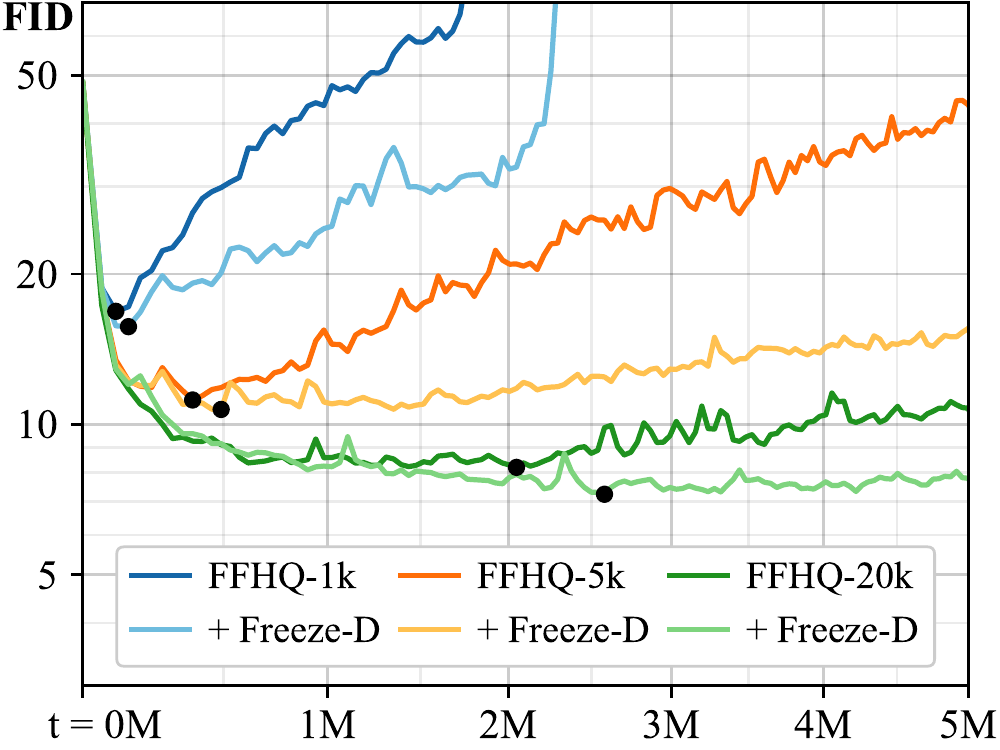}\hfill%
\includegraphics[width=\h]{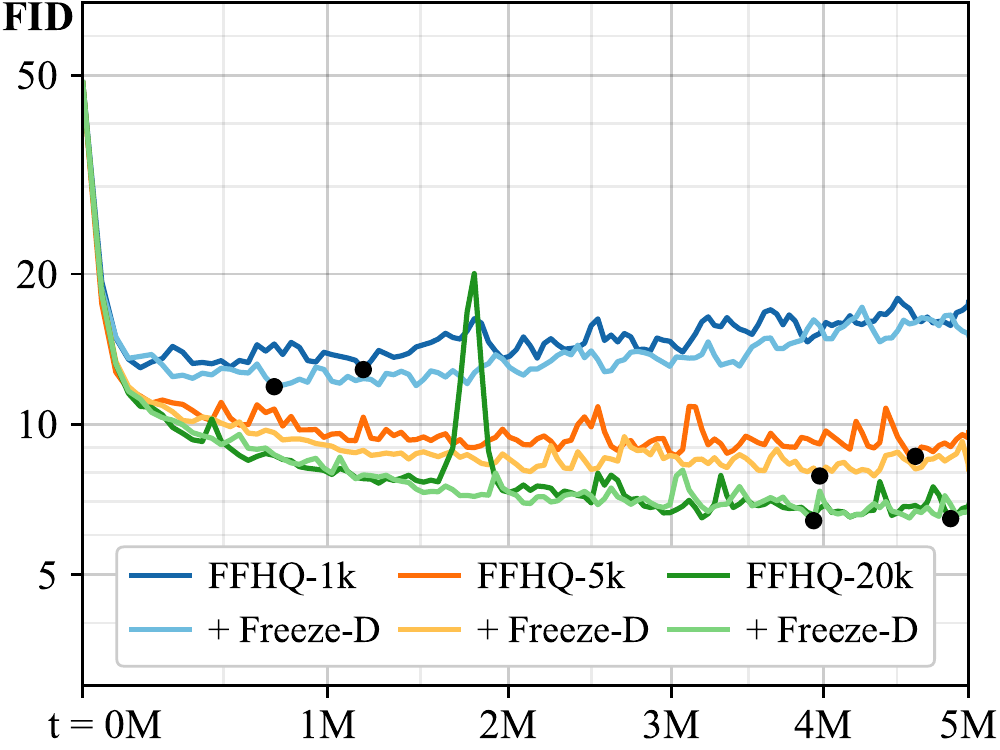}\hfill%
\includegraphics[width=\hh]{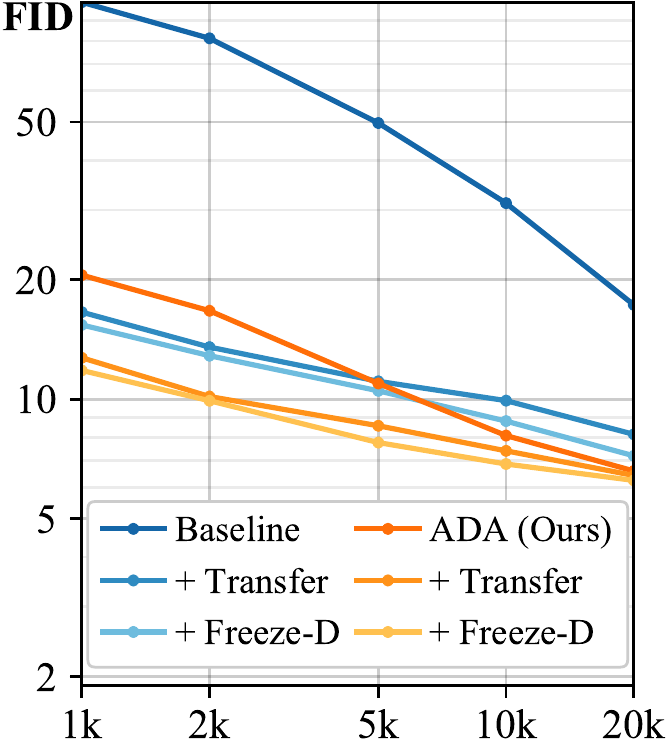}\hfill%
\includegraphics[width=\hh]{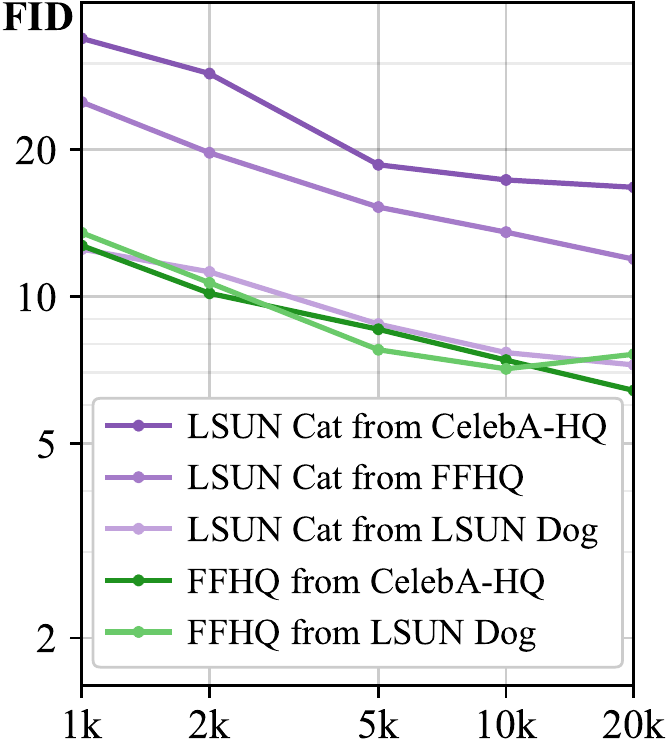}\\
\makebox[\h][c]{(a) Without ADA}\hfill%
\makebox[\h][c]{(b) With ADA}\hfill%
\makebox[\hh][c]{(c) Dataset sizes}\hfill%
\makebox[\hh][c]{(d) Datasets}%
\caption{
Transfer learning \textsc{FFHQ} starting from a pre-trained \textsc{CelebA-HQ} model, both $256\times256$.
(a) Training convergence for our baseline method and Freeze-D~\cite{Mo2020}.
(b) The same configura\-tions with ADA.
(c) FIDs as a function of dataset size.
(d) Effect of source and target datasets.
}
\label{#1}
\end{figure}
}

\newcommand{\figSmallDatasetImages}[1]{
\begin{figure}[t]
\footnotesize%
\renewcommand{\h}{0.187\linewidth}%
\renewcommand{\hh}{\h*\real{2}/\real{3}}%
\renewcommand{\hhh}{\h*\real{2}}%
\renewcommand{\hhhh}{\h*\real{2}*\real{6}/\real{20}}%
\renewcommand{\vv}{\vspace{-0.35mm}}%
\makebox[\hhh][c]{\textsc{MetFaces} (new dataset)}\hfill%
\makebox[\hh][c]{\textsc{BreCaHAD}}\hfill%
\makebox[\hh*\real{3}][c]{\textsc{AFHQ \FINAL{Cat}, Dog, \FINAL{Wild} (\scriptsize $512^2$)}}\hfill%
\makebox[\hhhh][c]{\textsc{CIFAR-10}}\\%
\makebox[\hhh][c]{\scriptsize 1336 img, $1024^2$, transfer learning from FFHQ}\hfill%
\makebox[\hh][c]{\scriptsize 1944 img, $512^2$}\hfill%
\makebox[\hh][c]{\scriptsize \FINAL{5153 img}}%
\makebox[\hh][c]{\scriptsize 4739 img}%
\makebox[\hh][c]{\scriptsize \FINAL{4738 img}}\hfill%
\makebox[\hhhh][c]{\scriptsize 50k, 10 cls, $32^2$}\\%
\parbox[b]{\h}{%
\includegraphics[width=\linewidth]{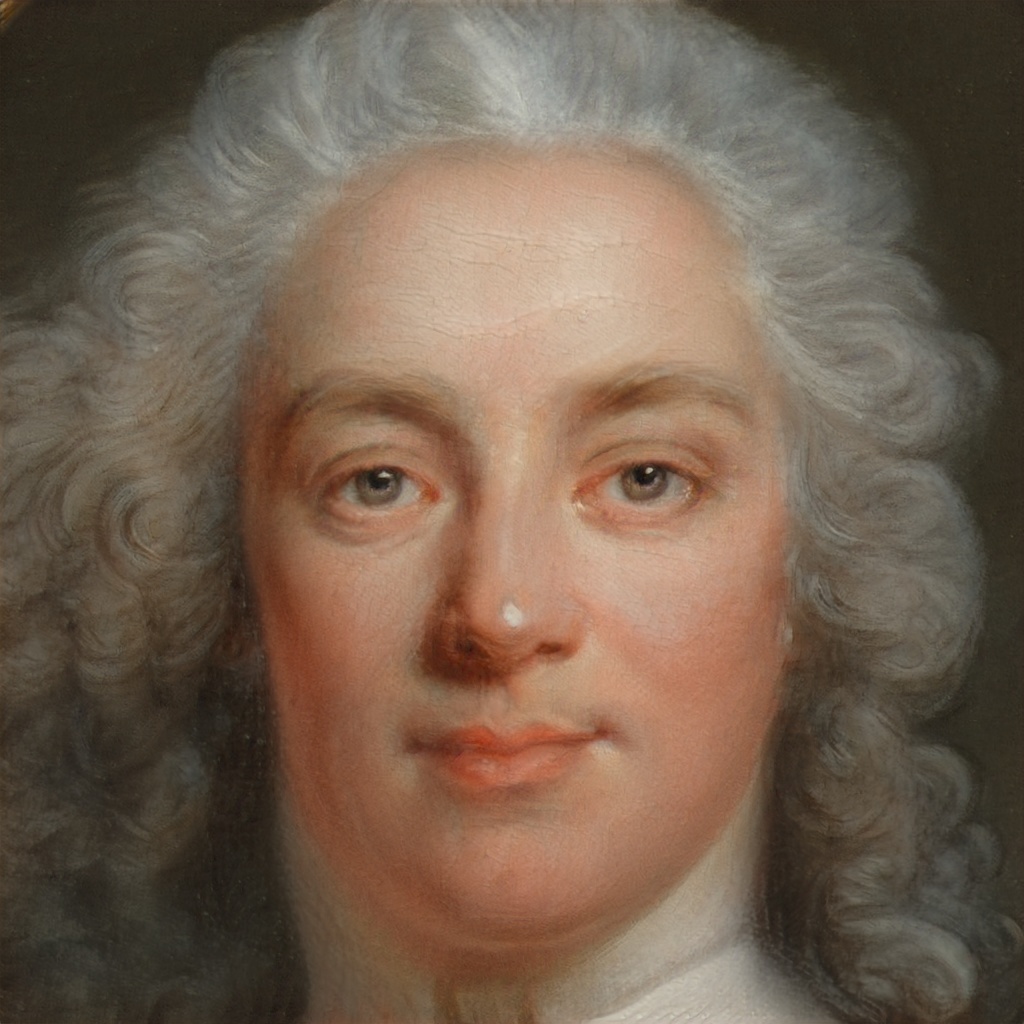}\vv\\
\includegraphics[width=\linewidth]{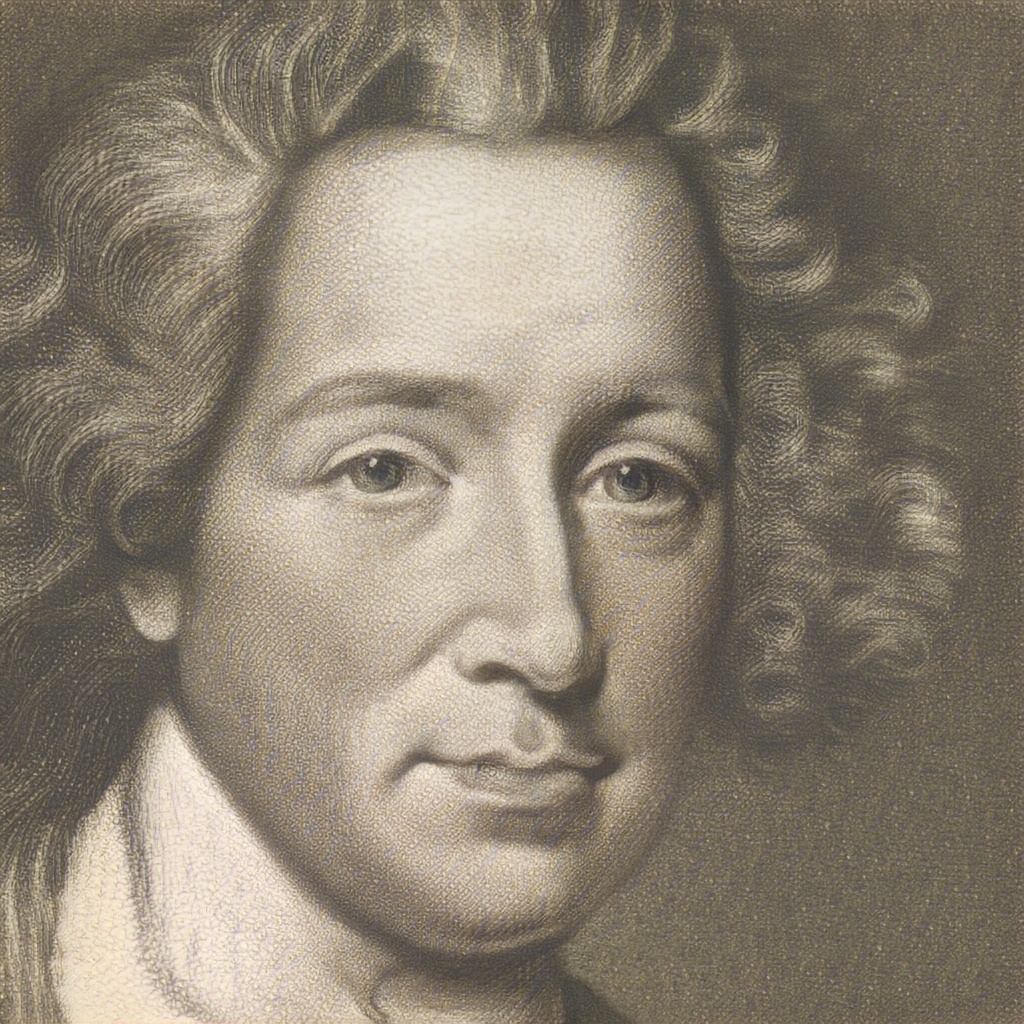}%
}%
\parbox[b]{\h}{%
\includegraphics[width=\linewidth]{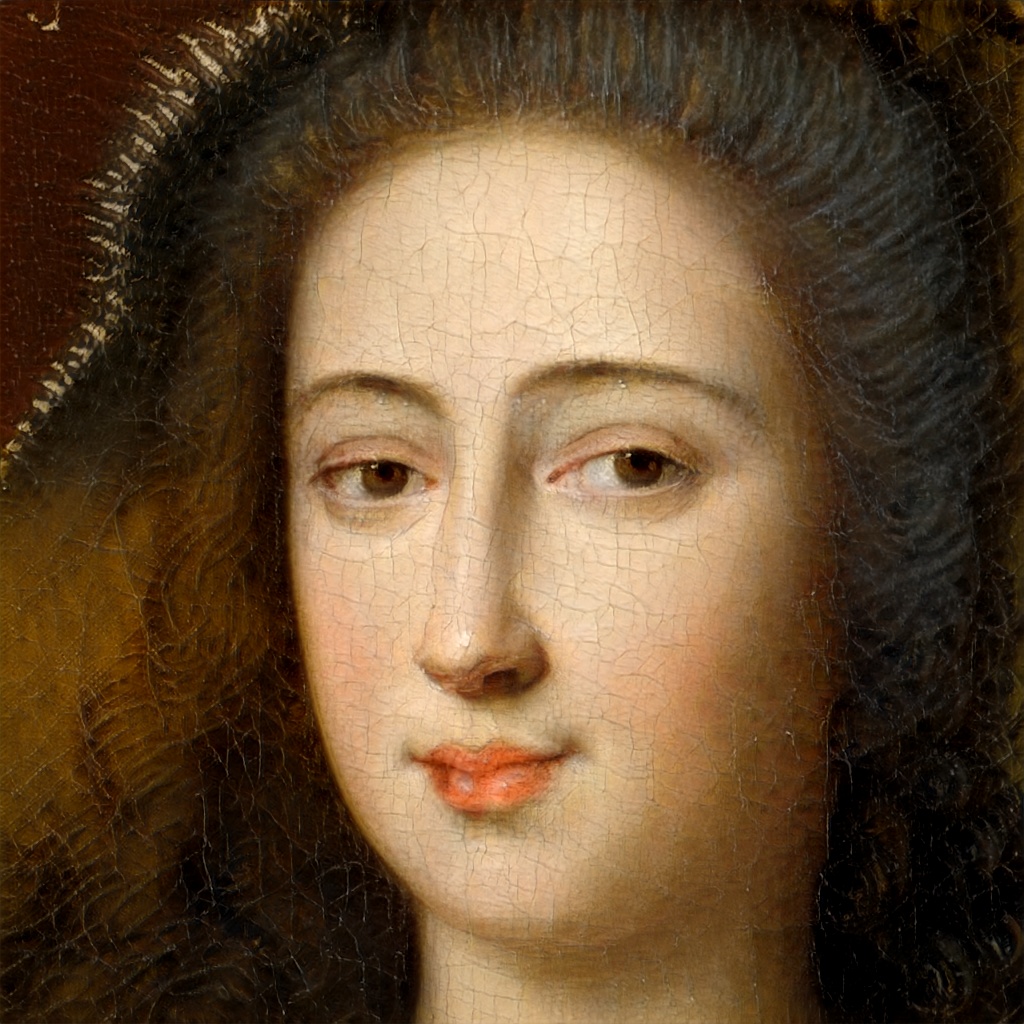}\vv\\
\includegraphics[width=\linewidth]{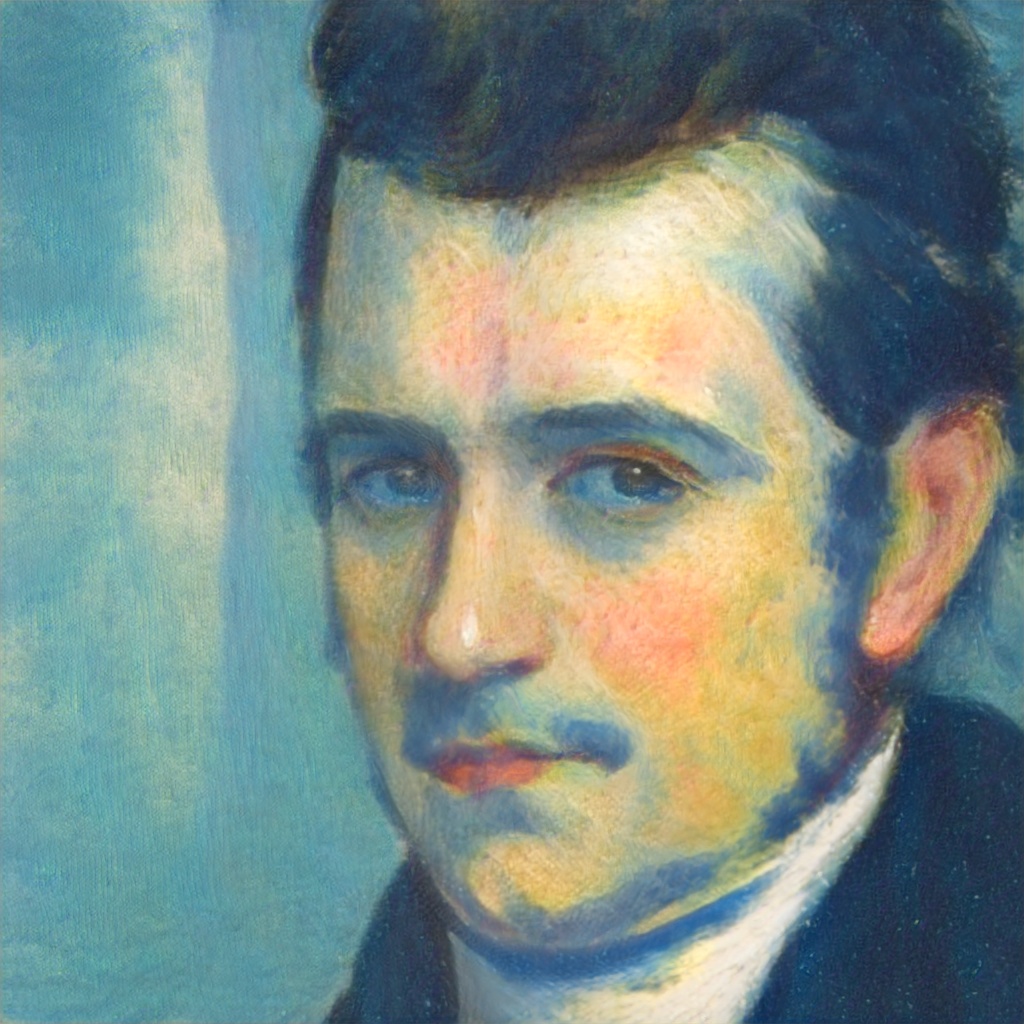}%
}\hfill%
\parbox[b]{\hh}{%
\includegraphics[width=\linewidth]{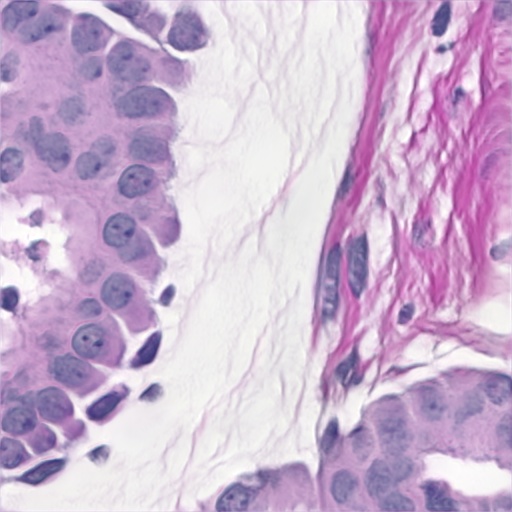}\vv\\
\includegraphics[width=\linewidth]{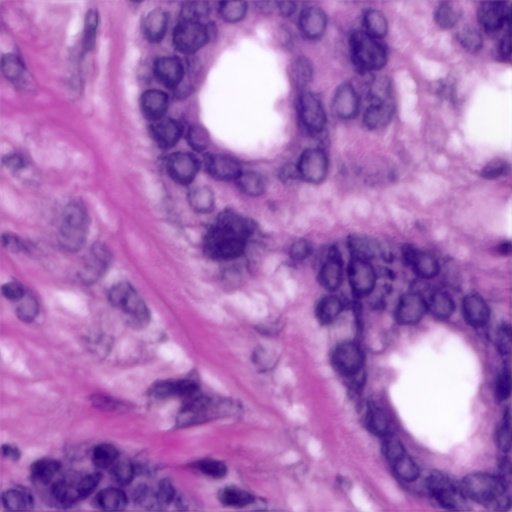}\vv\\
\includegraphics[width=\linewidth]{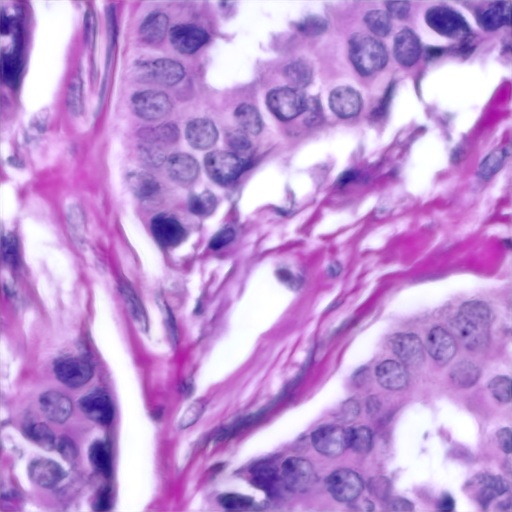}%
}\hfill%
\parbox[b]{\hh}{%
\includegraphics[width=\linewidth]{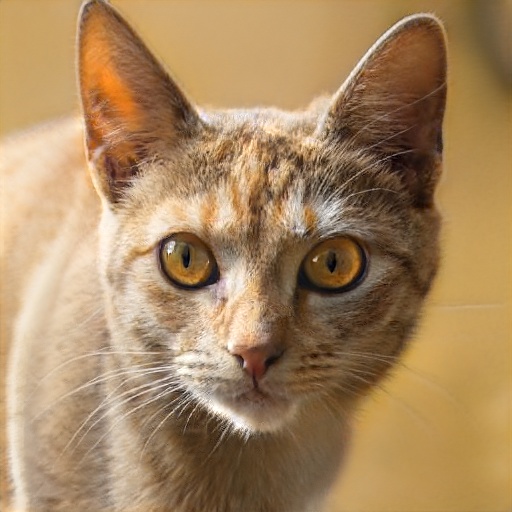}\vv\\
\includegraphics[width=\linewidth]{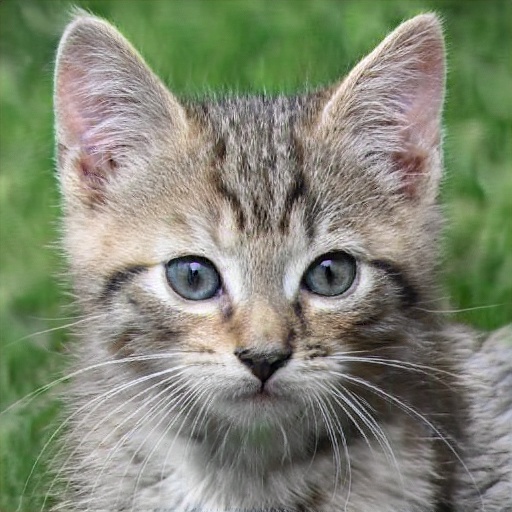}\vv\\
\includegraphics[width=\linewidth]{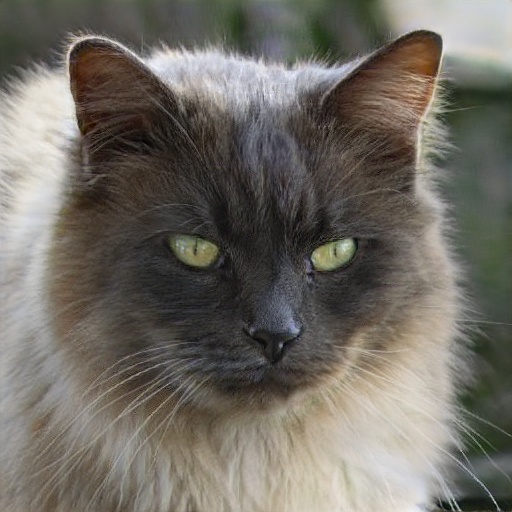}%
}%
\parbox[b]{\hh}{%
\includegraphics[width=\linewidth]{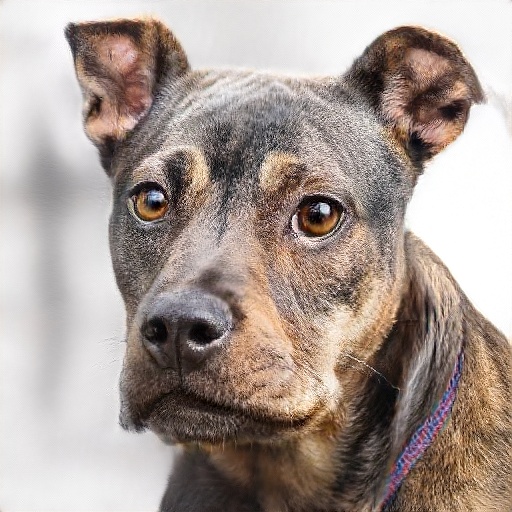}\vv\\
\includegraphics[width=\linewidth]{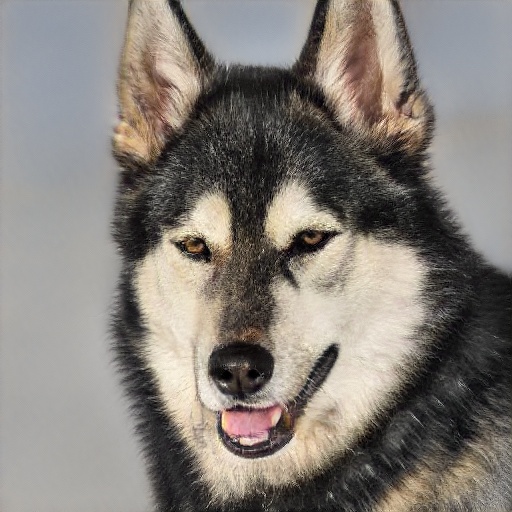}\vv\\
\includegraphics[width=\linewidth]{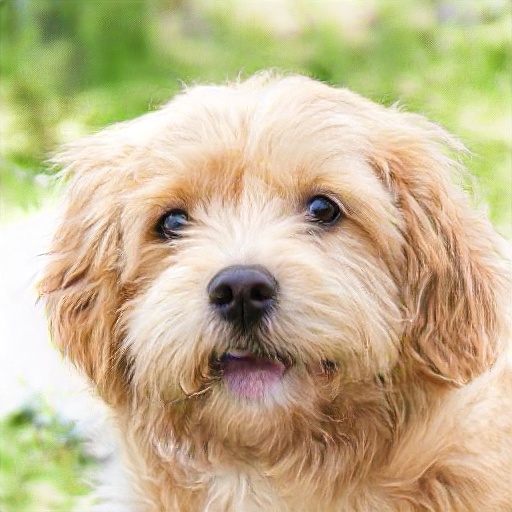}%
}%
\parbox[b]{\hh}{%
\includegraphics[width=\linewidth]{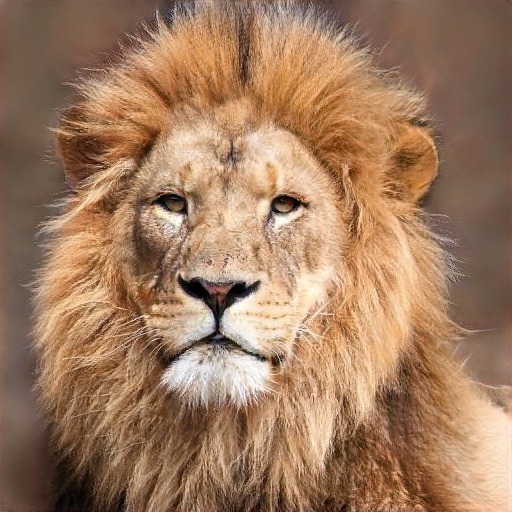}\vv\\
\includegraphics[width=\linewidth]{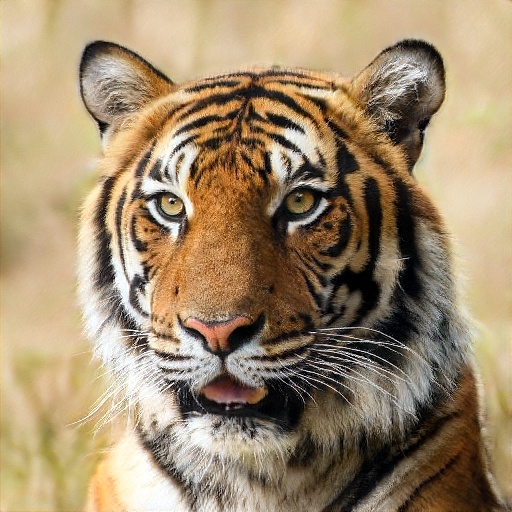}\vv\\
\includegraphics[width=\linewidth]{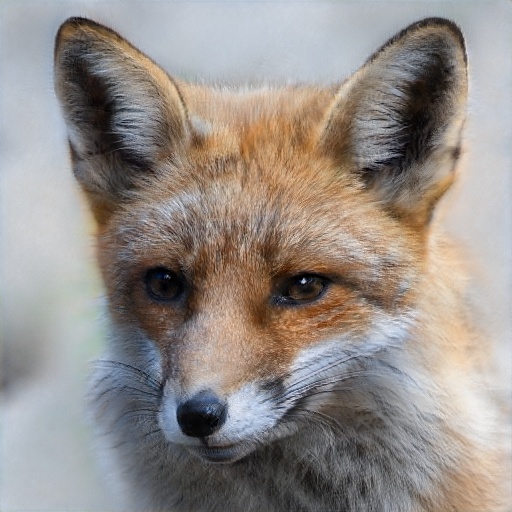}%
}\hfill%
\includegraphics[width=\hhhh]{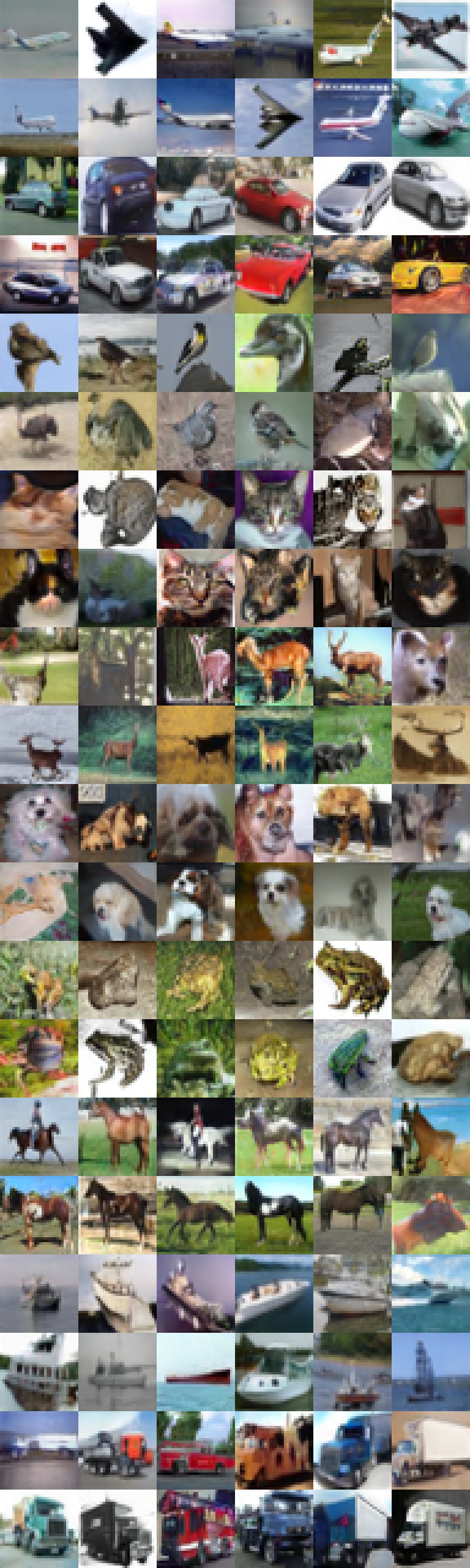}%
\caption{
Example generated images for several datasets with limited amount of training data, trained using ADA.
We use transfer learning with \textsc{MetFaces} and train other datasets from scratch.
See \refappResults{} for uncurated results and real images, and \refappImplementation{} for our training configurations.
}
\vspace{-1mm}
\label{#1}
\end{figure}
}

\newcommand{\tabSmallDatasets}{%
\newcolumntype{x}{>{\centering\arraybackslash\hspace{0pt}}p{9mm}}%
\newcolumntype{y}{>{\centering\arraybackslash\hspace{0pt}}p{13mm}}%
\newcolumntype{z}{>{\centering\arraybackslash\hspace{0pt}}p{15mm}}%
\renewcommand{\cs}{\hspace{1.4mm}}%
\tabulinesep=0.5mm%
\tabulinestyle{0.17mm}%
\FINAL{
\begin{tabu}{|l@{\hspace{1.6mm}}@{\cs}l@{\cs}|@{\cs}x@{\cs}x@{\cs}|@{\cs}y@{\cs}|@{\cs}z@{\cs}|}
\tabucline{-}
\multirow{3}{*}{\raisebox{0.9mm}{\bf Dataset}} & \multirow{3}{*}{\raisebox{0.9mm}{\bf Method}} & \multicolumn{2}{c|@{\cs}}{\bf Scratch} & {\bf Transfer} & {\bf + Freeze-D} \\
& & FID & KID & KID & KID \\[-1.3mm]
& & & \hspace{2.2mm}\raisebox{0mm}[0mm][0mm]{\tiny$\times10^3$} & \hspace{2.2mm}\raisebox{0mm}[0mm][0mm]{\tiny$\times10^3$} & \hspace{2.2mm}\raisebox{0mm}[0mm][0mm]{\tiny$\times10^3$} \\
\tabucline{-}
\multirow{2}{*}{\sc MetFaces}   & Baseline  & 57.26        & 35.66        & \s3.16        & 2.05      \\
                                & ADA       & \bf18.22     & \bf\s2.41    & \bf\s0.81     & \bf1.33   \\
\tabucline{-}
\multirow{2}{*}{\sc BreCaHAD}   & Baseline  & 97.72        & 89.76        & 18.07         & 6.94      \\
                                & ADA       & \bf15.71     & \bf\s2.88    & \bf\s3.36     & \bf1.91   \\
\tabucline{-}
\multirow{2}{*}{\sc AFHQ Cat}   & Baseline  & \s5.13       & \s1.54       & \s1.09        & 1.00      \\
                                & ADA       & \bf\s3.55    & \bf\s0.66    & \bf\s0.44     & \bf0.35   \\
\tabucline{-}
\multirow{2}{*}{\sc AFHQ Dog}   & Baseline  & 19.37        & \s9.62       & \s4.63        & 2.80      \\
                                & ADA       & \bf\s7.40    & \bf\s1.16    & \bf\s1.40     & \bf1.12   \\
\tabucline{-}
\multirow{2}{*}{\sc AFHQ Wild}  & Baseline  & \s3.48       & \s0.77       & \s0.31        & \bf0.12   \\
                                & ADA       & \bf\s3.05    & \bf\s0.45    & \bf\s0.15     & 0.14      \\
\tabucline{-}
\end{tabu}%
}%
}

\newcommand{\tabCIFAR}{%
\newcolumntype{x}{>{\centering\arraybackslash\hspace{0pt}}p{13.5mm}}%
\renewcommand{\cs}{\hspace{1.5mm}}%
\tabulinesep=0.795mm%
\tabulinestyle{0.17mm}%
\begin{tabu}{|l|@{\cs}x@{\cs}x@{\cs}|@{\cs}x@{\cs}x@{\cs}|}
\tabucline{-}
\multirow{2}{*}{\raisebox{-0.3mm}{\bf Method}} & \multicolumn{2}{c|@{\cs}}{\bf Unconditional\ \ \ \ \ } & \multicolumn{2}{c|}{\bf Conditional\ \ \ \ \ } \\
& FID $\downarrow$ & \s IS $\uparrow$ & FID $\downarrow$ & \s IS $\uparrow$ \\
\tabucline{-}
ProGAN      \hfill\cite{Karras2017}     & 15.52\nostd{}  & \s8.56\std{0.06}  & --\nostd{}        & --\nostd{}          \\
AutoGAN     \hfill\cite{Gong2019}       & 12.42\nostd{}  & \s8.55\std{0.10}  & --\nostd{}        & --\nostd{}          \\
BigGAN      \hfill\cite{Brock2018}      & --\nostd{}     & --\nostd{}        & 14.73\nostd{}     & \s9.22\nostd{}      \\
+ Tuning    \hfill\cite{Kavalerov2019}  & --\nostd{}     & --\nostd{}        & \s8.47\nostd{}    & \s9.07\std{0.13}    \\
MultiHinge  \hfill\cite{Kavalerov2019}  & --\nostd{}     & --\nostd{}        & \s6.40\nostd{}    & \s9.58\std{0.09}    \\
FQ-GAN      \hfill\cite{Zhao2020}       & --\nostd{}     & --\nostd{}        & \s5.59\std{0.12}  & \s8.48\nostd{0.03}  \\
\tabucline{-}
\FINAL{Baseline}         & \FINAL{\s8.32\std{0.09}}       & \FINAL{\s9.21\std{0.09}}       & \FINAL{\s6.96\std{0.41}}       & \FINAL{\s9.53\std{0.06}}       \\
\FINAL{+ ADA (Ours)}     & \FINAL{\s5.33\std{0.35}}       & \FINAL{{\bf10.02}\std{0.07}}   & \FINAL{\s3.49\std{0.17}}       & \FINAL{{\bf10.24}\std{0.07}}   \\
\FINAL{+ Tuning (Ours)}  & \FINAL{{\bf\s2.92}\std{0.05}}  & \FINAL{\s9.83\std{0.04}}       & \FINAL{{\bf\s2.42}\std{0.04}}  & \FINAL{10.14\std{0.09}}        \\
\tabucline{-}
\end{tabu}%
}

\newcommand{\figSmallDatasetResults}[1]{
\begin{figure}[t]
\centering%
\footnotesize%
\subtabletop{0.494\linewidth}{36mm}{0.75}{\tabSmallDatasets}\hfill%
\subtabletop{0.475\linewidth}{36mm}{0.75}{\tabCIFAR}\\%
\makebox[0.494\linewidth][c]{(a) Small datasets}\hfill%
\makebox[0.475\linewidth][c]{(b) CIFAR-10}\hfill%
\caption{
\FINAL{
(a) Several small datasets trained with StyleGAN2 baseline (config \textsc{f}) and ADA, from scratch and using transfer learning. We used \textsc{FFHQ-140k} with matching resolution as a starting point for all transfers. We report the best KID, and compute FID using the same snapshot.
(c) Mean and standard deviation for CIFAR-10, computed from the best scores of 5 training runs. For the comparison methods we report the average scores when available, and the single best score otherwise. The best IS and FID were searched separately \cite{Kavalerov2019}, and often came from different snapshots. We computed the FID for Progressive~GAN~\cite{Karras2017} using the publicly available pre-trained network.
}
}
\label{#1}
\end{figure}
}

\begin{document}

\maketitle
\confnotice

\ifarxiv
	\newcommand{\refappResults}{Appendix~\ref{app:results}}
	\newcommand{\refappResultsFirstContact}{Appendix~\ref{app:results}}
	\newcommand{\refappPipeline}{Appendix~\ref{app:pipeline}}
	\newcommand{\refappTheory}{Appendix~\ref{app:theory}}
	\newcommand{\refappImplementation}{Appendix~\ref{app:implementation}}
	\newcommand{\refappPower}{Appendix~\ref{app:power}}
\else
	\newcommand{\refappResults}{Appendix~A}
	\newcommand{\refappResultsFirstContact}{Appendix~A in the Supplement}
	\newcommand{\refappPipeline}{Appendix~B}
	\newcommand{\refappTheory}{Appendix~C}
	\newcommand{\refappImplementation}{Appendix~D}
	\newcommand{\refappPower}{Appendix~E}
\fi

\begin{abstract}

Training generative adversarial networks (GAN) using too little data typically leads to discriminator overfitting, causing training to diverge.
We propose an adaptive discriminator augmentation mechanism that significantly stabilizes training in limited data regimes. 
The approach does not require changes to loss functions or network architectures,
and is applicable both when training from scratch and when fine-tuning an existing GAN on another dataset.
We demonstrate, on several datasets, that good results are now possible using only a few thousand training images, often matching StyleGAN2 results with an order of magnitude fewer images.
We expect this to open up new application domains for GANs.
We also find that the widely used CIFAR-10 is, in fact, a limited data benchmark, and improve the record FID from 5.59 to \FINAL{2.42}.

\end{abstract}

\section{Introduction}
\label{sec:intro}

The increasingly impressive results of generative adversarial networks (GAN)  \cite{Goodfellow2014,Miyato2018,Miyato2018B,Brock2018,Karras2017,Karras2018,Karras2019} are fueled by the seemingly  unlimited supply of images available online.
Still, it remains challenging to collect a large enough set of images for a specific application that places constraints on subject type, image quality, geographical location, time period, privacy, copyright status, etc.
The difficulties are further exacerbated in applications that require the capture of a new, custom dataset: acquiring, processing, and distributing the $\sim10^5-10^6$ images required to train a modern high-quality, high-resolution GAN is a costly undertaking.
This curbs the increasing use of generative models in fields such as medicine \cite{Yi2018}. A significant reduction in the number of images required therefore has the potential to considerably help many applications.

The key problem with small datasets is that the discriminator overfits to the training examples; its feedback to the generator becomes meaningless and training starts to diverge~\cite{Arjovsky2017A, Khoreva2019}.
In almost all areas of deep learning~\cite{Shorten2019}, \emph{dataset augmentation} is the standard solution against overfitting.
For example, training an image classifier under rotation, noise, etc., leads to increasing invariance to these semantics-preserving distortions --- a highly desirable quality in a classifier \cite{He2018,Cubuk2018,Cubuk2019}.
In contrast, a GAN trained under similar dataset augmentations learns to generate the augmented distribution \cite{zhang2019cr,zhao2020improved}. In general, such ``leaking'' of augmentations to the generated samples is highly undesirable. For example, a noise augmentation leads to noisy results, even if there is none in the dataset.

In this paper, we demonstrate how to use a wide range of augmentations to prevent the discriminator from overfitting, while ensuring that none of the augmentations leak to the generated images.
We start by presenting a comprehensive analysis of the conditions that prevent the augmentations from leaking. We then design a diverse set of augmentations, and an adaptive control scheme that enables the same approach to be used regardless of the amount of training data, properties of the dataset, or the exact training setup (e.g., training from scratch or transfer learning \cite{Mo2020,Wang2019minegan,Wang2018transfer,Noguchi2019}).

We demonstrate, on several datasets, that good results are now possible using only a few thousand images, often matching StyleGAN2 results with an order of magnitude fewer images.
Furthermore, we show that the popular CIFAR-10 benchmark suffers from limited data \FINAL{and achieve} a new record Fr\'echet inception distance (FID) \cite{Heusel2017} of \FINAL{2.42}, significantly improving over the current state of the art of 5.59 \cite{Zhao2020}.
We also present \textsc{MetFaces}, a high-quality benchmark dataset for limited data scenarios.
\FINAL{Our implementation and models are available at {\small\url{https://github.com/NVlabs/stylegan2-ada}}}

\figOverfit{fig:Overfit} %

\section{Overfitting in GANs}
\label{sec:overfit}

We start by studying how the quantity of available training data affects GAN training. We approach this by artificially subsetting larger datasets (\textsc{FFHQ} and \textsc{LSUN cat}) and observing the resulting dynamics.
For our baseline, we considered StyleGAN2~\cite{Karras2019} and BigGAN~\cite{Brock2018,Schonfeld2020}. Based on initial testing, we settled on StyleGAN2 because it provided more predictable results with significantly lower variance between training runs (see \refappResultsFirstContact{}). For each run, we randomize the subset of training data, order of training samples, and network initialization. %
To facilitate extensive sweeps over dataset sizes and hyperparameters, we use a downscaled $256\times256$ version of \textsc{FFHQ} and a lighter-weight configuration that reaches the same quality as the official StyleGAN2 config \textsc{f} for this dataset, but runs $4.6\times$ faster on NVIDIA DGX-1.%
\footnote{We use $2\times$ fewer feature maps, $2\times$ larger minibatch, mixed-precision training for layers at \mbox{$\ge$ $32^2$}, \mbox{$\eta=0.0025$}, \mbox{$\gamma=1$}, and exponential moving average half-life of 20k images for generator weights.}
We measure quality by computing FID between 50k generated images and all available \FINAL{training} images, \FINAL{as recommended by Heusel et al.~\cite{Heusel2017}}, regardless of the subset \FINAL{actually} used for training.

Figure~\ref{fig:Overfit}a shows our baseline results for different subsets of \textsc{FFHQ}.
Training starts the same way in each case, but eventually the progress stops and FID starts to rise. The less training data there is, the earlier this happens.
Figure~\ref{fig:Overfit}b,c shows the discriminator output distributions for real and generated images during training.
The distributions overlap initially but keep drifting apart as the discriminator becomes more and more confident, and the point where FID starts to deteriorate is consistent with the loss of sufficient overlap between distributions.
This is a strong indication of overfitting, evidenced further by the drop in accuracy measured for a separate validation set.
We propose a way to tackle this problem by employing versatile augmentations that prevent the discriminator from becoming overly confident.

\subsection{Stochastic discriminator augmentation}
\label{sec:augment}

By definition, any augmentation that is applied to the training dataset will get inherited to the generated images \cite{Goodfellow2014}. %
Zhao et al.~\cite{zhao2020improved} recently proposed balanced consistency regularization (bCR) as a solution that is not supposed to leak augmentations to the generated images.
Consistency regularization states that two sets of augmentations, applied to the same input image, should yield the same output~\cite{sajjadi16,Laine2017}. 
Zhao et al.~add consistency regularization terms for the discriminator loss, and enforce discriminator consistency for both real and generated images, whereas
no augmentations or consistency loss terms are applied when training the generator (Figure~\ref{fig:Augment}a).
As such, their approach effectively strives to generalize the discriminator by making it blind to the augmentations used in the CR term.
However, meeting this goal opens the door for leaking augmentations, because the generator will be free to produce images containing them without any penalty.
In Section~\ref{sec:results}, we show experimentally that bCR indeed suffers from this problem, and thus its effects are fundamentally similar to dataset augmentation.

Our solution is similar to bCR in that we also apply a set of augmentations to all images shown to the discriminator.
However, instead of adding separate CR loss terms, we evaluate the discriminator \emph{only} using augmented images, and do this also when training the generator (Figure~\ref{fig:Augment}b).
This approach that we call \emph{stochastic discriminator augmentation} is therefore very straightforward. 
Yet, this possibility has received little attention, possibly because at first glance it is not obvious if it even works:
if the discriminator never sees what the training images really look like, it is not clear if it can guide the generator properly (Figure~\ref{fig:Augment}c).
We will therefore first investigate the conditions under which \FINAL{this approach} will not leak an augmentation to the generated images, and then build a full pipeline out of such transformations.

\figDiagramsAndAugmentExamples{fig:Augment} %

\vspace{-0.75mm}
\subsection{Designing augmentations that do not leak}
\label{sec:theory}
\vspace{-0.5mm}

Discriminator augmentation corresponds to putting distorting, perhaps even destructive goggles on the discriminator, and asking the generator to produce samples that cannot be distinguished from the training set when viewed through the goggles. %
Bora et al.~\cite{Bora2018} consider a similar problem in training GANs under corrupted measurements, %
and show that the training \emph{implicitly} undoes the corruptions and finds the correct distribution,
as long as the corruption process is represented by an invertible transformation of probability distributions over the data space. We call such augmentation operators \emph{non-leaking}. 

The power of these invertible transformations is that they allow conclusions about the equality or inequality of the underlying sets to be drawn by observing only the augmented sets. 
It is crucial to understand that this does \emph{not} mean that augmentations performed on individual images would need to be undoable. For instance, an augmentation as extreme as setting the input image to zero 90\% of the time is invertible in the probability distribution sense: it would be easy, even for a human, to reason about the original distribution by ignoring black images until only 10\% of the images remain. On the other hand, random rotations chosen uniformly from $\{0^\circ, 90^\circ, 180^\circ, 270^\circ\}$ are not invertible: it is impossible to discern differences among the orientations after the augmentation.

The situation changes if this rotation is only executed at a probability \mbox{$p<1$}: this increases the relative occurrence of $0^\circ$, and now the augmented distributions can match only if the generated images have correct orientation.
Similarly, many other stochastic augmentations can be designed to be non-leaking on the condition that they are skipped with a non-zero probability.
\refappTheory{} shows that this can be made to hold for a large class of widely used augmentations, including deterministic mappings (e.g.,~basis transformations), additive noise, transformation groups (e.g,~image or color space rotations, flips and scaling), and projections (e.g.,~cutout~\cite{Devries2017}).
Furthermore, composing non-leaking augmentations in a fixed order yields an overall non-leaking augmentation. %

In Figure~\ref{fig:Leaks} we validate our analysis by three practical examples.
Isotropic scaling with log-normal distribution is an example of an inherently safe augmentation that does not leak regardless of the value of $p$ (Figure~\ref{fig:Leaks}a).
However, the aforementioned rotation by a random multiple of 90$^\circ$ must be skipped at least part of the time (Figure~\ref{fig:Leaks}b).
When $p$ is too high, the generator cannot know which way the generated images should face and ends up picking one of the possibilities at random.
As could be expected, the problem does not occur exclusively in the limiting case of \mbox{$p=1$}.
In practice, the training setup is poorly conditioned for nearby values as well due to
finite sampling, finite representational power of the networks, inductive bias, and training dynamics. 
When $p$ remains below \mbox{$\sim\!0.85$}, the generated images are always oriented correctly.
Between these regions, the generator sometimes picks a wrong orientation initially, and then partially drifts towards the correct distribution.
The same observations hold for a sequence of continuous color augmentations (Figure~\ref{fig:Leaks}c).
This experiment suggests that as long as $p$ remains below $0.8$, leaks are unlikely to happen in practice.

\subsection{Our augmentation pipeline}

We start from the assumption that a maximally diverse set of augmentations is beneficial, given the success of RandAugment~\cite{Cubuk2019} in image classification tasks.
We consider a pipeline of 18 transformations that are grouped into 6 categories:
pixel blitting ($x$-flips, 90$^{\circ}$ rotations, integer translation), more general geometric transformations, color transforms, image-space filtering, additive noise~\cite{Sonderby2016}, and cutout~\cite{Devries2017}.
Details of the individual augmentations are given in \refappPipeline{}.
Note that we execute augmentations also when training the generator (Figure~\ref{fig:Augment}b), which requires the augmentations to be differentiable.
We achieve this by implementing them using standard differentiable primitives offered by the deep learning framework.

During training, we process each image shown to the discriminator using a pre-defined set of transformations in a fixed order.
The strength of augmentations is controlled by the scalar \mbox{$p\in[0,1]$}, so that each transformation is applied with probability $p$ or skipped with probability \mbox{$1-p$}. 
We always use the same value of $p$ for all transformations.  %
The randomization is done separately for each augmentation and for each image in a minibatch.
Given that there are many augmentations in the pipeline,
even fairly small values of $p$ make it very unlikely that the discriminator sees a clean image (Figure~\ref{fig:Augment}c).
Nonetheless, the generator is guided to produce only clean images as long as $p$ remains below the practical safety limit.

\figLeakGroups{fig:Leaks} %

In Figure~\ref{fig:FixedSweeps} we study the effectiveness of \FINAL{stochastic discriminator augmentation} by performing exhaustive sweeps over $p$ for different augmentation categories and dataset sizes. We observe that \FINAL{it} can improve the results significantly in many cases. However, the optimal augmentation strength depends heavily on the amount of training data, and not all augmentation categories are equally useful in practice.
\figFixedSweepsSimple{fig:FixedSweeps} %
With a 2k training set, the vast majority of the benefit came from pixel blitting and geometric transforms. Color transforms were modestly beneficial, while image-space filtering, noise, and cutout were not particularly useful.
In this case, the best results were obtained using strong augmentations.
The curves also indicate some of the augmentations becoming leaky when \mbox{$ p\rightarrow 1$}.
With a 10k training set, the higher values of $p$ were less helpful, and with 140k the situation was markedly different: all augmentations were harmful.
Based on these results, we choose to use only pixel blitting, geometric, and color transforms for the rest of our tests.
Figure~\ref{fig:FixedSweeps}d shows that while stronger augmentations reduce overfitting, they also slow down the convergence.

In practice, the sensitivity to dataset size mandates a costly grid search, and even so, relying on any fixed $p$ may not be the best choice.
Next, we address these concerns by making the process adaptive.

\section{Adaptive discriminator augmentation}
\label{sec:ada}

Ideally, we would like to avoid manual tuning of the augmentation strength and instead control it dynamically based on the degree of overfitting.
Figure~\ref{fig:Overfit} suggests a few possible approaches for this.
The standard way of quantifying overfitting is to use a separate validation set and observe its behavior relative to the training set.
From the figure we see that when overfitting kicks in, the validation set starts behaving increasingly like the generated images.
This is a quantifiable effect, albeit with the drawback of requiring a separate validation set when training data may already be in short supply.
We can also see that with the non-saturating loss \cite{Goodfellow2014} used by StyleGAN2, the discriminator outputs for real and generated images diverge symmetrically around zero as the situation gets worse.
This divergence can be quantified without a separate validation set.

Let us denote the discriminator outputs by $D_\textrm{train}$, $D_\textrm{validation}$, and $D_\textrm{generated}$ for the training set, validation set, and generated images, respectively, and their mean over $N$ consecutive minibatches by $\mathbb{E}[\cdot]$. In practice we use \mbox{$N=4$}, which corresponds to \mbox{$4\times64=256$} images.
We can now turn our observations about Figure~\ref{fig:Overfit} into two plausible overfitting heuristics:
\begin{equation}
r_v = \frac{\mathbb{E}[D_\textrm{train}] - \mathbb{E}[D_\textrm{validation}]}{\mathbb{E}[D_\textrm{train}] - \mathbb{E}[D_\textrm{generated}]}\hspace{15mm}%
r_t = \mathbb{E}[\textrm{sign}(D_\textrm{train})]
\end{equation}
For both heuristics, \mbox{$r=0$} means no overfitting and \mbox{$r=1$} indicates complete overfitting, and
our goal is to adjust the augmentation probability $p$ so that the chosen heuristic matches a suitable target value.
The first heuristic, $r_v$, expresses the output for a validation set relative to the training set and generated images.
Since it assumes the existence of a separate validation set, we include it mainly as a comparison method.
The second heuristic, $r_t$, estimates the portion of the training set that gets positive discriminator outputs.
We have found this to be far less sensitive to the chosen target value and other hyperparameters than the obvious alternative of looking at \mbox{$\mathbb{E}[D_\textrm{train}]$} directly.

We control the augmentation strength $p$ as follows. We initialize $p$ to zero and adjust its value once every four minibatches\footnote{\FINAL{This choice follows from StyleGAN2 training loop layout. The results are not sensitive to this parameter.}} based on the chosen overfitting heuristic. If the heuristic indicates too much/little overfitting, we counter by incrementing/decrementing $p$ by a fixed amount. We set the adjustment size so that $p$ can rise from 0 to 1 sufficiently quickly, e.g., in 500k images.
After every step we clamp $p$ from below to 0. %
We call this variant \emph{adaptive discriminator augmentation} (ADA).

In Figure~\ref{fig:AdaptiveSweeps}a,b we measure how the target value affects the quality obtainable using these heuristics.
We observe that $r_v$ and $r_t$ are both effective in preventing overfitting, and that they both improve the results over the best fixed $p$ found using grid search. We choose to use the more realistic $r_t$ heuristic in all subsequent tests, with $0.6$ as the target value.
Figure~\ref{fig:AdaptiveSweeps}c shows the resulting $p$ over time. With a 2k training set, augmentations were applied almost always towards the end.
This exceeds the practical safety limit after which some augmentations become leaky, indicating that the augmentations were not powerful enough. 
Indeed, FID started deteriorating after $p\approx0.5$ in this extreme case.
Figure~\ref{fig:AdaptiveSweeps}d shows the evolution of $r_t$ with adaptive vs fixed $p$, showing that a fixed $p$ tends to be too strong in the beginning and too weak towards the end.
\figAdaptiveSweeps{fig:AdaptiveSweeps} %

Figure~\ref{fig:NoOverfit} repeats the setup from Figure~\ref{fig:Overfit} using ADA.
Convergence is now achieved regardless of the training set size and overfitting no longer occurs.
Without augmentations, the gradients the generator receives from the discriminator become very simplistic over time\,---\,the discriminator starts to pay attention to only a handful of features, and the generator is free to create otherwise non-sensical images.
With ADA, the gradient field stays much more detailed which prevents such deterioration.
\FINAL{In an interesting parallel, it has been shown that loss functions can be made significantly more robust in regression settings by using similar image augmentation ensembles \cite{Kettunen2019}.}
\figNoOverfit{fig:NoOverfit} %

\vspace{-1.5mm}
\section{Evaluation}
\label{sec:results}
\vspace{-1.0mm}

We start by testing our method against a number of alternatives in \textsc{FFHQ} and \textsc{LSUN Cat}, first in a setting where a GAN is trained from scratch, then by applying transfer learning on a pre-trained GAN. We conclude with results for several smaller datasets.

\vspace{-1.0mm}
\subsection{Training from scratch}
\vspace{-1.0mm}
Figure~\ref{fig:MainSweep} shows our results in \textsc{FFHQ} and \textsc{LSUN Cat} across training set sizes, demonstrating that our adaptive discriminator augmentation (ADA) improves FIDs substantially in limited data scenarios.
We also show results for balanced consistency regularization (bCR)~\cite{zhao2020improved}, which has not been studied in the context of limited data before. 
We find that bCR can be highly effective when the lack of data is not too severe, but also that its set of augmentations leaks to the generated images. In this example, we used only $xy$-translations by integer offsets for bCR, and Figure~\ref{fig:MainSweep}d shows that the generated images get jittered as a result.
This means that bCR is essentially a dataset augmentation and needs to be limited to symmetries that actually benefit the training data, e.g., $x$-flip is often acceptable but $y$-flip only rarely.
Meanwhile, with ADA the augmentations do not leak, and thus the same diverse set of augmentations can be safely used in all datasets. 
We also find that the benefits for ADA and bCR are largely additive.
We combine ADA and bCR so that ADA is first applied to the input image (real or generated), and bCR then creates another version of this image using \emph{its own set of augmentations}.
Qualitative results are shown in \refappResults{}.
\figMainSweep{fig:MainSweep} %

In Figure~\ref{fig:ComparisonMethods}a we further compare our adaptive augmentation against a wider set of alternatives:
PA-GAN~\cite{Khoreva2019}, WGAN-GP~\cite{Gulrajani2017}, zCR~\cite{zhao2020improved}, auxiliary rotations~\cite{Chen2018aux}, and spectral normalization~\cite{Miyato2018B}. %
We also try modifying our baseline to use a shallower mapping network, which can be trained with less data, borrowing intuition from DeLiGAN~\cite{Gurumurthy2017}.
Finally, we try replacing our augmentations with multiplicative dropout~\cite{srivastava2014}, whose per-layer strength is driven by our adaptation algorithm.
We spent considerable effort tuning the parameters of all these methods, see \refappImplementation{}.
We can see that ADA gave significantly better results than the alternatives. While PA-GAN is somewhat similar to our method, its checksum task was not strong enough to prevent overfitting in our tests.
\FINAL{Figure~\ref{fig:ComparisonMethods}b shows that reducing the discriminator capacity is generally harmful and does not prevent overfitting.}
\figComparisonMethods{fig:ComparisonMethods} %

\subsection{Transfer learning}

Transfer learning reduces the training data requirements by starting from a model trained using some other dataset, instead of a random initialization. 
Several authors have explored this in the context of GANs \cite{Wang2019minegan,Wang2018transfer,Noguchi2019}, and 
Mo et al.~\cite{Mo2020} recently showed strong results by freezing the highest-resolution layers of the discriminator during transfer (Freeze-D).

We explore several transfer learning setups in Figure~\ref{fig:TransferLearning}, using the best Freeze-D configuration found for each case with grid search.
Transfer learning gives significantly better results than from-scratch training, and its
success seems to depend primarily on the diversity of the source dataset, instead of the similarity between subjects.
For example, \textsc{FFHQ} (human faces) can be trained equally well from \textsc{CelebA-HQ} (human faces, low diversity) or \textsc{LSUN Dog} (more diverse).
\textsc{LSUN Cat}, however, can only be trained from \textsc{LSUN Dog}, which has comparable diversity, but not from the less diverse datasets.
With small target dataset sizes, our baseline achieves reasonable FID quickly, but the progress soon reverts as training continues.
ADA is again able to prevent the divergence almost completely.
Freeze-D provides a small but reliable improvement when used together with ADA but is not able to prevent the divergence on its own.

\figTransferLearning{fig:TransferLearning} %

\vspace{-0.4mm}%
\subsection{Small datasets}
\label{sec:datasets}
\vspace{-0.4mm}%

We tried our method with several datasets that consist of a limited number of training images (Figure~\ref{fig:SmallDatasetImages}).
\textsc{MetFaces} is our new dataset of 1336 high-quality faces extracted from the collection of Metropolitan Museum of Art ({\small\tt https://metmuseum.github.io/}).
\textsc{BreCaHAD} \cite{BreCaHAD} consists of only 162 breast cancer histopathology images ($1360\times1024$); we reorganized these into 1944 partially overlapping crops of $512^2$.
\FINAL{Animal faces (\textsc{AFHQ})~\cite{AFHQ} includes $\sim$5k closeups per category for dogs, cats, and wild life; we treated these as three separate datasets and trained a separate network for each of them.}
\textsc{CIFAR-10} includes 50k tiny images in 10 categories \cite{CIFAR10}.
\figSmallDatasetImages{fig:SmallDatasetImages} %

\FINAL{Figure~\ref{fig:SmallDatasetResults}} reveals that FID is not an ideal metric for small datasets, because it becomes dominated by the inherent bias when the number of real images is insufficient.
We find that kernel inception distance (KID)~\cite{KID2018}\,---\,that is unbiased by design\,---\,is more descriptive in practice and see that ADA provides a dramatic improvement over baseline StyleGAN2.
\FINAL{This is especially true when training from scratch, but transfer learning also benefits from ADA.}
In the widely used CIFAR-10 benchmark, we improve the SOTA FID from 5.59 to \FINAL{2.42} and inception score (IS)~\cite{Salimans2016B}
 from 9.58 to \FINAL{10.24} in the class-conditional setting \FINAL{(Figure~\ref{fig:SmallDatasetResults}b)}. This large improvement portrays CIFAR-10 as a limited data benchmark. We also note that CIFAR-specific architecture tuning had a significant effect.
\figSmallDatasetResults{fig:SmallDatasetResults} %

\vspace{0mm}%
\section{Conclusions}
\label{sec:conclusion}
\vspace{0mm}%

We have shown that our adaptive discriminator augmentation reliably stabilizes training and vastly improves the result quality when training data is in short supply.
Of course, augmentation is not a substitute for real data\,---\,one should always try to collect a large, high-quality set of training data first, and only then fill the gaps using augmentation.
As future work, it would be worthwhile to search for the most effective set of augmentations, and to see if recently published techniques, such as the U-net discriminator~\cite{Schonfeld2020} or multi-modal generator~\cite{Sendik2020}, could also help with limited data.

\FINAL{Enabling ADA has a negligible effect on the energy consumption of training a single model. As such, using it does not increase the cost of training models for practical use or developing methods that require large-scale exploration. For reference, \refappPower{} provides a breakdown of all computation that we performed related to this paper; the project consumed a total of 325 MWh of electricity, or 135 single-GPU years, the majority of which can be attributed to extensive comparisons and sweeps.}

\FINAL{
Interestingly, the core idea of discriminator augmentations was independently discovered by three other research groups in parallel work: Z.~Zhao~et~al.~\cite{Zhao2020image}, Tran~et~al.~\cite{Tran2020}, and S.~Zhao~et~al.~\cite{ShengyuZhao2020}. We recommend these papers as they all offer a different set of intuition, experiments, and theoretical justifications.
While two of these papers \cite{Zhao2020image,ShengyuZhao2020} propose essentially the same augmentation mechanism as we do, they study the absence of leak artifacts only empirically. The third paper \cite{Tran2020} presents a theoretical justification based on invertibility, but arrives at a different argument that leads to a more complex network architecture, along with significant restrictions on the set of possible augmentations. None of these works consider the possibility of tuning augmentation strength adaptively.
Our experiments in Section~\ref{sec:ada} show that the optimal augmentation strength not only varies between datasets of different content and size, but also over the course of training\,---\,even an optimal set of fixed augmentation parameters is likely to leave performance on the table.

A direct comparison of results between the parallel works is difficult because the only dataset used in all papers is CIFAR-10. Regrettably, the other three papers compute FID using 10k generated images and 10k \emph{validation} images (FID-10k), while we use follow the original recommendation of Heusel et al.~\cite{Heusel2017} and use 50k generated images and all \emph{training} images. Their FID-10k numbers are thus not comparable to the FIDs in Figure~\ref{fig:SmallDatasetResults}b. For this reason we also computed FID-10k for our method, obtaining $7.01\pm0.06$ for unconditional and $6.54\pm0.06$ for conditional. These compare favorably to parallel work's unconditional 9.89 \cite{ShengyuZhao2020} or 10.89 \cite{Tran2020}, and conditional 8.30 \cite{Zhao2020image} or 8.49 \cite{ShengyuZhao2020}.
It seems likely that some combination of the ideas from all four papers could further improve our results. For example, more diverse set of augmentations or contrastive regularization \cite{Zhao2020image} might be worth testing.
}

\paragraph{Acknowledgements}
We thank David Luebke for helpful comments; Tero Kuosmanen \FINAL{and Sabu Nadarajan for their support with} compute infrastructure; and Edgar Sch{\"o}nfeld for \FINAL{guidance} on setting up unconditional BigGAN.

\newpage
\section*{Broader impact}

Data-driven generative modeling means learning a computational recipe for generating complicated data based purely on examples. This is a foundational problem in machine learning. In addition to their fundamental nature, generative models have several uses within applied machine learning research as priors, regularizers, and so on. In those roles, they advance the capabilities of computer vision and graphics algorithms for analyzing and synthesizing realistic imagery.

The methods presented in this work enable high-quality generative image models to be trained using significantly less data than required by existing approaches. It thereby primarily contributes to the deep technical question of how much data is enough for generative models to succeed in picking up the necessary commonalities and relationships in the data.

From an applied point of view, this work contributes to efficiency; it does not introduce fundamental new capabilities. Therefore, it seems likely that the advances here will not substantially affect the overall themes\,---\,surveillance, authenticity, privacy, etc.\,---\,in the active discussion on the broader impacts of computer vision and graphics.

Specifically, generative models' implications on image and video authenticity is a topic of active discussion. Most attention revolves around conditional models that allow semantic control and sometimes manipulation of existing images. Our algorithm does not offer direct controls for high-level attributes (e.g., identity, pose, expression of people) in the generated images, nor does it enable direct modification of existing images. However, over time and through the work of other researchers, our advances will likely lead to improvements in these types of models as well.

The contributions in this work make it easier to train high-quality generative models with custom sets of images. By this, we eliminate, or at least significantly lower, the barrier for applying GAN-type models in many applied fields of research. We hope and believe that this will accelerate progress in several such fields. For instance, modeling the space of possible appearance of biological specimens (tissues, tumors, etc.) is a growing field of research that appears to chronically suffer from limited high-quality data. Overall, generative models hold promise for increased understanding of the complex and hard-to-pinpoint relationships in many real-world phenomena; our work hopefully increases the breadth of phenomena that can be studied.

\renewcommand\thefigure{F\arabic{figure}}
\setcounter{figure}{0}

\renewcommand\thefigure{\arabic{figure}}

{\small
	\bibliographystyle{ieee}
	\bibliography{paper}
}

\ifarxiv
	\appendix
	\newcommand{\refpaper}[1]{\ref{#1}}
	\vfil\vfil\vfil\vfil\vfil\vfil\vfil\vfil\vfil\vfil
	\vfil\vfil\vfil\vfil\vfil\vfil\vfil\vfil\vfil\vfil
	\vfil\vfil\vfil\vfil\vfil\vfil\vfil\vfil\vfil\vfil
	\newcommand{\figUncuratedMetFaces}[1]{
\begin{figure}[p]
\footnotesize%
\renewcommand{\h}{0.485\linewidth}%
\makebox[\h][c]{ADA (Ours), truncated \raisebox{0.1mm}{\scalebox{0.9}{$(\psi=0.7)$}}}\hfill%
\makebox[\h][c]{Real images from the training set}%
\vspace{0mm}\\
\includegraphics[width=\h]{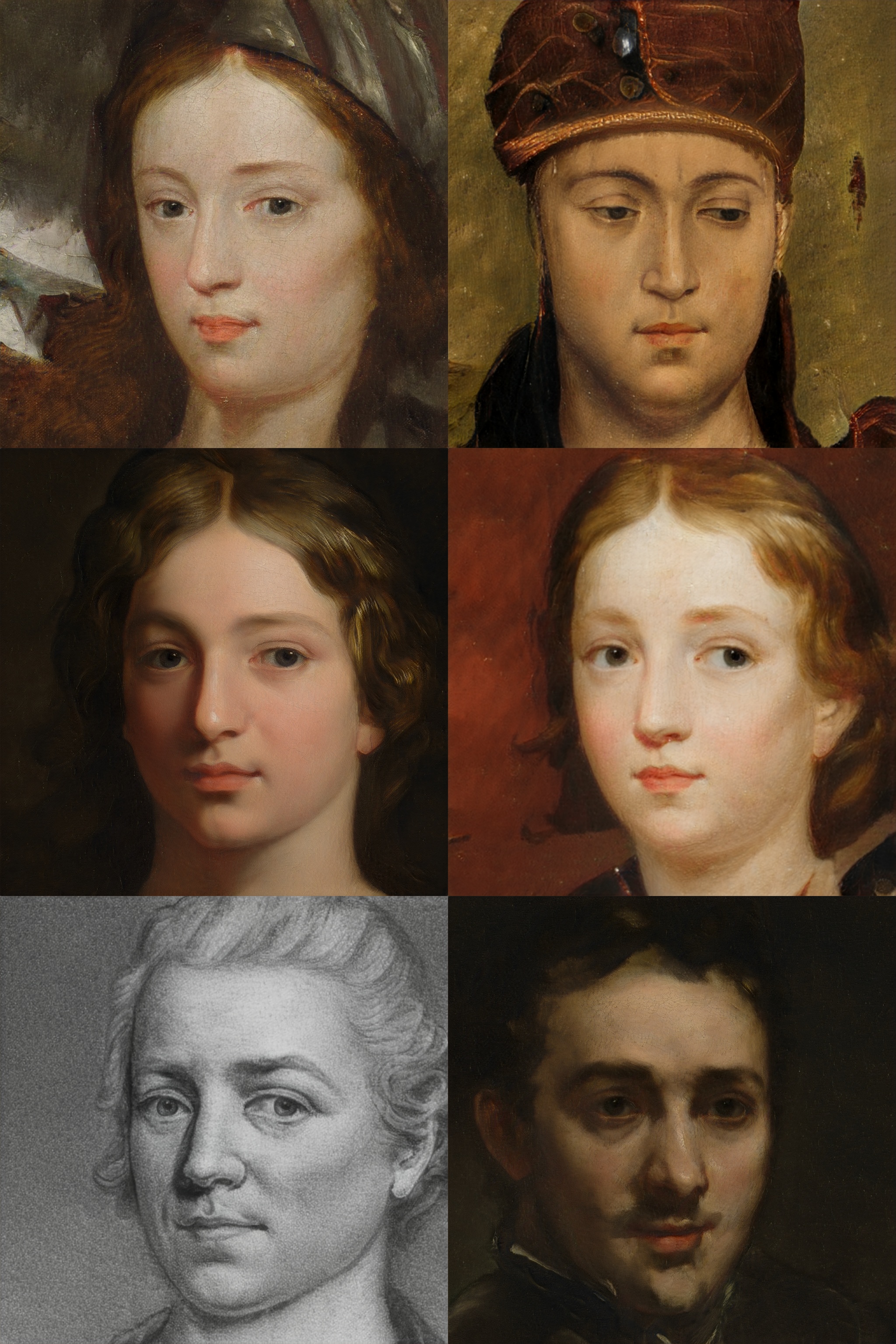}\hfill%
\includegraphics[width=\h]{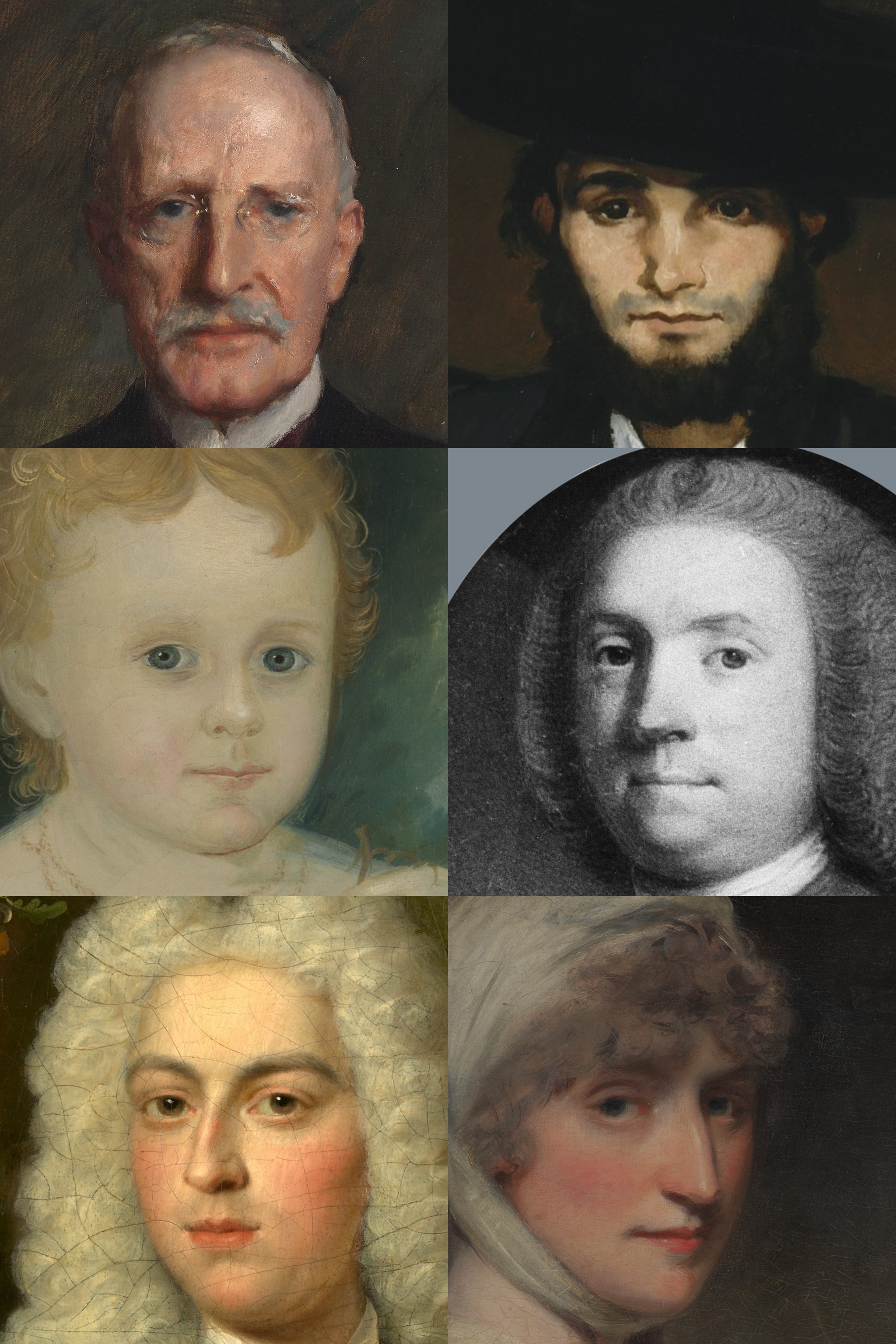}%
\vspace{1.1mm}\\
\makebox[\h][c]{ADA (Ours), untruncated}\hfill%
\makebox[\h][c]{Original StyleGAN2 config \textsc{f}, untruncated}%
\vspace{0mm}\\
\includegraphics[width=\h]{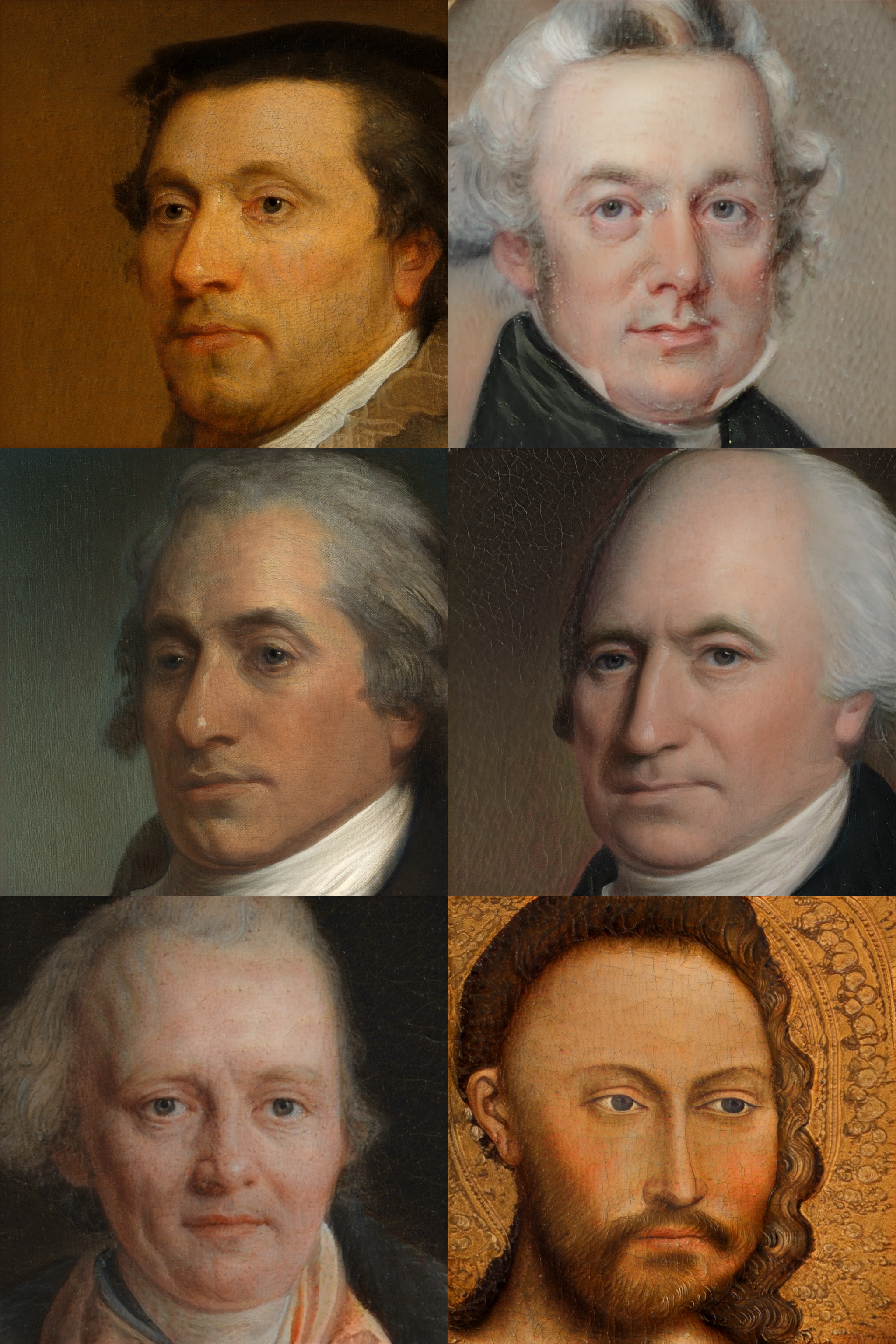}\hfill%
\includegraphics[width=\h]{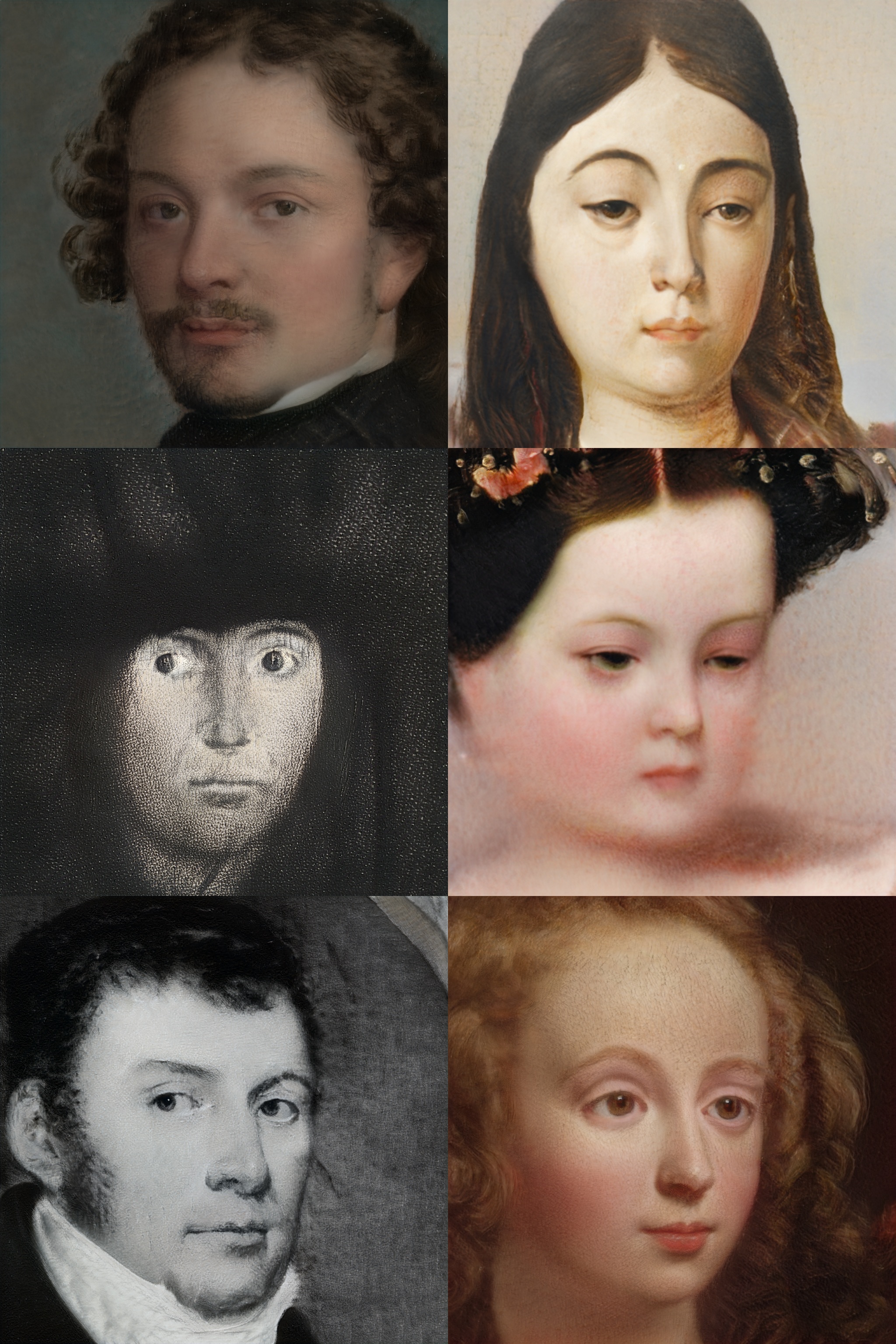}%
\vspace{-1.2mm}\\
\makebox[\h][c]{\scriptsize FID {\bf15.34} -- KID {\bf0.81}\raisebox{0.25mm}{\scalebox{0.7}{$\times10^3$}} -- Recall 0.261}\hfill%
\makebox[\h][c]{\scriptsize FID 19.47 -- KID 3.16\raisebox{0.25mm}{\scalebox{0.7}{$\times10^3$}} -- Recall {\bf0.350}}%
\vspace{0mm}%
\caption{
Uncurated 1024$\times$1024 results generated for \textsc{MetFaces} (1336 images) with and without ADA, along with real images from the training set.
Both generators were trained using transfer learning, starting from the pre-trained StyleGAN2 for FFHQ.
We recommend zooming in.
}
\label{#1}
\end{figure}
}

\newcommand{\figUncuratedBreCaHAD}[1]{
\begin{figure}[p]
\footnotesize%
\renewcommand{\h}{0.493\linewidth}%
\makebox[\h][c]{\FINAL{ADA (Ours), truncated \raisebox{0.1mm}{\scalebox{0.9}{$(\psi=0.7)$}}}}\hfill%
\makebox[\h][c]{\FINAL{Real images from the training set}}%
\vspace{0.2mm}\\
\includegraphics[width=\h]{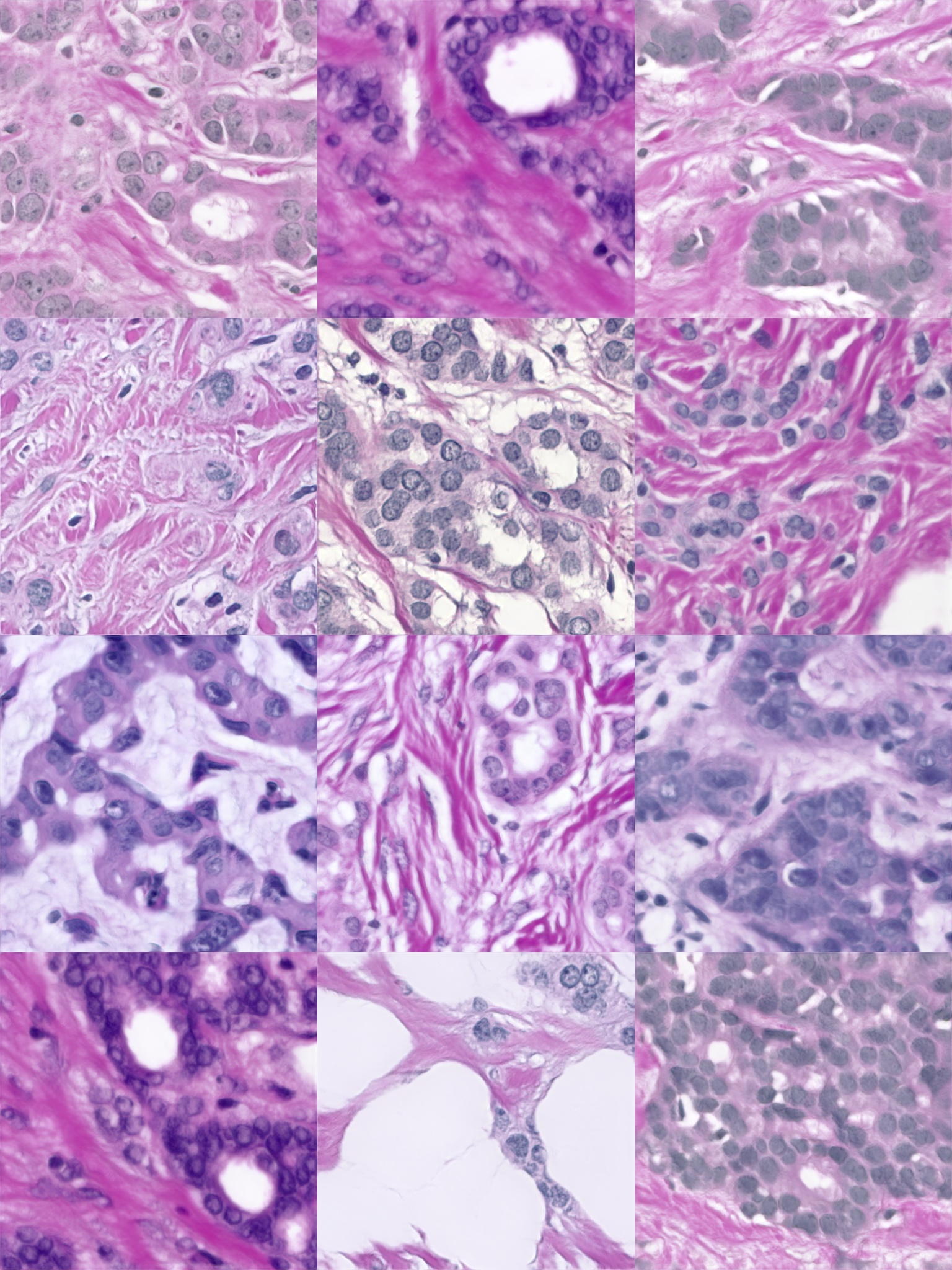}\hfill%
\includegraphics[width=\h]{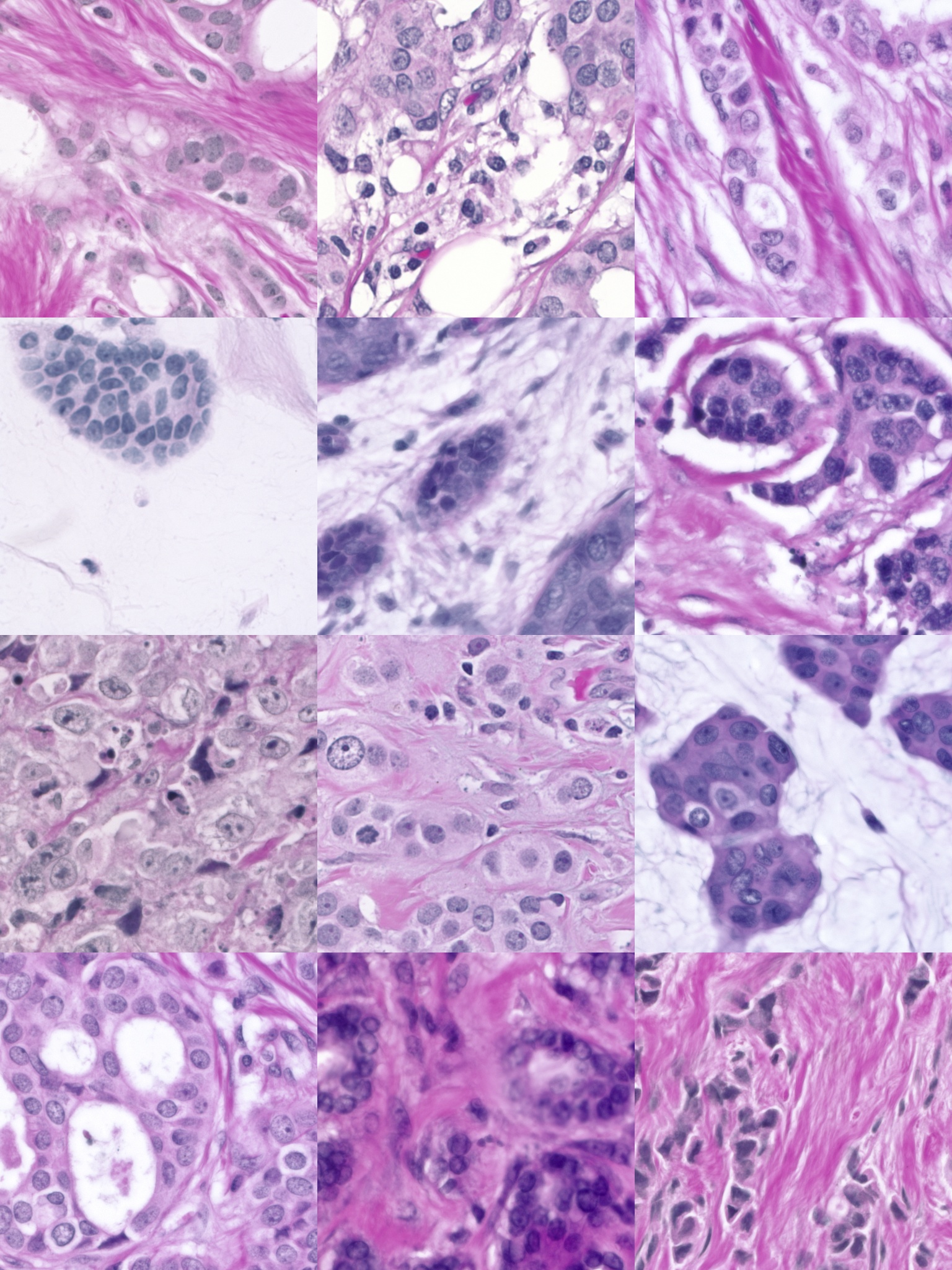}%
\vspace{2mm}\\
\makebox[\h][c]{\FINAL{ADA (Ours), untruncated}}\hfill%
\makebox[\h][c]{\FINAL{Original StyleGAN2 config \textsc{f}, untruncated}}%
\vspace{0.5mm}\\
\includegraphics[width=\h]{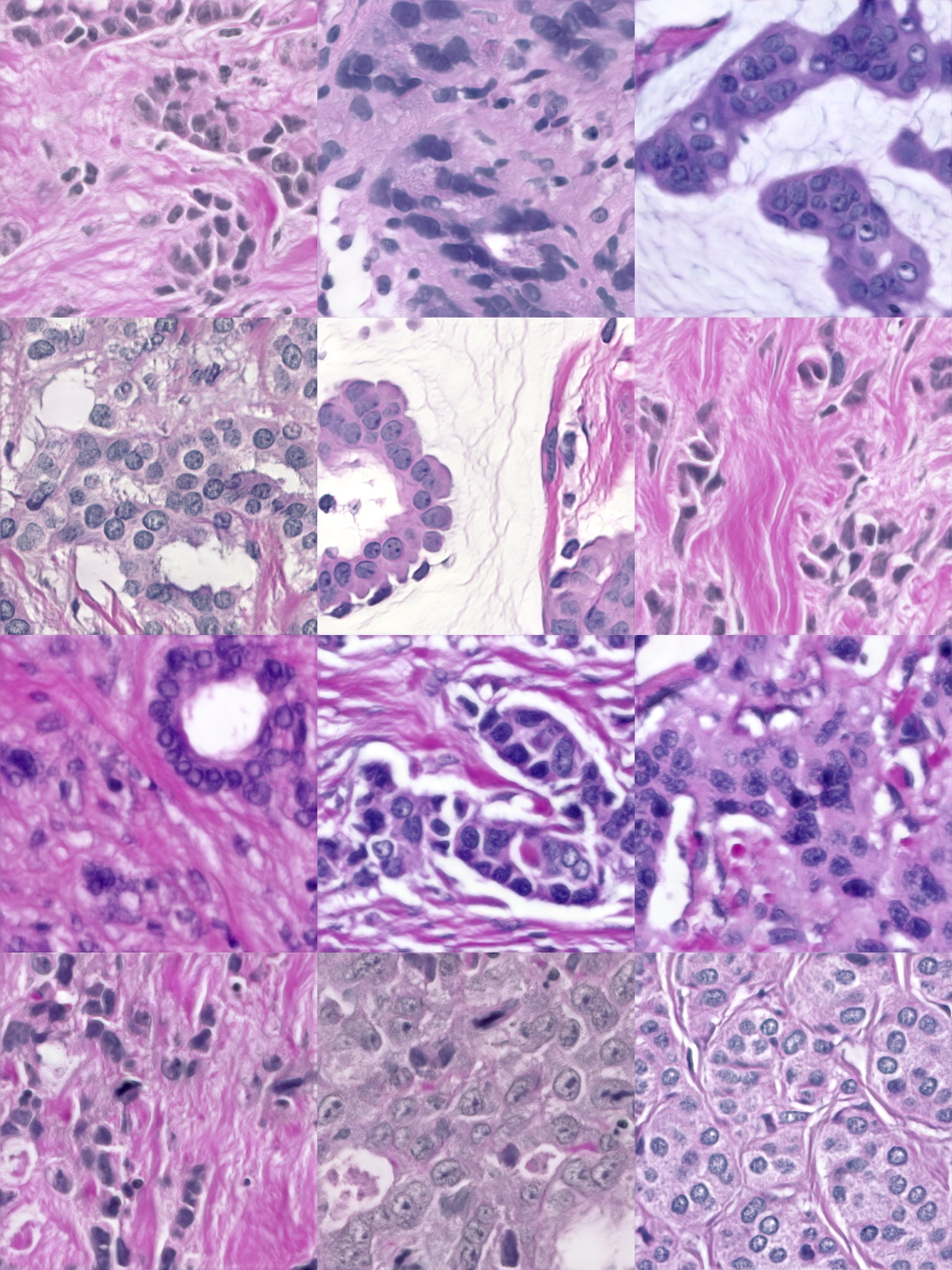}\hfill%
\includegraphics[width=\h]{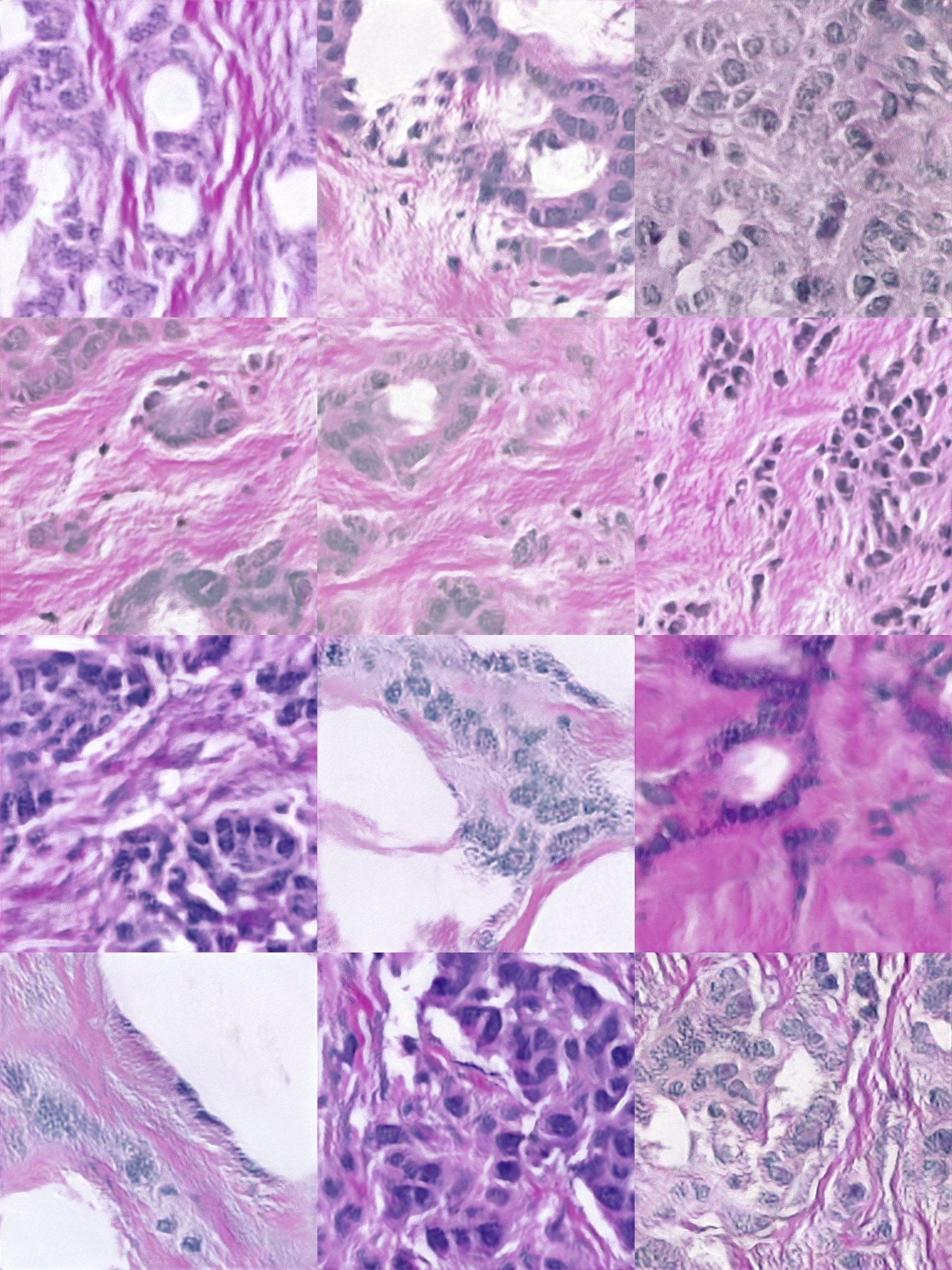}%
\vspace{-1.2mm}\\
\makebox[\h][c]{\scriptsize FID {\bf15.71} -- KID {\bf2.88}\raisebox{0.25mm}{\scalebox{0.7}{$\times10^3$}} -- Recall {\bf0.340}}\hfill%
\makebox[\h][c]{\scriptsize FID 97.72 -- KID 89.76\raisebox{0.25mm}{\scalebox{0.7}{$\times10^3$}} -- Recall 0.027}%
\vspace{2mm}%
\caption{
Uncurated 512$\times$512 results generated for \textsc{BreCaHAD} \cite{BreCaHAD} (1944 images) with and without ADA, along with real images from the training set.
Both generators were trained from scratch.
We recommend zooming in to inspect the image quality in detail.
}
\label{#1}
\end{figure}
}

\newcommand{\figUncuratedCats}[1]{
\begin{figure}[p]
\footnotesize%
\renewcommand{\h}{0.493\linewidth}%
\makebox[\h][c]{\FINAL{ADA (Ours), truncated \raisebox{0.1mm}{\scalebox{0.9}{$(\psi=0.7)$}}}}\hfill%
\makebox[\h][c]{\FINAL{Real images from the training set}}%
\vspace{0.2mm}\\
\includegraphics[width=\h]{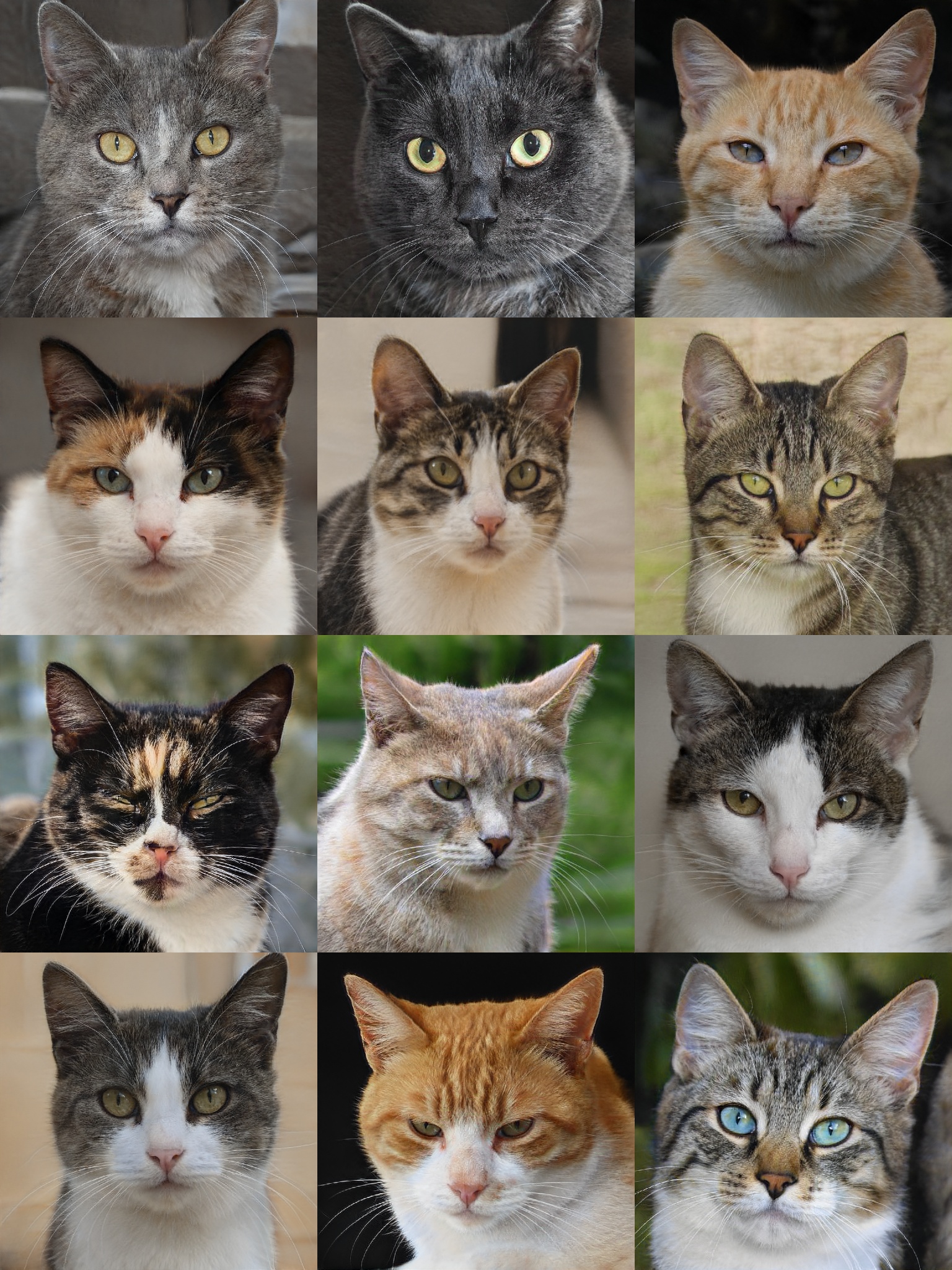}\hfill%
\includegraphics[width=\h]{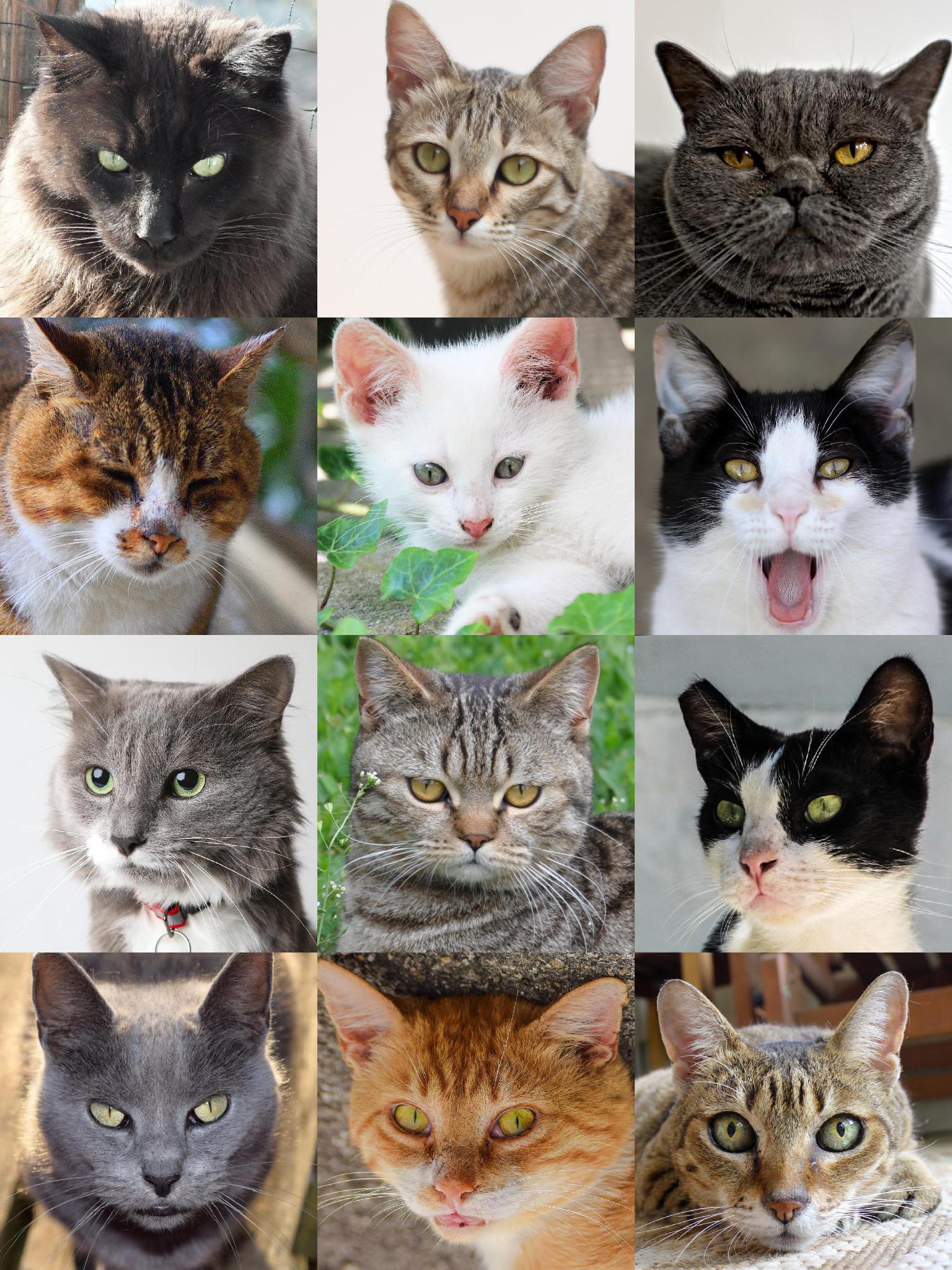}%
\vspace{2mm}\\
\makebox[\h][c]{\FINAL{ADA (Ours), untruncated}}\hfill%
\makebox[\h][c]{\FINAL{Original StyleGAN2 config \textsc{f}, untruncated}}%
\vspace{0.5mm}\\
\includegraphics[width=\h]{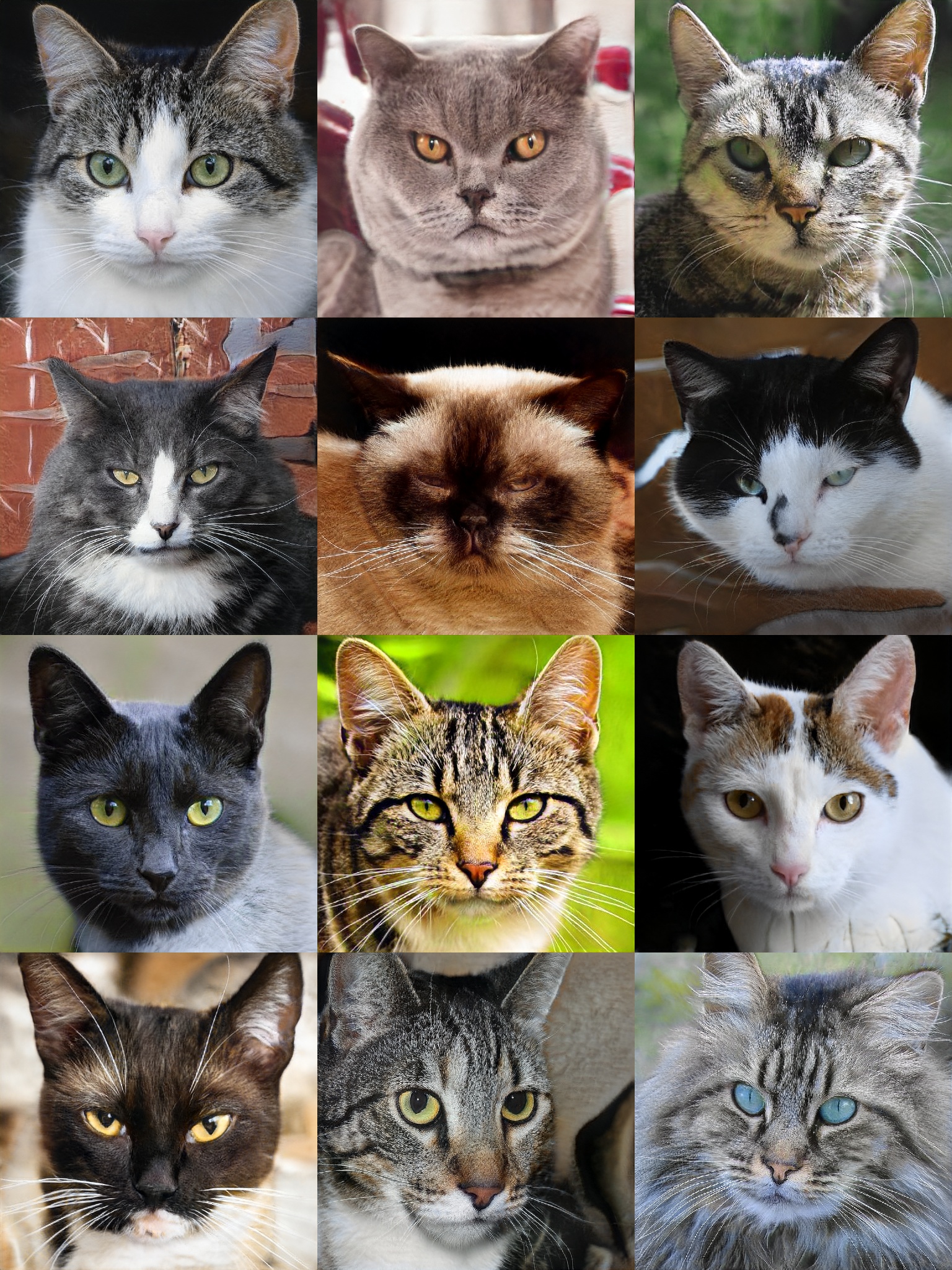}\hfill%
\includegraphics[width=\h]{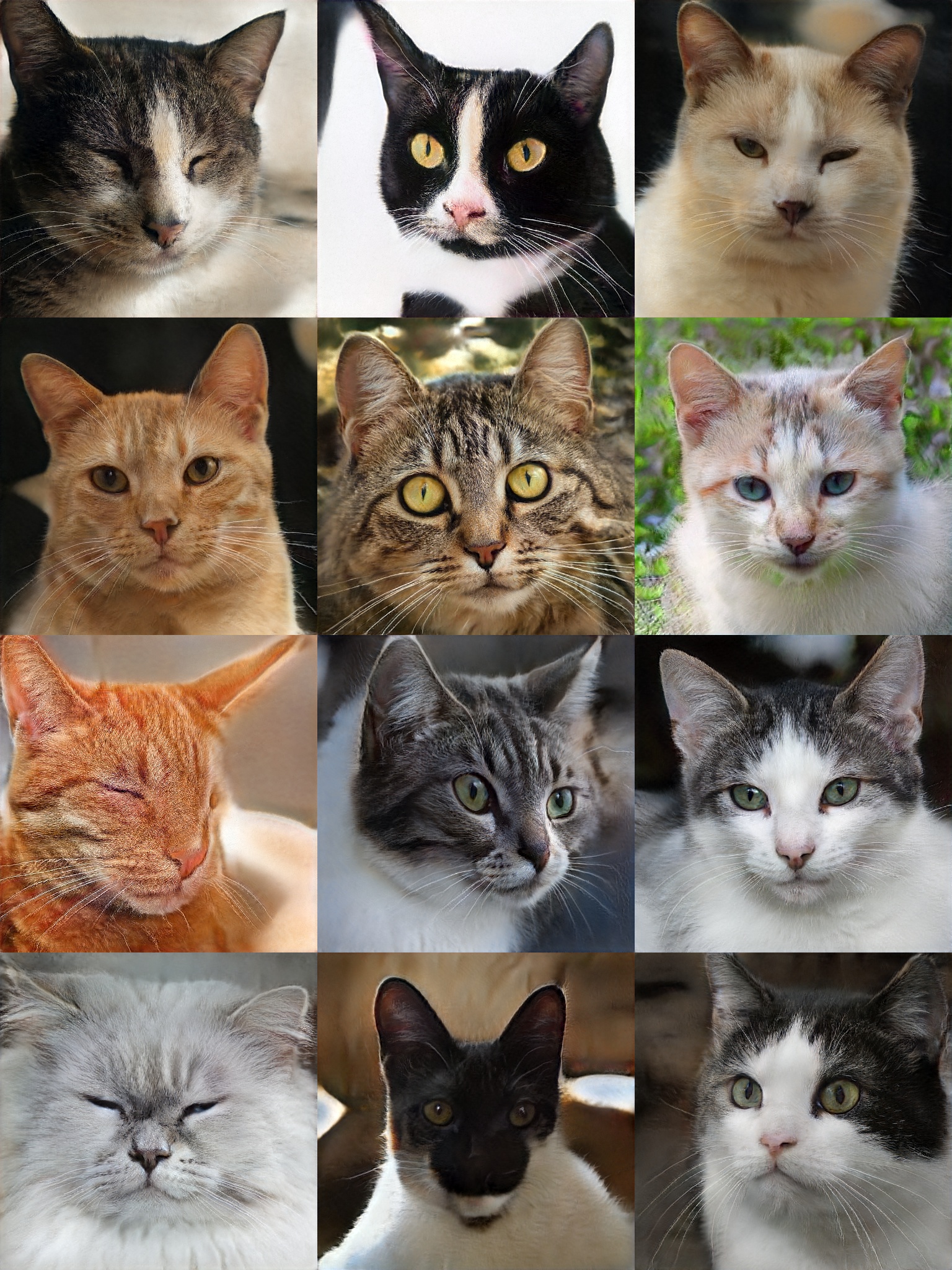}%
\vspace{-1.2mm}\\
\makebox[\h][c]{\scriptsize FID {\bf3.55} -- KID {\bf0.66}\raisebox{0.25mm}{\scalebox{0.7}{$\times10^3$}} -- Recall {\bf 0.430}}\hfill%
\makebox[\h][c]{\scriptsize FID 5.13 -- KID 1.54\raisebox{0.25mm}{\scalebox{0.7}{$\times10^3$}} -- Recall 0.215}%
\vspace{2mm}%
\caption{
\FINAL{%
Uncurated 512$\times$512 results generated for \textsc{AFHQ Cat} \cite{AFHQ} (5153 images) with and without ADA, along with real images from the training set.
Both generators were trained from scratch.
We recommend zooming in to inspect the image quality in detail.
}}
\label{#1}
\end{figure}
}

\newcommand{\figUncuratedDogs}[1]{
\begin{figure}[p]
\footnotesize%
\renewcommand{\h}{0.493\linewidth}%
\makebox[\h][c]{ADA (Ours), truncated \raisebox{0.1mm}{\scalebox{0.9}{$(\psi=0.7)$}}}\hfill%
\makebox[\h][c]{Real images from the training set}%
\vspace{0.2mm}\\
\includegraphics[width=\h]{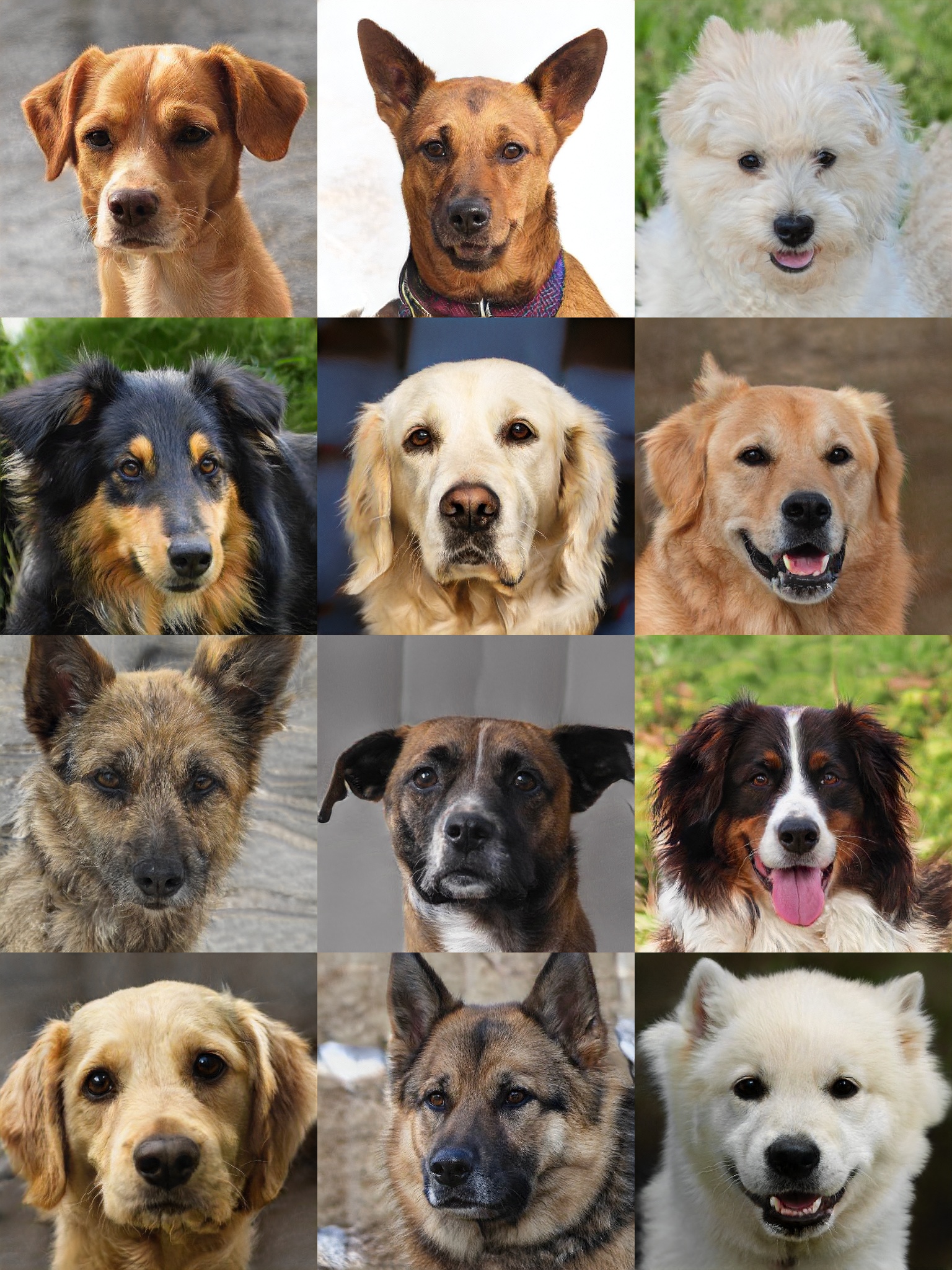}\hfill%
\includegraphics[width=\h]{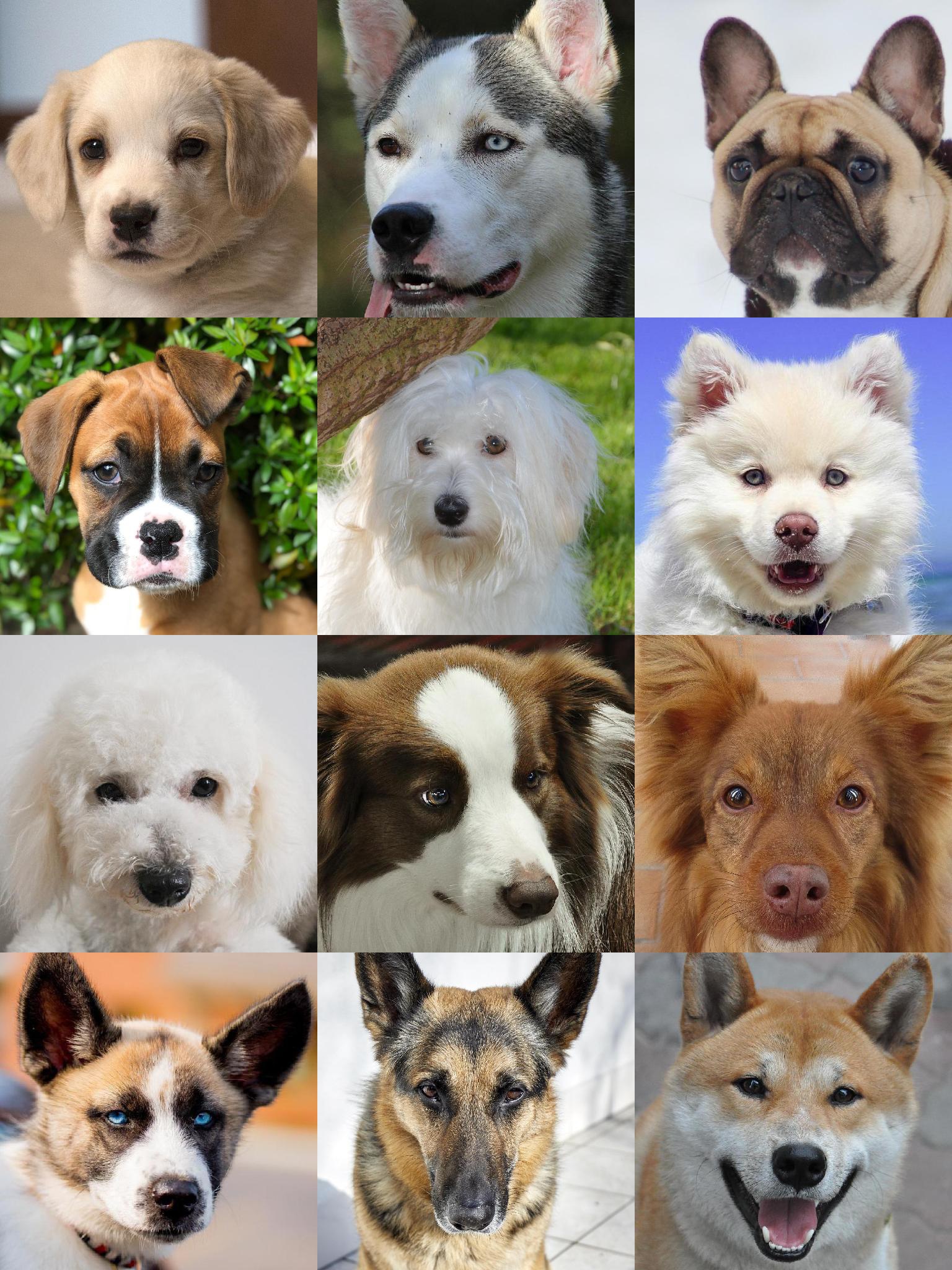}%
\vspace{2mm}\\
\makebox[\h][c]{ADA (Ours), untruncated}\hfill%
\makebox[\h][c]{Original StyleGAN2 config \textsc{f}, untruncated}%
\vspace{0.5mm}\\
\includegraphics[width=\h]{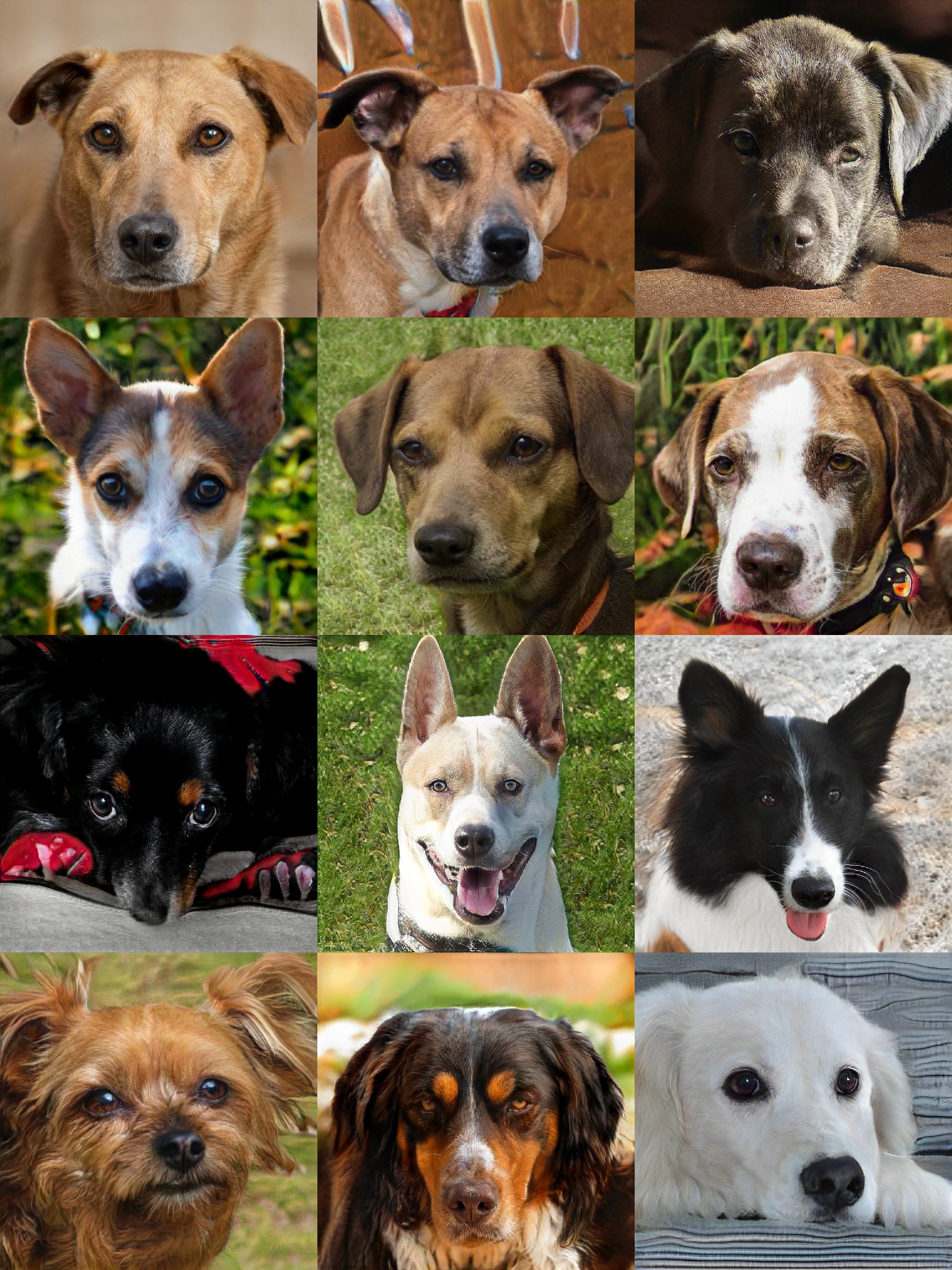}\hfill%
\includegraphics[width=\h]{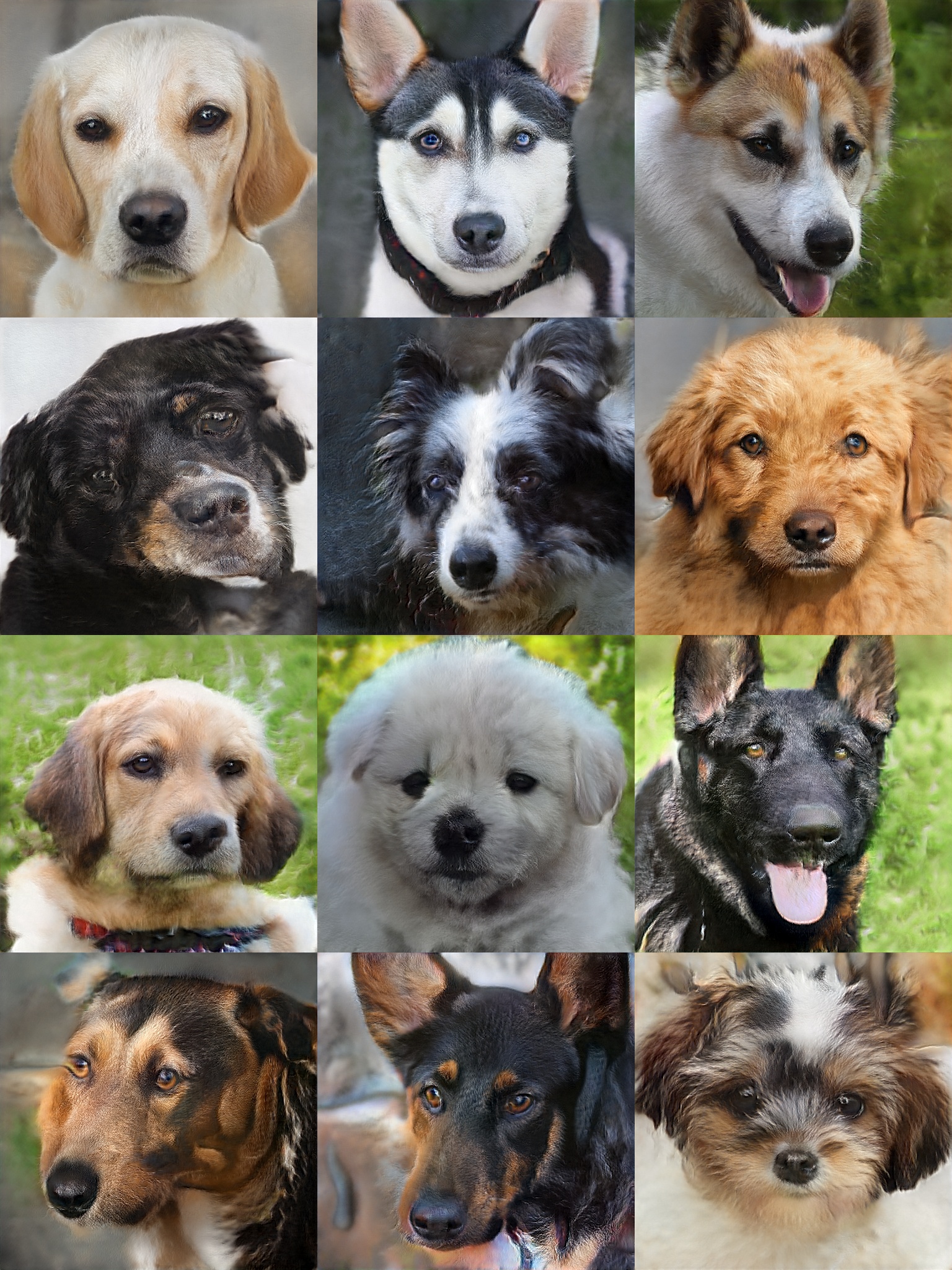}%
\vspace{-1.2mm}\\
\makebox[\h][c]{\scriptsize FID {\bf7.40} -- KID {\bf1.16}\raisebox{0.25mm}{\scalebox{0.7}{$\times10^3$}} -- Recall {\bf0.454}}\hfill%
\makebox[\h][c]{\scriptsize FID 19.37 -- KID 9.62\raisebox{0.25mm}{\scalebox{0.7}{$\times10^3$}} -- Recall 0.196}%
\vspace{2mm}%
\caption{
Uncurated 512$\times$512 results generated for \textsc{AFHQ Dog} \cite{AFHQ} (4739 images) with and without ADA, along with real images from the training set.
Both generators were trained from scratch.
We recommend zooming in to inspect the image quality in detail.
}
\label{#1}
\end{figure}
}

\newcommand{\figUncuratedWild}[1]{
\begin{figure}[p]
\footnotesize%
\renewcommand{\h}{0.493\linewidth}%
\makebox[\h][c]{\FINAL{ADA (Ours), truncated \raisebox{0.1mm}{\scalebox{0.9}{$(\psi=0.7)$}}}}\hfill%
\makebox[\h][c]{\FINAL{Real images from the training set}}%
\vspace{0.2mm}\\
\includegraphics[width=\h]{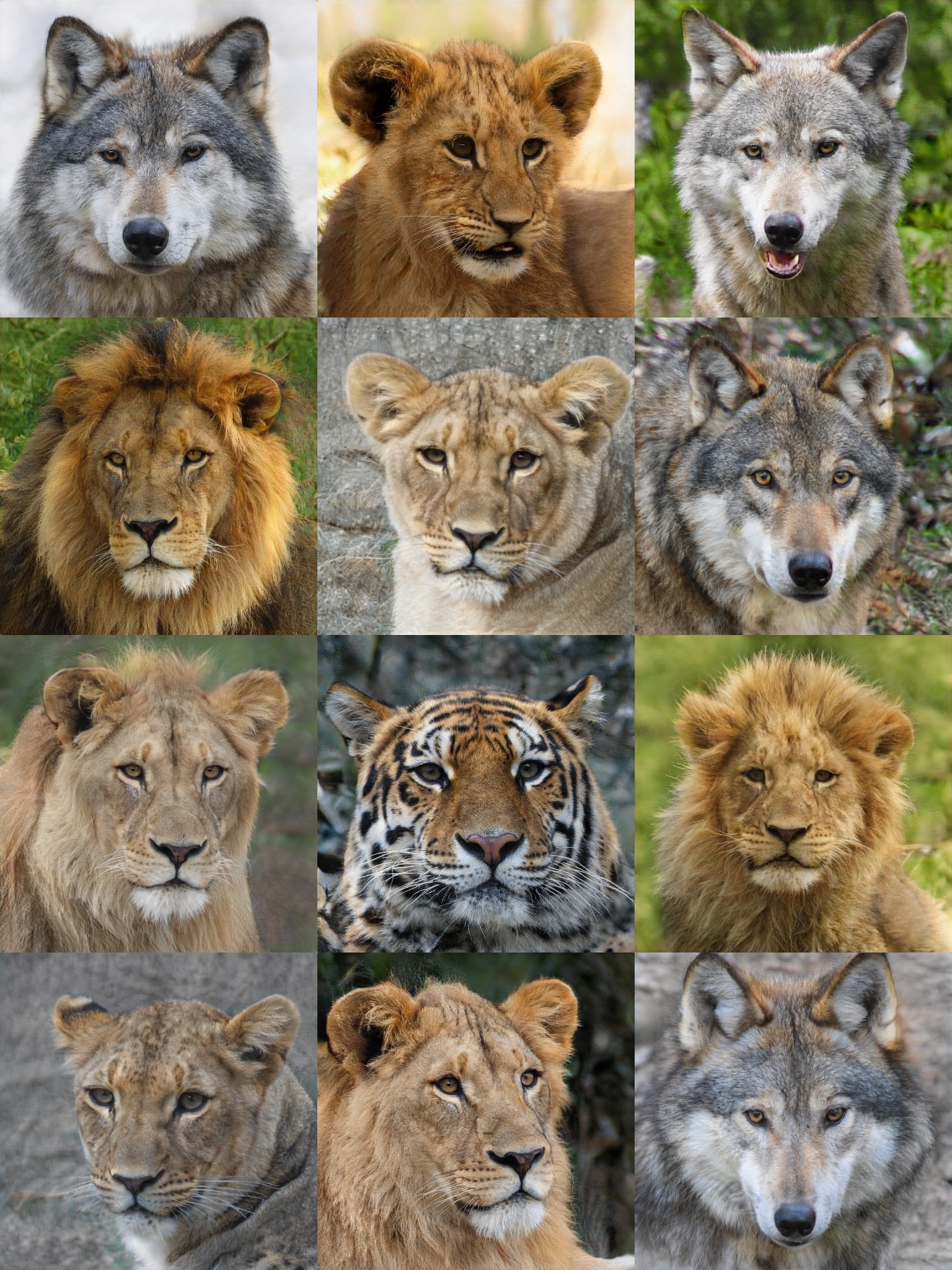}\hfill%
\includegraphics[width=\h]{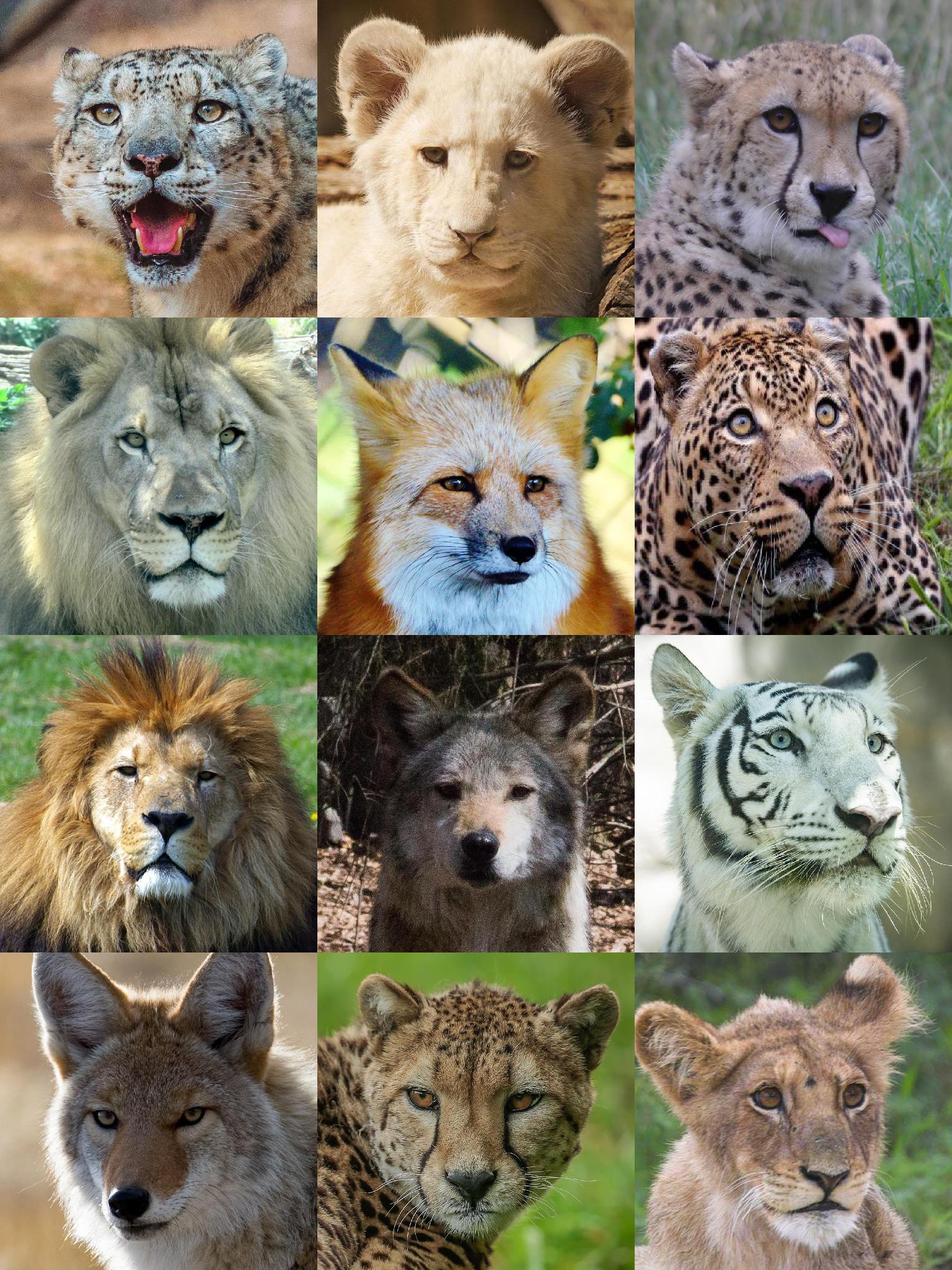}%
\vspace{2mm}\\
\makebox[\h][c]{\FINAL{ADA (Ours), untruncated}}\hfill%
\makebox[\h][c]{\FINAL{Original StyleGAN2 config \textsc{f}, untruncated}}%
\vspace{0.5mm}\\
\includegraphics[width=\h]{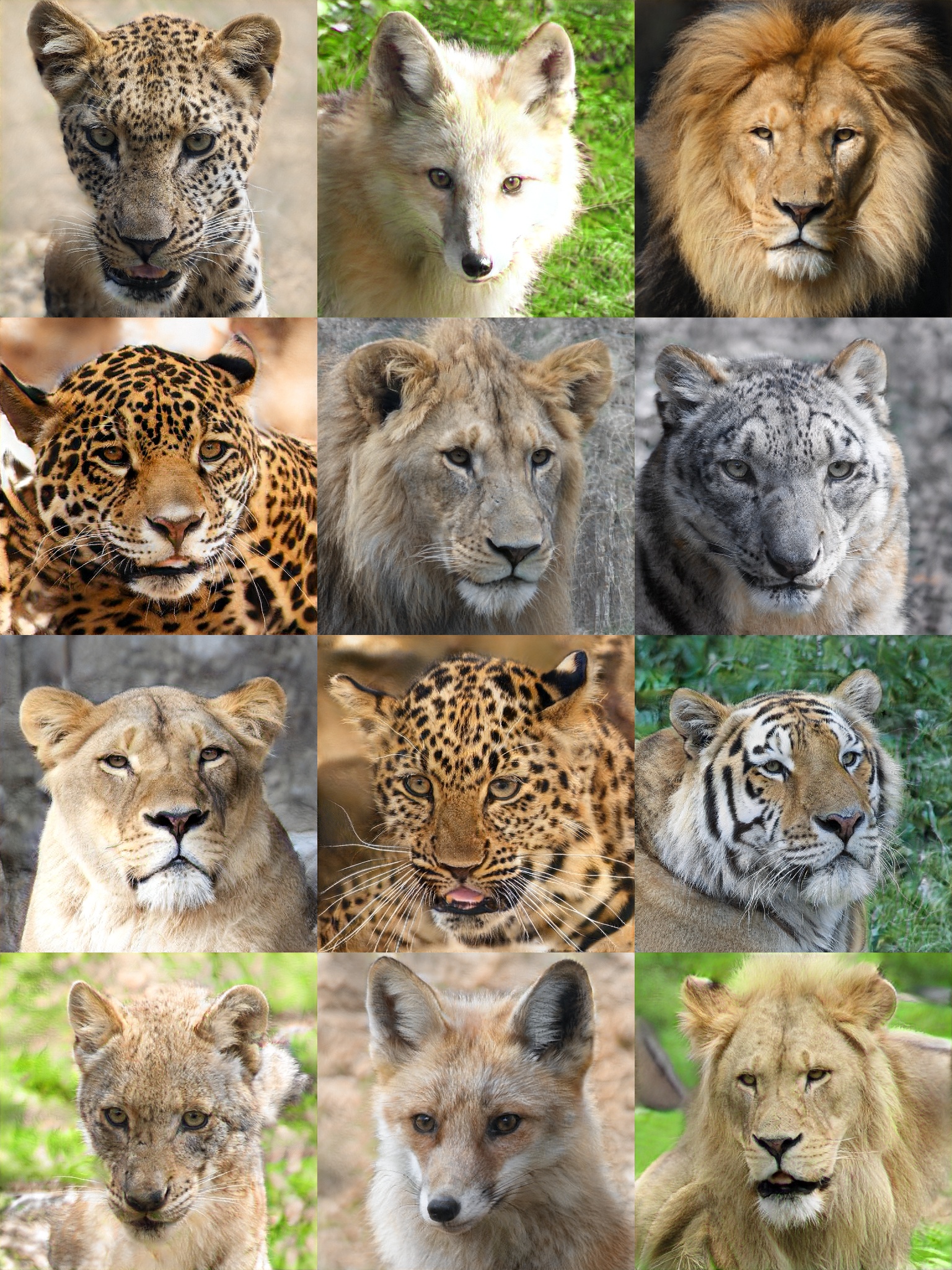}\hfill%
\includegraphics[width=\h]{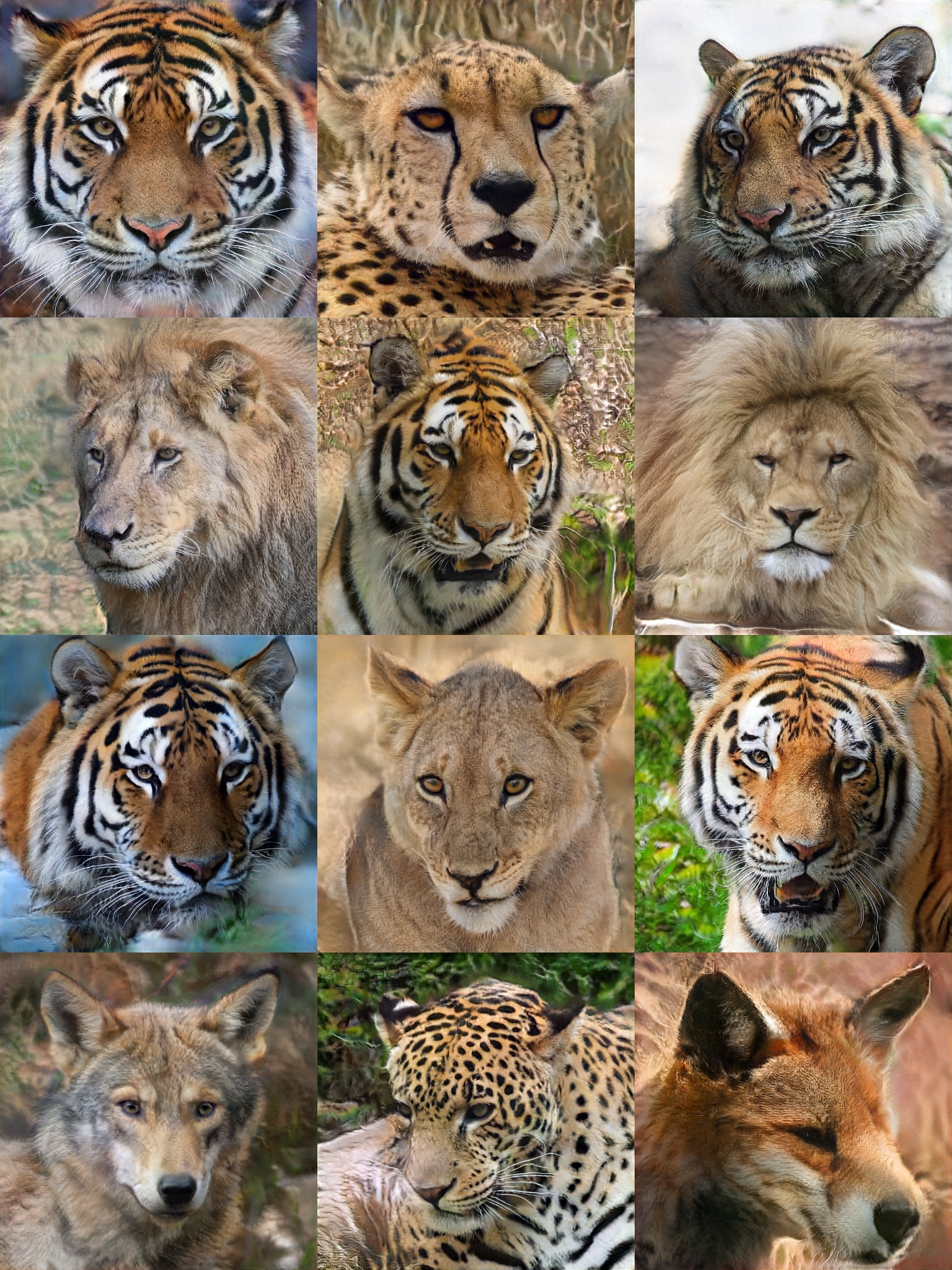}%
\vspace{-1.2mm}\\
\makebox[\h][c]{\scriptsize FID {\bf3.05} -- KID {\bf0.45}\raisebox{0.25mm}{\scalebox{0.7}{$\times10^3$}} -- Recall {\bf 0.147}}\hfill%
\makebox[\h][c]{\scriptsize FID 3.48 -- KID 0.77\raisebox{0.25mm}{\scalebox{0.7}{$\times10^3$}} -- Recall 0.143}%
\vspace{2mm}%
\caption{
\FINAL{%
Uncurated 512$\times$512 results generated for \textsc{AFHQ Wild} \cite{AFHQ} (4738 images) with and without ADA, along with real images from the training set.
Both generators were trained from scratch.
We recommend zooming in to inspect the image quality in detail.
}}
\label{#1}
\end{figure}
}

\newcommand{\figUncuratedCIFAR}[1]{
\begin{figure}[p]
\footnotesize%
\renewcommand{\h}{3.3mm}%
\renewcommand{\hh}{0.318\linewidth}%
\renewcommand{\vv}{\hh/\real{12}*\real{12}}%
\renewcommand{\vvv}{\hh/\real{12}*\real{3}}%
\renewcommand{\vvvv}{0.2mm}%
\makebox[0mm][l]{}\hspace{\h}%
\makebox[\hh][c]{\FINAL{Generator with best FID}}\hfill%
\makebox[\hh][c]{Real images}\hfill%
\makebox[\hh][c]{\FINAL{Generator with best IS}}\\
\makebox[0mm][l]{\rotatebox{90}{\makebox[\vv][c]{Unconditional}}}\hspace{\h}%
\includegraphics[width=\hh]{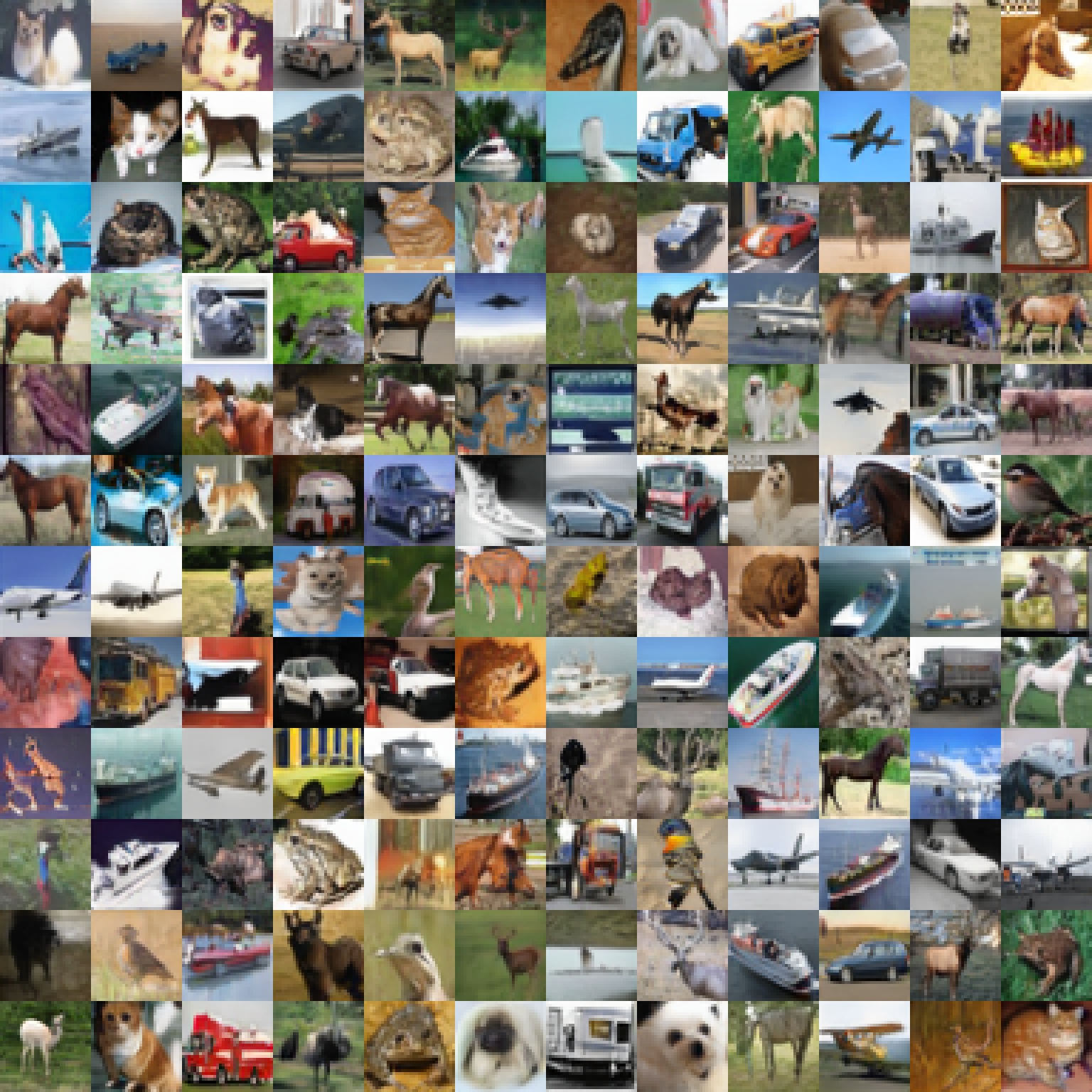}\hfill%
\includegraphics[width=\hh]{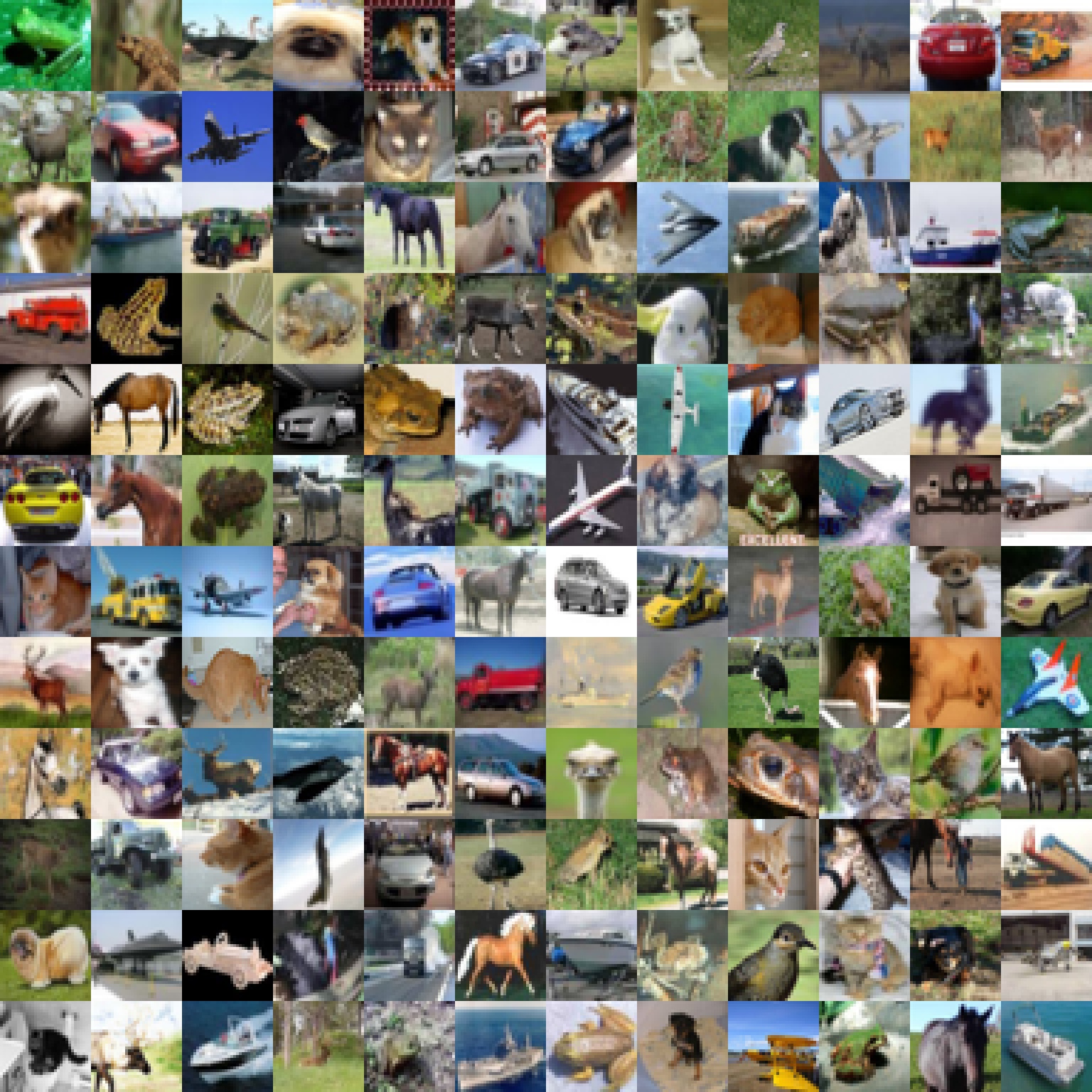}\hfill%
\includegraphics[width=\hh]{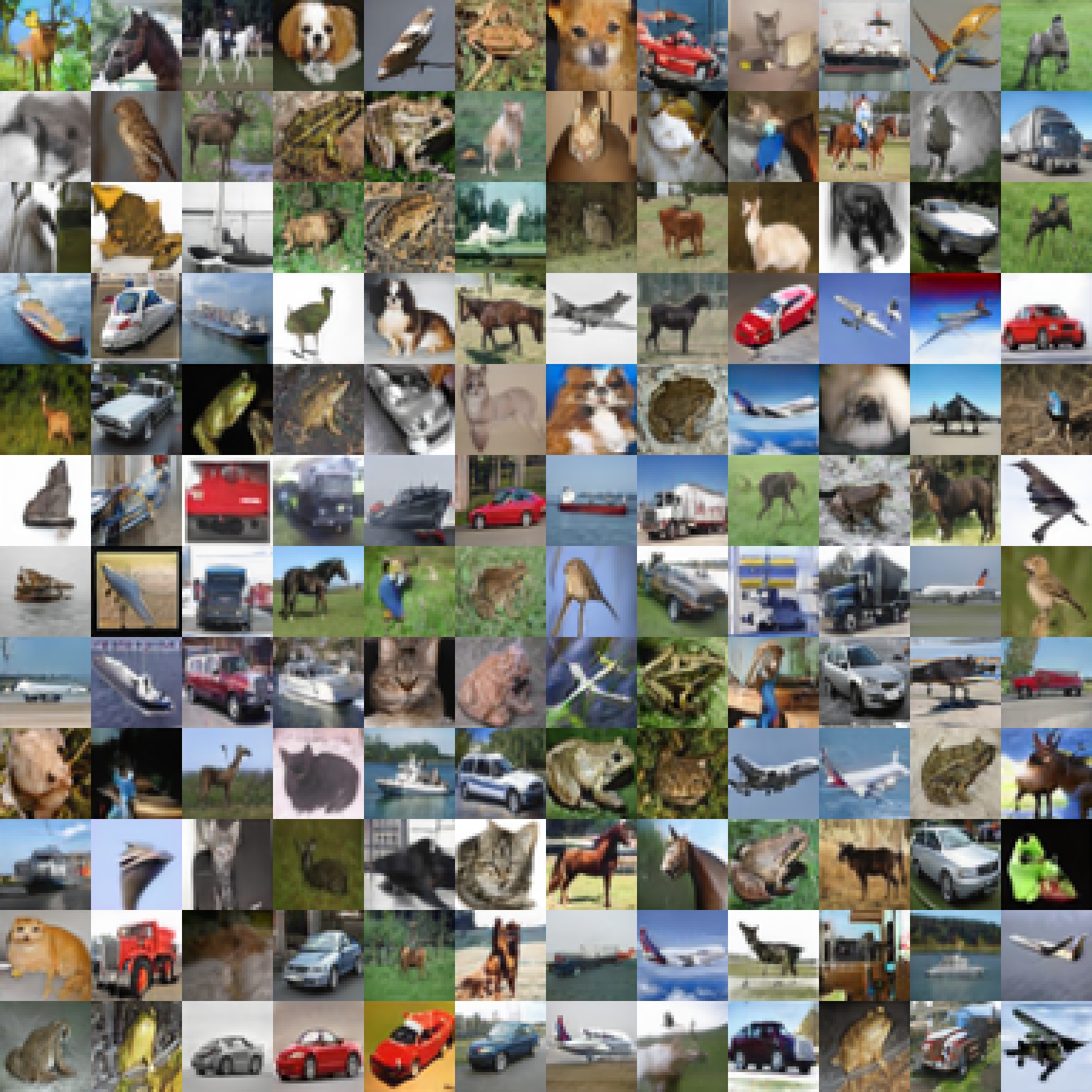}\vspace{-1.2mm}\\
\makebox[0mm][l]{}\hspace{\h}%
\makebox[\hh][c]{\scriptsize \FINAL{FID {\bf2.85} -- IS 9.74}}\hfill%
\makebox[\hh][c]{\scriptsize IS 11.24}\hfill%
\makebox[\hh][c]{\scriptsize \FINAL{FID 5.70 -- IS {\bf10.08}}}\vspace{1.4mm}\\
\makebox[0mm][l]{\rotatebox{90}{\makebox[\vvv][c]{Plane}}}\hspace{\h}%
\includegraphics[width=\hh]{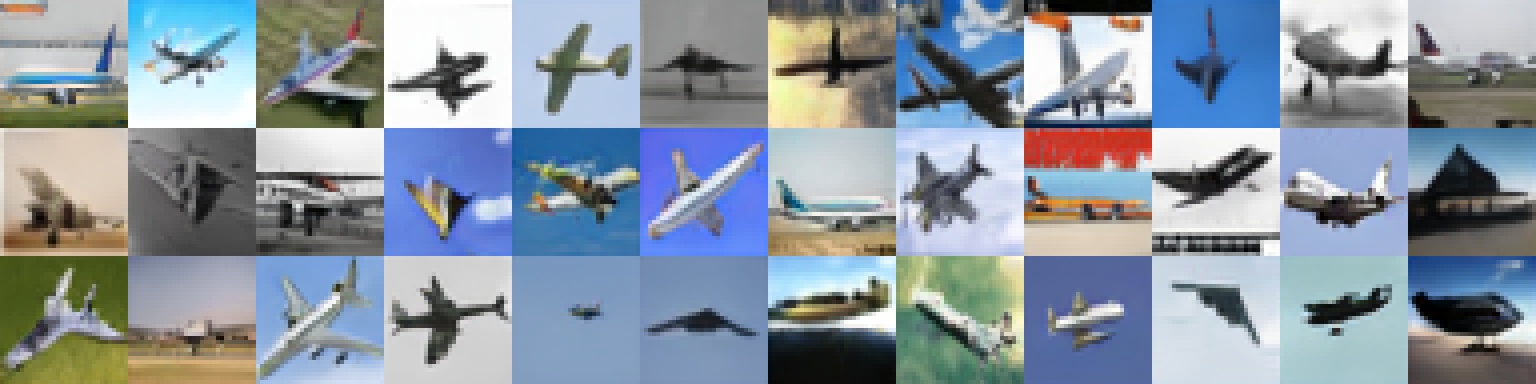}\hfill%
\includegraphics[width=\hh]{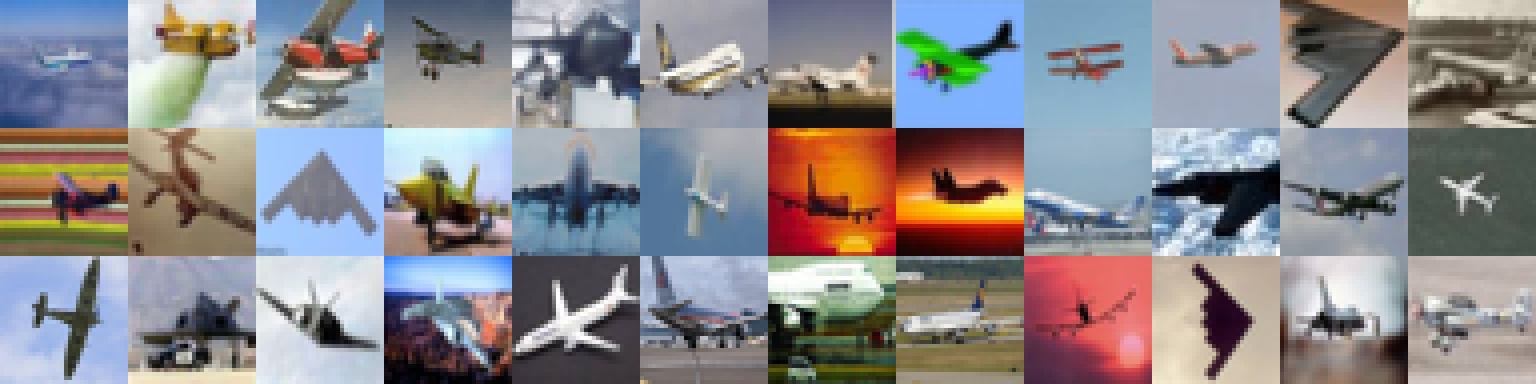}\hfill%
\includegraphics[width=\hh]{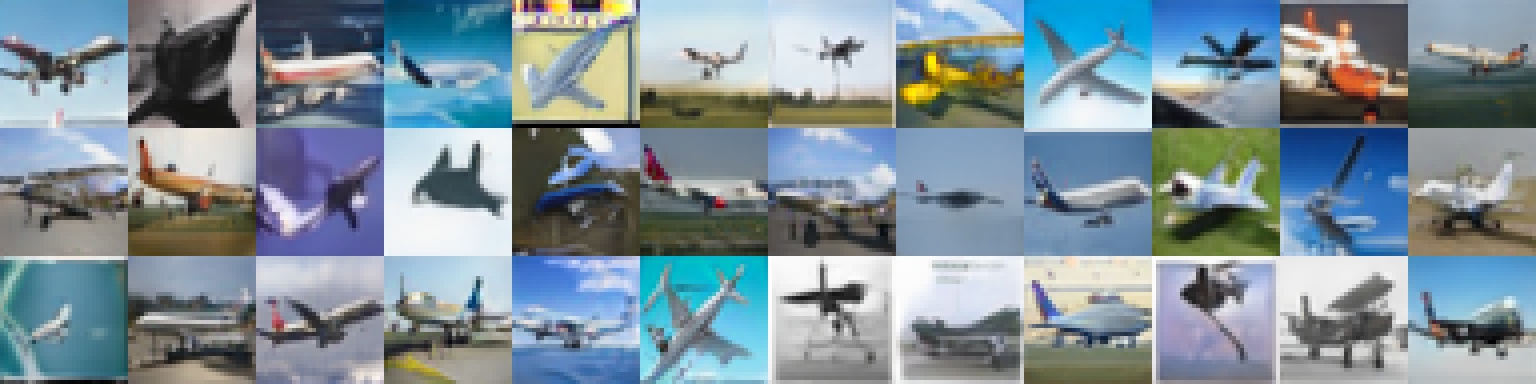}\vspace{\vvvv}\\
\makebox[0mm][l]{\rotatebox{90}{\makebox[\vvv][c]{Car}}}\hspace{\h}%
\includegraphics[width=\hh]{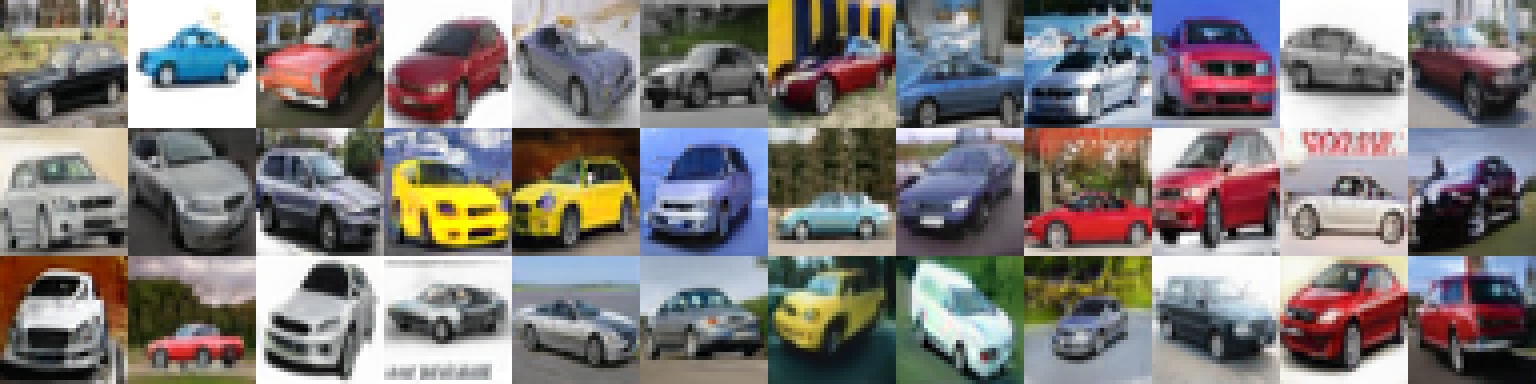}\hfill%
\includegraphics[width=\hh]{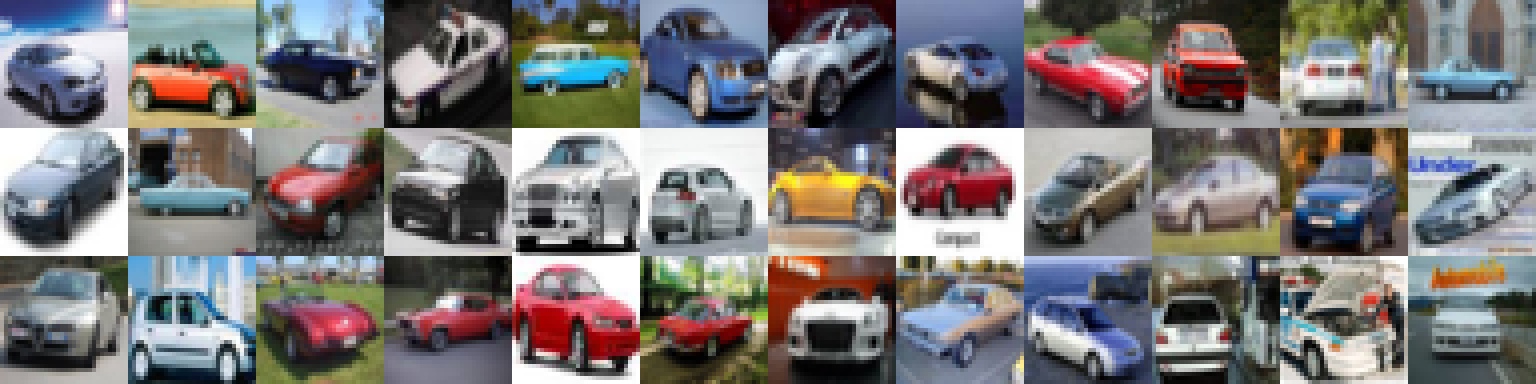}\hfill%
\includegraphics[width=\hh]{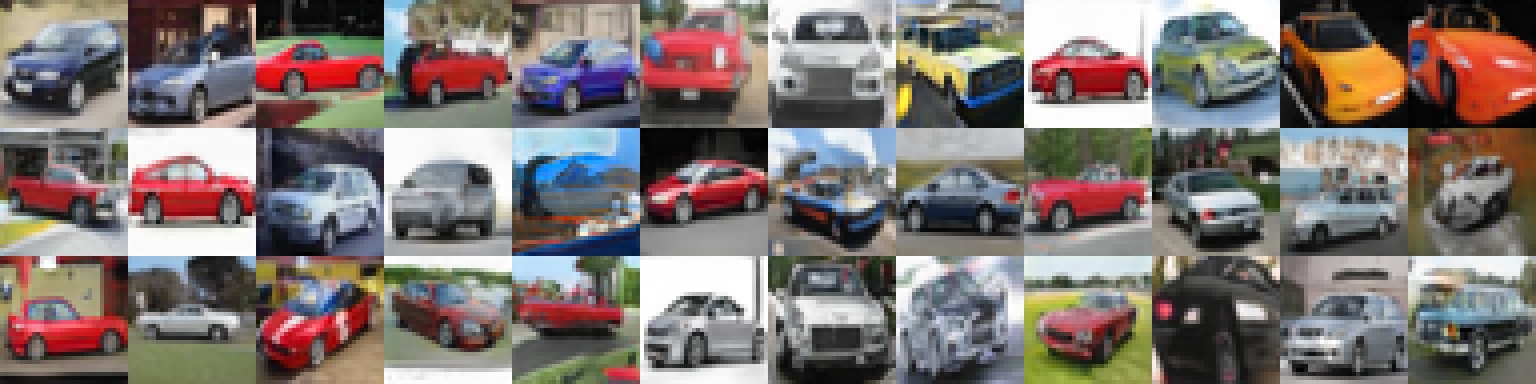}\vspace{\vvvv}\\
\makebox[0mm][l]{\rotatebox{90}{\makebox[\vvv][c]{Bird}}}\hspace{\h}%
\includegraphics[width=\hh]{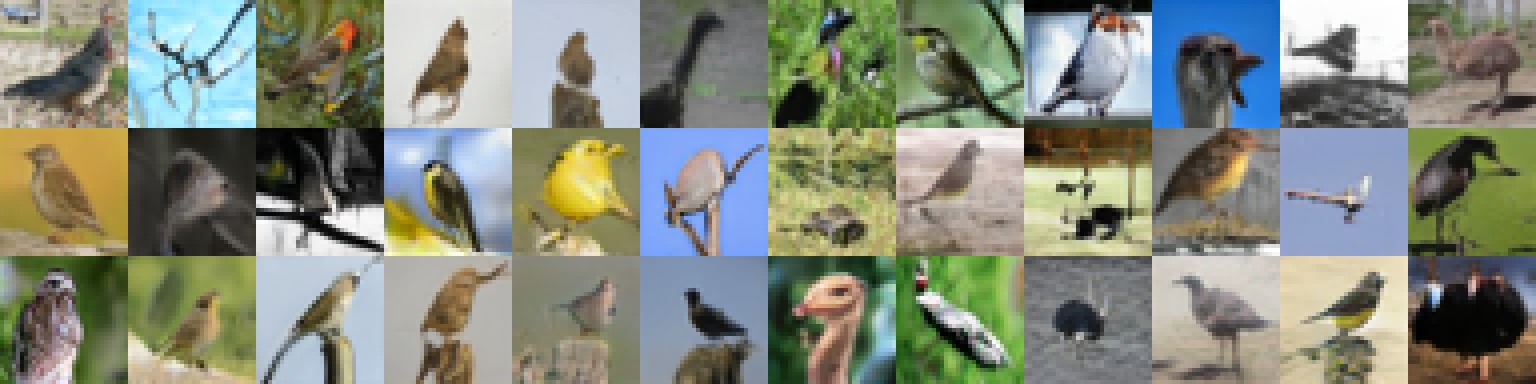}\hfill%
\includegraphics[width=\hh]{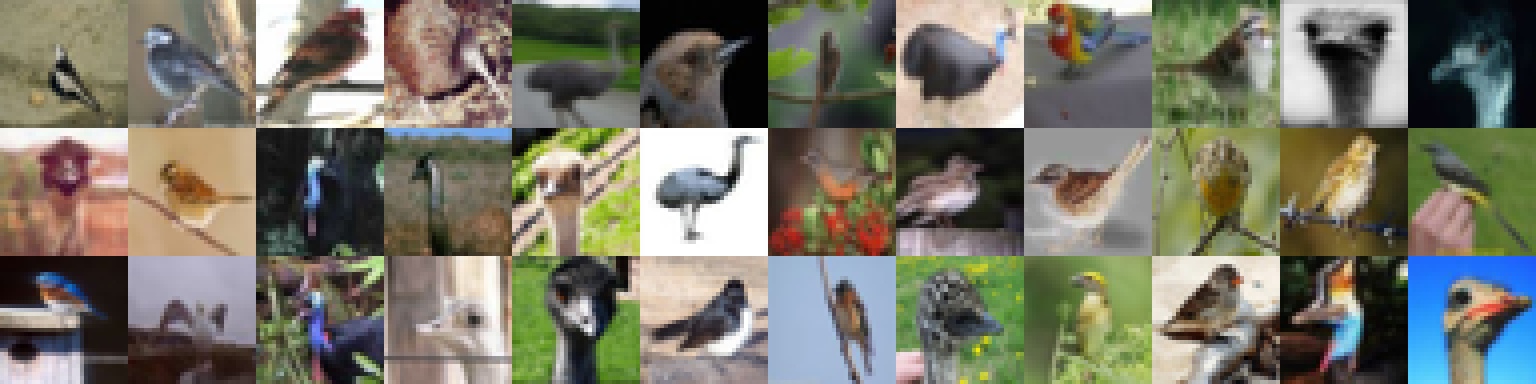}\hfill%
\includegraphics[width=\hh]{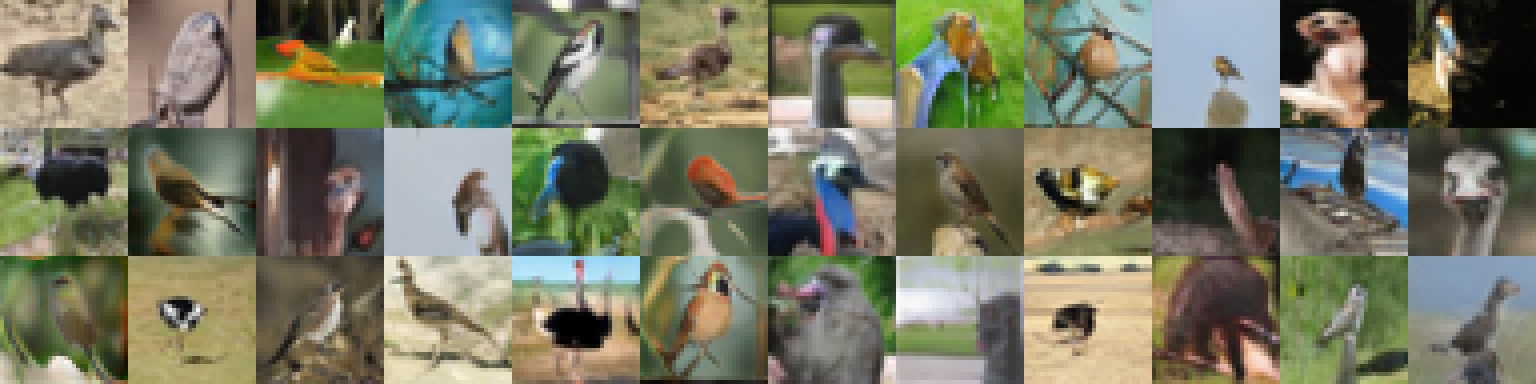}\vspace{\vvvv}\\
\makebox[0mm][l]{\rotatebox{90}{\makebox[\vvv][c]{Cat}}}\hspace{\h}%
\includegraphics[width=\hh]{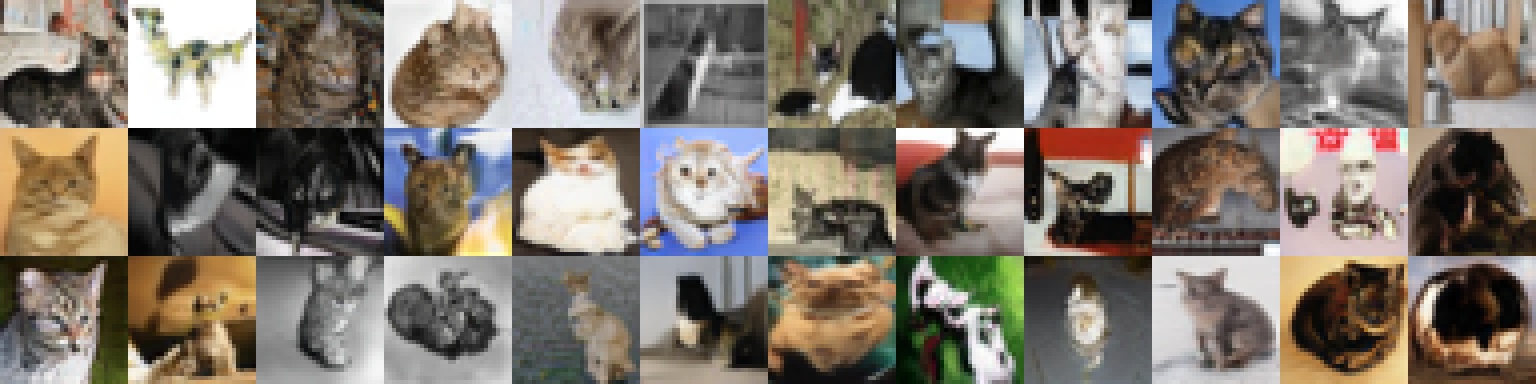}\hfill%
\includegraphics[width=\hh]{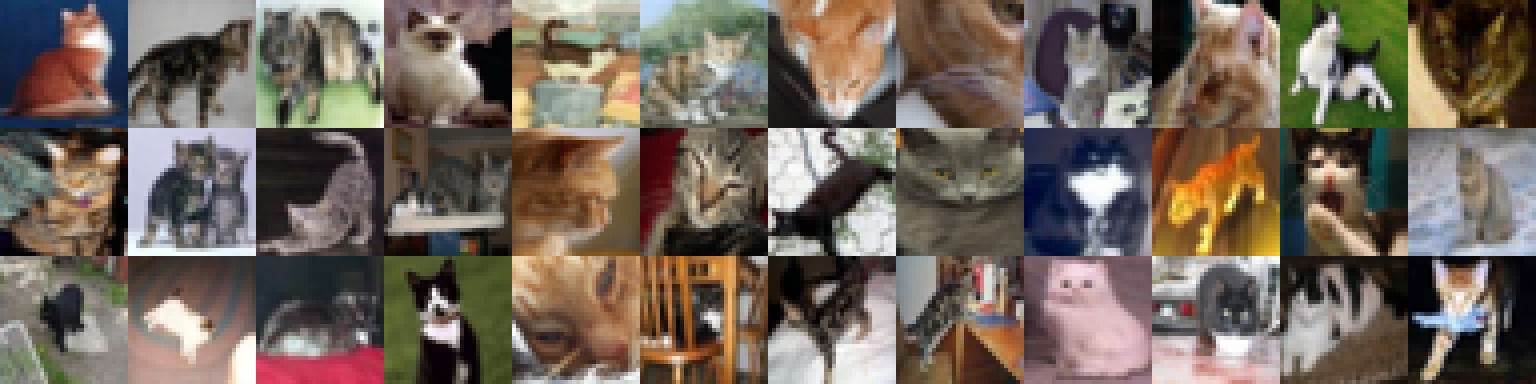}\hfill%
\includegraphics[width=\hh]{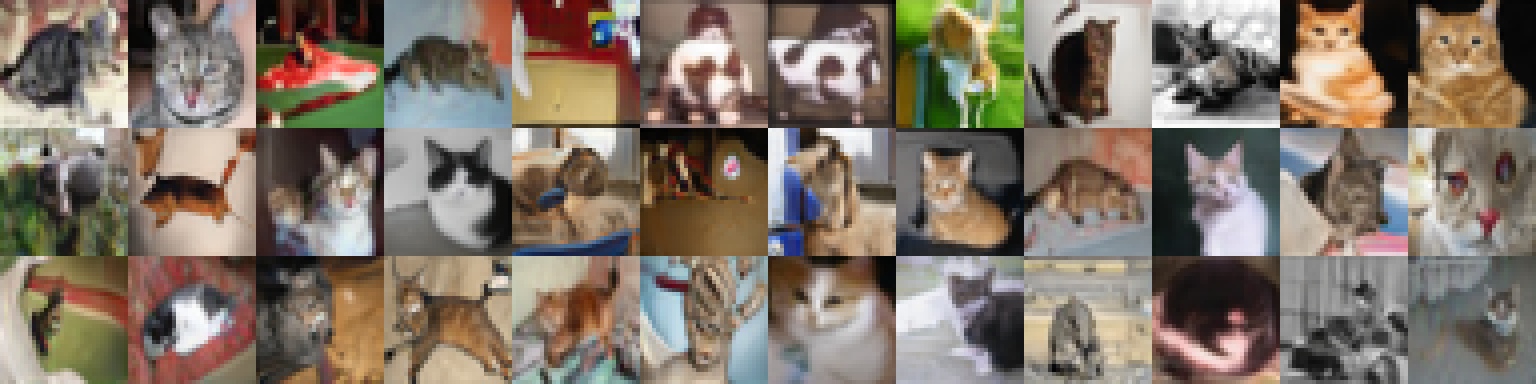}\vspace{\vvvv}\\
\makebox[0mm][l]{\rotatebox{90}{\makebox[\vvv][c]{Deer}}}\hspace{\h}%
\includegraphics[width=\hh]{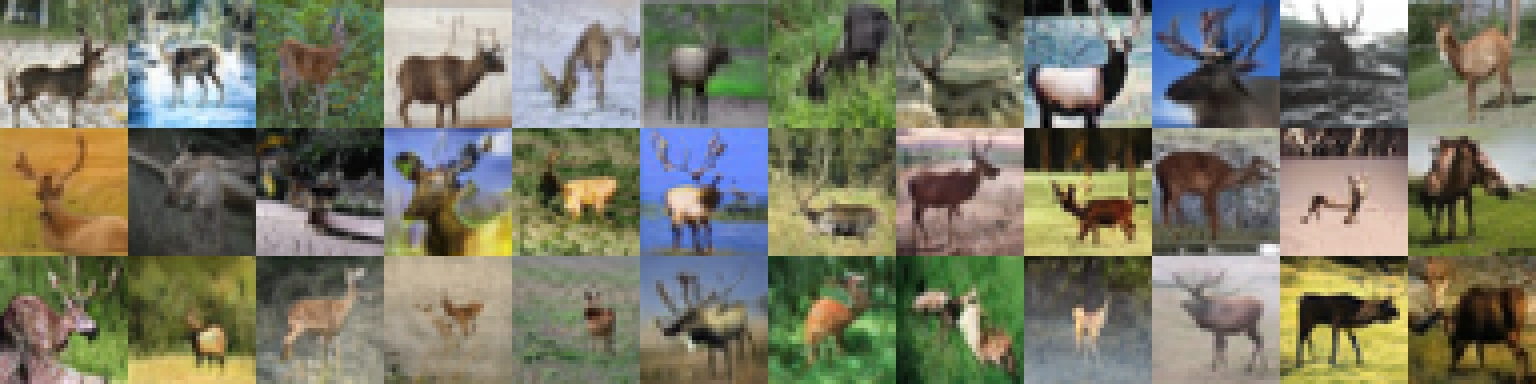}\hfill%
\includegraphics[width=\hh]{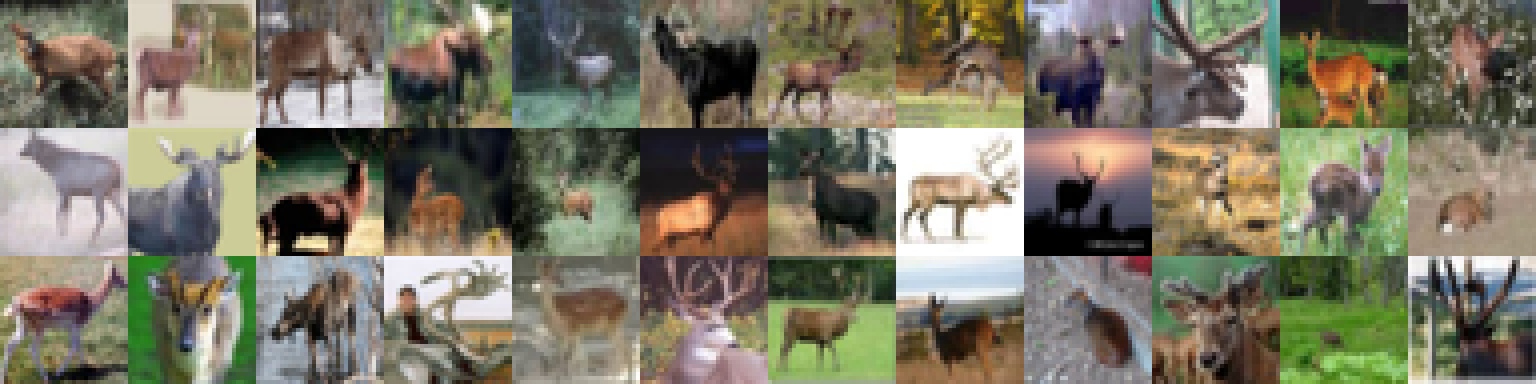}\hfill%
\includegraphics[width=\hh]{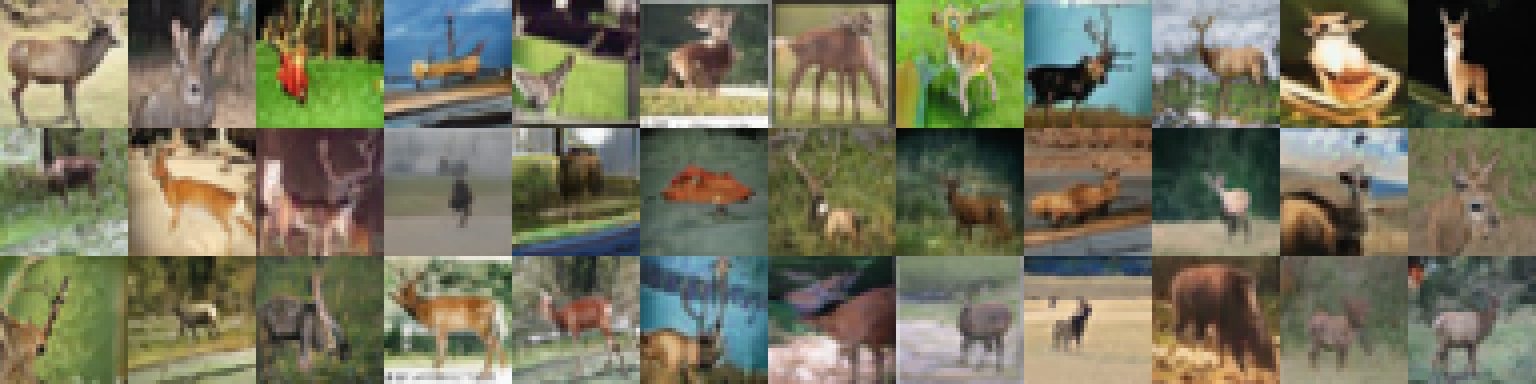}\vspace{\vvvv}\\
\makebox[0mm][l]{\rotatebox{90}{\makebox[\vvv][c]{Dog}}}\hspace{\h}%
\includegraphics[width=\hh]{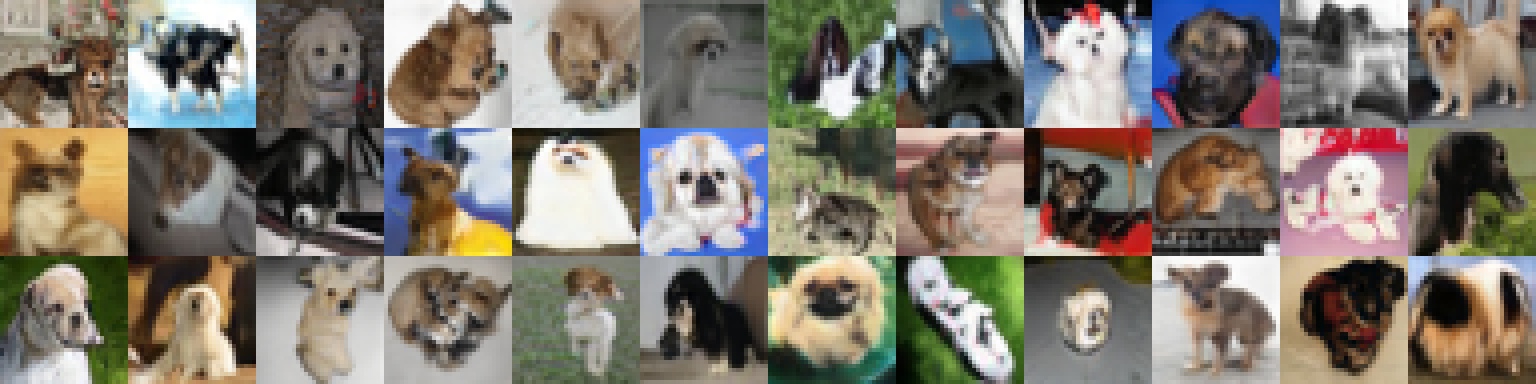}\hfill%
\includegraphics[width=\hh]{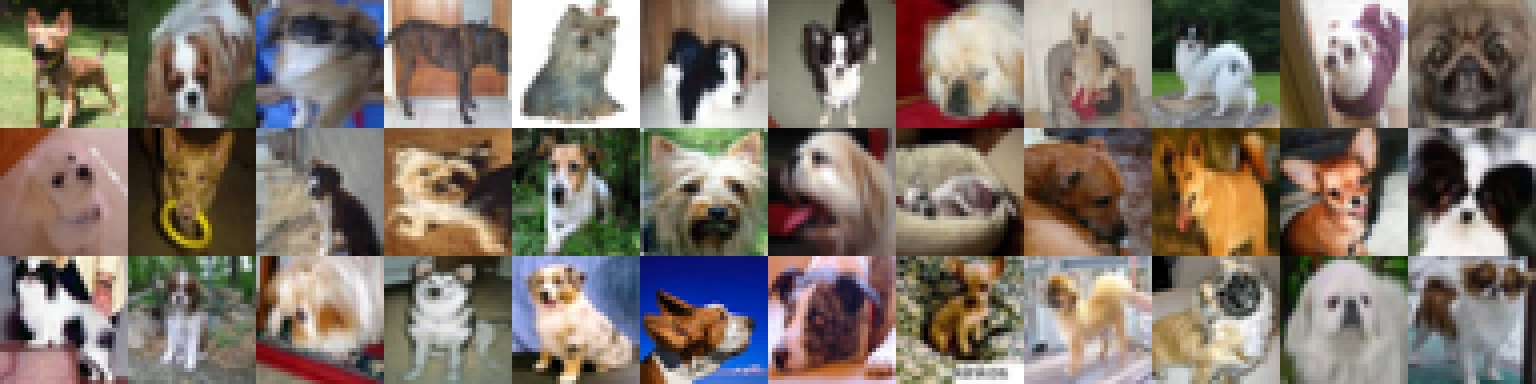}\hfill%
\includegraphics[width=\hh]{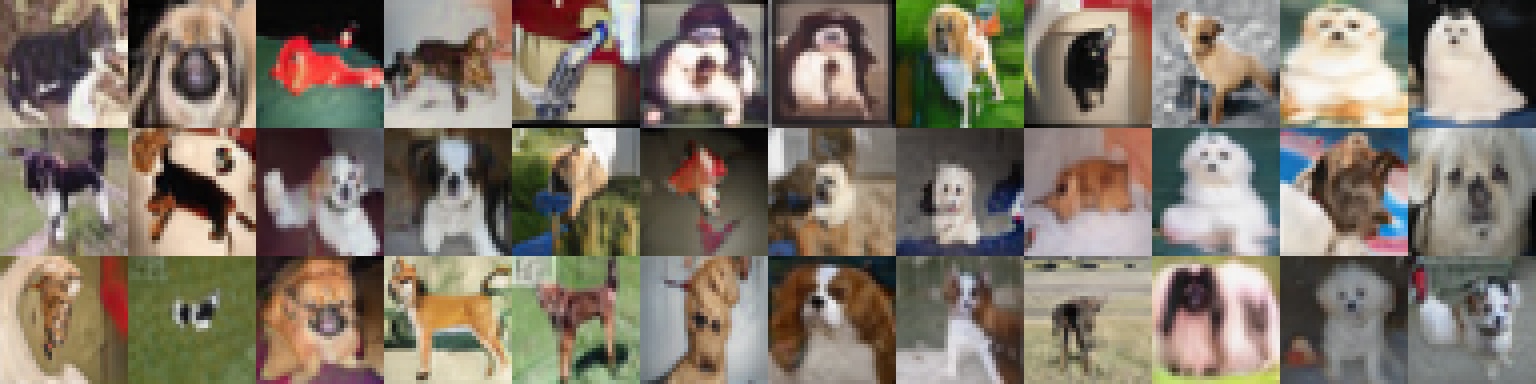}\vspace{\vvvv}\\
\makebox[0mm][l]{\rotatebox{90}{\makebox[\vvv][c]{Frog}}}\hspace{\h}%
\includegraphics[width=\hh]{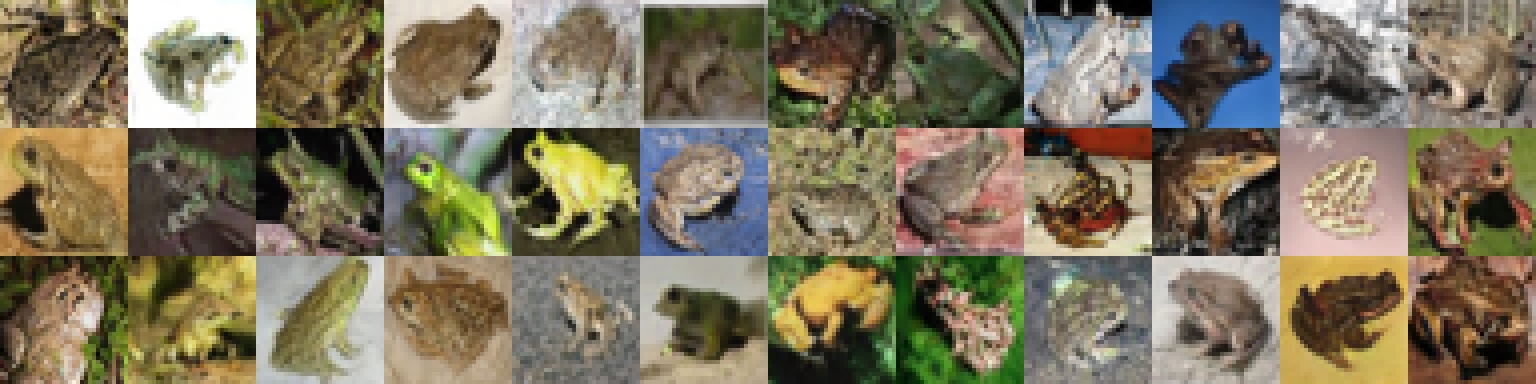}\hfill%
\includegraphics[width=\hh]{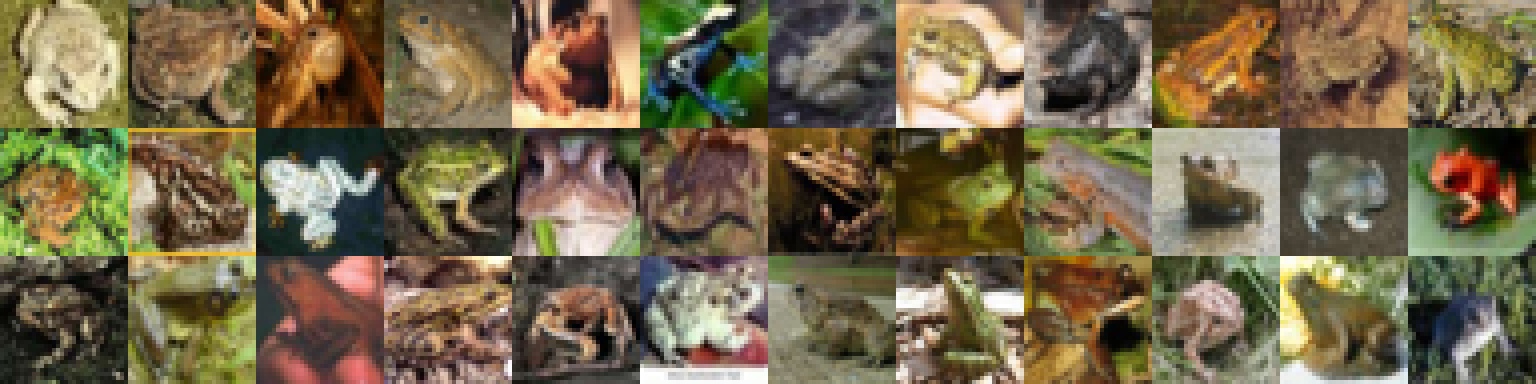}\hfill%
\includegraphics[width=\hh]{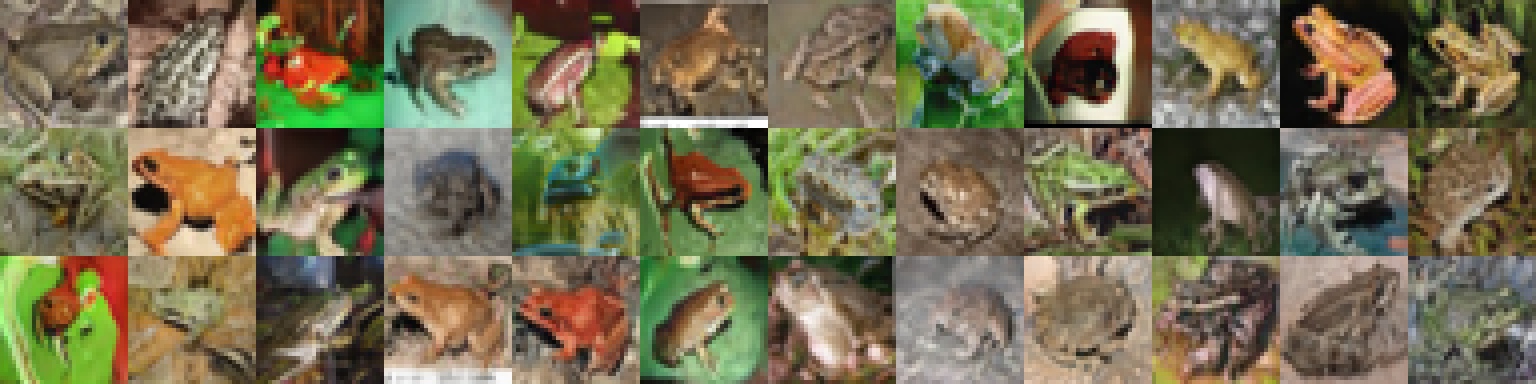}\vspace{\vvvv}\\
\makebox[0mm][l]{\rotatebox{90}{\makebox[\vvv][c]{Horse}}}\hspace{\h}%
\includegraphics[width=\hh]{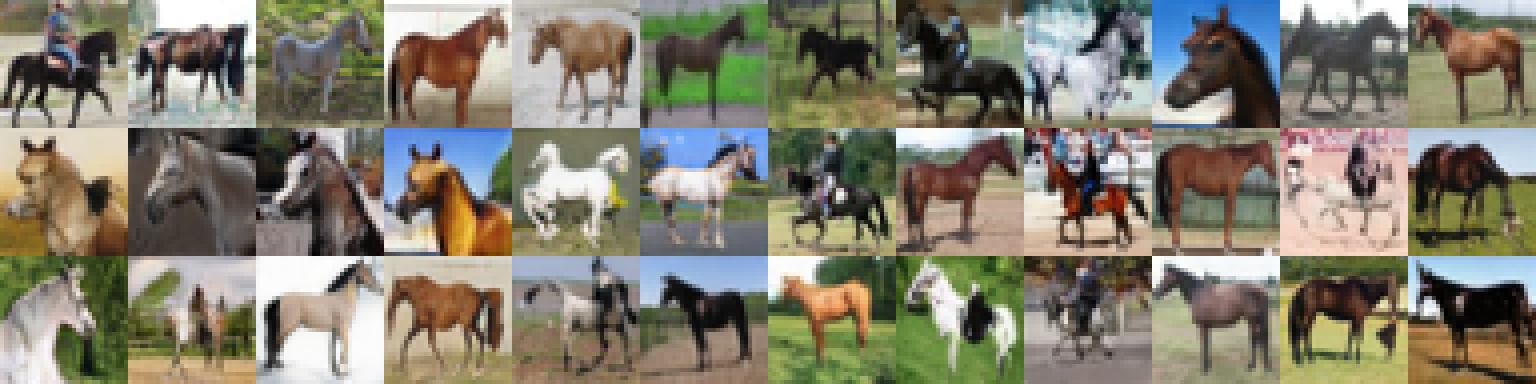}\hfill%
\includegraphics[width=\hh]{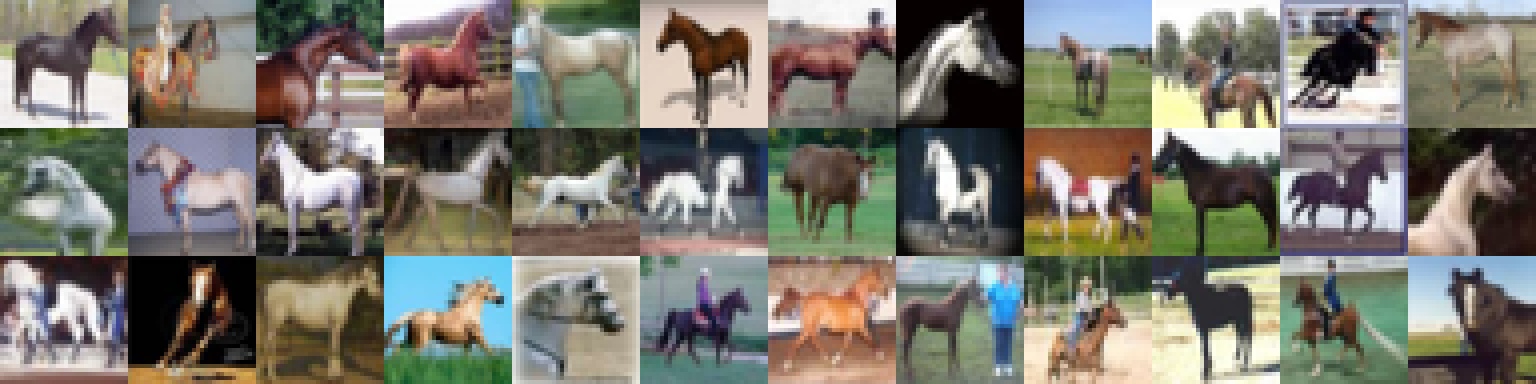}\hfill%
\includegraphics[width=\hh]{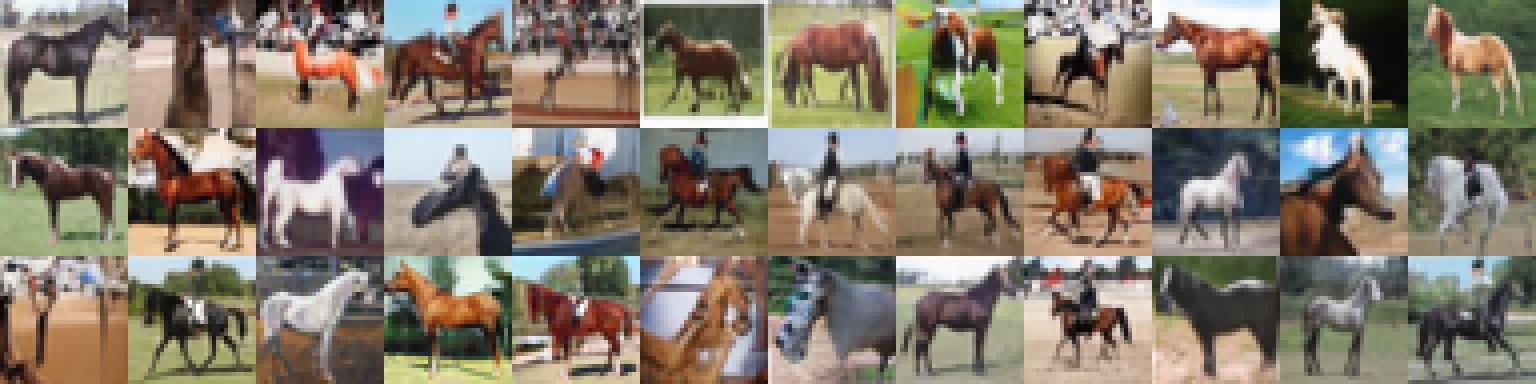}\vspace{\vvvv}\\
\makebox[0mm][l]{\rotatebox{90}{\makebox[\vvv][c]{Ship}}}\hspace{\h}%
\includegraphics[width=\hh]{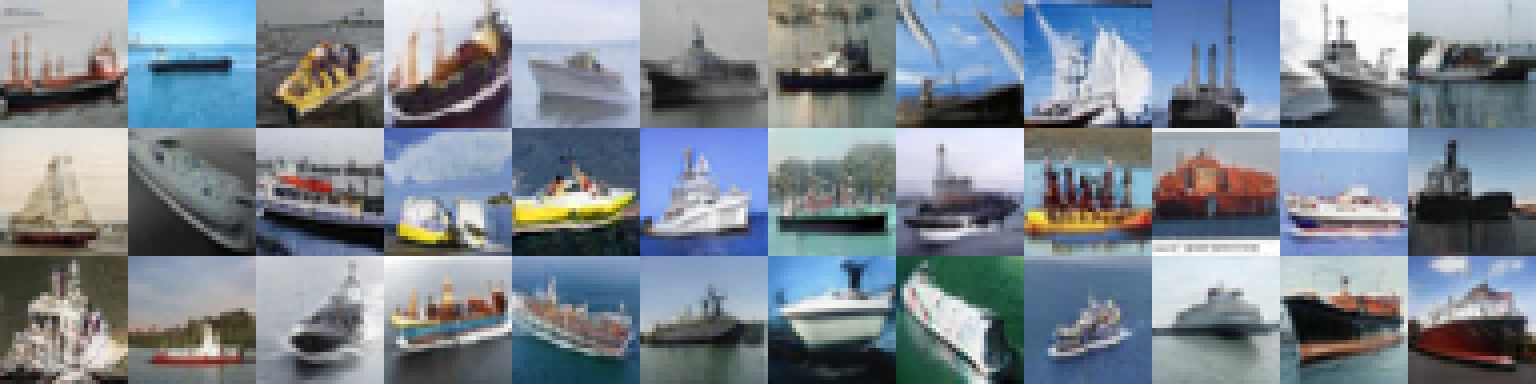}\hfill%
\includegraphics[width=\hh]{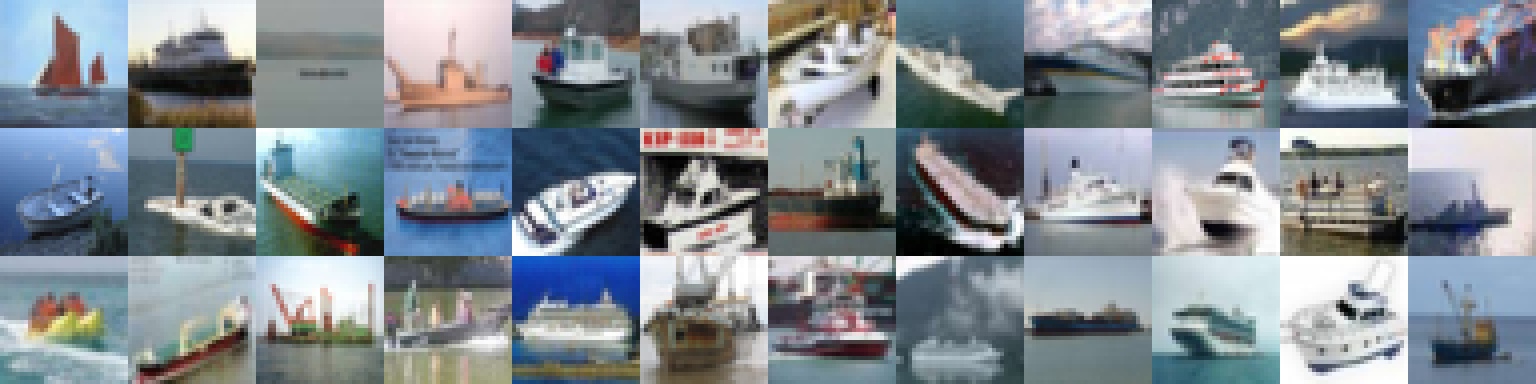}\hfill%
\includegraphics[width=\hh]{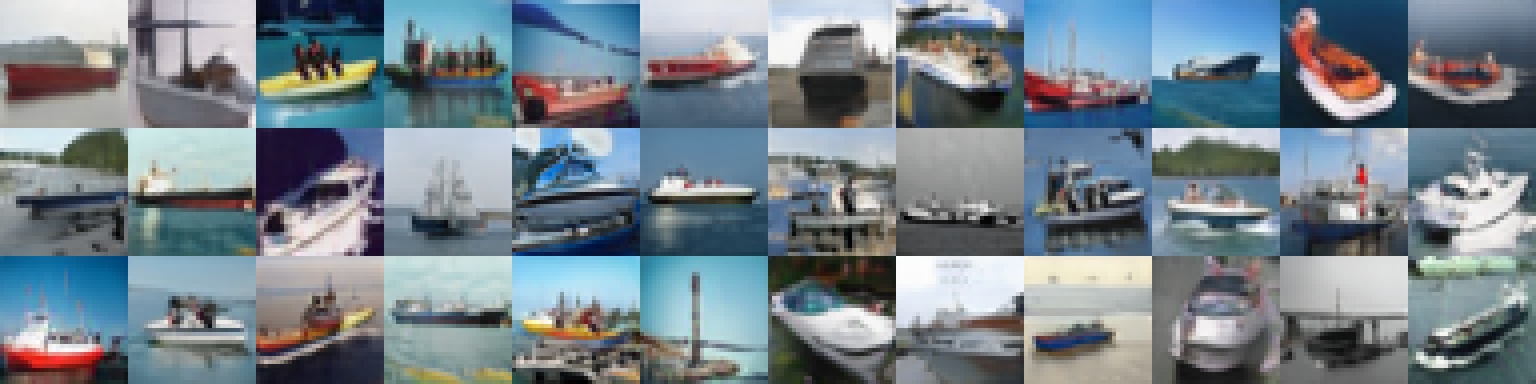}\vspace{\vvvv}\\
\makebox[0mm][l]{\rotatebox{90}{\makebox[\vvv][c]{Truck}}}\hspace{\h}%
\includegraphics[width=\hh]{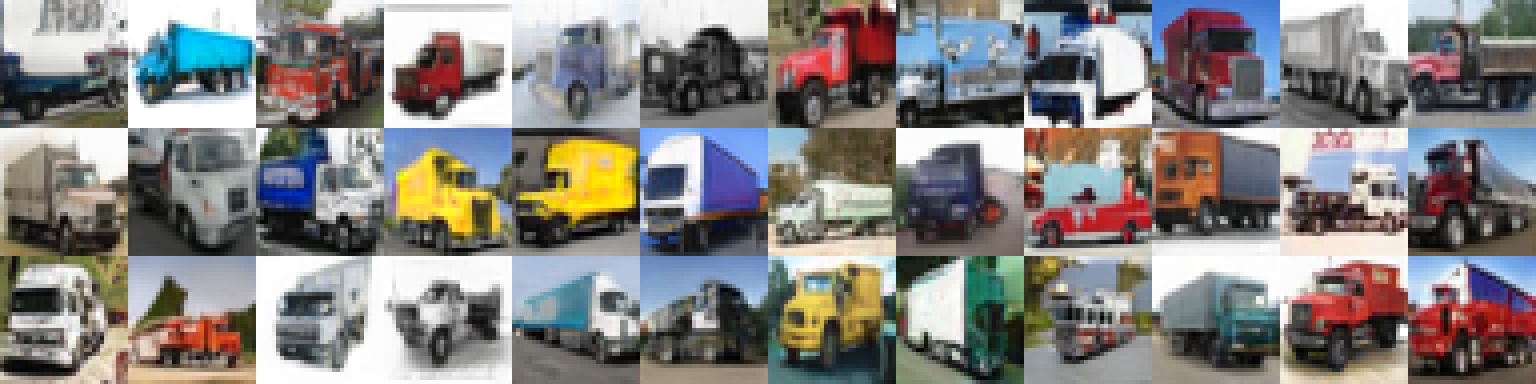}\hfill%
\includegraphics[width=\hh]{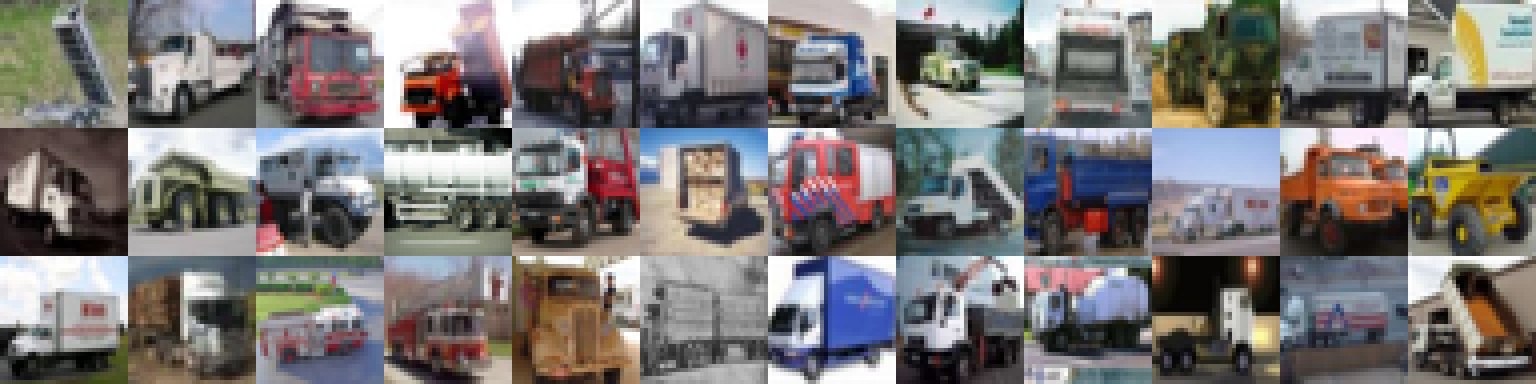}\hfill%
\includegraphics[width=\hh]{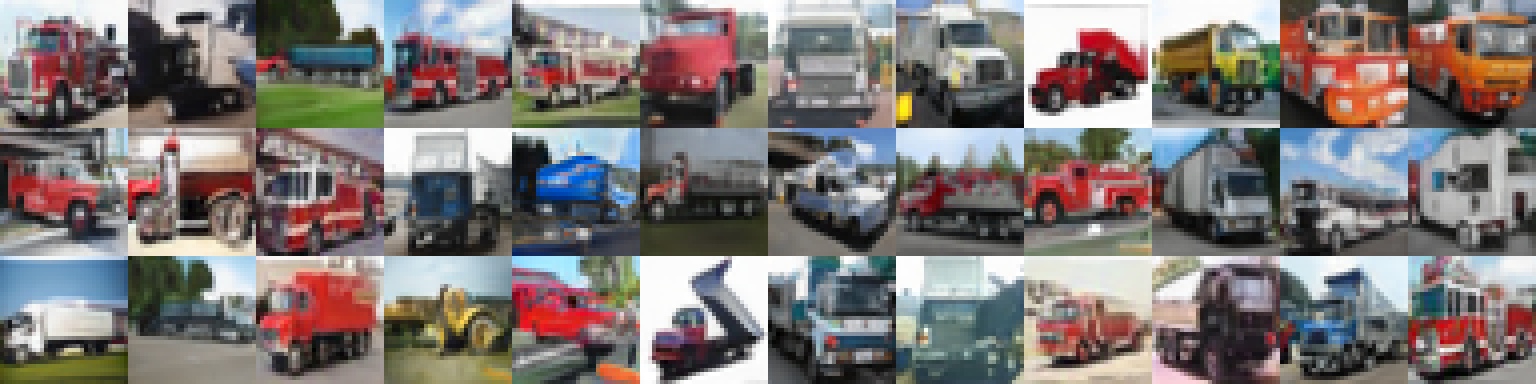}\vspace{-1.2mm}\\
\makebox[0mm][l]{}\hspace{\h}%
\makebox[\hh][c]{\scriptsize \FINAL{FID {\bf2.38} -- IS 10.00}}\hfill%
\makebox[\hh][c]{\scriptsize IS 11.24}\hfill%
\makebox[\hh][c]{\scriptsize \FINAL{FID 3.62 -- IS {\bf10.33}}}%
\caption{
Generated and real images for \textsc{CIFAR-10} in the unconditional setting (top) and each class in the conditional setting (bottom).
We show the results for the best generators trained in the context of \FINAL{Figure~\refpaper{fig:SmallDatasetResults}b}, selected according to either FID or IS.
The numbers refer to the single best model and are therefore slightly better than the averages quoted in the result table.
\FINAL{
It can be seen that the model with the lowest FID produces images with a wider variation in coloring and poses compared to the model with highest IS. 
}
This is in line with the common approximation (e.g.,~\cite{Brock2018}) that FID roughly corresponds to Recall and IS to Precision, two independent aspects of result quality \cite{Sajjadi2018,Tuomas2019}.
}
\label{#1}
\end{figure}
}

\renewcommand{\h}{0.23\linewidth}
\renewcommand{\hh}{3.4mm}
\renewcommand{\vv}{-1.5mm}
\renewcommand{\vvv}{1.4mm}
\newcommand{\colA}[1]{} %
\newcommand{\colB}[1]{\hspace{0.3mm}#1} %
\newcommand{\colC}[1]{\hfill#1} %
\newcommand{\colD}[1]{} %
\newcommand{\colE}[1]{} %
\newcommand{\colF}[1]{\hfill#1} %
\newcommand{\colG}[1]{} %
\newcommand{\colH}[1]{} %
\newcommand{\colI}[1]{\hfill#1} %

\newcommand{\figSubsetImagesFFHQ}[1]{
\begin{figure}[p]
\makebox[0mm][l]{}\hspace{\hh}%
\colA{\makebox[\h][c]{1k training set}}%
\colB{\makebox[\h][c]{2k training set}}%
\colC{\makebox[\h][c]{5k training set}}%
\colD{\makebox[\h][c]{10k training set}}%
\colE{\makebox[\h][c]{20k training set}}%
\colF{\makebox[\h][c]{30k training set}}%
\colG{\makebox[\h][c]{50k training set}}%
\colH{\makebox[\h][c]{70k training set}}%
\colI{\makebox[\h][c]{140k training set}}%
\\
\makebox[0mm][l]{\rotatebox{90}{\makebox[\h][c]{BigGAN}}}\hspace{\hh}%
\colA{\includegraphics[width=\h]{generated_images/appendix/subset_sweep_ffhq_biggan/2x2_sub1k_seed0.\ext}}%
\colB{\includegraphics[width=\h]{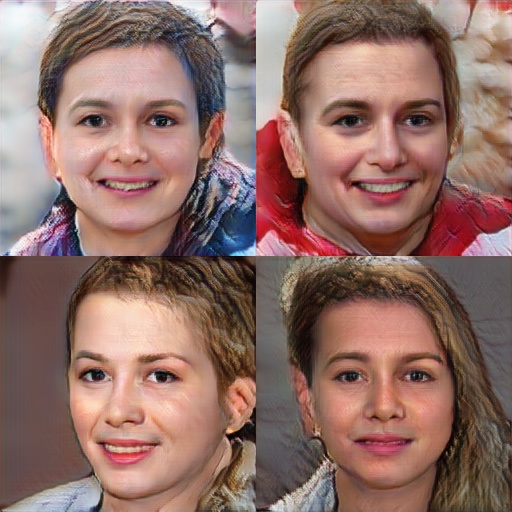}}%
\colC{\includegraphics[width=\h]{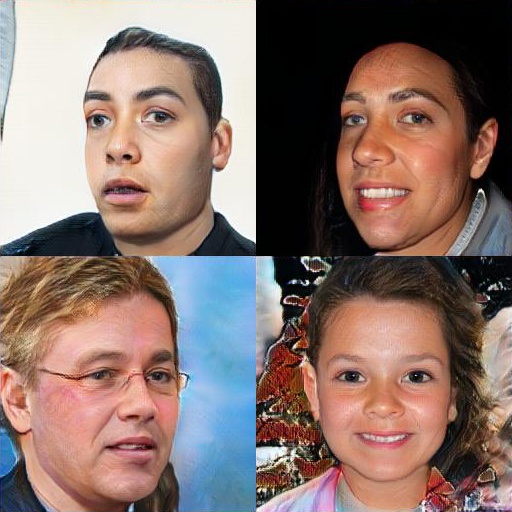}}%
\colD{\includegraphics[width=\h]{generated_images/appendix/subset_sweep_ffhq_biggan/2x2_sub10k_seed8.\ext}}%
\colE{\includegraphics[width=\h]{generated_images/appendix/subset_sweep_ffhq_biggan/2x2_sub20k_seed16.\ext}}%
\colF{\includegraphics[width=\h]{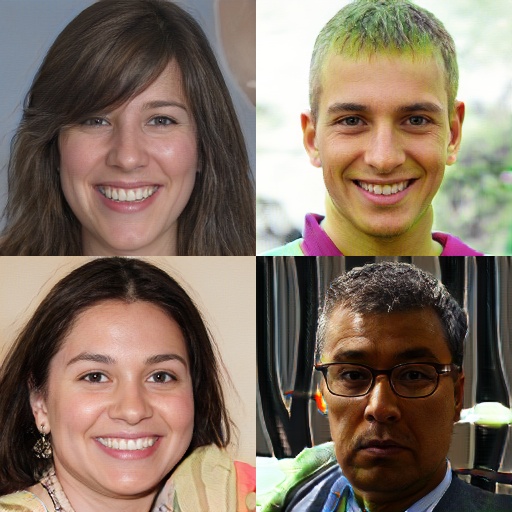}}%
\colG{\includegraphics[width=\h]{generated_images/appendix/subset_sweep_ffhq_biggan/2x2_sub50k_seed8.\ext}}%
\colH{\includegraphics[width=\h]{generated_images/appendix/subset_sweep_ffhq_biggan/2x2_sub70k_seed8.\ext}}%
\colI{\includegraphics[width=\h]{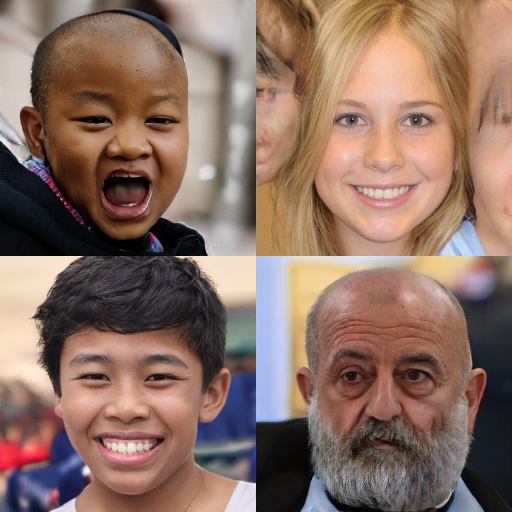}}%
\vspace{\vv}\\
\makebox[0mm][l]{}\hspace{\hh}%
\colA{\makebox[\h][c]{\scriptsize FID 85.03}}%
\colB{\makebox[\h][c]{\scriptsize FID 60.47}}%
\colC{\makebox[\h][c]{\scriptsize FID 32.34}}%
\colD{\makebox[\h][c]{\scriptsize FID 23.17}}%
\colE{\makebox[\h][c]{\scriptsize FID 17.22}}%
\colF{\makebox[\h][c]{\scriptsize FID 15.84}}%
\colG{\makebox[\h][c]{\scriptsize FID 13.25}}%
\colH{\makebox[\h][c]{\scriptsize FID 12.50}}%
\colI{\makebox[\h][c]{\scriptsize FID 11.08}}%
\vspace{\vvv}\\
\makebox[0mm][l]{\rotatebox{90}{\makebox[\h][c]{StyleGAN2}}}\hspace{\hh}%
\colA{\includegraphics[width=\h]{generated_images/appendix/subset_sweep_ffhq_cfgF/2x2_sub1k_seed80.\ext}}%
\colB{\includegraphics[width=\h]{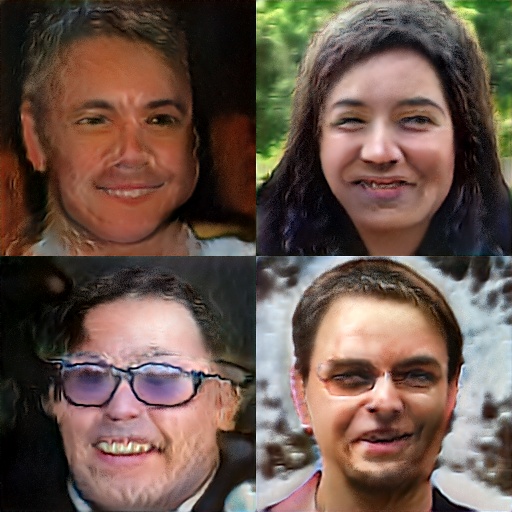}}%
\colC{\includegraphics[width=\h]{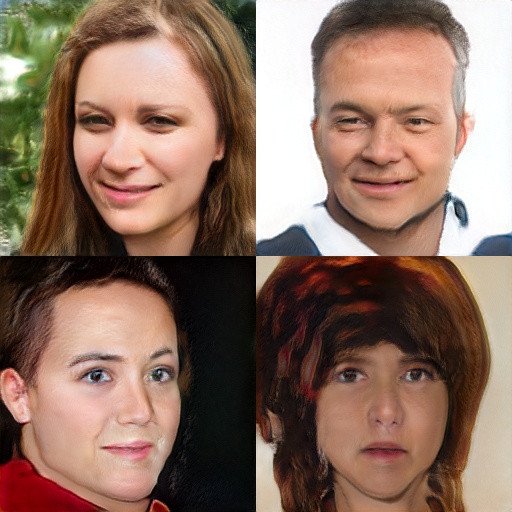}}%
\colD{\includegraphics[width=\h]{generated_images/appendix/subset_sweep_ffhq_cfgF/2x2_sub10k_seed50.\ext}}%
\colE{\includegraphics[width=\h]{generated_images/appendix/subset_sweep_ffhq_cfgF/2x2_sub20k_seed0.\ext}}%
\colF{\includegraphics[width=\h]{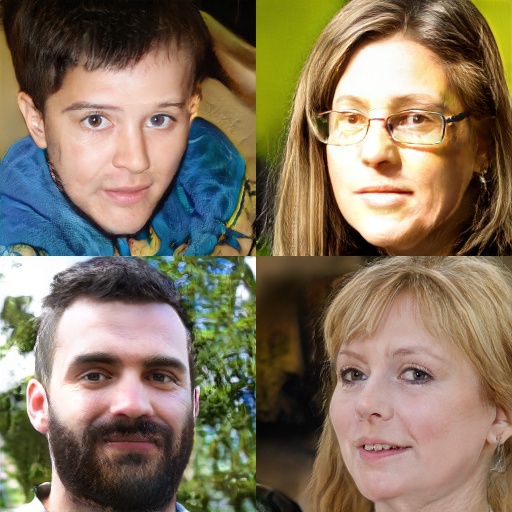}}%
\colG{\includegraphics[width=\h]{generated_images/appendix/subset_sweep_ffhq_cfgF/2x2_sub50k_seed90.\ext}}%
\colH{\includegraphics[width=\h]{generated_images/appendix/subset_sweep_ffhq_cfgF/2x2_sub70k_seed70.\ext}}%
\colI{\includegraphics[width=\h]{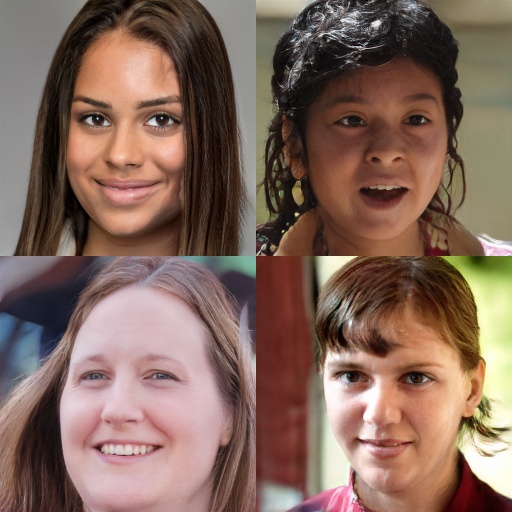}}%
\vspace{\vv}\\
\makebox[0mm][l]{}\hspace{\hh}%
\colA{\makebox[\h][c]{\scriptsize FID 77.24 -- Recall 0.000}}%
\colB{\makebox[\h][c]{\scriptsize FID 66.77 -- Recall 0.002}}%
\colC{\makebox[\h][c]{\scriptsize FID 39.42 -- Recall 0.030}}%
\colD{\makebox[\h][c]{\scriptsize FID 19.82 -- Recall 0.116}}%
\colE{\makebox[\h][c]{\scriptsize FID 11.60 -- Recall 0.239}}%
\colF{\makebox[\h][c]{\scriptsize FID 8.80 -- Recall 0.283}}%
\colG{\makebox[\h][c]{\scriptsize FID 5.96 -- Recall 0.370}}%
\colH{\makebox[\h][c]{\scriptsize FID 4.88 -- Recall 0.394}}%
\colI{\makebox[\h][c]{\scriptsize FID 3.81 -- Recall {\bf 0.452}}}%
\vspace{\vvv}\\
\makebox[0mm][l]{\rotatebox{90}{\makebox[\h][c]{Our baseline}}}\hspace{\hh}%
\colA{\includegraphics[width=\h]{generated_images/appendix/subset_sweep_ffhq_baseline/2x2_sub1k_seed90.\ext}}%
\colB{\includegraphics[width=\h]{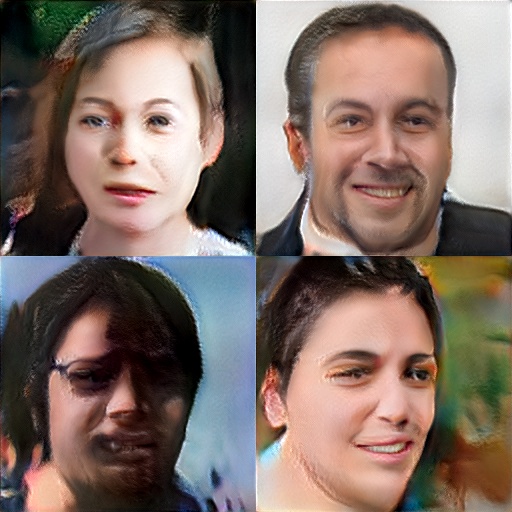}}%
\colC{\includegraphics[width=\h]{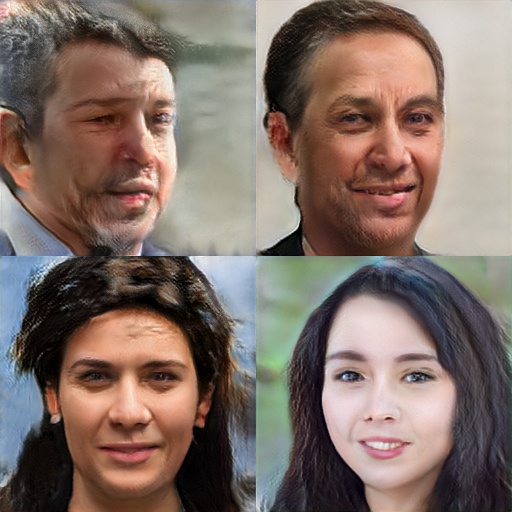}}%
\colD{\includegraphics[width=\h]{generated_images/appendix/subset_sweep_ffhq_baseline/2x2_sub10k_seed30.\ext}}%
\colE{\includegraphics[width=\h]{generated_images/appendix/subset_sweep_ffhq_baseline/2x2_sub20k_seed50.\ext}}%
\colF{\includegraphics[width=\h]{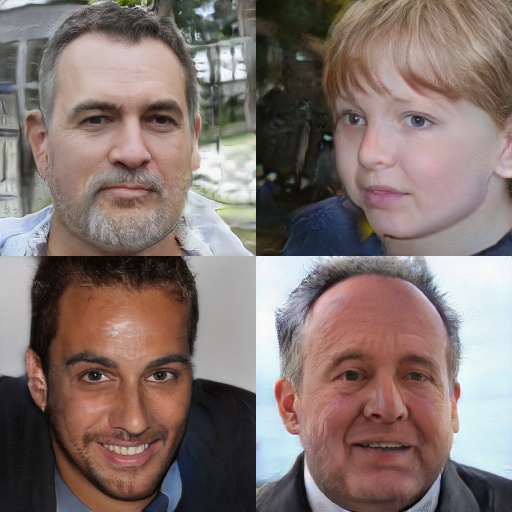}}%
\colG{\includegraphics[width=\h]{generated_images/appendix/subset_sweep_ffhq_baseline/2x2_sub50k_seed0.\ext}}%
\colH{\includegraphics[width=\h]{generated_images/appendix/subset_sweep_ffhq_baseline/2x2_sub70k_seed50.\ext}}%
\colI{\includegraphics[width=\h]{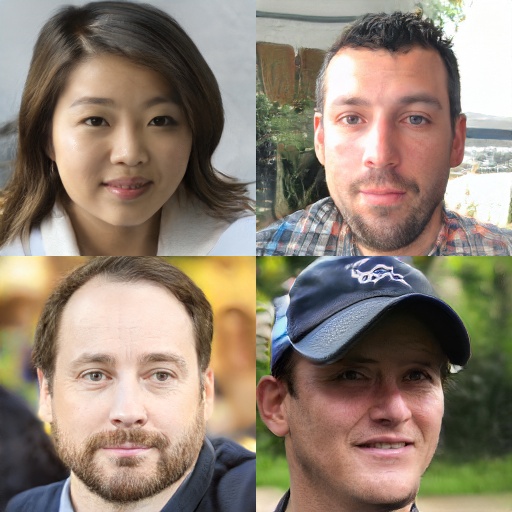}}%
\vspace{\vv}\\
\makebox[0mm][l]{}\hspace{\hh}%
\colA{\makebox[\h][c]{\scriptsize FID 88.43 -- Recall 0.000}}%
\colB{\makebox[\h][c]{\scriptsize FID 76.61 -- Recall 0.000}}%
\colC{\makebox[\h][c]{\scriptsize FID 43.72 -- Recall 0.010}}%
\colD{\makebox[\h][c]{\scriptsize FID 30.25 -- Recall 0.077}}%
\colE{\makebox[\h][c]{\scriptsize FID 17.30 -- Recall 0.156}}%
\colF{\makebox[\h][c]{\scriptsize FID 11.40 -- Recall 0.258}}%
\colG{\makebox[\h][c]{\scriptsize FID 6.72 -- Recall 0.346}}%
\colH{\makebox[\h][c]{\scriptsize FID 5.12 -- Recall 0.371}}%
\colI{\makebox[\h][c]{\scriptsize FID {\bf 3.54} -- Recall {\bf 0.452}}}%
\vspace{\vvv}\\
\makebox[0mm][l]{\rotatebox{90}{\makebox[\h][c]{ADA (Ours)}}}\hspace{\hh}%
\colA{\includegraphics[width=\h]{generated_images/appendix/subset_sweep_ffhq_augment/2x2_sub1k_seed160.\ext}}%
\colB{\includegraphics[width=\h]{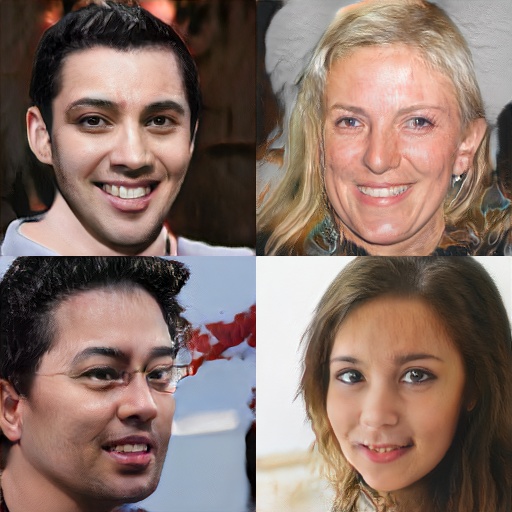}}%
\colC{\includegraphics[width=\h]{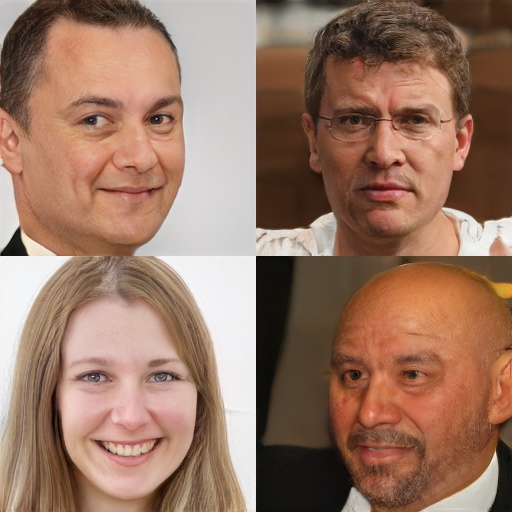}}%
\colD{\includegraphics[width=\h]{generated_images/appendix/subset_sweep_ffhq_augment/2x2_sub10k_seed0.\ext}}%
\colE{\includegraphics[width=\h]{generated_images/appendix/subset_sweep_ffhq_augment/2x2_sub20k_seed30.\ext}}%
\colF{\includegraphics[width=\h]{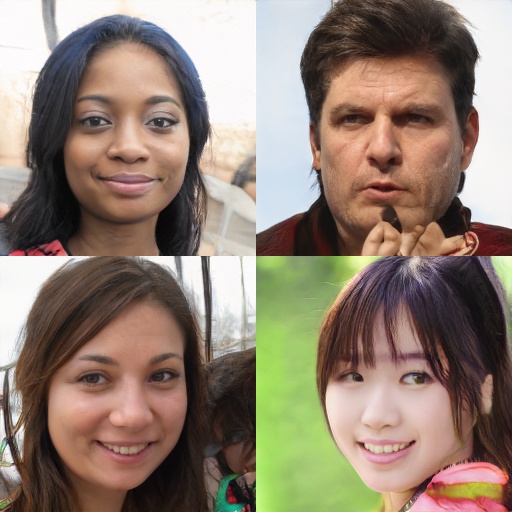}}%
\colG{\includegraphics[width=\h]{generated_images/appendix/subset_sweep_ffhq_augment/2x2_sub50k_seed30.\ext}}%
\colH{\includegraphics[width=\h]{generated_images/appendix/subset_sweep_ffhq_augment/2x2_sub70k_seed70.\ext}}%
\colI{\includegraphics[width=\h]{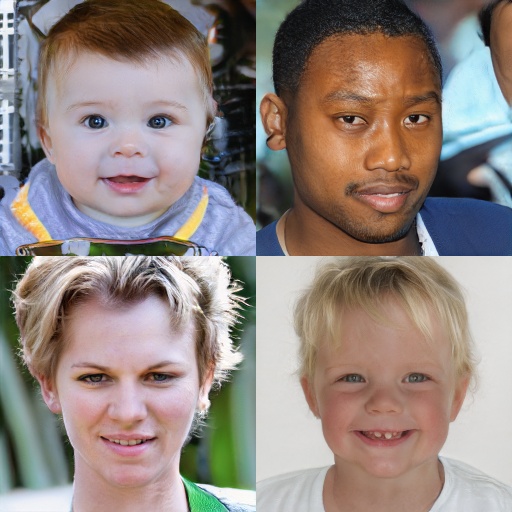}}%
\vspace{\vv}\\
\makebox[0mm][l]{}\hspace{\hh}%
\colA{\makebox[\h][c]{\scriptsize FID {\bf 20.55} -- Recall {\bf 0.058}}}%
\colB{\makebox[\h][c]{\scriptsize FID {\bf 15.76} -- Recall {\bf 0.135}}}%
\colC{\makebox[\h][c]{\scriptsize FID 10.78 -- Recall {\bf 0.185}}}%
\colD{\makebox[\h][c]{\scriptsize FID 8.10 -- Recall {\bf 0.266}}}%
\colE{\makebox[\h][c]{\scriptsize FID 6.30 -- Recall 0.301}}%
\colF{\makebox[\h][c]{\scriptsize FID 5.40 -- Recall {\bf 0.354}}}%
\colG{\makebox[\h][c]{\scriptsize FID 4.53 -- Recall {\bf 0.392}}}%
\colH{\makebox[\h][c]{\scriptsize FID 4.27 -- Recall {\bf 0.399}}}%
\colI{\makebox[\h][c]{\scriptsize FID 3.79 -- Recall 0.440}}%
\vspace{\vvv}\\
\makebox[0mm][l]{\rotatebox{90}{\makebox[\h][c]{ADA + bCR}}}\hspace{\hh}%
\colA{\includegraphics[width=\h]{generated_images/appendix/subset_sweep_ffhq_cr_da/2x2_sub1k_seed160.\ext}}%
\colB{\includegraphics[width=\h]{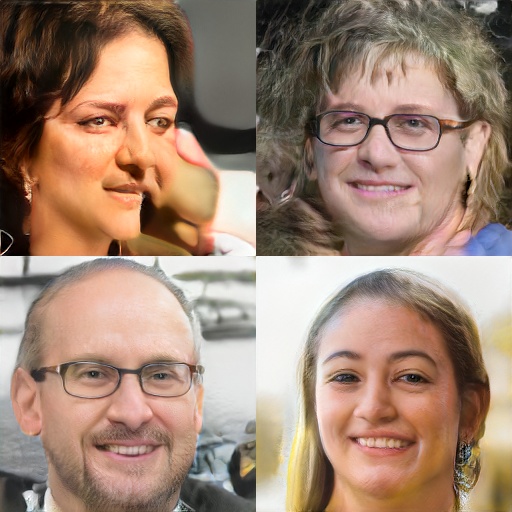}}%
\colC{\includegraphics[width=\h]{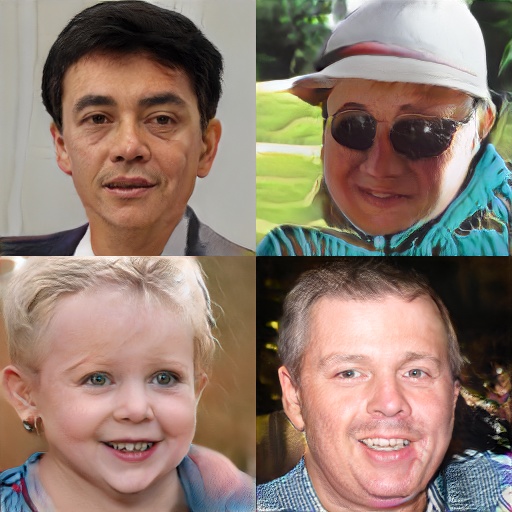}}%
\colD{\includegraphics[width=\h]{generated_images/appendix/subset_sweep_ffhq_cr_da/2x2_sub10k_seed60.\ext}}%
\colE{\includegraphics[width=\h]{generated_images/appendix/subset_sweep_ffhq_cr_da/2x2_sub20k_seed20.\ext}}%
\colF{\includegraphics[width=\h]{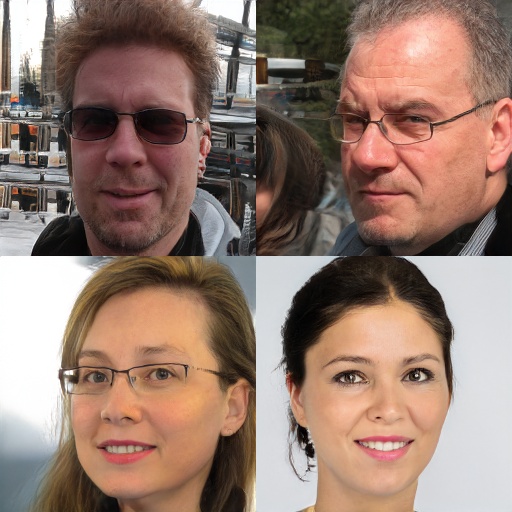}}%
\colG{\includegraphics[width=\h]{generated_images/appendix/subset_sweep_ffhq_cr_da/2x2_sub50k_seed90.\ext}}%
\colH{\includegraphics[width=\h]{generated_images/appendix/subset_sweep_ffhq_cr_da/2x2_sub70k_seed60.\ext}}%
\colI{\includegraphics[width=\h]{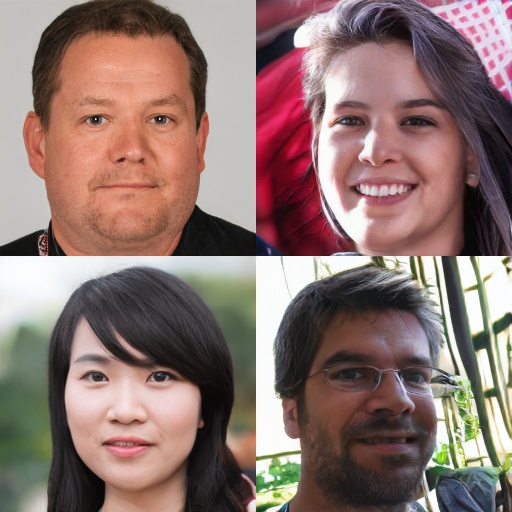}}%
\vspace{\vv}\\
\makebox[0mm][l]{}\hspace{\hh}%
\colA{\makebox[\h][c]{\scriptsize FID 21.44 -- Recall 0.029}}%
\colB{\makebox[\h][c]{\scriptsize FID 17.05 -- Recall 0.076}}%
\colC{\makebox[\h][c]{\scriptsize FID {\bf 10.21} -- Recall 0.155}}%
\colD{\makebox[\h][c]{\scriptsize FID {\bf 7.46} -- Recall 0.233}}%
\colE{\makebox[\h][c]{\scriptsize FID {\bf 5.27} -- Recall {\bf 0.317}}}%
\colF{\makebox[\h][c]{\scriptsize FID {\bf 4.55} -- Recall 0.327}}%
\colG{\makebox[\h][c]{\scriptsize FID {\bf 3.96} -- Recall 0.380}}%
\colH{\makebox[\h][c]{\scriptsize FID {\bf 3.87} -- Recall 0.402}}%
\colI{\makebox[\h][c]{\scriptsize FID 3.55 -- Recall 0.412}}%
\caption{
Images generated for different subsets of FFHQ at 256$\times$256 resolution using the training setups from Figures~\refpaper{fig:MainSweep} and \ref{fig:BigGANComparison}.
We show the best snapshot of the best training run for each case, selected according to FID, so the numbers are slightly better than the medians reported in Figure~\refpaper{fig:MainSweep}c.
In addition to FID, we also report the Recall metric~\cite{Tuomas2019} as a more direct way to estimate image diversity.
The bolded numbers indicate the lowest FID and highest Recall for each training set size.
``BigGAN'' corresponds to the unconditional variant of BigGAN~\cite{Brock2018} proposed by Sch\"onfeld~et~al.~\cite{Schonfeld2020}, and ``StyleGAN2'' corresponds to config \textsc{f} of the official TensorFlow implementation by Karras~et~al.~\cite{Karras2019}.
}
\label{#1}
\end{figure}
}

\newcommand{\figBigGANComparison}[1]{
\begin{figure}[p]
\footnotesize%
\renewcommand{\h}{0.497\linewidth}%
\includegraphics[width=\h]{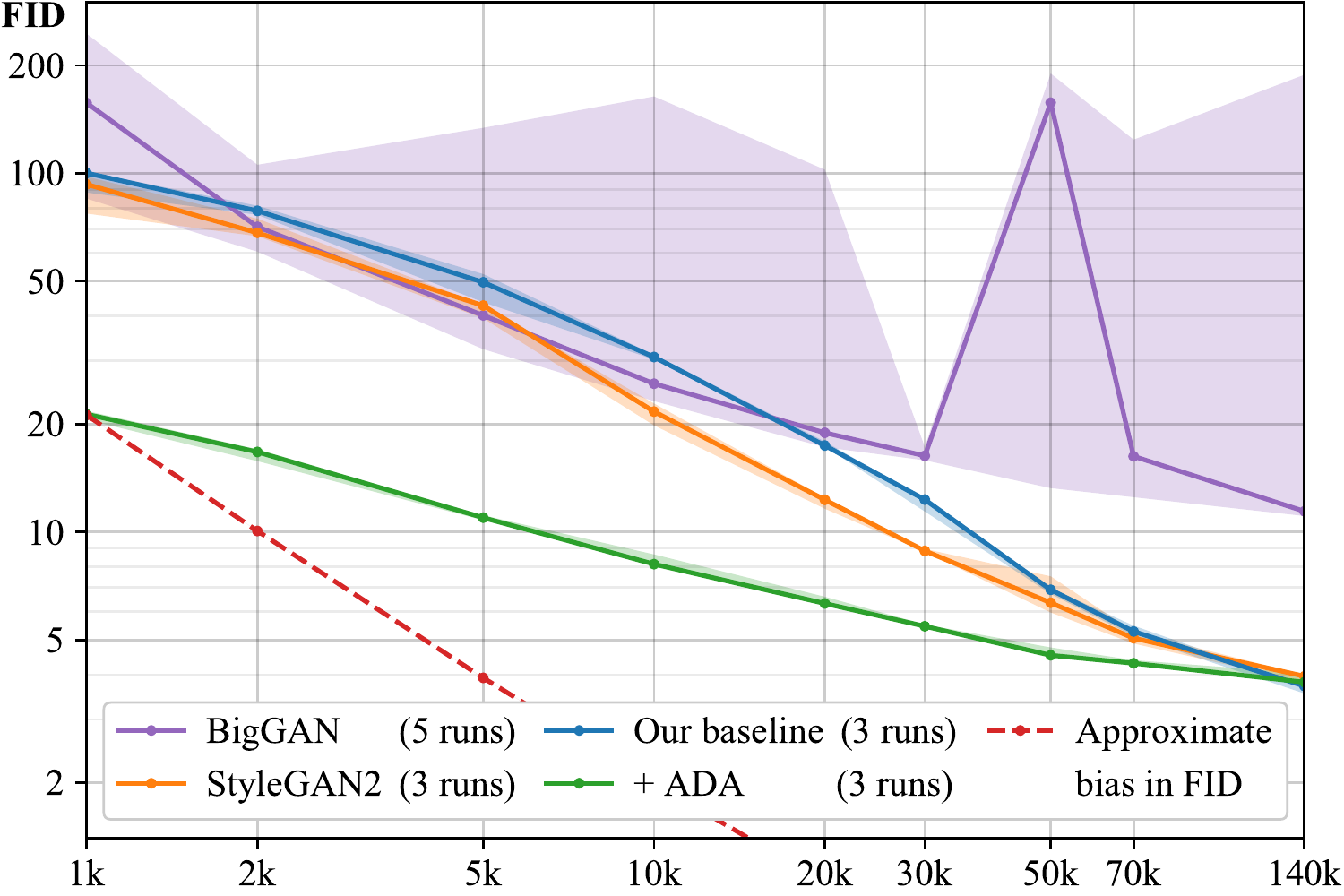}\hfill%
\includegraphics[width=\h]{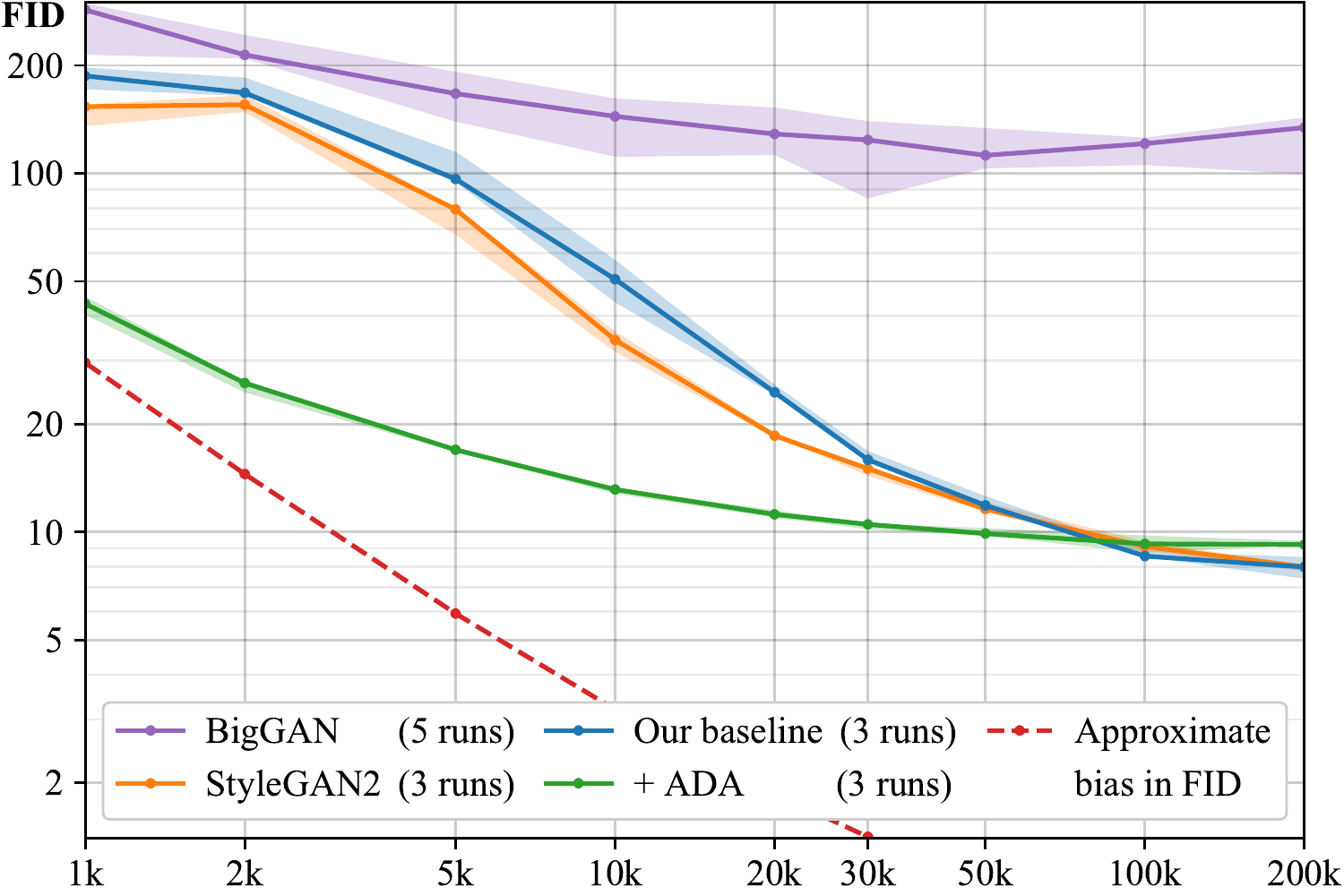}\vspace{0.8mm}\\
\makebox[\h][c]{(a) Different subsets of \textsc{FFHQ} at 256$\times$256}\hfill%
\makebox[\h][c]{(b) Different subsets of \textsc{LSUN Cat} at 256$\times$256}%
\caption{
Comparison of our results with unconditional BigGAN~\cite{Brock2018,Schonfeld2020} and StyleGAN2 config~\textsc{f}~\cite{Karras2019}.
We report the median/min/max FID as a function of training set size, calculated over multiple independent training runs.
The dashed red line illustrates the expected bias of the FID metric, computed using a hypothetical generator that outputs random images from the training set as-is.
}
\label{#1}
\end{figure}
}

\newcommand{\smallpm}{\raisebox{0.2mm}{\scalebox{0.7}{$\pm$}}\hspace{0.2mm}}

\newcommand{\figLeakyComparison}[1]{
\begin{figure}[p]
\footnotesize%
\renewcommand{\h}{0.420\linewidth}%
\renewcommand{\hh}{0.285\linewidth}%
\renewcommand{\hhh}{15.48mm}%
\renewcommand{\hhhh}{\hhh*\real{170}/\real{256}}%
\parbox[b]{\h}{%
\centering%
{\scriptsize\bf Integer translation}\\[-0.3mm]
\includegraphics[width=\hhhh,trim={43 0 43 0},clip]{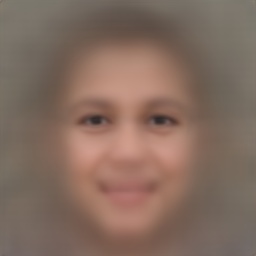}\hfill%
\includegraphics[width=\hhhh,trim={43 0 43 0},clip]{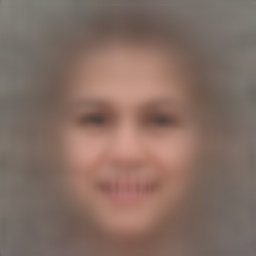}\hfill%
\includegraphics[width=\hhhh,trim={43 0 43 0},clip]{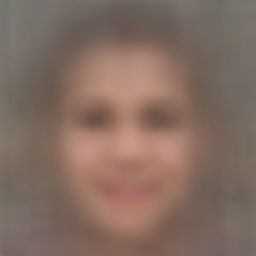}\hfill%
\includegraphics[width=\hhhh,trim={43 0 43 0},clip]{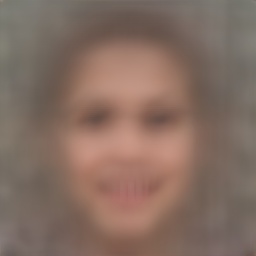}\hfill%
\includegraphics[width=\hhh]{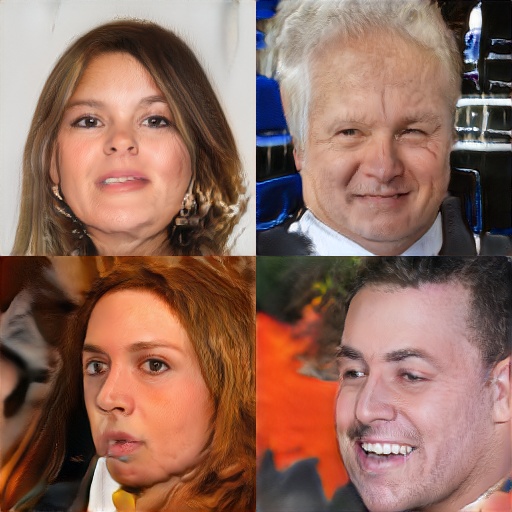}\\[-1.5mm]
\makebox[\hhhh][c]{\scriptsize\smallpm{}0px}\hfill%
\makebox[\hhhh][c]{\scriptsize\smallpm{}4px}\hfill%
\makebox[\hhhh][c]{\scriptsize\smallpm{}8px}\hfill%
\makebox[\hhhh][c]{\scriptsize\smallpm{}16px}\hfill%
\makebox[\hhh][c]{\scriptsize\smallpm{}16px samples}\\[0mm]
{\scriptsize\bf Arbitrary rotation}\\[-0.3mm]
\includegraphics[width=\hhhh,trim={43 0 43 0},clip]{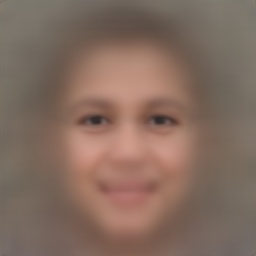}\hfill%
\includegraphics[width=\hhhh,trim={43 0 43 0},clip]{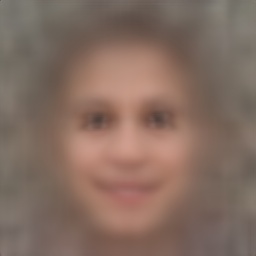}\hfill%
\includegraphics[width=\hhhh,trim={43 0 43 0},clip]{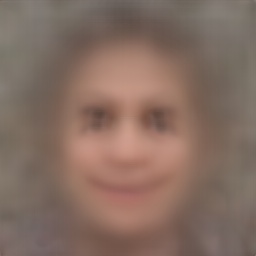}\hfill%
\includegraphics[width=\hhhh,trim={43 0 43 0},clip]{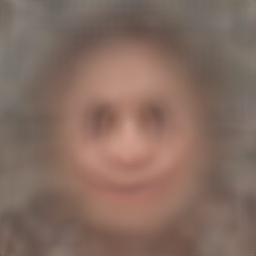}\hfill%
\includegraphics[width=\hhh]{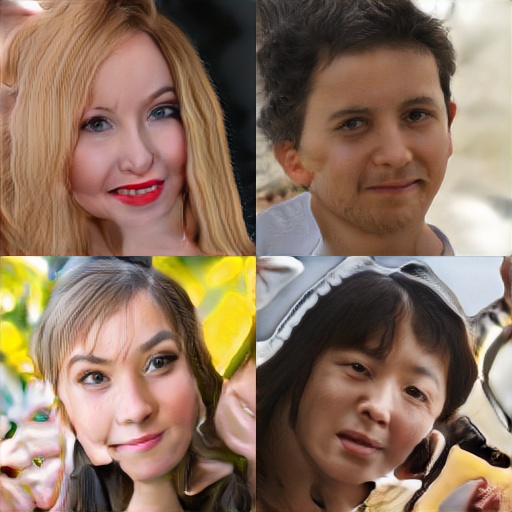}\\[-1.5mm]
\makebox[\hhhh][c]{\scriptsize\smallpm{}0$^{\circ}$}\hfill%
\makebox[\hhhh][c]{\scriptsize\smallpm{}10$^{\circ}$}\hfill%
\makebox[\hhhh][c]{\scriptsize\smallpm{}20$^{\circ}$}\hfill%
\makebox[\hhhh][c]{\scriptsize\smallpm{}45$^{\circ}$}\hfill%
\makebox[\hhh][c]{\scriptsize\smallpm{}45$^{\circ}$ samples}\vspace{0.1mm}%
}%
\hspace{0.009\linewidth}\hfill%
\includegraphics[width=\hh]{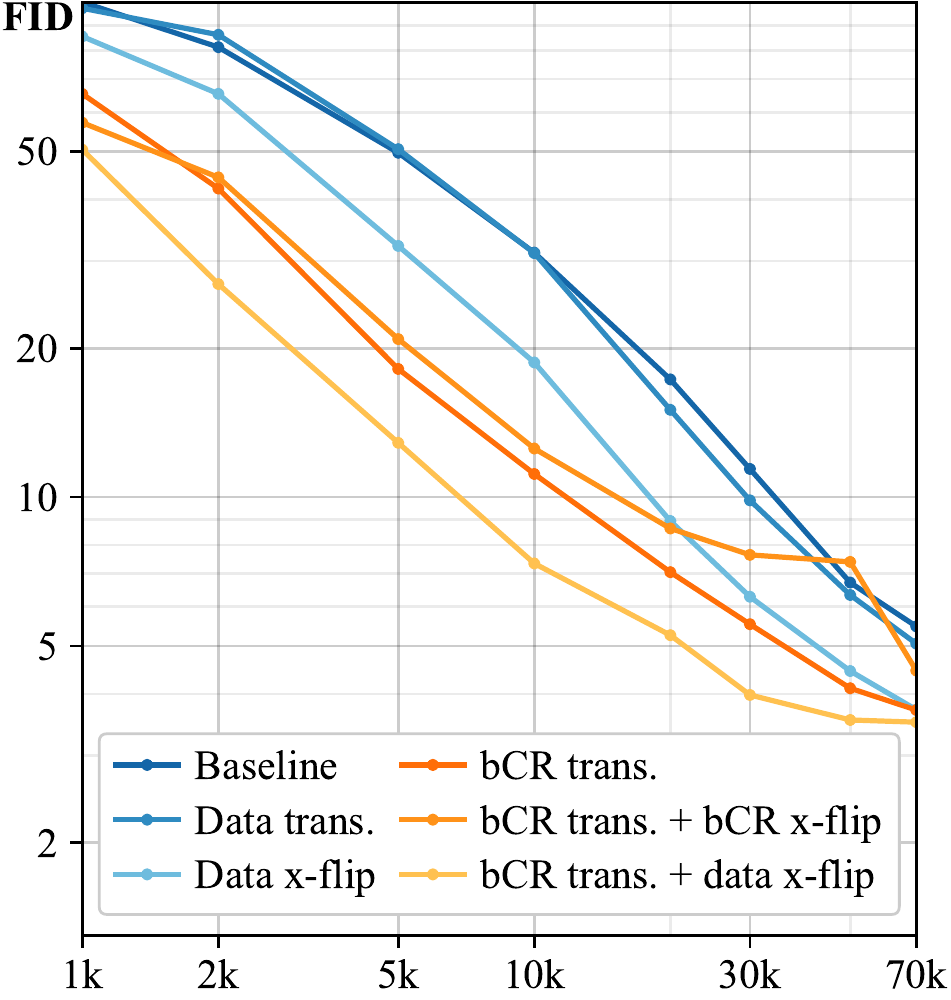}\hfill%
\includegraphics[width=\hh]{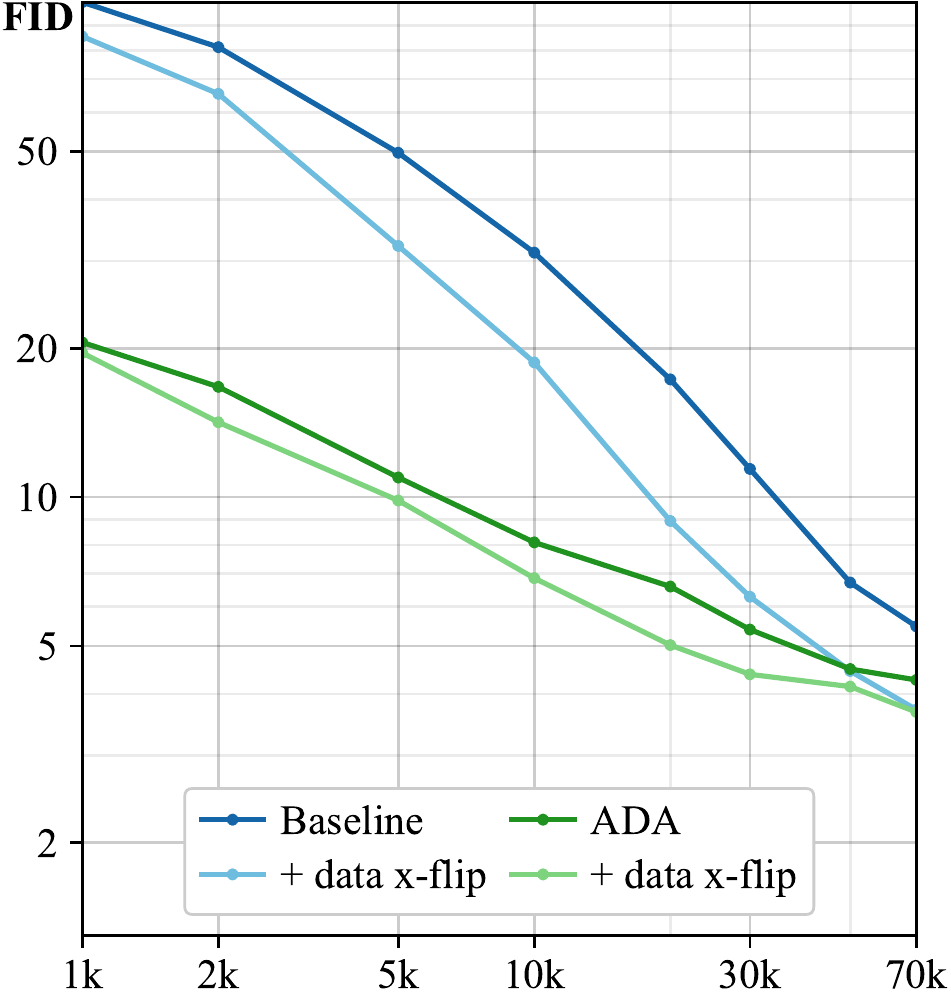}\\
\makebox[\h][c]{(a) Mean images for bCR with FFHQ-5k}%
\hspace{0.009\linewidth}\hfill%
\makebox[\hh][c]{(b) bCR vs. dataset augment}\hfill%
\makebox[\hh][c]{(c) Effect of dataset $x$-flips}%
\caption{
(a)
Examples of bCR leaking to generated images.
(b)
Comparison between dataset augmentation and bCR using $\pm$8px translations and $x$-flips.
(c)
In general, dataset $x$-flips can provide a significant boost to FID in cases where they are appropriate.
For baseline, the effect is almost equal to doubling the size of training set, as evidenced by the consistent 2$\times$ horizontal offset between the blue curves.
With ADA the effect is somewhat weaker.
}
\label{#1}
\end{figure}
}

\newcommand{\figFullSweep}[1]{
\begin{figure}[p]
\renewcommand{\h}{0.244\linewidth}%
\renewcommand{\hh}{3mm}%
\renewcommand{\hhh}{3.9mm}%
\renewcommand{\hhhh}{1.2mm}%
\renewcommand{\vv}{\h*\real{3}/\real{3.2}}%
\renewcommand{\vvv}{2.6mm}%
\makebox[0mm][l]{}\hspace{\hh}%
\hspace{\hhh}\makebox[\h-\hhh-\hhhh][c]{2k training set}\hspace{\hhhh}\hfill%
\hspace{\hhh}\makebox[\h-\hhh-\hhhh][c]{10k training set}\hspace{\hhhh}\hfill%
\hspace{\hhh}\makebox[\h-\hhh-\hhhh][c]{50k training set}\hspace{\hhhh}\hfill%
\hspace{\hhh}\makebox[\h-\hhh-\hhhh][c]{140k training set}\hspace{\hhhh}\vspace{0.5mm}\\
\makebox[0mm][l]{\rotatebox{90}{\hspace{\vvv}\makebox[\vv-\vvv][c]{Individual}}}\hspace{\hh}%
\includegraphics[width=\h]{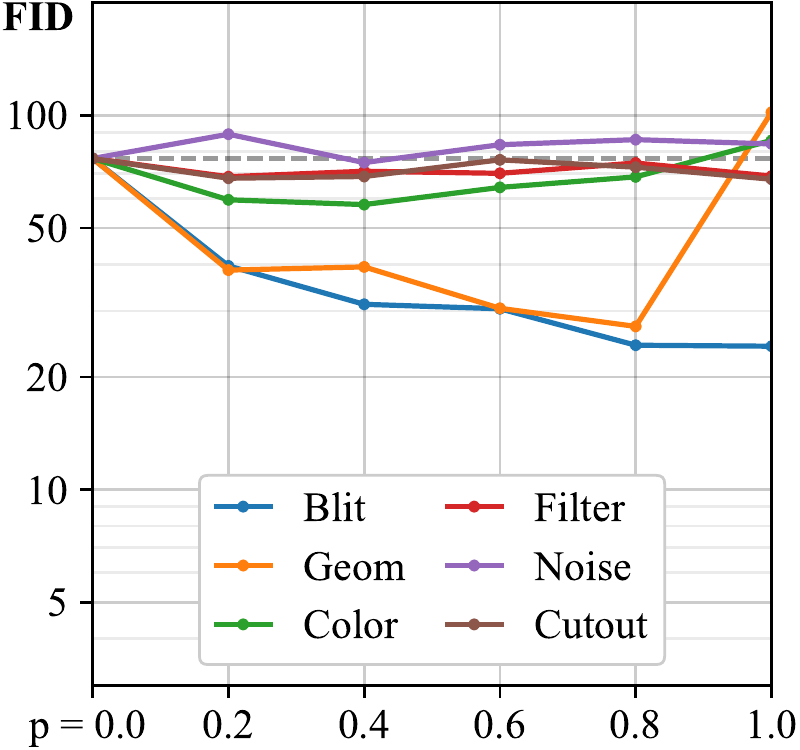}\hfill%
\includegraphics[width=\h]{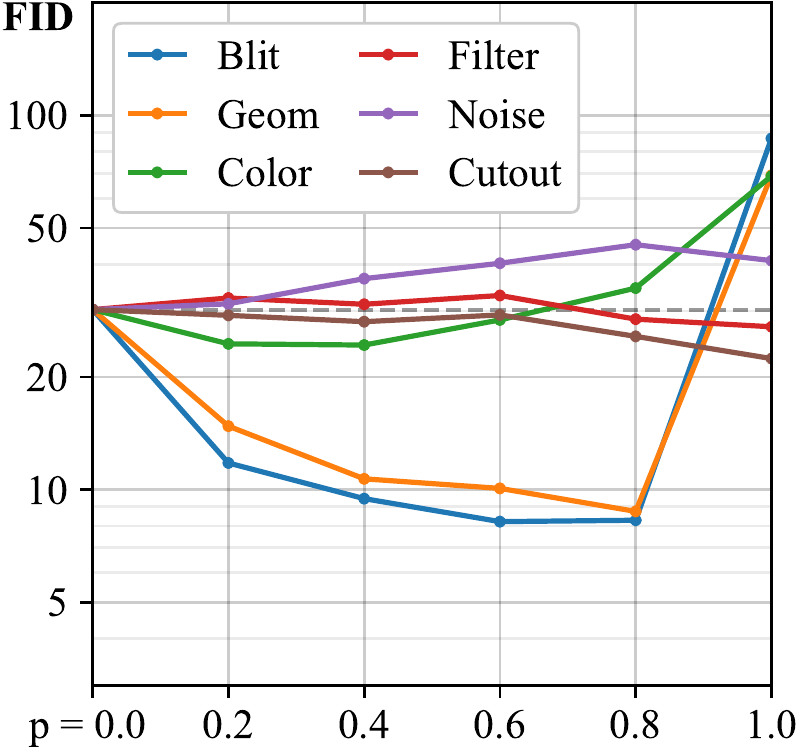}\hfill%
\includegraphics[width=\h]{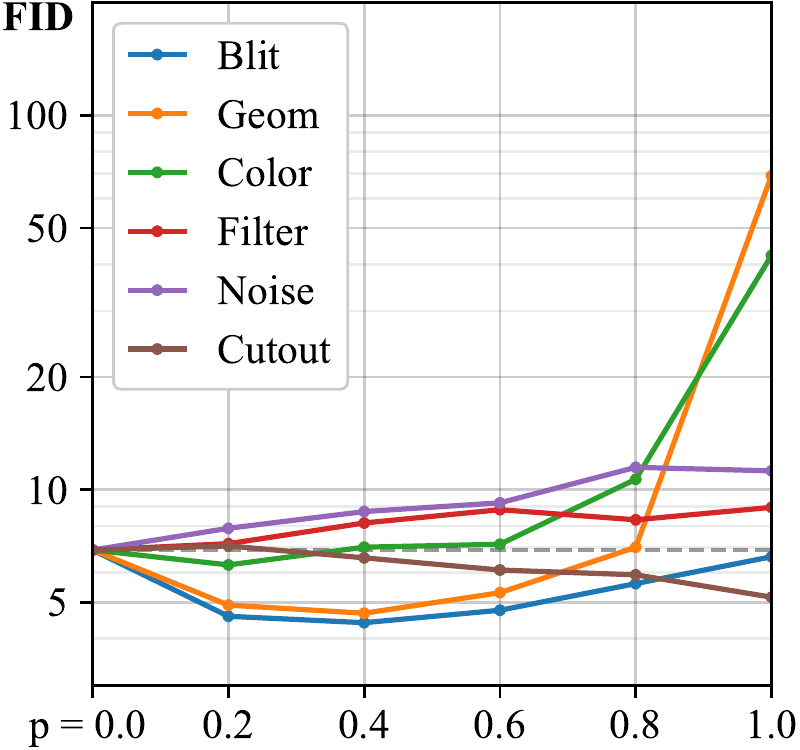}\hfill%
\includegraphics[width=\h]{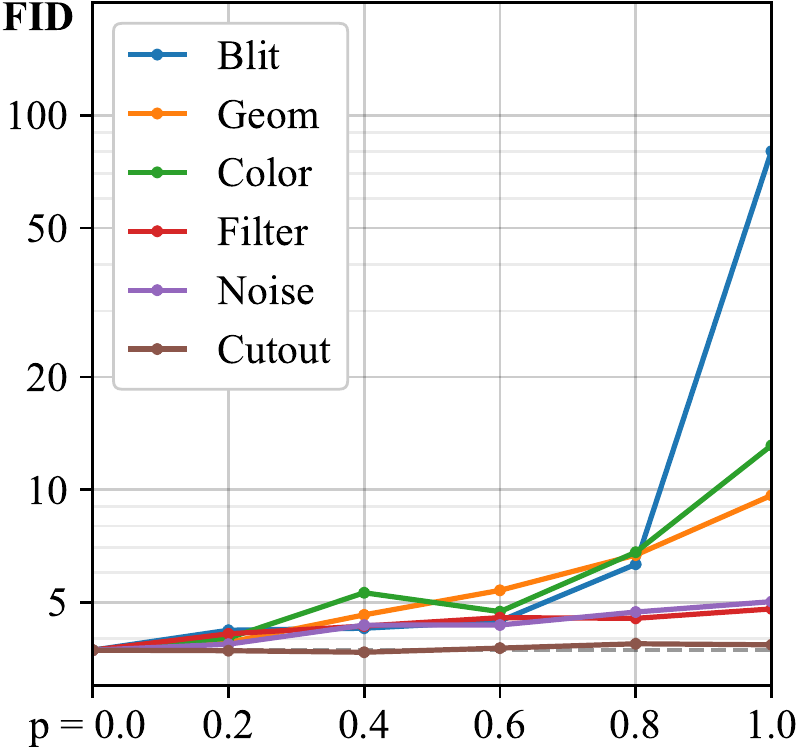}\vspace{1mm}\\
\makebox[0mm][l]{\rotatebox{90}{\hspace{\vvv}\makebox[\vv-\vvv][c]{Cumulative}}}\hspace{\hh}%
\includegraphics[width=\h]{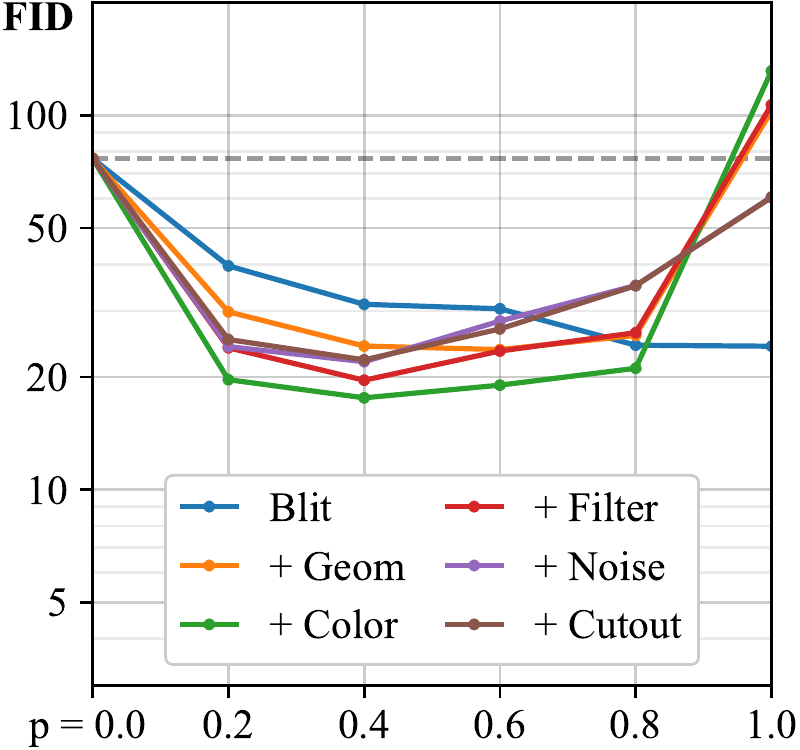}\hfill%
\includegraphics[width=\h]{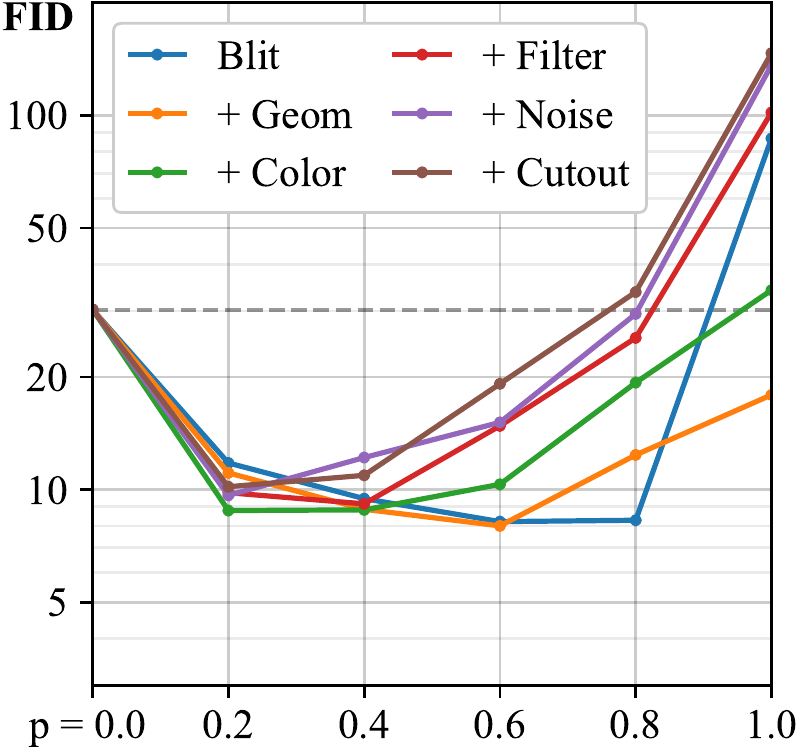}\hfill%
\includegraphics[width=\h]{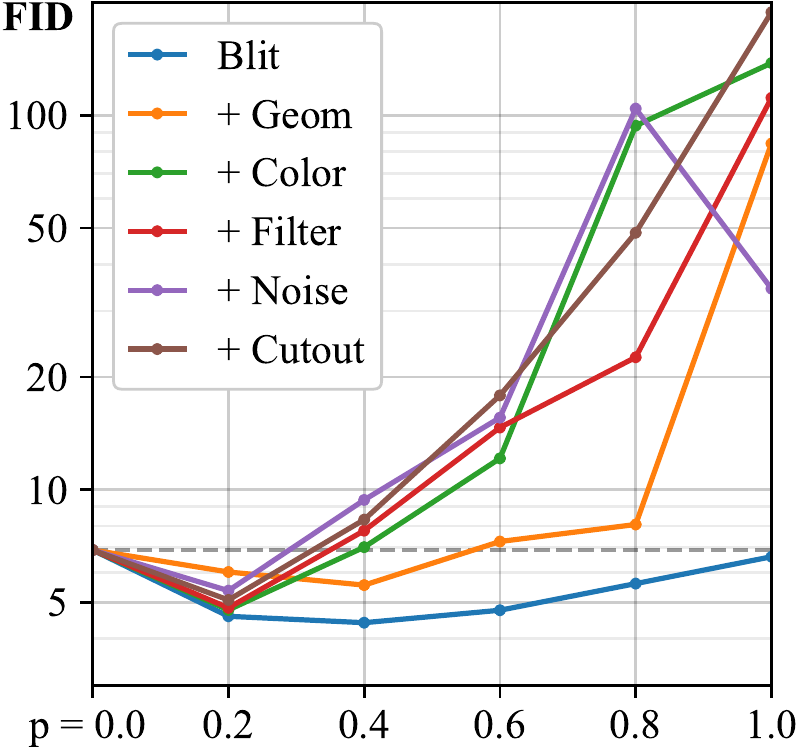}\hfill%
\includegraphics[width=\h]{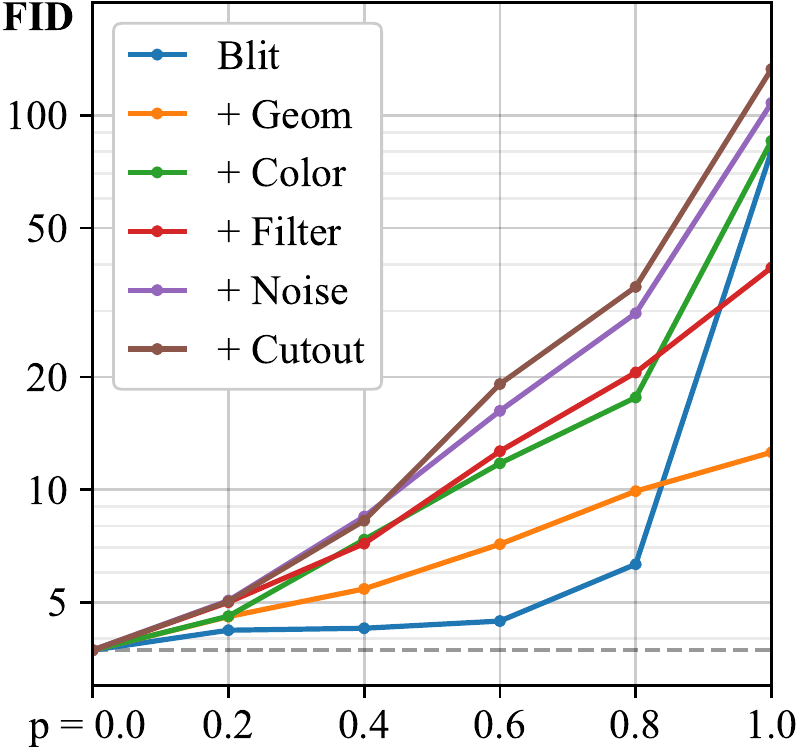}%
\caption{
Extended version of Figure~\refpaper{fig:FixedSweeps}, illustrating the individual and cumulative effect of different augmentation categories with increasing augmentation probability $p$.
}
\label{#1}
\end{figure}
}

\newcounter{plinenum}
\setcounter{plinenum}{0}
\newlength{\pindent}
\setlength{\pindent}{3mm}

\newcommand{\pl}[1]{\addtocounter{plinenum}{1}\makebox[3.1mm][r]{\color{gray}{\scriptsize\arabic{plinenum}:}}\hspace*{1.5mm}\hspace*{#1\pindent}} %
\newcommand{\plabel}[1]{\newcounter{alg:#1}\setcounter{alg:#1}{\value{plinenum}}} %
\newcommand{\pref}[1]{\arabic{alg:#1}} %

\newcommand{\pk}[1]{\textbf{#1}} %
\newcommand{\pf}[1]{\textsc{#1}} %
\newcommand{\ps}[1]{\mathit{#1}} %
\newcommand{\pc}[1]{\color{gray}{$\triangleright$ #1}} %
\newcommand{\pa}{\leftarrow} %
\newcommand{\pifaug}[2]{\pk{apply} #1 \pk{with} probability #2} %

\newcommand{\algBlitGeomColor}{%
\scriptsize%
\setcounter{plinenum}{0}%
\begin{tabbing}%
\pl{0}  \pk{input:} original image $X$, augmentation probability $p$ \\
\pl{0}  \pk{output:} augmented image $Y$ \\
\pl{0}  $(w, h) \pa \pf{Size}(X)$ \\
\pl{0}  $Y \pa \pf{Convert}(X, \pf{float})$ \hspace*{0.3mm} \pc{$Y_{x,y} \in [-1,+1]^3$} \\[1.5mm]
\pl{0}  \pc{Select parameters for pixel blitting} \plabel{blitbegin} \\
\pl{0}  $G \pa I_3$ \hspace*{0.8mm} \pc{Homogeneous 2D transformation matrix} \\
\pl{0}  \pifaug{$x$-flip}{$p$} \plabel{xflip} \\
\pl{1}      \pk{sample} $i \sim \mathcal{U}\{0,1\}$ \\
\pl{1}      $G \pa \pf{Scale2D}(1-2i, 1) \cdot G$ \\
\pl{0}  \pifaug{$90^{\circ}$ rotations}{$p$} \plabel{blitrot} \\
\pl{1}      \pk{sample} $i \sim \mathcal{U}\{0,3\}$ \\
\pl{1}      $G \pa \pf{Rotate2D}\big( -\frac{\pi}{2} \cdot i \big) \cdot G$ \\
\pl{0}  \pifaug{integer translation}{$p$} \plabel{xint} \\
\pl{1}      \pk{sample} $t_x, t_y \sim \mathcal{U}(-0.125, +0.125)$ \\
\pl{1}      $G \pa \pf{Translate2D}\big( \text{round}(t_x w), \text{round}(t_y h) \big) \cdot G$ \plabel{blitend} \\[1.5mm]
\pl{0}  \pc{Select parameters for general geometric transformations} \plabel{geombegin} \\
\pl{0}  \pifaug{isotropic scaling}{$p$} \plabel{scale} \\
\pl{1}      \pk{sample} $s \sim \mathrm{Lognormal}\big( 0, (0.2\cdot\ln2)^2 \big)$ \\
\pl{1}      $G \pa \pf{Scale2D}(s, s) \cdot G$ \\
\pl{0}  $p_\ps{rot} \pa 1-\sqrt{1-p}$ \hspace*{1.5mm} \pc{$P(\ps{pre} \cup \ps{post}) = p$} \\
\pl{0}  \pifaug{pre-rotation}{$p_\ps{rot}$} \plabel{prerot} \\
\pl{1}      \pk{sample} $\theta \sim \mathcal{U}(-\pi,+\pi)$ \\
\pl{1}      $G \pa \pf{Rotate2D}(-\theta) \cdot G$ \hspace{0.5mm} \=\pc{Before anisotropic scaling} \\
\pl{0}  \pifaug{anisotropic scaling}{$p$} \plabel{aniso} \\
\pl{1}      \pk{sample} $s \sim \mathrm{Lognormal}\big( 0, (0.2\cdot\ln2)^2 \big)$ \\
\pl{1}      $G \pa \pf{Scale2D}\big( s, \frac{1}{s} \big) \cdot G$ \\
\pl{0}  \pifaug{post-rotation}{$p_\ps{rot}$} \plabel{postrot} \\
\pl{1}      \pk{sample} $\theta \sim \mathcal{U}(-\pi,+\pi)$ \\
\pl{1}      $G \pa \pf{Rotate2D}(-\theta) \cdot G$  \>\pc{After anisotropic scaling} \\
\pl{0}  \pifaug{fractional translation}{$p$} \plabel{xfrac} \\
\pl{1}      \pk{sample} $t_x, t_y \sim \mathcal{N}\big( 0, (0.125)^2 \big)$ \\
\pl{1}      $G \pa \pf{Translate2D}( t_x w, t_y h) \cdot G$ \plabel{geomend} \\[1.5mm]
\pl{0}  \pc{Pad image and adjust origin} \plabel{padbegin} \\
\pl{0}  $H(z) \pa \pf{Wavelet}(\pf{sym6})$ \hspace*{0.5mm} \pc{Orthogonal lowpass filter} \plabel{gwavelet} \\
\pl{0}  $(m_\ps{lo}, m_\ps{hi}) \pa \pf{CalculatePadding}\big( G, w, h, H(z) \big)$ \plabel{gmargin} \\
\pl{0}  $Y \pa \pf{Pad}(Y, m_\ps{lo}, m_\ps{hi}, \pf{reflect})$ \\
\pl{0}  \FINAL{$T \pa \pf{Translate2D}\big( \frac{1}{2}w \hspace{-0.6mm}-\hspace{-0.6mm} \frac{1}{2} \hspace{-0.6mm}+\hspace{-0.6mm} m_{\ps{lo},x}, \frac{1}{2}h \hspace{-0.6mm}-\hspace{-0.6mm} \frac{1}{2} \hspace{-0.6mm}+\hspace{-0.6mm} m_{\ps{lo},y} \big)$} \\
\pl{0}  $G \pa T \cdot G \cdot T^{-1}$ \hspace*{0.5mm} \=\pc{Place origin at image center} \plabel{padend} \\[1.5mm]
\pl{0}  \pc{Execute geometric transformations} \plabel{gexecbegin} \\
\pl{0}  $Y' \pa \pf{Upsample2x2}\big( Y, H(z^{-1}) \big)$ \plabel{gupsample} \\
\pl{0}  $S \pa \pf{Scale2D}(2,2)$ \\
\pl{0}  $G \pa S \cdot G \cdot S^{-1}$                 \>\pc{Account for the upsampling} \\
\pl{0}  \pk{for each} pixel $(x_o, y_o) \in Y'$ \pk{do} \\
\pl{1}      $[x_i, y_i, z_i]^T \pa G^{-1} \cdot [x_o, y_o, 1]^T$ \\
\pl{1}      $Y_{x_o, y_o} \pa \pf{BilinearLookup}(Y', x_i, y_i)$ \plabel{gbilinear} \\
\pl{0}  $Y \pa \pf{Downsample2x2}\big( Y, H(z) \big)$ \\
\pl{0}  $Y \pa \pf{Crop}(Y, m_\ps{lo}, m_\ps{hi})$ \hspace*{1mm} \pc{Undo the padding} \plabel{gexecend} \\[1.5mm]
\pl{0}  \pc{Select parameters for color transformations} \plabel{colorbegin} \\
\pl{0}  $C \pa I_4$ \hspace*{0.8mm} \pc{Homogeneous 3D transformation matrix} \\
\pl{0}  \pifaug{brightness}{$p$} \plabel{brightness} \\
\pl{1}      \pk{sample} $b \sim \mathcal{N}\big( 0, (0.2)^2 \big)$ \\
\pl{1}      $C \pa \pf{Translate3D}(b, b, b) \cdot C$ \\
\pl{0}  \pifaug{contrast}{$p$} \plabel{contrast} \\
\pl{1}      \pk{sample} $c \sim \mathrm{Lognormal}\big( 0, (0.5\cdot\ln2)^2 \big)$ \\
\pl{1}      $C \pa \pf{Scale3D}(c, c, c) \cdot C$ \\
\pl{0}  $v \pa [1, 1, 1, 0]$\,$/$\,\raisebox{-0.1mm}{\scalebox{0.9}{$\sqrt{3}$}} \hspace*{10mm} \=\pc{Luma axis} \\
\pl{0}  \pifaug{luma flip}{$p$} \plabel{lumaflip} \\
\pl{1}      \pk{sample} $i \sim \mathcal{U}\{0,1\}$ \\
\pl{1}      $C \pa \big( I_4 - 2v^Tv \cdot i \big) \cdot C$     \>\pc{Householder reflection} \\
\pl{0}  \pifaug{hue rotation}{$p$} \plabel{hue} \\
\pl{1}      \pk{sample} $\theta \sim \mathcal{U}(-\pi,+\pi)$ \\
\pl{1}      $C \pa \pf{Rotate3D}(v, \theta) \cdot C$            \>\pc{Rotate around $v$} \\
\pl{0}  \pifaug{saturation}{$p$} \plabel{saturation} \\
\pl{1}      \pk{sample} $s \sim \mathrm{Lognormal}\big( 0, (1\cdot\ln2)^2 \big)$ \\
\pl{1}      $C \pa \Big( v^Tv + \big( I_4 - v^Tv \big) \cdot s \Big) \cdot C$ \plabel{colorend} \\[1.5mm]
\pl{0}  \pc{Execute color transformations} \plabel{cexecbegin} \\
\pl{0}  \pk{for each} pixel $(x, y) \in Y$ \pk{do} \\
\pl{1}      $(r_i, g_i, b_i) \pa Y_{x,y}$ \\
\pl{1}      $[r_o, g_o, b_o, a_o]^T \pa C \cdot [r_i, g_i, b_i, 1]^T $ \\
\pl{1}      $Y_{x,y} \pa (r_o, g_o, b_o)$ \plabel{cexecend} \\
\pl{0}  \pk{return} $Y$
\end{tabbing}%
}

\newcommand{\augsep}{\begin{tikzpicture}\draw[gray] (0,0) -- (\linewidth,0);\end{tikzpicture}}
\newcommand{\augrow}[3]{%
\parbox[b][#3]{\linewidth-#3*\real{4}}{\centering\vfill{#2}\vfill}\hfill%
\includegraphics[height=#3]{generated_images/appendix/augment_params/#1.\ext}%
}

\newcommand{\figBlitGeomColor}[1]{
\begin{figure}[p]
\scriptsize%
\renewcommand{\h}{0.472\linewidth}%
\renewcommand{\hh}{0.5\linewidth}%
\renewcommand{\vv}{214.5mm}%
\renewcommand{\vvv}{14.1mm}%
\renewcommand{\vvvv}{1.5mm}%
\parbox[b][\vv]{\h}{\vspace{-1.5mm}\algBlitGeomColor\vspace{-3.9mm}\hfill\vfill}%
\hfill\begin{tikzpicture}\draw[gray] (0,0) -- (0,\vv);\end{tikzpicture}\hfill%
\parbox[b][\vv]{\hh}{
\setlength{\lineskip}{0mm}%
\setlength{\baselineskip}{0mm}%
\makebox[\linewidth-\vvv*\real{4}][c]{Percentile:}\hfill%
\makebox[\vvv][c]{5\textsuperscript{th}}%
\makebox[\vvv][c]{35\textsuperscript{th}}%
\makebox[\vvv][c]{65\textsuperscript{th}}%
\makebox[\vvv][c]{95\textsuperscript{th}}%
\vfill\augsep\vfill%
\makebox[\linewidth][c]{\footnotesize\bf Pixel blitting}%
\vfill\augsep\vfill%
\augrow{00_xflip}{$x$-flip}{\vvv}\\[\vvvv]%
\augrow{01_rotate90}{$90^{\circ}$\\rotations}{\vvv}\\[\vvvv]%
\augrow{02_xint}{Integer\\translation}{\vvv}%
\vfill\augsep\vfill%
\makebox[\linewidth][c]{\footnotesize\bf General geometric transformations}%
\vfill\augsep\vfill%
\augrow{03_scale}{Isotropic\\scaling}{\vvv}\\[\vvvv]%
\augrow{04_rotate}{Arbitrary\\rotation}{\vvv}\\[\vvvv]%
\augrow{05_aniso}{Anisotropic\\scaling}{\vvv}\\[\vvvv]%
\augrow{06_xfrac}{Fractional\\translation}{\vvv}%
\vfill\augsep\vfill%
\makebox[\linewidth][c]{\footnotesize\bf Color transformations}%
\vfill\augsep\vfill%
\augrow{07_brightness}{Brightness}{\vvv}\\[\vvvv]%
\augrow{08_contrast}{Contrast}{\vvv}\\[\vvvv]%
\augrow{09_lumaflip}{Luma\\flip}{\vvv}\\[\vvvv]%
\augrow{10_hue}{Hue\\rotation}{\vvv}\\[\vvvv]%
\augrow{11_saturation}{Saturation}{\vvv}%
}%
\vspace{0.4mm}%
\caption{
Pseudocode and example images for geometric and color transformations (Appendix~\ref{app:BlitGeomColor}).
We illustrate the effect of each individual transformation (\raisebox{0.25mm}{\scriptsize\bf apply}) using four sets of parameter values, representing the 5\textsuperscript{th}, 35\textsuperscript{th}, 65\textsuperscript{th}, and 95\textsuperscript{th} percentiles of their corresponding distributions (\raisebox{0.25mm}{\scriptsize\bf sample}).
\vspace{-0.6mm}%
}
\label{#1}
\end{figure}
}

\newcommand{\algFilterNoiseCutout}{%
\scriptsize%
\setcounter{plinenum}{0}%
\begin{tabbing}%
\pl{0}  \pk{input:} original image $X$, augmentation probability $p$ \\
\pl{0}  \pk{output:} augmented image $Y$ \\
\pl{0}  $(w, h) \pa \pf{Size}(X)$ \\
\pl{0}  $Y \pa \pf{Convert}(X, \pf{float})$ \hspace*{0.3mm} \pc{$Y_{x,y} \in [-1,+1]^3$} \\[1.5mm]
\pl{0}  \pc{Select parameters for image-space filtering} \plabel{filterbegin} \\
\pl{0}  $b \pa \Big[ \big[ 0, \frac{\pi}{8} \big], \big[ \frac{\pi}{8}, \frac{\pi}{4} \big], \big[ \frac{\pi}{4}, \frac{\pi}{2} \big], \big[ \frac{\pi}{2}, \pi \big] \Big]$ \pc{Freq. bands} \plabel{fbands} \\
\pl{0}  $g \pa [1, 1, 1, 1]$     \hspace*{7mm} \=\pc{Global gain vector (identity)} \plabel{fgain} \\
\pl{0}  $\lambda \pa [10, 1, 1, 1]$\,$/$\,$13$ \>\pc{Expected power spectrum $(1/f)$} \plabel{flambda} \\
\pl{0}  \pk{for} $i = 1, 2, 3, 4$ \pk{do} \plabel{floop} \\
\pl{1}      \pifaug{amplification for $b_i$}{p} \plabel{fapply} \\
\pl{2}          $t \pa [1, 1, 1, 1]$          \>\pc{Temporary gain vector} \\
\pl{2}          \pk{sample} $t_i \sim \mathrm{Lognormal}\big( 0, (1\cdot\ln2)^2 \big)$ \plabel{fsample} \\
\pl{2}          $t \pa t \big/ \sqrt{\sum_j \lambda_j t^2_j}$ \hspace*{0.5mm} \pc{Normalize power} \plabel{fnormalize} \\
\pl{2}          $g \pa g \odot t$             \>\pc{Accumulate into global gain} \plabel{filterend} \\[1.5mm]
\pl{0}  \pc{Execute image-space filtering} \plabel{fexecbegin} \\
\pl{0}  $H(z) \pa \pf{Wavelet}(\pf{sym2})$ \hspace*{0.5mm} \pc{Orthogonal 4-tap filter bank} \\
\pl{0}  $H'(z) \pa 0$                          \>\pc{Combined amplification filter} \plabel{fprimebegin} \\
\pl{0}  \pk{for} $i = 1, 2, 3, 4$ \pk{do} \\
\pl{1}      $H'(z) \pa H'(z) + \pf{Bandpass}\big( H(z), b_i \big) \cdot g_i$ \plabel{fbandpass} \plabel{fprimeend} \\
\pl{0}  $(m_\ps{lo}, m_\ps{hi}) \pa \pf{CalculatePadding}\big( H'(z) \big)$ \\
\pl{0}  $Y \pa \pf{Pad}(Y, m_\ps{lo}, m_\ps{hi}, \pf{reflect})$ \plabel{fconvbegin} \\
\pl{0}  $Y \pa \pf{SeparableConv2D}\big( Y, H'(z) \big)$ \plabel{fseparable} \\
\pl{0}  $Y \pa \pf{Crop}(Y, m_\ps{lo}, m_\ps{hi})$ \plabel{fexecend} \plabel{fconvend} \\[1.5mm]
\pl{0}  \pc{Additive RGB noise} \plabel{noisebegin} \\
\pl{0}  \pifaug{noise}{p} \\
\pl{1}      \pk{sample} $\sigma \sim \mathrm{Halfnormal}\big( (0.1)^2 \big)$ \plabel{nsample} \\
\pl{1}      \pk{for each} pixel $(x,y) \in Y$ \pk{do} \\
\pl{2}          \pk{sample} $n_r, n_g, n_b \sim \mathcal{N}(0, \sigma^2)$ \\
\pl{2}          $Y_{x,y} \pa Y_{x,y} + [n_r, n_g, n_b]$ \plabel{noiseend} \\[1.5mm]
\pl{0}  \pc{Cutout} \plabel{cutoutbegin} \\
\pl{0}  \pifaug{cutout}{p} \\
\pl{1}      \pk{sample} $c_x, c_y \sim \mathcal{U}(0, 1)$ \\
\pl{1}      $r_\ps{lo} \pa \text{round}\Big( \Big[ \big( c_x - \frac{1}{4} \big) \cdot w, \big( c_y - \frac{1}{4} \big) \cdot h \Big] \Big)$ \\[0.2mm]
\pl{1}      $r_\ps{hi} \pa \text{round}\Big( \Big[ \big( c_x + \frac{1}{4} \big) \cdot w, \big( c_y + \frac{1}{4} \big) \cdot h \Big] \Big)$ \\
\pl{1}      $Y \pa Y \odot \big( 1 - \pf{RectangularMask}(r_\ps{lo}, r_\ps{hi}) \big)$ \plabel{cutoutend} \\
\pl{0}  \pk{return} $Y$
\end{tabbing}%
}

\newcommand{\figFilterNoiseCutout}[1]{
\begin{figure}[t]
\scriptsize%
\renewcommand{\h}{0.472\linewidth}%
\renewcommand{\hh}{0.5\linewidth}%
\renewcommand{\vv}{113.3mm}%
\renewcommand{\vvv}{14.1mm}%
\renewcommand{\vvvv}{1.5mm}%
\vspace{0.23mm}%
\parbox[b][\vv]{\h}{\vspace{-1.5mm}\algFilterNoiseCutout\vspace{-6mm}\hfill\vfill}%
\hfill\begin{tikzpicture}\draw[gray] (0,0) -- (0,\vv);\end{tikzpicture}\hfill%
\parbox[b][\vv]{\hh}{
\setlength{\lineskip}{0mm}%
\setlength{\baselineskip}{0mm}%
\makebox[\linewidth-\vvv*\real{4}][c]{Percentile:}\hfill%
\makebox[\vvv][c]{5\textsuperscript{th}}%
\makebox[\vvv][c]{35\textsuperscript{th}}%
\makebox[\vvv][c]{65\textsuperscript{th}}%
\makebox[\vvv][c]{95\textsuperscript{th}}%
\vfill\augsep\vfill%
\makebox[\linewidth][c]{\footnotesize\bf Image-space filtering}%
\vfill\augsep\vfill%
\augrow{12_filter0}{Frequency\\band $b_1$\\[0.5mm]$\left[ 0, \frac{\pi}{8} \right]$}{\vvv}\\[\vvvv]%
\augrow{13_filter1}{Frequency\\band $b_2$\\[0.5mm]$\left[ \frac{\pi}{8}, \frac{\pi}{4} \right]$}{\vvv}\\[\vvvv]%
\augrow{14_filter2}{Frequency\\band $b_3$\\[0.5mm]$\left[ \frac{\pi}{4}, \frac{\pi}{2} \right]$}{\vvv}\\[\vvvv]%
\augrow{15_filter3}{Frequency\\band $b_4$\\[0.5mm]$\left[ \frac{\pi}{2}, \pi \right]$}{\vvv}%
\vfill\augsep\vfill%
\makebox[\linewidth][c]{\footnotesize\bf Image-space corruptions}%
\vfill\augsep\vfill%
\augrow{16_noise}{Additive\\RGB noise}{\vvv}\\[\vvvv]%
\augrow{17_cutout}{Cutout}{\vvv}%
}%
\vspace{1.5mm}%
\caption{
Pseudocode and example images for image-space filtering and corruptions (Appendix~\ref{app:FilterNoiseCutout}).
$x \odot y$ denotes element-wise multiplication.
}
\label{#1}
\end{figure}
}

\newcommand{\YES}{\checkmark}
\newcommand{\NO}{ --}
\newcommand{\REZ}[1]{#1$\times$#1}
\newcommand{\FMfull}{$1\times$}
\newcommand{\FMhalf}{$\frac{1}{2}\times$}
\newcommand{\POT}[1]{\raisebox{0.2mm}{\scalebox{0.8}{$\times10^{#1}$}}}

\newcommand{\figHyperparams}[1]{
\begin{figure}[t]
\footnotesize%
\centering%
\newcolumntype{x}{>{\centering\arraybackslash\hspace{0pt}}m{17mm}}%
\renewcommand{\cs}{\hspace{1.1mm}}%
\tabulinesep=0.7mm%
\tabulinestyle{0.17mm}%
\begin{tabu}{|l|@{\cs}x@{\cs}x@{\cs}x@{\cs}x@{\cs}x@{\cs}x@{\cs}|}
\tabucline{-}
& & & & & & \\[-2.5mm]
{\bf Parameter} & {\bf StyleGAN2 config F} & {\bf Our baseline} & {\bf BreCaHAD, \FINAL{AFHQ}} & {\bf MetFaces} & \FINAL{\bf CIFAR-10} & \FINAL{\bf + Tuning} \\
& & & & & & \\[-2.5mm]
\tabucline{-}
Resolution                     & \REZ{1024}  & \REZ{256}  & \REZ{512}  & \REZ{1024}  & \REZ{32}  & \REZ{32}  \\
Number of GPUs                 & 8           & 8          & 8          & 8           & 2         & 2         \\
Training length                & 25M         & 25M        & 25M        & 25M         & 100M      & 100M      \\
Minibatch size                 & 32          & 64         & 64         & 32          & 64        & 64        \\
\FINAL{Minibatch stddev}       & \FINAL{4}   & \FINAL{8}  & \FINAL{8}  & \FINAL{4}   & \FINAL{32} & \FINAL{32} \\
Dataset $x$-flips              & \YES$/$\NO  & \NO        & \YES       & \YES        & \NO       & \NO       \\
\tabucline{-}
Feature maps                   & \FMfull     & \FMhalf    & \FMfull    & \FMfull     & 512       & 512       \\
Learning rate $\eta$\POT{3}    & 2           & 2.5        & 2.5        & 2           & 2.5       & 2.5       \\
$R_1$ regularization $\gamma$  & 10          & 1          & 0.5        & 2           & 0.01      & 0.01      \\
G moving average               & 10k         & 20k        & 20k        & 10k         & \FINAL{500k} & \FINAL{500k} \\ %
Mixed-precision                & \NO         & \YES       & \YES       & \YES        & \YES      & \YES      \\
\tabucline{-}
Mapping net depth              & 8           & 8          & 8          & 8           & 8         & 2         \\
Style mixing reg.              & \YES        & \YES       & \YES       & \YES        & \YES      & \NO       \\
Path length reg.               & \YES        & \YES       & \YES       & \YES        & \YES      & \NO       \\
Resnet D                       & \YES        & \YES       & \YES       & \YES        & \YES      & \NO       \\
\tabucline{-}
\end{tabu}%
\caption{
Hyperparameters used in each experiment.
}
\label{#1}
\end{figure}
}

\newcommand{\SUBT}[2]{{\color{grey}\hspace{4mm}\makebox[12.5mm][l]{#1}#2}}
\newcommand{\SUBV}[1]{{\color{grey}#1}}
\newcommand{\figPower}[1]{
\begin{figure}[t]
\centering%
\footnotesize%
\newcolumntype{x}{>{\arraybackslash\hspace{0pt}}p{52.9mm}}%
\newcolumntype{y}{>{\centering\arraybackslash\hspace{0pt}}p{23mm}}%
\begin{tabular}{|@{\hspace{2.5mm}}x|y|y|y|}
\hline
\multirow{2}{*}{\bf Item}               & {\bf Number of}       & {\bf GPU years}   & {\bf Electricity} \\
                                        & {\bf training runs}   & {\bf (Volta)}     & {\bf (MWh)}       \\\hline
Early exploration                       & \s253                 & \s22.65           & \s52.05           \\
Paper exploration                       & 1116                  & \s36.54           & \s87.39           \\
Setting up the baselines                & \s251                 & \s12.19           & \s30.70           \\
Paper figures                           & \s960                 & \s50.53           & 108.02            \\[0.7mm]
\SUBT{Fig.1}{Baseline convergence}      & \SUBV{\s\s21}         & \SUBV{\s\s1.01}   & \SUBV{\s\s2.27}   \\
\SUBT{Fig.3}{Leaking behavior}          & \SUBV{\s\s78}         & \SUBV{\s\s3.62}   & \SUBV{\s\s7.93}   \\
\SUBT{Fig.4}{Augmentation categories}   & \SUBV{\s\s90}         & \SUBV{\s\s4.45}   & \SUBV{\s\s9.40}   \\
\SUBT{Fig.5}{ADA heuristics}            & \SUBV{\s\s61}         & \SUBV{\s\s3.16}   & \SUBV{\s\s6.87}   \\
\SUBT{Fig.6}{ADA convergence}           & \SUBV{\s\s15}         & \SUBV{\s\s0.78}   & \SUBV{\s\s1.70}   \\
\SUBT{Fig.7}{Training set sweeps}       & \SUBV{\s174}          & \SUBV{\s10.82}    & \SUBV{\s22.70}    \\
\SUBT{Fig.8a}{Comparison methods}       & \SUBV{\s\s69}         & \SUBV{\s\s4.18}   & \SUBV{\s\s8.64}   \\
\SUBT{Fig.8b}{Discriminator capacity}   & \SUBV{\s144}          & \SUBV{\s\s7.70}   & \SUBV{\s15.93}    \\
\SUBT{Fig.9}{Transfer learning}         & \SUBV{\s\s40}         & \SUBV{\s\s0.71}   & \SUBV{\s\s1.67}   \\
\SUBT{Fig.11a}{Small datasets}          & \SUBV{\s\s30}         & \SUBV{\s\s1.71}   & \SUBV{\s\s4.15}   \\
\SUBT{Fig.11b}{CIFAR-10}                & \SUBV{\s\s30}         & \SUBV{\s\s0.93}   & \SUBV{\s\s2.71}   \\
\SUBT{Fig.19}{BigGAN comparison}        & \SUBV{\s\s54}         & \SUBV{\s\s3.34}   & \SUBV{\s\s7.12}   \\
\SUBT{Fig.20}{bCR leaks}                & \SUBV{\s\s40}         & \SUBV{\s\s2.19}   & \SUBV{\s\s4.57}   \\
\SUBT{Fig.21}{Cumulative augmentations} & \SUBV{\s114}          & \SUBV{\s\s5.93}   & \SUBV{\s12.36}    \\[0.7mm]
Results intentionally left out          & \s177                 & \s\s5.51          & \s11.78           \\
Wasted due to technical issues          & \s255                 & \s\s3.86          & \s\s8.39          \\
Code release                            & \s375                 & \s12.49           & \s26.71           \\\hline
Total                                   & 3387                  & 143.76            & 325.06            \\\hline
\end{tabular}
\caption{
\FINAL{
Computational effort expenditure and electricity consumption data for this project. The unit for computation is GPU-years on a single NVIDIA V100 GPU\,---\,it would have taken approximately 135 years to execute this project using a single GPU. See the text for additional details about the computation and energy consumption estimates.
\emph{Early exploration} includes all training runs that affected our decision to start this project.
\emph{Paper exploration} includes all training runs that were done specifically for this project, but were not intended to be used in the paper as-is.
\emph{Setting up the baselines} includes all hyperparameter tuning for the baselines.
\emph{Figures} provides a per-figure breakdown, and underlines that just reproducing all the figures would require over 50 years of computation on a single GPU.
\emph{Results intentionally left out} includes additional results that were initially planned, but then left out to improve focus and clarity.
\emph{Wasted due to technical issues} includes computation wasted due to code bugs and infrastructure issues.
\emph{Code release} covers testing and benchmarking related to the public release.
}}
\label{#1}
\end{figure}
}

	\setcounter{figure}{11}

\ifarxiv
\else
We present additional result images and comparisons in Appendix~\ref{app:results}.
We then proceed to describe our augmentation pipeline in Appendix~\ref{app:pipeline} and analyze the augmentations from a theoretical standpoint in Appendix~\ref{app:theory}.
Finally, we detail our training setup in Appendix~\ref{app:implementation} \FINAL{and present a detailed breakdown of our total energy consumption in Appendix~\ref{app:power}.}
\fi

\section{Additional results}
\label{app:results}

\figUncuratedMetFaces{fig:UncuratedMetFaces} %
\figUncuratedBreCaHAD{fig:UncuratedBreCaHAD} %
\figUncuratedCats{fig:UncuratedCats} %
\figUncuratedDogs{fig:UncuratedDogs} %
\figUncuratedWild{fig:UncuratedWild} %
\FINAL{In Figures \ref{fig:UncuratedMetFaces}, \ref{fig:UncuratedBreCaHAD}, \ref{fig:UncuratedCats}, \ref{fig:UncuratedDogs}, and \ref{fig:UncuratedWild}, we show generated images for \textsc{MetFaces}, \textsc{BreCaHAD}, and \textsc{AFHQ Cat, Dog, Wild}}, respectively, along with real images from the respective training sets (Section~\refpaper{sec:datasets} and \FINAL{Figure~\refpaper{fig:SmallDatasetResults}a}).
The images were selected at random; we did not perform any cherry-picking besides choosing one global random seed.
We can see that ADA yields excellent results in all cases, and with slight truncation~\cite{Marchesi2017,Karras2018}, virtually all of the images look convincing.
Without ADA, the convergence is hampered by discriminator overfitting, leading to inferior image quality for the original StyleGAN2, \FINAL{especially in \textsc{MetFaces}, \textsc{AFHQ Dog}, and \textsc{BreCaHAD}}.

\figUncuratedCIFAR{fig:UncuratedCIFAR} %
Figure~\ref{fig:UncuratedCIFAR} shows examples of the generated CIFAR-10 images in both unconditional and class-conditional setting (See Appendix~\ref{app:cifarconditioning} for details on the conditional setup).
\figSubsetImagesFFHQ{fig:SubsetImagesFFHQ} %
Figure~\ref{fig:SubsetImagesFFHQ} shows qualitative results for different methods using subsets of FFHQ at 256$\times$256 resolution.
Methods that do not employ augmentation (BigGAN, StyleGAN2, and our baseline) degrade noticeably as the size of the training set decreases, generally yielding poor image quality and diversity with fewer than 30k training images.
With ADA, the degradation is much more graceful, and the results remain reasonable even with a 5k training set.

\figBigGANComparison{fig:BigGANComparison} %
Figure~\ref{fig:BigGANComparison} compares our results with unconditional BigGAN~\cite{Brock2018,Schonfeld2020} and StyleGAN2 config \textsc{f}~\cite{Karras2019}.
BigGAN was very unstable in our experiments: while some of the results were quite good, approximately 50\% of the training runs failed to converge.
StyleGAN2, on the other hand, behaved predictably, with different training runs resulting in nearly identical FID.
We note that FID has a general tendency to increase as the training set gets smaller\,---\,not only because of the lower image quality, but also due to inherent bias in FID itself~\cite{KID2018}.
In our experiments, we minimize the impact of this bias by always computing FID between 50k generated images and all available real images, regardless of which subset was used for training.
To estimate the magnitude of bias in FID, we simulate a hypothetical generator that replicates the training set as-is, and compute the average FID over 100 random trials with different subsets of training data; the standard deviation was $\le$2\% in all cases.
We can see that the bias remains negligible with $\ge$20k training images but starts to dominate with $\le$2k.
Interestingly, ADA reaches the same FID as the best-case generator with FFHQ-1k, indicating that FID is no longer able to differentiate between the two in this case.

\figLeakyComparison{fig:LeakyComparison} %
Figure~\ref{fig:LeakyComparison} shows additional examples of bCR leaking to generated images and compares bCR with dataset augmentation.
In particular, rotations in range $[-45^{\circ}, +45^{\circ}]$ (denoted $\pm45^{\circ}$) serve as a very clear example that attempting to make the discriminator blind to certain transformations opens up the possibility for the generator to produce similarly transformed images with no penalty.
In applications where such leaks are acceptable, one can employ either bCR or dataset augmentation\,---\,%
we find that it is difficult to predict which method is better. For example, with translation augmentations bCR was significantly better than dataset augmentation, whereas $x$-flip was much more effective when implemented as a dataset augmentation.

\figFullSweep{fig:FullSweep} %
Finally, Figure~\ref{fig:FullSweep} shows an extended version of Figure~\refpaper{fig:FixedSweeps}, illustrating the effect of different augmentation categories with increasing augmentation probability $p$.
Blit + Geom + Color yielded the best results with a 2k training set and remained competitive with larger training sets as well.

\newpage
\section{Our augmentation pipeline}
\label{app:pipeline}

We designed our augmentation pipeline based on three goals.
First, the entire pipeline must be strictly non-leaking (Appendix~\ref{app:theory}).
Second, we aim for a maximally diverse set of augmentations, inspired by the success of RandAugment~\cite{Cubuk2019}.
Third, we strive for the highest possible image quality to reduce unintended artifacts such as aliasing.
In total, our pipeline consists of 18 transformations: geometric (7), color (5), filtering (4), and corruption (2).
We implement it entirely on the GPU in a differentiable fashion, with full support for batching.
All parameters are sampled independently for each image.

\subsection{Geometric and color transformations}
\label{app:BlitGeomColor}

\figBlitGeomColor{fig:BlitGeomColor} %
Figure~\ref{fig:BlitGeomColor} shows pseudocode for our geometric and color transformations, along with example images.
In general, geometric transformations tend to lose high-frequency details of the input image due to uneven resampling, which may reduce the capability of the discriminator to detect pixel-level errors in the generated images.
We alleviate this by introducing a dedicated sub-category, \emph{pixel blitting}, that only copies existing pixels as-is, without blending between neighboring pixels.
Furthermore, we avoid gradual image degradation from multiple consecutive transformations by collapsing all geometric transformations into a single combined operation.

The parameters for pixel blitting are selected on lines~\pref{blitbegin}--\pref{blitend}, consisting of $x$-flips (line~\pref{xflip}), 90$^{\circ}$ rotations (line~\pref{blitrot}), and integer translations (line~\pref{xint}).
The transformations are accumulated into a homogeneous $3\times3$ matrix $G$, defined so that input pixel $(x_i,y_i)$ is placed at \mbox{$[x_o,y_o,1]^T = G \cdot [x_i,y_i,1]^T$} in the output.
The origin is located at the center of the image and neighboring pixels are spaced at unit intervals.
We apply each transformation with probability $p$ by sampling its parameters from uniform distribution, either discrete $\mathcal{U}\{\cdot\}$ or continuous $\mathcal{U}(\cdot)$, and updating $G$ using elementary transforms:
\begin{equation}
\resizebox{0.94\hsize}{!}{$
\pf{\footnotesize Scale2D}(s_x, s_y) =
\left[\begin{matrix}
	s_x & 0   & 0 \\
	0   & s_y & 0 \\
	0   & 0   & 1 \\
\end{matrix}\right]%
,\hspace{1mm}
\pf{\footnotesize Rotate2D}(\theta) =
\left[\begin{matrix}
	\cos\theta & -\sin\theta & 0 \\
	\sin\theta & \cos\theta  & 0 \\
	0 & 0 & 1 \\
\end{matrix}\right]%
,\hspace{1mm}
\pf{\footnotesize Translate2D}(t_x, t_y) =
\left[\begin{matrix}
	1 & 0 & t_x \\
	0 & 1 & t_y \\
	0 & 0 & 1   \\
\end{matrix}\right]%
$}
\end{equation}
General geometric transformations are handled in a similar way on lines~\pref{geombegin}--\pref{geomend}, consisting of isotropic scaling (line~\pref{scale}), arbitrary rotation (lines~\pref{prerot} and \pref{postrot}), anisotropic scaling (line~\pref{aniso}), and fractional translation (line~\pref{xfrac}).
Since both of the scaling transformations are multiplicative in nature, we sample their parameter, $s$, from a log-normal distribution so that $\ln s \sim \mathcal{N}\big( 0, (0.2\cdot\ln2)^2 \big)$.
In practice, this can be done by first sampling $t \sim \mathcal{N}(0,1)$ and then calculating $s = \exp_2(0.2t)$.
We allow anisotropic scaling to operate in other directions besides the coordinate axes by breaking the rotation into two independent parts, one applied before the scaling (line~\pref{prerot}) and one after it (line~\pref{postrot}).
We apply the rotations slightly less frequently than other transformations, so that the probability of applying \emph{at least one} rotation is equal to $p$.
Note that we also have two translations in our pipeline (lines~\pref{xint} and \pref{xfrac}), one applied at the beginning and one at the end.
To increase the diversity of our augmentations, we use $\mathcal{U}(\cdot)$ for the former and $\mathcal{N}(\cdot)$ for the latter.

Once the parameters are settled, the combined geometric transformation is executed on lines~\pref{padbegin}--\pref{gexecend}.
We avoid undesirable effects at image borders by first padding the image with reflection.
The amount of padding is calculated dynamically based on $G$ so that none of the output pixels are affected by regions outside the image (line~\pref{gmargin}).
We then upsample the image to a higher resolution (line~\pref{gupsample}) and transform it using bilinear interpolation (line~\pref{gbilinear}).
Operating at a higher resolution is necessary to reduce aliasing when the image is minified, e.g., as a result of isotropic scaling\,---\,%
interpolating at the original resolution would fail to correctly filter out frequencies above Nyquist in this case, no matter which interpolation filter was used.
The choice of the upsampling filter requires some care, however, because we must ensure that an identity transform does not modify the image in any way (e.g., when $p=0$).
In other words, we need to use a lowpass filter $H(z)$ with cutoff $f_c=\frac{\pi}{2}$ that satisfies $\pf{Downsample2D}\big( \pf{Upsample2D}\big( Y, H(z^{-1}) \big), H(z) \big) = Y$.
Luckily, existing literature on wavelets~\cite{Daubechies1992} offers a wide selection of such filters; we choose 12-tap symlets (\textsc{sym6}) to strike a balance between resampling quality and computational cost.

Finally, color transformations are applied to the resulting image on lines~\pref{colorbegin}--\pref{cexecend}.
The overall operation is similar to geometric transformations: we collect the parameters of each individual transformation into a homogeneous $4\times4$ matrix $C$ that we then apply to each pixel by computing \mbox{$[r_o,g_o,b_o,1]^T = C \cdot [r_i,g_i,b_i,1]^T$}.
The transformations include adjusting brightness (line~\pref{brightness}), contrast (line~\pref{contrast}), and saturation (line~\pref{saturation}), as well as flipping the luma axis while keeping the chroma unchanged (line~\pref{lumaflip}) and rotating the hue axis by an arbitrary amount (line~\pref{hue}).

\subsection{Image-space filtering and corruptions}
\label{app:FilterNoiseCutout}

\figFilterNoiseCutout{fig:FilterNoiseCutout} %
Figure~\ref{fig:FilterNoiseCutout} shows pseudocode for our image-space filtering and corruptions.
The parameters for image-space filtering are selected on lines~\pref{filterbegin}--\pref{filterend}.
The idea is to divide the frequency content of the image into 4 non-overlapping bands and amplify/weaken each band in turn via a sequence of 4 transformations, so that each transformation is applied independently with probability $p$ (lines~\pref{floop}--\pref{fapply}).
Frequency bands $b_2$, $b_3$, and $b_4$ correspond to the three highest octaves, respectively, while the remaining low frequencies are attributed to $b_1$ (line~\pref{fbands}).
We track the overall gain of each band using vector $g$ (line~\pref{fgain}) that we update after each transformation (line~\pref{filterend}).
We sample the amplification factor for a given band from log-normal distribution (line~\pref{fsample}), similar to geometric scaling, and normalize the overall gain so that the total energy is retained on expectation.
For the normalization, we assume that the frequency content obeys $1/f$ power spectrum typically seen in natural images (line~\pref{flambda}).
While this assumption is not strictly true in our case, especially when some of the previous frequency bands have already been amplified, it is sufficient to keep the output pixel values within reasonable bounds.

The filtering is executed on lines~\pref{fexecbegin}--\pref{fexecend}.
We first construct a combined amplification filter $H'(z)$ (lines~\pref{fprimebegin}--\pref{fprimeend}) and then perform separable convolution for the image using reflection padding (lines~\pref{fconvbegin}--\pref{fconvend}).
We use a zero-phase filter bank derived from 4-tap symlets (\textsc{sym2})~\cite{Daubechies1992}.
Denoting the wavelet scaling filter by $H(z)$, the corresponding bandpass filters are obtained as follows (line~\pref{fbandpass}):
\FINAL{%
\begin{eqnarray}
\pf{Bandpass}\big( H(z), b_1 \big) &=& H(z) H(z^{-1}) H(z^2) H(z^{-2}) H(z^4) H(z^{-4}) / 8 \\
\pf{Bandpass}\big( H(z), b_2 \big) &=& H(z) H(z^{-1}) H(z^2) H(z^{-2}) H(-z^4) H(-z^{-4}) / 8 \\
\pf{Bandpass}\big( H(z), b_3 \big) &=& H(z) H(z^{-1}) H(-z^2) H(-z^{-2}) / 4 \\
\pf{Bandpass}\big( H(z), b_4 \big) &=& H(-z) H(-z^{-1}) / 2
\end{eqnarray}
}%
Finally, we apply additive RGB noise on lines~\pref{noisebegin}--\pref{noiseend} and cutout on lines~\pref{cutoutbegin}--\pref{cutoutend}.
We vary the strength of the noise by sampling its standard deviation from half-normal distribution, i.e., $\mathcal{N}(\cdot)$ restricted to non-negative values (line~\pref{nsample}).
For cutout, we match the original implementation of DeVries~and~Taylor~\cite{Devries2017} by setting pixels to zero within a rectangular area of size $\left( \frac{w}{2}, \frac{h}{2} \right)$, with the center point selected from uniform distribution over the entire image.

\section{Non-leaking augmentations}
\label{app:theory}

The goal of GAN training is to find a generator function $G$ whose output probability distribution $\probx$ (under suitable stochastic input) matches a given target distribution $\proby$.

When augmenting both the dataset and the generator output, the key safety principle is that if $\probx$ and $\proby$ do not match, then their augmented versions must not match either. If the augmentation pipeline violates this principle, the generator is free to learn some different output distribution than the dataset, as these look identical after the augmentations -- we say that the augmentations \emph{leak}. Conversely, if the principle holds, then the only option for the generator is to learn the correct distribution: no other choice results in a post-augmentation match.

In this section, we study the conditions on the augmentation pipeline under which this holds and demonstrate the safety and caveats of various common augmentations and their compositions.

\paragraph{Notation} Throughout this section, we denote probability distributions (and their generalizations) with lowercase bold-face letters (e.g., $\probx$), operators acting on them by calligraphic letters ($\aug$), and variates sampled from probability distributions by upper-case letters ($X$).

\subsection{Augmentation operator}

A very general model for augmentations is as follows. Assume a fixed but arbitrarily complicated non-linear and stochastic augmentation pipeline. To any image $X$, it assigns a \emph{distribution} of augmented images, such as demonstrated in Figure~\refpaper{fig:Augment}c. 
This idea is captured by an \emph{augmentation operator} $\aug$ that maps probability distributions to probability distributions (or, informally, datasets to augmented datasets). A distribution with the lone image $X$ is the Dirac point mass $\delta_X$, which is mapped to some distribution $\aug \delta_X$ of augmented images.\footnote{These distributions are \emph{probability measures} over a non-discrete high dimensional space: for example, in our experiments with $256 \times 256$ RGB images, this space is $\real^{256*256*3} = \real^{196608}$.} In general, applying $\aug$ to an arbitrary distribution $\probx$ yields the linear superposition $\aug \probx$ of such augmented distributions. 

It is important to understand that $\aug$ is different from a function $f(X;\phi)$ that actually applies the augmentation on any individual image $X$ sampled from $\probx$ (parametrized by some $\phi$, e.g., angle in case of a rotation augmentation). It captures the \emph{aggregate} effect of applying this function on all images in the distribution and subsumes the randomization of the function parameters. $\aug$ is always linear and deterministic, regardless of non-linearity of the function $f$ and stochasticity of its parameters $\phi$. We will later discuss \emph{invertibility} of $\aug$. Here it is also critical to note that its invertibility is not equivalent with the invertibility of the function $f$ it is based on; for an example, refer to the discussion in Section~\refpaper{sec:theory}.

Specifically, $\aug$ is a \emph{(Markov) transition operator}. Intuitively, it is an (uncountably) infinite-dimensional generalization of a Markov transition matrix (i.e. a stochastic matrix), with nonnegative entries that sum to $1$ along columns. In this analogy, probability distributions upon which $\aug$ operates are vectors, with nonnegative entries summing to $1$. %
More generally, the distributions have a vector space structure and they can be arbitrarily linearly combined (in which case they may lose their validity as probability distributions and are viewed as arbitrary \emph{signed measures}). Similarly, we can do algebra with the with the operators by linearly combining and composing them like matrices. Concepts such as null space and invertibility carry over to this setting, with suitable technical care. In the following, we will be somewhat informal with the measure theoretical and functional analytic details of the problem, and draw upon this analogy as appropriate.\footnote{The addition and scalar multiplication of measures is taken to mean that for any set $S$ to which $\probx$ and $\proby$ assign a measure, $[\alpha \probx + \beta \proby](S) = \alpha \probx(S) + \beta \proby(S)$. When the measures are represented by density functions, this simplifies to the usual pointwise linear combination of the functions. We always mean addition and scalar multiplication of probability distributions in this sense (as opposed to e.g. addition of random variables), unless otherwise noted.

Technically, one can consider the vector space of finite signed measures on $\real^N$, which is a Banach space under the Total Variation norm. Markov operators form a convex subset of linear operators acting on this space, and general linear combinations thereof form a subspace (and a subalgebra). The exact mathematical conditions under which some of the following findings apply may be intricate but have limited practical significance given the approximate nature of GAN training.}%

\subsection{Invertibility implies non-leaking augmentations}

Within this framework, our question can be stated as follows. Given a target distribution $\proby$ and an augmentation operator $\aug$, we train for a generated distribution $\probx$ such that the augmented distributions match, namely 
\begin{equation}
	\aug \probx = \aug \proby.
	\label{eq:match}
\end{equation}
The desired outcome is that this equation is satisfied only by the correct target distribution, namely $\probx = \proby$. We say that $\aug$ \emph{leaks} if there exist distributions $\probx \neq \proby$ that satisfy the above equation, and the goal is to find conditions that guarantee the absence of leaks. %

There are obviously no such leaks in classical non-augmented training, where $\aug$ is the identity $\augeye$, whence $\aug \probx = \aug \proby \Rightarrow \augeye \probx = \augeye \proby \Rightarrow \probx = \proby$. For arbitrary augmentations, the desired outcome $\probx = \proby$ does always satisfy Eq.~\ref{eq:match}; however, if also other choices of $\probx$ satisfy it, then it cannot be guaranteed that the training lands on the desired solution. A trivial example is an augmentation that maps every image to black (in other words, $\aug \probz = \delta_0$ for any $\probz$). Then, $\aug \probx = \aug \proby$ does not imply that $\probx = \proby$, as indeed any choice of $\probx$ produces the same set of black images that satisfies Eq.~\ref{eq:match}. In this case, it is vanishingly unlikely that the training finds the solution $\probx = \proby$. 

More generally, assume that $\aug$ has a non-trivial null space, namely there exists a signed measure $\probn \neq 0$ such that $\aug \probn = 0$, that is, $\probn$ is in the null space of $\aug$. Equivalently, $\aug$ is not invertible, because $\probn$ cannot be recovered from $\aug \probn$. Then, $\probx = \proby + \alpha \probn$ for any $\alpha \in \real$ satisfies Eq.~\ref{eq:match}. Therefore non-invertibility of $\aug$ implies that measures in its null space may freely leak into the learned distribution (as long as the sum remains a valid probability distribution that assigns non-negative mass to all sets). Conversely, assume that some $\probx \neq \proby$ satisfies Eq.~\ref{eq:match}. Then $\aug (\probx - \proby) = \aug \proby - \aug \proby = 0$, so $\probx - \proby$ is in null space of $\aug$ and therefore $\aug$ is not invertible. %

Therefore, leaking augmentations imply non-invertibility of the augmentation operator, which conversely implies the central principle: \textbf{if the augmentation operator $\aug$ is invertible, it does not leak.} Such a non-leaking operator further satisfies the requirements of Lemma 5.1. of Bora et al.~\cite{Bora2018}, where the invertibility is shown to imply that a GAN learns the correct distribution.

The invertibility has an intuitive interpretation: the training process can implicitly ``undo'' the augmentations, as long as probability mass is merely shifted around and not squashed flat.

\subsection{Compositions and mixtures}
We only access the operator $\aug$ indirectly: it is implemented as a procedure, rather than a matrix-like entity whose null space we could study directly (even if we know that such a thing exists in principle). Showing invertibility for an arbitrary procedure is likely to be impossible. Rather, we adopt a \emph{constructive} approach, and build our augmentation pipeline from combinations of simple known-safe augmentations, in a way that can be shown to not leak. This calls for two components: a set of combination rules that preserve the non-leaking guarantee, and a set of elementary augmentations that have this property. In this subsection we address the former.

By elementary linear algebra: assume $\aug$ and $\augu$ are invertible. Then the composition $\aug \augu$ is invertible, as is any finite chain of such compositions. Hence, \textbf{sequential composition of non-leaking augmentations is non-leaking}. We build our pipeline on this observation.

The other obvious combination of augmentations is obtained by probabilistic mixtures: given invertible augmentations $\aug$ and $\augu$, perform $\aug$ with probability $\alpha$ and $\augu$ with probability $1-\alpha$. The operator corresponding to this augmentation is the ``pointwise'' convex blend $\alpha \aug + (1-\alpha) \augu$. More generally, one can mix e.g. a continuous family of augmentations $\aug_\phi$ with weights given by a non-negative unit-sum function $\alpha(\phi)$, as $\int \alpha(\phi) \aug_\phi ~ \mathrm{d}\phi$. Unfortunately, \textbf{stochastically choosing among a set of augmentations is \emph{not} guaranteed to preserve the non-leaking property}, and must be analyzed case by case (which is the content of the next subsection). To see this, consider an extremely simple discrete probability space with only two elements. The augmentation operator $\aug = \big( \begin{smallmatrix} 0 & 1 \\ 1 & 0 \end{smallmatrix} \big)$ flips the elements. Mixed with probability $\alpha = \frac{1}{2}$ with the identity augmentation $\augeye$ (which keeps the distribution unchanged), we obtain the augmentation $\frac{1}{2} \aug + \frac{1}{2}\augeye = \frac{1}{2} \big( \begin{smallmatrix} 1 & 1 \\ 1 & 1 \end{smallmatrix} \big)$ which is a singular matrix and therefore not invertible. Intuitively, this operator smears any probability distribution into a degenerate equidistribution, from which the original can no longer be recovered. Similar considerations carry over to arbitrarily complicated linear operators. 

\subsection{Non-leaking elementary augmentations}
In the following, we construct several examples of relatively large classes of elementary augmentations that do not leak and can therefore be used to form a chain of augmentations. Importantly, most of these classes are not inherently safe, as they are stochastic mixtures of even simpler augmentations, as discussed above. However, in many cases we can show that the degenerate situation only arises with specific choices of mixture distribution, which we can then avoid.

Specifically, for every type of augmentation, we identify a configuration where applying it with probability strictly less than 1 results in an invertible transformation. From the standpoint of this analysis, we interpret this stochastic skipping as modifying the augmentation operator itself, in a way that boosts the probability of leaving the input unchanged and reduces the probability of other outcomes.

\subsubsection{Deterministic mappings}

The simplest form of augmentation is a deterministic mapping, where the operator $\aug_f$ assigns to every image $X$ a unique image $f(X)$. %
In the most general setting $f$ is any measurable function and $\aug_f \probx$ is the corresponding pushforward measure. When $f$ is a diffeomorphism,
$\aug_f$ acts by the usual change of variables formula with a density correction by a Jacobian determinant.
These mappings are invertible as long as $f$ itself is invertible. %
Conversely, if $f$ is not invertible, then neither is $\aug_f$.

Here it may be instructive to highlight the difference between $f$ and $\aug_f$. The former transforms the underlying space on which the probability distributions live -- for example, if we are dealing with images of just two pixels (with continuous and unconstrained values), $f$ is a nonlinear ``warp'' of the two-dimensional plane. In contrast, $\aug_f$ operates on distributions defined on this space -- think of a continuous 2-dimensional function (density) on the aforementioned plane. The action of $\aug_f$ is to move the density around according to $f$, while compensating for thinning and concentration of the mass due to stretching. As long as $f$ maps every distinct point to a distinct point, this warp can be reversed.

An important special case is that where $f$ is a linear transformation of the space. Then the invertibility of $\aug_f$ becomes a simpler question of the invertibility of a finite-dimensional matrix that represents $f$.

Note that when an invertible deterministic transformation is skipped probabilistically, the determinism is lost, and very specific choices of transformation could result in non-invertibility (see e.g. the example of flipping above). We only use deterministic mappings as building blocks of other augmentations, and never apply them in isolation with stochastic skipping.

\subsubsection{Transformation group augmentations}

Many commonly used augmentations are built from transformations that act as a \emph{group} under sequential composition. %
Examples of this are flips, translations, rotations, scalings, shears, and many color and intensity transformations. We show that a stochastic mixture of transformations within a finitely generated abelian group is non-leaking as long as the mixture weights are chosen from a non-degenerate distribution.

As an example, the four deterministic augmentations $\{\augr_{0}, \augr_{90}, \augr_{180}, \augr_{270}\}$ that rotate the images to every one of the 90-degree increment orientations constitute a group. This is seen by checking that the set satisfies the axiomatic definition of a group. Specifically, the set is \emph{closed}, as composing two of elements always results in an element of the same set, e.g. $\augr_{270} \augr_{180} = \augr_{90}$. It is also obviously associative, and has an identity element $\augr_{0} = \augeye$. Finally, every element has an inverse, e.g. $\augr_{90}^{-1} = \augr_{270}$. We can now simply speak of powers of the single generator element, whereby the four group elements are written as $\{\augr_{90}^0, \augr_{90}^1, \augr_{90}^2, \augr_{90}^3\}$ and further (as well as negative) powers ``wrap over'' to the same elements. This group is isomorphic to $\integer_4$, the additive group of integers modulo $4$.

A group of rotations is \emph{compact} due to the wrap-over effect. An example of a \emph{non-compact} group is that of translations (with non-periodic boundary conditions): compositions of translations are still translations, but one cannot wrap over. Furthermore, more than one generator element can be present (e.g. y-translation in addition to x-translation), but we require that these commute, i.e. the order of applying the transformations must not matter (in which case the group is called \emph{abelian}).

Similar considerations extend to continuous \emph{Lie groups}, e.g. that of rotations by any angle; here the generating element is replaced by an infinitesimal generator from the corresponding \emph{Lie algebra}, and the discrete powers by the continuous exponential mapping. For example, continuous rotation transformations are isomorphic to the group $\mathrm{SO}(2)$, or $\mathrm{U}(1)$.

In the following subsections show that \textbf{for finitely generated abelian groups whose identity element matches the identity augmentation, stochastic mixtures of augmentations within the group are invertible, as long as the appropriate Fourier transform of the probability distribution over the elements has no zeros.}%

\paragraph{Discrete compact one-parameter groups}
We demonstrate the key points in detail with the simple but relevant case of a discrete compact one-parameter group and generalize later. Let $\augg$ be a deterministic augmentation that generates the finite cyclic group $\{\augg^i\}_{i=0}^{N-1}$ of order $N$ (e.g. the four 90-degree rotations above), such that the element $\augg^0$ is the identity mapping that leaves its input unchanged.

Consider a stochastic augmentation $\aug$ that randomly applies an element of the group, with the probability of choosing each element given by the probability vector $p \in \real^N$ (where $p$ is nonnegative and sums to $1$):
\begin{equation}
\aug = \sum_{i=0}^{N-1} p_i \augg^i
\end{equation}

To show the conditions for invertibility of $\aug$, we build an operator $\augu$ that explicitly inverts $\aug$, namely $\augu \aug = \augeye = \augg^0$. Whenever this is possible, $\aug$ is invertible and non-leaking. We build $\augu$ from the same group elements with a different weighting\footnote{Unlike with $p$, there is no requirement for $q$ to represent a nonnegative probability density that sums to $1$, as we are establishing the general invertibility of $\aug$ without regard to its probabilistic interpretation. Note that $\augu$ is never actually constructed or evaluated when applying our method in practice, and does not need to represent an operation that can be algorithmically implemented; our interest is merely to identify the conditions for its existence.} vector $q \in \real^N$:
\begin{equation}
	\augu = \sum_{j=0}^{N-1} q_j \augg^j
\end{equation}

We now seek a vector $q$ for which $\augu \aug = \augeye$, that is, for which $\augu$ is the desired inverse. Now,
\begin{eqnarray}
	\augu \aug &=& \left( \sum_{i=0}^{N-1} p_i \augg^{i} \right) \left( \sum_{j=0}^{N-1} q_j \augg^{j} \right) \\
	&=& \sum_{i,j=0}^{N-1} p_i q_j \augg^{i+j}
\end{eqnarray}
The powers of the group operation, as well as the indices of the weight vectors, are taken as modulo $N$ due to the cyclic wrap-over of the group element. Collecting the terms that correspond to each $\augg^k$ in this range and changing the indexing accordingly, we arrive at:
\newcommand{\fourier}{\mathbf{F}}
\begin{eqnarray}
	&=& \sum_{k=0}^{N-1} \left[\sum_{l=0}^{N-1} p_{l} q_{k-l} \right] \augg^{k} \\
	&=& \sum_{k=0}^{N-1} [p \otimes q]_k \augg^{k}
\end{eqnarray}
where we observe that the multiplier in front of each $\augg^k$ is given by the cyclic convolution of the elements of the vectors $p$ and $q$. This can be written as a pointwise product in terms of the Discrete Fourier Transform $\fourier$, denoting the DFT's of $p$ and $q$ by a hat:
\begin{eqnarray}
	&=& \sum_{k=0}^{N-1} [\fourier^{-1} (\hat p \odot \hat q)]_k \augg^{k}
\end{eqnarray}
To recover the sought after inverse, assuming every element of $\hat p$ is nonzero, we set $\hat q_i = \frac{1}{\hat p_i}$ for all $i$:
\begin{eqnarray}
	&=& \sum_{k=0}^{N-1} [\fourier^{-1} (\hat p \odot \hat p^{-1})]_k \augg^{k} \\
	&=& \sum_{k=0}^{N-1} [\fourier^{-1} \mathbf{1}]_k \augg^{k} \\
	&=& \augg^{0} \\
	&=& \augeye
\end{eqnarray}
Here, we take advantage of the fact that the inverse DFT of a constant vector of ones is the vector $[1,~0,~...,~0]$.

In summary, the product of $\augu$ and $\aug$ effectively computes a convolution between their respective group element weights. This convolution assigns all of the weight to the identity element precisely when one has $\hat q_i = \frac{1}{\hat p_i}$, for all $i$, whereby $\augu$ is the inverse of $\aug$. This inverse only exists when the Fourier transform $\hat p_i$ of the augmentation probability weights has no zeros.

The intuition is that the mixture of group transformations ``smears'' probability mass among the different transformed versions of the distribution. Analogously to classical deconvolution, this smearing can be undone (``deconvolved'') as long as the convolution does not destroy any frequencies by scaling them to zero.

Some noteworthy consequences of this are:
\begin{itemize}
	\item Assume $p$ is a constant vector $\frac{1}{N} \mathbf{1}$, that is, the augmentation applies the group elements with uniform probability. In this case $\hat p = \delta_0$ and convolution with any zero-mean weight vector is zero. This case is almost certain to cause leaks of the group elements themselves. To see this directly, the mixed augmentation operator is now $\aug := \frac{1}{N} \sum_{j=0}^{N-1} \augg^j$. Consider the true distribution of training samples $\proby$, and a version $\proby' = \augg^k \proby$ into which some element of the transformation group has leaked. Now,
	\begin{equation}
		\aug \proby' = \aug (\augg^k \proby) = \frac{1}{N} \sum_{j=0}^{N-1} \augg^j \augg^k \proby = \frac{1}{N} \sum_{j=0}^{N-1} \augg^{j+k} \proby = \frac{1}{N} \sum_{j=0}^{N-1} \augg^j \proby = \aug \proby
		\label{eq:groupleak}
	\end{equation}
	(recalling the modulo arithmetic in the group powers). By Eq.~\ref{eq:match}, this is a leak, and the training may equally well learn the distribution $\augg^k \proby$ rather than $\proby$. By the same reasoning, any mixture of transformed elements may be learned (possibly even a different one for each image). 
	
	\item Similarly, if $p$ is periodic (with period that is some integer factor of $N$, other than $N$ itself), the Fourier transform is a sparse sequence of spikes separated by zeros. Another viewpoint to this is that the group has a subgroup, whose elements are chosen uniformly. Similar to above, this is almost certain to cause leaks with elements of that subgroup.

	\item With more sporadic zero patterns, the leaks can be seen as ``conditional'': while the augmentation operator has a null space, it is not generally possible to write an equivalent of Eq.~\ref{eq:groupleak} without setting conditions on the distribution $\proby$ itself. In these cases, leaks only occur for specific kinds of distributions, e.g., when a sufficient amount of group symmetry is already present in the distribution itself. 
	
	For example, consider a dataset where all four 90 degree orientations of any image are equally likely, and an augmentation that performs either a $0$ or $90$ degree rotation at equal probability. This corresponds to the probability vector $p = [0.5,~0.5,~0,0]$ over the four elements of the 90-degree rotation group. This distribution has a single zero in its Fourier transform. The associated leak might manifest as the generator only learning to produce images in orientations $0$ and $180$ degrees, and relying on the augmentation to fill the gaps. 
	
	\FINAL{Such a leak could not happen in e.g. a dataset depicting upright faces, and the failure of invertibility would be harmless in this case. However, this may no longer hold when the augmentation is a part of a composed pipeline, as other augmentations may have introduced partial invariances that were not present in the original data.} %
\end{itemize}

In our augmentations involving compact groups (\textbf{rotations and flips}), we always choose the elements with a uniform probability, but importantly, only perform the augmentation with some probability less than one. This combination can be viewed as increasing the probability of choosing the group identity element. The probability vector $p$ is then constant, except for having a higher value at $p_0$; the Fourier transform of such a vector has no zeros.

\paragraph{Non-compact discrete one-parameter groups}
The above reasoning can be extended to groups which are not compact, in particular \textbf{translations by integer offsets} (without periodic boundaries). In the discrete case, such a group is necessarily isomorphic to the additive group $\integer$ of all integers, and no modulo integer arithmetic is performed. The mixture density is then a two-sided sequence $\{p_i\}$ with $i \in \integer$, and the appropriate Fourier transform maps this to a periodic function. By an analogous reasoning with the previous subsection, the invertibility holds as long as this spectrum has no zeros.

\paragraph{Continuous one-parameter groups}
With suitable technical care, these arguments can be extended to continuous groups with elements $\augg_\phi$ indexed by a continuous parameter $\phi$. In the compact case (e.g. \textbf{continuous rotation}), the group elements wrap over at some period $L$, such that $\augg_{\phi+L} = \augg_\phi$. In the non-compact case (e.g. \textbf{translation (addition) and scaling (multiplication) by real-valued amounts}) no such wrap-over occurs. The compact and non-compact groups are isomorphic to $\mathrm{U}(1)$, and the additive group $\real$, respectively. Stochastic mixtures of these group elements are expressed by probability density functions $p(\phi)$, with $\phi \in [0,L)$ if the group is compact, and $\phi \in \real$ otherwise. The Fourier transforms are replaced by the appropriate generalizations, and the invertibility holds when the spectrum has no zeros.

Here it is important to use the correct parametrization of the group. Note that one could in principle parametrize e.g. rotations in arbitrary ways, and it may seem ambiguous as to what parametrization to use, which would appear to render concepts like uniform distribution meaningless. The issue arises when replacing the sums in the earlier formulas with integrals, whereby one needs to choose a measure of integration. These findings apply specifically to the natural \emph{Haar measure} and the associated parametrization -- essentially, the measure that accumulates at constant rate when taking small steps in the group by applying the infinitesimal generator. For rotation groups, the usual ``area'' measure over the angular parametrization coincides with the Haar measure, and therefore e.g. uniform distribution is taken to mean that all angles are chosen equally likely. For translation, the natural Euclidian distance is the correct parametrization. For other groups, such as scaling, the choice is a bit more nuanced: when composing scaling operations, the scale factor combines by multiplication instead of addition, so the natural parametrization is the \emph{logarithm} of the scale factor.

For continuous compact groups (rotation), we use the same scheme as in the discrete case: uniform probability mixed with identity at a probability greater than zero.

For continuous non-compact groups, the Fourier transform of the normal distribution has no zeros and results in an invertible augmentation when used to choose among the group elements. Other distributions with this property are at least the $\alpha$-stable and more generally the infinitely divisible family of distributions. When the parametrization is logarithmic, we may instead use exponentiated values from these distributions (e.g. the log-normal distribution). Finally, stochastically mixing zero-mean normal distributed variables with identity does not introduce zeros to the FT, as it merely lifts the already positive values of the spectrum.

\paragraph{Multi-parameter abelian groups}
Finally, these findings generalize to groups that are products of a finite number of single-parameter groups, provided that the elements of the different groups commute among each other (in other words, finitely generated abelian groups). An example of this is the group of 2-dimensional translations obtained by considering x- and y-translations simultaneously.\footnote{However, for example the non-abelian group of 3-dimensional rotations, $\mathrm{SO}(3)$, is \emph{not} obtained as a product of the single-parameter ``Euler angle'' rotations along three axes, and therefore is not covered by the present formulation of our theory. The reason is that the three different rotations do not commute. One may of course still freely compose the three single-parameter rotation augmentations in sequence, but note that the combined effect can only induce a subset of possible probability distributions on $\mathrm{SO}(3)$.}

The Fourier transforms are replaced with suitable multi-dimensional generalizations, and the probability distributions and their Fourier transforms obtain multidimensional domains accordingly.

\paragraph{Discussion} Invertibility is a \emph{sufficient} condition to ensure the absence of leaks. However, it may not always be \emph{necessary}: in the case of \emph{non-compact} groups, a hypothesis could be made that even a technically non-invertible operator does not leak. 
For example, a shift augmentation with uniform distributed offset on a continuous interval is not invertible, as the Fourier transform of its density is a sinc function with periodic zeros (except at $0$). This only allows for leaks of zero-mean functions whose FT is supported on this evenly spaced set of frequencies -- in other words, infinitely periodic functions. Even though such functions are in the null space of the augmentation operator, they cannot be added to any density in an infinite domain without violating non-negativity, and so we may hypothesize that no leak can in fact occur. In practice, however, the near-zero spectrum values might allow for a periodic leak modulated by a wide window function to occur for very specific (and possibly contrived) data distributions.

In contrast, straightforward examples and practical demonstrations of leaks are easily found for compact groups, e.g. with uniform or periodic rotations.

\subsubsection{Noise and image filter augmentations}
We refer to Theorem 5.3. of Bora et al.~\cite{Bora2018}, where it is shown that in a setting effectively identical to ours, \textbf{addition of noise that is independent of the image is an invertible operation as long as the Fourier spectrum of the noise distribution does not contain zeros}. The reason is that addition of mutually independent random variables results in a convolution of their probability distributions. Similar to groups, this is a multiplication in the Fourier domain, and the zeros correspond to irrevocable loss of information, making the inversion impossible. The inverse can be realized by ``deconvolution'', or division in the Fourier domain.

A potential source of confusion is that the Fourier transform is commonly used to describe spatial correlations of noise in signal processing. We refer to a different concept, namely the Fourier transform of the probability density of the noise, often called the \emph{characteristic function} in probability literature (although correlated noise is also subsumed by this analysis).

\paragraph{Gaussian product noise} In our setting, we also randomize the magnitude parameter of the noise, in effect stochastically mixing between different noise distributions. The above analysis subsumes this case, as the mixture is also a random noise, with a density that is a weighted blend between the densities of the base noises. However, the noise is no longer independent across points, so its joint distribution is no longer separable to a product of marginals, and one must consider the joint Fourier transform in full dimension. 

Specifically, we draw the per-pixel noise from a normal distribution and modulate this entire noise field by a multiplication with a single (half-)normal random number. The resulting distribution has an everywhere nonzero Fourier transform and hence is invertible. To see this, first consider two standard normal distributed random scalars $X$ and $Y$, and their product $Z = XY$ (taken in the sense of multiplying the random variables, not the densities). Then $Z$ is distributed according to the density $p_Z(Z) = \frac{K_0(|Z|)}{\pi}$, where $K_0$ is a modified Bessel function, and has the characteristic function (Fourier transform) $\hat p_Z(\omega) = \frac{1}{\sqrt{\omega^2+1}}$, which is everywhere positive \cite{Wishart1932}.

Then, considering our situation with a product of a normal distributed scalar $X$ and an independent normal distributed vector $\mathbf{Y} \in \real^N$, the $N$ entries of the product $\mathbf{Z} = X \mathbf{Y}$ become mutually dependent. The \emph{marginal} distribution of each entry is nevertheless exactly the above product distribution $p_Z$. By Fourier slice theorem, all one-dimensional slices through the main axes of the characteristic function of $\mathbf{Z}$ must then coincide with the characteristic function $\hat p_Z$ of this marginal distribution. Finally, because the joint distribution is radially symmetric, so is the characteristic function, and this must apply to \emph{all} slices through the origin, yielding the everywhere positive Fourier transform $\hat p_\mathbf{Z}(\bm{\omega}) = \frac{1}{\sqrt{|\bm{\omega}|^2+1}}$. When stochastically mixed with identity (as is our random skipping procedure), the Fourier Transform values are merely lifted towards $1$ and no new zero-crossings are introduced.

\paragraph{Additive noise in transformed bases} Similar notes apply to additive noise in a different basis: one can consider the noise augmentation as being flanked by an invertible deterministic (possibly also nonlinear) basis transformation and its inverse. It then suffices to show that the additive noise has a non-zero spectrum in isolation. In particular, multiplicative noise with a non-negative distribution can be viewed as additive noise in logarithmic space and is invertible if the logarithmic version of the noise distribution has no zeros in its Fourier transform. The \textbf{image-space filters} are a combination of a linear basis transformation to the wavelet basis, and additive Gaussian noise under a non-linear logarithmic transformation. %

\subsubsection{Random projection augmentations}

The \textbf{cutout augmentation} (as well as e.g. the pixel and patch blocking in AmbientGAN~\cite{Bora2018}) can be interpreted as projecting a random subset of the dimensions to zero.

Let $\augp_1, \augp_2, ..., \augp_N$ be a set of deterministic projection augmentation operators with the defining property that $\augp_j^2 = \augp_j$. %
For example, each one of these operators can set a different fixed rectangular region to zero. Clearly the individual projections have a null space (unless they are the identity projection) and they are not invertible in isolation.

Consider a stochastic augmentation that randomly applies one of these projections, or the identity. Let $p_0,~p_1,~...,~p_N$ denote the discrete probabilities of choosing the identity operator $\augeye$ for $p_0$, and $\augp_k$ for the remaining $p_k$. Define the mixture of the projections as:
\begin{equation}
	\aug = p_0 \augeye + \sum_{j=1}^N p_j \augp_j
\end{equation}
Again, $\aug$ is a mixture of operators, however unlike in earlier examples, some (but not all) of the operators are non-invertible. Under what conditions on the probability distribution $p$ is $\aug$ invertible?

Assume that $\aug$ is not invertible, i.e. there exists a probability distribution $\probx \ne 0$ such that $\aug \probx = 0$. Then
\begin{equation}
	0 = \aug \probx = p_0 \probx + \sum_{j=1}^N p_j \augp_j \probx
\end{equation}
and rearranging,
\begin{equation}
	\sum_{j=1}^N p_j \augp_j \probx = -p_0 \probx
\end{equation}
Under reasonable technical assumptions (e.g. discreteness of the pixel intensity values, such as justified in Theorem 5.4. of Bora et al.~\cite{Bora2018}), we can consider the inner product of both sides of this equation with $\probx$:
\begin{equation}
	\sum_{j=1}^N p_j \langle \probx, \augp_j \probx \rangle = -p_0 \langle \probx, \probx \rangle
\end{equation}
The right side of this equation is strictly negative if the probability $p_0$ of identity is greater than zero, as $\probx \ne 0$. The left side is a non-negative sum of non-negative terms, as the inner product of a vector with its projection is non-negative. %
Therefore, the assumption leads to a contradiction unless $p_0 = 0$; conversely, \textbf{random projection augmentation does not leak if there is a non-zero probability that it produces the identity.}

\subsection{Practical considerations}

\subsubsection{Conditioning}
In practical numerical computation, an operator that is technically invertible may nevertheless be so close to a non-invertible configuration that inversion fails in practice. Assuming a finite state space, this notion is captured by the condition number, which is infinite when the matrix is singular, and large when it is singular for all practical purposes. The same consideration applies to infinite state spaces, but the appropriate technical notion of conditioning is less clear.

The practical value of the analysis in this section is in identifying the conditions where exact non-invertibility happens, so that appropriate safety margin can be kept. We achieve this by regulating the probability $p$ of performing a given augmentation, and keeping it at a safe distance from $p=1$ which for many of the augmentations corresponds to a non-invertible condition (e.g. uniform distribution over compact group elements).

For example, consider applying transformations from a finite group with a uniform probability distribution, where the augmentation is applied with probability $p$. In a finite state space, a matrix corresponding to this augmentation has $1-p$ for its smallest singular value, and $1$ for the largest, resulting in condition number $1/(1-p)$ which approaches infinity as $p$ approaches one.

\subsubsection{Pixel-level effects and boundaries}
When dealing with images represented on finite pixel grids, naive practical implementations of some of the group transformations do not strictly speaking form groups. For example, a composition of two continuous rotations of an image with angles $\phi$ and $\theta$ %
\FINAL{does not generally} reproduce the same image as a single rotation by angle $\phi+\theta$, %
\FINAL{if the transformed image is resampled} to the rectangular pixel grid twice. Furthermore, parts of the image may fall outside the boundaries of the grid, whereby their values are lost and cannot be restored even if a reverse transformation is made afterwards, unless special care is taken. These effects may become significant when multiple transformations are composed. %

In our implementation, we mitigate these issues as much as possible by accumulating the chain of transformations into a matrix and a vector representing the total affine transformation implemented by all the grouped augmentations, and only then applying it on the image. This is possible because all the augmentations we use are affine transformations in the image (or color) space. Furthermore, prior to applying the geometric transformations, the images are reflection padded and scaled to double resolution (and conversely, cropped and downscaled afterwards). Effectively the image is then treated as an infinite tiling of suitably reflected finer-resolution copies of itself, and a practical target-resolution crop is only sampled at augmentation time.

\section{Implementation details}
\label{app:implementation}

We implemented our techniques on top of the \FINAL{StyleGAN2 official TensorFlow implementation}\footnote{\texttt{\small https://github.com/NVlabs/stylegan2}}.
We kept most of the details unchanged, including
	network architectures \cite{Karras2019},
	weight demodulation \cite{Karras2019},
	path length regularization \cite{Karras2019},
	lazy regularization \cite{Karras2019},
	style mixing regularization \cite{Karras2018},
	bilinear filtering in all up/downsampling layers \cite{Karras2018},
	equalized learning rate for all trainable parameters \cite{Karras2017},
	minibatch standard deviation layer at the end of the discriminator \cite{Karras2017},
	exponential moving average of generator weights \cite{Karras2017},
	non-saturating logistic loss \cite{Goodfellow2014} with $R_1$ regularization \cite{Mescheder2018},
	and Adam optimizer \cite{adam} with $\beta_1=0$, $\beta_2=0.99$, and $\epsilon=10^{-8}$.

We ran our experiments on a computing cluster with a few dozen NVIDIA DGX-1s, each containing 8 Tesla V100 GPUs, using TensorFlow 1.14.0, PyTorch 1.1.0 (for comparison methods), CUDA 10.0, and cuDNN 7.6.3.
We used the official pre-trained Inception network\footnote{\texttt{\small http://download.tensorflow.org/models/image/imagenet/inception-2015-12-05.tgz}} to compute FID, KID, and Inception score.

\subsection{Hyperparameters and training configurations}
\label{app:cifarconditioning}

\figHyperparams{fig:Hyperparams} %
Figure~\ref{fig:Hyperparams} shows the hyperparameters that we used in our experiments, as well as the original StyleGAN2 config \textsc{f}~\cite{Karras2019}.
We performed all training runs using 8 GPUs and continued the training until the discriminator had seen a total of 25M real images, except for \textsc{CIFAR-10}, where we used 2 GPUs and 100M images.
We used minibatch size of 64 when possible, but reverted to 32 for \textsc{MetFaces} in order to avoid running out of GPU memory.
\FINAL{Similar to StyleGAN2, we evaluated the minibatch standard deviation layer independently over the images processed by each GPU.}

\paragraph{Dataset augmentation}
We did not use dataset augmentation in any of our experiments with \textsc{FFHQ}, \mbox{\textsc{LSUN Cat}}, or \textsc{CIFAR-10}, except for the FFHQ-140k case and in Figure~\ref{fig:LeakyComparison}.
In particular, we feel that leaky augmentations are inappropriate for \textsc{CIFAR-10} given its status as a standard benchmark dataset, where dataset/leaky augmentations would unfairly inflate the results.
\textsc{MetFaces}, \textsc{BreCaHad}, and \textsc{AFHQ Dog} are horizontally symmetric in nature, so we chose to enable dataset $x$-flips for these datasets to maximize result quality.

\paragraph{Network capacity}
We follow the original StyleGAN2 configuration for high-resolution datasets ($\ge512^2$):
a layer operating on \mbox{$N = w \times h$} pixels uses \mbox{$\min\big( 2^{16}/\sqrt{N}, 512 \big)$} feature maps.
With \textsc{CIFAR-10} we use 512 feature maps for all layers. %
In the $256\times256$ configuration used with \textsc{FFHQ} and \mbox{\textsc{LSUN Cat}}, we facilitate extensive sweeps over dataset sizes by decreasing the number of feature maps to \mbox{$\min\big( 2^{15}/\sqrt{N}, 512 \big)$}.

\paragraph{Learning rate and weight averaging}
We selected the optimal learning rates using grid search and found that it is generally beneficial to use the highest learning rate that does not result in training instability.
We also found that larger minibatch size allows for a slightly higher learning rate.
For the moving average of generator weights~\cite{Karras2017}, the natural choice is to parameterize the decay rate with respect to minibatches\,---\,not individual images\,---\,so that increasing the minibatch size results in a longer decay.
\FINAL{Furthermore, we observed that a very long moving average consistently gave the best results on CIFAR-10. To reduce startup bias, we linearly ramp up the length parameter from 0 to 500k over the first 10M images.}

\paragraph{R1 regularization}
Karras~et~al.~\cite{Karras2019} postulated that the best choice for the $R_1$ regularization weight $\gamma$ is highly
dependent on the dataset.
We thus performed extensive grid search for each column in Figure~\ref{fig:Hyperparams}, considering $\gamma \in \{0.001, 0.002, 0.005, \ldots, 20, 50, 100\}$.
Although the optimal $\gamma$ does vary wildly, from 0.01 to 10, it seems to scale almost linearly with the resolution of the dataset.
\FINAL{In practice, we have found that a good initial guess is given by $\gamma_0 = 0.0002 \cdot N / M$, where $N = w \times h$ is the number of pixels and $M$ is the minibatch size. Nevertheless, the optimal value of $\gamma$ tends to vary depending on the dataset, so we recommend experimenting with different values in the range $\gamma \in [\gamma_0 / 5, \gamma_0 \cdot 5]$.}

\paragraph{Mixed-precision training}
We utilize the high-performance Tensor Cores available in Volta-class GPUs by employing mixed-precision FP16/FP32 training in all of our experiments (with two exceptions, discussed in Appendix~\ref{app:comparisonmethods}).
We store the trainable parameters with full FP32 precision for the purposes of optimization but cast them to FP16 before evaluating $G$ and $D$.
The main challenge with mixed-precision training is that the numerical range of FP16 is limited to $\sim\pm2^{16}$, as opposed to $\sim\pm2^{128}$ for FP32.
Thus, any unexpected spikes in signal magnitude\,---\,no matter how transient\,---\,will immediately collapse the training dynamics.
We found that the risk of such spikes can be reduced drastically using three tricks:
first, by limiting the use of FP16 to only the 4 highest resolutions, i.e., layers for which $N_\ps{layer} \ge N_\ps{dataset} / (2\times2)^4$;
second, by pre-normalizing the style vector $s$ and each row of the weight tensor $w$ before applying weight modulation and demodulation\footnote{Note that our pre-normalization only affects the intermediate results; it has no effect on the final output of the convolution layer due to the subsequent post-normalization performed by weight demodulation.};
and third, by clamping the output of every convolutional layer to $\pm2^8$, i.e., an order of magnitude wider range than is needed in practice.
We observed about 60\% end-to-end speedup from using FP16 and verified that the results were virtually identical to FP32 on our baseline configuration.

\paragraph{CIFAR-10}
\FINAL{
We enable class-conditional image generation on CIFAR-10 by extending the original StyleGAN2 architecture as follows.
For the generator, we embed the class identifier into a 512-dimensional vector that we concatenate with the original latent code after normalizing each, i.e., $z' = \mathrm{concat}\big(\mathrm{norm}(z), \mathrm{norm}(\mathrm{embed}(c))\big)$, where $c$ is the class identifier.
For the discriminator, we follow the approach of Miyato~and~Koyama~\cite{Miyato2018} by evaluating the final discriminator output as $D(x) = \mathrm{norm}\big(\mathrm{embed}(c)\big) \cdot D'(x)^T$, where $D'(x)$ corresponds to the feature vector produced by the last layer of $D$.
}
\FINAL{To compute FID, we generate 50k images using randomly selected class labels and compare their statistics against the 50k images from the training set. For IS, we compute the mean over 10 independent trials using 5k generated images per trial.}
As illustrated in \FINAL{Figures~\refpaper{fig:SmallDatasetResults}b} and \ref{fig:Hyperparams}, we found that we can improve the FID considerably by disabling style mixing regularization~\cite{Karras2018}, path length regularization~\cite{Karras2019}, and residual connections in $D$~\cite{Karras2019}.
Note that all of these features are highly beneficial on higher-resolution datasets such as FFHQ.
We find it somewhat alarming that they have precisely the opposite effect on CIFAR-10\,---\,this suggests that some previous conclusions reached in the literature using CIFAR-10 may fail to generalize to other datasets.

\subsection{Comparison methods}
\label{app:comparisonmethods}

We implemented the comparison methods shown in \FINAL{Figures~\refpaper{fig:ComparisonMethods}a} on top of our baseline configuration, identifying the best-performing hyperparameters for each method via extensive grid search.
Furthermore, we inspected the resulting network weights and training dynamics in detail to verify correct behavior, e.g., that with the discriminator indeed learns to correctly handle the auxiliary tasks with PA-GAN and auxiliary rotations.
We found zCR and WGAN-GP to be inherently incompatible with our mixed-precision training setup due to their large variation in gradient magnitudes.
We thus reverted to full-precision FP32 for these methods.
Similarly, we found lazy regularization to be incompatible with bCR, zCR, WGAN-GP, and auxiliary rotations.
Thus, we included their corresponding loss terms directly into our main training loss, evaluated on every minibatch.

\paragraph{bCR}
We implement balanced consistency regularization proposed by Zhao~et~al.~\cite{zhao2020improved} by introducing two new loss terms as shown in Figure~\refpaper{fig:Augment}a.
We set $\lambda_\text{real} = \lambda_\text{fake} = 10$ and use integer translations on the range of $[-8, +8]$ pixels.
In Figure~\ref{fig:LeakyComparison}, we also perform experiments with $x$-flips and arbitrary rotations.

\paragraph{zCR}
In addition to bCR, Zhao~et~al.~\cite{zhao2020improved} also propose latent consistency regularization (zCR) to improve the diversity of the generated images.
We implement zCR by perturbing each component of the latent ${\bf z}$ by $\sigma_\text{noise} = 0.1$ and encouraging the generator to maximize the $L_2$ difference between the generated images, measured as an average over the pixels, with weight $\lambda_\text{gen} = 0.02$.
Similarly, we encourage the discriminator to minimize the $L_2$ difference in $D(x)$ with weight $\lambda_\text{dis} = 0.2$.

\paragraph{PA-GAN}
Zhang~and~Khoreva~\cite{Khoreva2019} propose to reduce overfitting by requiring the discriminator to learn an auxiliary checksum task.
This is done by providing a random bit string as additional input to $D$, requiring that the sign of the output is flipped based on the parity of bits that were set, and dynamically increasing the number of bits when overfitting is detected.
We select the number of bits using our $r_t$ heuristic with target 0.95.
Given the value of $p$ produced by the heuristic, we calculate the number of bits as $k = \lceil p \cdot 16 \rceil$.
Similar to Zhang~and~Khoreva, we fade in the effect of newly added bits smoothly over the course of training.
In practice, we use a fixed string of 16 bits, where the first $k-1$ bits are sampled from $\mathrm{Bernoulli}(0.5)$, the $k$\textsuperscript{th} bit is sampled from $\mathrm{Bernoulli}\big(\mathrm{min}(p \cdot 16 - k + 1, 0.5) \big)$, and the remaining $16-k$ bits are set to zero.

\paragraph{WGAN-GP}
For WGAN-GP, proposed by Gulrajani~et~al.~\cite{Gulrajani2017}, we reuse the existing implementation included in the StyleGAN2 codebase with $\lambda = 10$.
We found WGAN-GP to be quite unstable in our baseline configuration, which necessitated us to disable mixed-precision training and lazy regularization, as well as to settle for a considerably lower learning rate $\eta = 0.0010$.

\paragraph{Auxiliary rotations}
Chen~et~al.~\cite{Chen2018aux} propose to improve GAN training by introducing an auxiliary rotation loss for $G$ and $D$.
In addition the main training objective, the discriminator is shown real images augmented with 90$^{\circ}$ rotations and asked to detect their correct orientation.
Similarly, the generator is encouraged to produce images whose orientation is easy for the discriminator to detect correctly.
We implement this method by introducing two new loss terms that are evaluated on a $4\times$ larger minibatch, consisting of rotated versions of the images shown to the discriminator as a part of the main loss.
We extend the last layer of $D$ to output 5 scalar values instead of one and interpret the last 4 components as raw logits for softmax cross-entropy loss.
We weight the additional loss terms using $\alpha = 10$ for $G$, and $\beta = 5$ for $D$.

\paragraph{Spectral normalization}
Miyato~et~al.~\cite{Miyato2018B} propose to regularize the discriminator by explicitly enforcing an upper bound for its Lipschitz constant, and several follow-up works \cite{Zhang2018sagan,Brock2018,zhao2020improved,Schonfeld2020} have found it to be beneficial.
Given that spectral normalization is effectively a no-op when applied to the StyleGAN2 generator \cite{Karras2019}, we apply it only to the discriminator.
We ported the original Chainer implementation\footnote{\texttt{\small https://github.com/pfnet-research/sngan\_projection}} to TensorFlow, and applied it to the main convolution layers of $D$.
We found it beneficial to not use spectral normalization with the {\tt FromRGB} layer, residual skip connections, or the last fully-connected layer.

\paragraph{Freeze-D}
Mo~et~al.~\cite{Mo2020} propose to freeze the first $k$ layers of the discriminator to improve results with transfer learning.
We tested several different choices for $k$\FINAL{; the best results were given by $k=10$ in Figure~\refpaper{fig:TransferLearning} and by $k=13$ in Figure~\refpaper{fig:SmallDatasetResults}b.}
In practice, this corresponds to freezing all layers operating at the \FINAL{3 or 4 highest resolutions, respectively.}

\paragraph{BigGAN}
BigGAN results in Figures~\ref{fig:BigGANComparison} and \ref{fig:SubsetImagesFFHQ} were run on a modified version of the original BigGAN PyTorch implementation\footnote{\texttt{\small https://github.com/ajbrock/BigGAN-PyTorch}}.
The implementation was adapted for unconditional operation following Sch\"onfeld~et~al.~~\cite{Schonfeld2020} by matching their hyperparameters, replacing class-conditional BatchNorm
with self-modulation, where the BatchNorm parameters are conditioned only on the latent vector $z$, and not using class projection in the discriminator.

\paragraph{Mapping network depth}
For the ``Shallow mapping'' case in \FINAL{Figure~\refpaper{fig:ComparisonMethods}a}, we reduced the depth of the mapping network from 8 to 2.
Reducing the depth further than 2 yielded consistently inferior results, confirming the usefulness of the mapping network.
\FINAL{In general, we found depth 2 to yield slightly better results than depth 8, making it a good default choice for future work.}

\paragraph{Adaptive dropout}
Dropout~\cite{srivastava2014} is a well-known technique for combating overfitting in practically all areas of machine learning.
In \FINAL{Figure~\refpaper{fig:ComparisonMethods}a}, we employ multiplicative Gaussian dropout for all layers of the discriminator, similar to the approach employed by Karras~et~al.~\cite{Karras2017} in the context of LSGAN loss~\cite{Mao2016}.
We adjust the standard deviation dynamically using our $r_t$ heuristic with target 0.6, so that the resulting $p$ is used directly as the value for $\sigma$.

\subsection{MetFaces dataset}

We have collected a new dataset, MetFaces, by extracting images of human faces from the Metropolitan Museum of Art online collection.
Dataset images were searched using terms such as `paintings', `watercolor' and `oil on canvas', and downloaded via the \texttt{\small https://metmuseum.github.io/} API.
This resulted in a set of source images that depicted paintings, drawings, and statues.
Various automated heuristics, such as face detection and image quality metrics, were used to narrow down the set of images
to contain only human faces.  A manual selection pass over the remaining images was performed to weed out poor quality images
not caught by automated filtering.  Finally, faces were cropped and aligned to produce 1,336 high quality images
at $1024^2$ resolution.

The whole dataset, including the unprocessed images, is available at\\
{\small\url{https://github.com/NVlabs/metfaces-dataset}}

\section{Energy consumption}
\label{app:power}

\FINAL{
Computation is a core resource in any machine learning project: its availability and cost, as well as the associated energy consumption, are key factors in both choosing research directions and practical adoption. We provide a detailed breakdown for our entire project in Table~\ref{fig:Power} in terms of both GPU time and electricity consumption.
We report expended computational effort as single-GPU years (Volta class GPU). We used a varying number of NVIDIA DGX-1s for different stages of the project, and converted each run to single-GPU equivalents by simply scaling by the number of GPUs used.

We followed the Green500 power measurements guidelines \cite{Ge2020} similarly to Karras et al.~\cite{Karras2019}. 
The entire project consumed approximately 300 megawatt hours (MWh) of electricity. 
Almost half of the total energy was spent on exploration and shaping the ideas before the actual paper production started. Subsequently the majority of computation was targeted towards the extensive sweeps shown in various figures. 
Given that ADA does not significantly affect the cost of training a single model, e.g., training StyleGAN2~\cite{Karras2019} with $1024\times1024$ FFHQ still takes approximately 0.7 MWh.
}

\figPower{fig:Power} %

\fi

\end{document}